\documentclass[aos]{imsart}

\usepackage{latexsym}
\usepackage{verbatim}                 %%
\usepackage{graphicx,psfrag,epstopdf}%[dvips]
\usepackage{enumerate}
\usepackage{enumitem}
\usepackage{makecell,multirow}
\usepackage{mathrsfs}
\usepackage{amsfonts,amsmath,amssymb,amsthm, amssymb} %%
\usepackage[all]{xy}
\usepackage{xcolor,bm}
\usepackage[numbers]{natbib}
\usepackage{dsfont}

\usepackage[english]{babel}

\usepackage{tikz}   %The tikz library is not imported by default. So, it must be imported manually in the preamble of the document (after documentclass, but before the begin document statements).

\usetikzlibrary{fit, positioning, arrows, automata, calc, bayesnet, shapes, matrix}
\tikzset{
->, % makes the edges directed
>=stealth', % makes the arrow heads bold
node distance=3cm, % specifies the minimum distance between two nodes. Change if necessary.
every state/.style={thick, fill=gray!10}, % sets the properties for each 'state' node
initial text=$ $, % sets the text that appears on the start arrow
}

\usepackage[normalem]{ulem}               % to striketrhourhg text
\newcommand\redout{\bgroup\markoverwith
{\textcolor{red}{\rule[0.5ex]{2pt}{0.8pt}}}\ULon}

\theoremstyle{plain}

\newtheorem*{theorem*}{Theorem}

\newtheorem{thm}{Theorem}[section]
\newtheorem{lem}{Lemma}[section] %{Lemma}
\newtheorem{col}{Corollary}[section]
\newtheorem{defn}{Definition}[section]
\newtheorem{comm}{comment}[section]

\begin{document}

\begin{frontmatter}
\title{Measure-Theoretic Probability of Complex Co-occurrence and $E$-Integral}

\begin{aug}
%%%%%%%%%%%%%%%%%%%%%%%%%%%%%%%%%%%%%%%%%%%%%%%
%% Only one address is permitted per author. %%
%% Only division, organization and e-mail is %%
%% included in the address.                  %%
%% Additional information can be included in %%
%% the Acknowledgments section if necessary. %%
%%%%%%%%%%%%%%%%%%%%%%%%%%%%%%%%%%%%%%%%%%%%%%%
\author[A]{\fnms{Jian-Yong} \snm{Wang}\ead[label=e1]{jywang@xmu.edu.cn}}
\and
\author[B]{\fnms{Han} \snm{Yu}\ead[label=e2,mark]{han.yu@unco.edu}}
%%%%%%%%%%%%%%%%%%%%%%%%%%%%%%%%%%%%%%%%%%%%%%
%% Addresses                                %%
%%%%%%%%%%%%%%%%%%%%%%%%%%%%%%%%%%%%%%%%%%%%%%
\address[A]{School of Mathematical Sciences,
Xiamen University, Xiamen 361005, China.
\printead{e1}}

\address[B]{Department of Applied Statistics and Research Methods, University of Northern Colorado, Greeley, CO 80639, USA.
\printead{e2}}
\end{aug}

\begin{abstract}
Complex high-dimensional co-occurrence data are increasingly popular from a complex system of interacting physical, biological and social processes in discretely indexed modifiable areal units or continuously indexed locations of a study region for landscape-based mechanism. Modeling, predicting and interpreting complex co-occurrences are very general and fundamental problems of statistical and machine learning in a broad variety of real-world modern applications. Probability and conditional probability of co-occurrence are introduced by being defined in a general setting with set functions to develop a rigorous measure-theoretic foundation for the inherent challenge of data sparseness. The data sparseness is a main challenge inherent to probabilistic modeling and reasoning of co-occurrence in statistical inference. The behavior of a class of natural integrals called E-integrals is investigated based on the defined conditional probability of co-occurrence. The results on the properties of E-integral are presented. The paper offers a novel measure-theoretic framework where E-integral as a basic measure-theoretic concept can be the starting point for the expectation functional approach preferred by Whittle (1992) and Pollard (2001) to the development of probability theory for the inherent challenge of  co-occurrences emerging in modern high-dimensional co-occurrence data problems and opens the doors to more sophisticated and interesting research in complex high-dimensional co-occurrence data science.
\end{abstract}



\end{frontmatter}

\section {Introduction}\label{section1}

A large number of events occurs simultaneously in practice from a complex system of interacting physical, biological and social processes in discretely indexed modifiable areal units or continuously indexed locations of a study region.  Heterogeneous high-dimensional data are nowadays rule in health, medicine, epidemiology, technology, econometrics, business, finance, sociology, and political science, to name just a few. In public health study, multimorbidity, the co-existence of multiple chronic conditions in an individual in different time periods \cite{Glynn:2011, Marengoni:2011, Marengoni:2008}, has been identified as one of the major health system concerns of the twenty-first century \cite{McPhail:2016, Oliver:2012}. Not only older adults but also a substantial number of young and middle-aged people also have multimorbidity \cite{Barnett:2012, Agborsangaya:2012, Fortin:2005}. In computer vision and text mining, words and images are collected as co-occurrence data to match textual information hidden in words and visible information hidden in images defined by a network for interpretation \cite{Barnard:1995, Mori:1999, Tate:2001}. In social sciences, the semantic network of words appearing together in the texts, such as newspaper articles, political speeches, novels and fiction, or transcripts of debates, has been emerging as a basis to test existing hypotheses, formulate theories, understand the prevailing discourses, and tailor their messages more effectively within our society \cite{Segev:2021}.

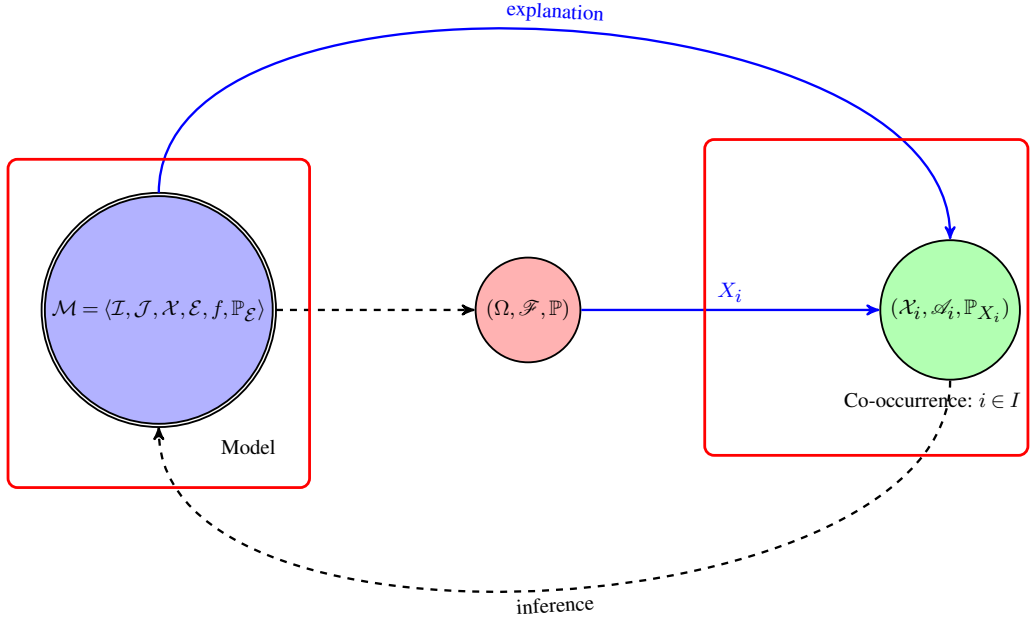
\begin{figure}
\begin{center}
\resizebox{.6\textheight}{!}{%
\begin{tikzpicture}
\tikzstyle{main}=[circle, minimum size = 14mm, thick, draw =black!80, node distance = 16mm]
\node[state, fill=red!30]        (O) {$(\Omega, \mathscr{F}, \mathbb{P})$};
\node[state, left= of O, accepting, fill=blue!30] (M1) {$\mathcal{M}= \langle \mathcal{I, J, X, E}, f, \mathbb{P}_{\mathcal{E}}\rangle$};
\node[state,  right=4.5cm of O, fill=green!30]                 (O1) {$(\mathcal{X}_i, \mathscr{A}_i, \mathbb{P}_{X_i})$};
\draw[line width= 1,dashed]
      (M1) edge[below] node[sloped]{} (O)
      (O1) to  [below, out=270,in=270, looseness=.8]   node[sloped]{inference} (M1);
\draw[->,blue, line width= 1]
      (O) edge[above] node[sloped]{$X_i$} (O1)
      (M1) to  [above, out=90,in=90, looseness=.8]   node[sloped]{explanation} (O1);
\plate [color=red, very thick, inner sep=1.1cm, xshift=-1cm, yshift=.4cm] {plate1} {(O1)} {Co-occurrence: $i\in I$};
\plate [color=red, very thick, inner sep=0.5cm, xshift=0cm, yshift=0cm] {plate2} {(M1)} {Model};
\end{tikzpicture}\label{MAE}
}\caption{Mechanism-based explanation and interpretation of co-occurrences from a fixed probability space $(\Omega, \mathscr{F}, \mathbb{P})$ that sits in the background.}
\end{center}
\end{figure}

Modeling, interpreting and predicting co-occurrences of events is a general and fundamental problem of statistical and machine learning. It has a wide variety of real-world modern applications in information retrieval, natural language processing,  computer vision, data mining, remote sensing data analysis, joint models of different types of distributions, measurement error models, space-time coregionalization models, spatio-temporal dynamic models. Much of research questions in these modern applications are an attempt to infer co-occurrence relationships in the context of probabilistic modeling and analysis of association and causation with statistical evidence collected under conditions where fully randomized controlled experiments are not possible. The ultimate goal of statistical modeling is to interpret and explain the co-occurrence data with a probabilistic model illustrated in Figure~\ref{MAE}. Furthermore, one is often considerably interested in inferring causal and counterfactual links between variables beyond association in sciences. The causal relationships between the variables are expressed in the configuration of deterministic, functional relationships and probabilities are introduced through the assumption that certain variables are exogenous latent random variables. When working with mechanism-based interpretation and explanation of the co-occurrence data \cite{Hedstrom:2010}, advanced spatio-temporal structural equation models (SEM) can be considered as fundamental representations of the complex co-occurrence problems. In order to formulate such a question in general, we consider structural causal models (SCMs),  also known as nonparametric structural equation models (NPSEMs). A structural causal model is a tuple
\begin{equation}\mathcal{M}= \langle \mathcal{I, J, X, E}, f, \mathbb{P}_{\mathcal{E}}\rangle, \label{eqn:scm}
\end{equation}
where $\mathcal{I}$ is an index set of endogenous variables, $\mathcal{J}$ is an index set of exogenous variables, $\mathcal{X} =\Pi_{i\in \mathcal{I}}\mathcal{X}_i$, or just $\mathcal{X}_{\mathcal{I}}$ for short, is the product space of the endogenous variables with $\mathcal{X}_i$ being a measurable space to set interventions, $\mathcal{E} =\Pi_{j\in \mathcal{J}}\mathcal{E}_j$, or just $\mathcal{E}_{\mathcal{J}}$ for short, is the product space of the exogenous variables with $\mathcal{E}_j$ being a measurable space, $\mathbb{P}_\mathcal{E}$ is a probability measure on $\mathcal{E}$  for the exogenous unstructural disturbance noise, and $f: \mathcal{X}\times \mathcal{E} \rightarrow\mathcal{X}$ is a measurable function that specifies the causal mechanism encoded in structural equations \cite{Bongers:2021}. This modeling allows us to represent interventions in an unambiguous way by changing the causal mechanisms that target specific endogenous variables as well as encode the structural properties of the functional relations in a graph with index sets. A pair of random variables $(X_{\mathcal{I}}, E_{\mathcal{J}})$ is a solution of the SCM $\mathcal{M}= \langle \mathcal{I, J, X, E}, f, \mathbb{P}_{\mathcal{E}}\rangle$ if the perturbation is equal to $\mathbb{P}_{\mathcal{E}_{\mathcal{J}}}$ (i.e., $\mathbb{P}_{E_{\mathcal{J}}} = \mathbb{P}_{\mathcal{E}_{\mathcal{J}}}$) and the structural equations are satisfied (i.e., $X_{\mathcal{I}} = f(X_{\mathcal{I}}, E_{\mathcal{J}}) \ a.s.$).  The endogenous $X_{\mathcal{I}}$ is observable, while the exogenous random variables $E_{\mathcal{J}}$ are latent. For a solution $X_{\mathcal{I}}$, we call the distribution $\mathbb{P}_{X_{\mathcal{I}}}$ the observational distribution of $\mathcal{M}$ associated to $X_{\mathcal{I}}$. SCMs arise in genetics \cite{Wright:1921}, econometrics \cite{Haavelmo:1943}, electrical engineering \cite{Mason:1953, Mason:1956} and the social sciences \cite{Duncan:1975, Duncan:1973}.  SCMs are widely used for causal modeling \cite{Bollen:1989, Pearl:2009, Peters:2017, Spirtes:2000} and the corresponding statistical methods are developed for causal inference \cite{Buhlmann:2014, Maathuis:2009, Mooij:2013, Mooij:2016, Peters:2014}.

The types of research questions in these areas continues to increase in their complexity with the advance of technology. As the higher order co-occurrences, e.g. co-occurrence in triples, quadruples, etc, are observed, the intrinsic data sparseness problem of co-occurrence data becomes more urgent than that of co-occurrence in pairs \cite{Hofmann:1998}.  The complexity of encoding the model and describing the raw data conditioned on that model is encoded in a class of index sets. The class of index sets includes parents of variables and sufficient set or admissible set for adjustment in the causal inference context \cite{Pearl:2009}. The class of index sets plays a critical role in the identifying assumptions underlying all causal inferences, the languages used in formulating the assumptions, the conditional nature of all causal and counterfactual claims, and the methods developed for the assessment of such claims \cite{Peters:2016}. Index sets are also critical in capturing the significant variations important to the process being modeled and understanding what is measured and perceived \cite{Gine:2016}. A physical, biological or social process cannot be modeled and identified successfully to answer causal questions unless data are available at appropriate indices and their structure. Many researchers have shown that different scales of index set---hence aggregations--- often lead to contradictory interpretations, such paradoxes being referred to as ``ecological fallacies" \cite{Robinson:1950, King:1997} or known as the modifiable areal unit problem (MAUP) in spatial analysis and geographical information science \cite{Openshaw:1983, Anselin:2000, Tate:2001}. The measure-theoretic treatment of the sparseness problem provides insights into the problem of co-occurrence data in a unified foundation. In the culture of data science, we pursue the fundamental interpretation and understanding of scientific problems arising from co-occurrence events.

Let $(\Omega,\mathscr{F},P)$ be a given probability space. The space might in practice be geographic space, or socio-economic space, or more generally network space as abstraction of reality. We usually call an element $A\in\mathscr{F}$ an event and $P(A)$ is the probability of occurrence of event $A$ or abbreviated as the probability of  event $A$. In many classical books on probability or measure (see, e.g., \cite{Billingsley:1995},  \cite{Durrett:2019}, \cite{Bogachev:2007}, \cite{yan:2004}, \cite{Kailai:2000}, \cite{Robert:2000}), the definitions of conditional probability and conditional expectation are well given. For example, given $A,B\in\mathscr{F}$ with $P(B)>0$, the conditional probability of event $A$ given event $B$ is $P(A\cap B)/P(B)$. Note that, since $A\cap B\in\mathscr{F}$, $A\cap B$ is also an event that implies events $A$ and $B$ occur simultaneously. $P(A\cap B)$ is thus naturally called the probability of the event that events $A$ and $B$ occur simultaneously. We will extend $P(A\cap B)$ as the probability of co-occurrence of $A$ and $B$ in Section \ref{section2} to accommodate the complex problems of co-occurrence emerging in modern data science and proceed to the fundamental measure-theoretic treatment.

We deal with lots of $\sigma$-fields in modern applications, not just the one $\sigma$-field which is the concern of measure theory. Consider a random object  $X(\omega)$ on a given probability space $(\Omega,\mathscr{F},P)$. There exists an objective measurable space $(\Omega_1,\mathscr{F}_1)$ such that $X:\Omega \rightarrow \Omega_1$ is $\mathscr{F}\setminus\mathscr{F}_1$-measurable \cite{Robert:2000}. As an illustration, take $X$ as the identity mapping $I$  on $\Omega$, then $I$ is the unique mapping from $\Omega$ to $\Omega$ such that $I(\omega)=\omega,\ \forall \omega\in\Omega$. Further let $I$ be an identity random object on $\Omega$ equipped with $\mathscr{F}$, then there exists an objective measurable space $(\Omega,\mathscr{G})$, where $\mathscr{G}$ is a sub $\sigma$-field of $\mathscr{F}$, such that the identity mapping $I$ on $(\Omega,\mathscr{F},P)$ is $\mathscr{F}\setminus\mathscr{G}$-measurable. Whenever a clear distinction is needed, the identity random object from $(\Omega,\mathscr{F})$ to $(\Omega,\mathscr{F})$ is denoted by $I({\mathscr{F}}, {\mathscr{F}})$, just $I_{\mathscr{F}}$ for short or $I_0$ for short when emphasized as a reference, and from $(\Omega,\mathscr{F})$ to $(\Omega,\mathscr{G})$ $(\mathscr{G}\subset\mathscr{F})$ denoted by $I({\mathscr{F}}, {\mathscr{G}})$, or just $I_{\mathscr{G}}$ for short. In other words, the identity random object on $(\Omega,\mathscr{F},P)$ is not unique since it is obvious that $I(\mathscr{F}, {\mathscr{G}})$ is different from $I(\mathscr{F}, \mathscr{F})$. If $X(\omega)$ is an random object from $(\Omega,\mathscr{F})$ to $(\Omega^*,\mathscr{F}^*)$, then for a sub $\sigma$-field $\mathscr{G}$ of $\mathscr{F}^*$, we denote $X_{\mathscr{G}}$ the random object from $(\Omega,\mathscr{F})$ to $(\Omega^*,\mathscr{G})$ such that $X_{\mathscr{G}}(\omega)=X(\omega),\ \forall \omega\in\Omega$. Then $X_{\mathscr{G}}$ is different from $X_\mathscr{F}$.

\emph{Contributions}. Motivated by the important role of sets of indices in formulating the identifying assumptions underlying causal inferences, the languages used in articulating the assumptions, the conditional nature of all causal and counterfactual claims and the methods developed for the assessment of such claims \cite{Pearl:2009}, we introduce in a general setting the intuitive definitions of probability of co-occurrence and conditional probability of co-occurrence, which is in the science-friendly vocabulary and interpretation for explicating structural assumptions in practice. Technically speaking, the best treatment of complex co-occurrence starts with classes of indices for the conditional nature of inference about a fixed probability space $(\Omega, \mathscr{F}, \mathbb{P})$ and thus can fill the nontrivial gap between the measure-theoretic foundation and the structural causal model (\ref{eqn:scm}) proposed by Bongers et al. \cite{Bongers:2021} and hierarchical mixture models (HMMs) proposed by Hofmann and Puzicha \cite{Hofmann:1998} for co-occurrence data. On top of that, many probabilistic ideas are greatly simplified by reformulation as properties of sigma-fields \cite{Pollard:2001}. It has been shown that the best treatment of independence starts with the concept of independent sub $\sigma$-fields of $\mathscr{F}$ for a fixed probability space $(\Omega, \mathscr{F}, \mathbb{P})$.  Then the corresponding results with respect to conditional probability and conditional expectation can be described in a unified semantic and mathematical framework in the presence of the heterogeneity of random objects. Take high-dimensional data as an example, a target random object in the middle plate of Figure~\ref{MAE} can be heterogeneous with a set of indices $I_1$ for visible information in images, a set of indices $I_2$ for audible information in speech, a set of indices $I_3$ for textual information in words from a complex system; a target random object in the right plate of Figure~\ref{MAE} can be heterogeneous with a set of indices $I_1$ for observed variables and a set of indices $I_2$ for counterfactual variables for augmented probability measure in the Neyman-Rubin potential outcome framework for causal inference; a target random object in the left plate of Figure~\ref{MAE} can be heterogeneous with a set of indices $I_1$ for endogenous variables and a set of indices $I_2$ for exogenous variables for augmented probability measure in the structural theory of causation. Furthermore, as an extension of conditional probability, a class of natural integrals called E-integrals is introduced as being the preferred expectation operator approach to develop the theory of probability for a wider variety of important modern applications in optimization problems, quantum mechanics, information theory and statistical mechanics \cite{Whittle:1992, Whittle:2000, Pollard:2001}. The properties of $E$-integral are investigated. With the rigorous measure-theoretic approach, certain measure problems can be cleanly resolved regarding the complex problem of many heterogenous co-occurrences emerging in modern data science in general and data spaces, loss functions, and statistical risk arising in machine learning in particular. For its own sake, the paper provides new insights into the inherent challenge of data sparseness underlying the modeling complex co-occurrence data problems in higher order co-occurrences and opens exciting new research directions. As an extended conditionality principle, the paper provides a rigorous foundation by presenting the measure-theoretic results for the active areas of research on the modeling complex co-occurrence data with SCMs and HMMs.

\emph{Outline}. The paper is structured as follows.  We introduce the definition and properties of probability of co-occurrence and conditional probability of co-occurrence as set functions in the most elementary form, without imposing any parameterizations on them, to pursue the fundamental measure-theoretic treatment of the complex co-occurrence problems in Section 2. In Section 3, we examine the transformation of conditional probability of co-occurrence defined in Definition \ref{definition2.12}. In Section 4, we provide the definition of the probability density of co-occurrence with respect to its product probability measure. In section 5, we further devote to investigate the integrals with respect to the measure for probability and conditional probability of co-occurrence.  Then $E$-integral is developed for the occurrence of many events to deal with the complexity of the co-occurrence problems in Section 6. Our goal is to provide a digestible narrative and postpone for this reason all proofs and most of the technical material to an appendix. The Supplementary Material \cite{Wang:2021} provides the complete proofs of all the theoretical results in the main text. Appendix A contains the lemmas and theorems that are used in several proofs. Appendix B contains all the other proofs of all the theoretical results in the main text.

\section{Probability and Conditional Probability of Co-occurrence}\label{section2}

\subsection{Notation}
Let $\mathbb{N}$ be the set of natural numbers, $\mathbb{R}$ the set of real numbers, $\mathbb{R}^+$ the set of nonnegative real numbers, and $\overline{\mathbb{R}}=[-\infty,+\infty]$. Let $\Lambda$ be a nonempty index set and $I_1,I_2\subset\Lambda$, then $I_1+I_2:=I_1\cup I_2$ provided that $I_1\cap I_2=\emptyset$.

Let $(\Omega,\mathscr{F},P)$ be a given probability space,  $(\Omega_i,\mathscr{F}_i), i\in\Lambda$ be measurable spaces and $X_i:(\Omega,\mathscr{F}) \rightarrow(\Omega_i,\mathscr{F}_i),i\in \Lambda$ be random objects. For any $I \subset \Lambda$, denote the product space $\Omega_I:=\prod_{i\in I}\Omega_i$, where $\omega_I:=(\omega_i: \omega_i\in\Omega_i, i\in I)\in\Omega_I$ in general and $\omega_I:=(\omega_i,i\in I)$ provided that $I$ is a countable index set in particular, and $\mathscr{F}_I:=\prod_{i\in I}\mathscr{F}_i$. Then the measurable product space $(\Omega_I,\mathscr{F}_I):=(\prod_{i\in I}\Omega_i,\prod_{i\in I}\mathscr{F}_i)$. Denote $X_{I}(\omega):=(X_i(\omega):i\in I)$ as a random object from $(\Omega,\mathscr{F})$ to $(\Omega_{I},\mathscr{F}_{I})$.

Let $(\Omega_i,\mathscr{F}_i)$, $i\in I$, be measurable spaces. If $I=\{1, 2\}$,  let $X:(\Omega,\mathscr{F})\rightarrow(\Omega_1,\mathscr{F}_1)$ and $Y:(\Omega,\mathscr{F})\rightarrow(\Omega_2,\mathscr{F}_2)$ be random objects, i.e., $X(\omega)$ and $Y(\omega)$ are $\mathscr{F}\setminus \mathscr{F}_1$-measurable and $\mathscr{F}\setminus\mathscr{F}_2$-measurable, respectively. The probability measures induced  by $X$ on $(\Omega_1,\mathscr{F}_1)$ and $Y$ on $(\Omega_2,\mathscr{F}_2)$ are denoted as $P[X]$ and $P[Y]$, respectively. The functional relationships among $X_I$  hinge on the graph of $\Lambda$ that encodes a semantic network. Thus in our presentation involving a set of $I$'s, the notation $P[\bullet]$ is a flexible alternative to $P_{\bullet}$ for symbolic manipulation whenever $X_{I}$ involves complex structure of $I$ for $P_{X_I}$ and will be used whenever operation is needed on the high-dimensional random object $\boldsymbol{X}$ with complex heterogenous structure. In other words, a pair of square brackets serves to describe and manipulate the complexity of the index set $I$ inherent in a probability measure while a pair of round brackets is kept for events in a $\sigma$-field. In addition, "almost everywhere with respect to the measure $\mu$" is abbreviated to "a.e.\{$\mu$\}".

\subsection{Definitions and Theorems}

\begin{defn}\label{definition2.1} Let $(\Omega,\mathscr{F},P)$ be a given probability space and $I\subset \mathbb{N}$. Let $ A_i\in \mathscr{F}, i\in I$, then $P(\bigcap_{i\in I}A_i)$ is called the probability of co-occurrence of events $A_i, i\in I$, and denoted by $P[A_i,i\in I]$, i.e., $$P[A_i,i\in I]=P(\bigcap\limits_{i\in I}A_i).$$
Given $ A_i\in \mathscr{F}, i\in I_1$ and $ B_j\in \mathscr{F}, j\in I_2$, where $I_1, I_2\subset \mathbb{N}$, then the probability of co-occurrence of all events $A_i,i\in I_1$ and $B_j,j\in I_2$ is denoted by $P[A_i,i\in I_1; B_j,j\in I_2]$, i.e., $$P[A_i,i\in I_1; B_j,j\in I_2]=P((\bigcap_{i\in I_1}A_i)\bigcap(\bigcap_{j\in I_2}B_j)).$$
Given $A_i\in \mathscr{F},i=1,2,3,\cdots,n$, then $$P[A_1,A_2,\cdots,A_n]=P[A_i,i=1,2,3,\cdots,n]=P(\bigcap\limits_{i=1}^{n}A_i).$$
\end{defn}

\begin{comm}\label{comment2.1.1}Without any ambiguity we also take
$$[A_i,i\in I]:=\bigcap\limits_{i\in I}A_i,$$
$$[A_i,i\in I_1; B_j,j\in I_2]:=(\bigcap_{i\in I_1}A_i)\bigcap(\bigcap_{j\in I_2}B_j),$$
$$[A_1,A_2,\cdots,A_n]:=\bigcap\limits_{i=1}^{n}A_i.$$
\end{comm}

\begin{defn}\label{definition2.2} Let $(\Omega,\mathscr{F},P)$ be a given probability space. Let $ A_i\in \mathscr{F}, i\in I_1$, $ B_j\in \mathscr{F}, j\in I_2$, where $I_1, I_2\subset \mathbb{N}$. Given $P[B_j,j\in I_2]>0$, we call $$P[A_i,i\in I_1|B_j,j\in I_2]:=\frac{P[A_i,i\in I_1;B_j,j\in I_2]}{P[B_j,j\in I_2]}$$ the conditional probability of co-occurrence of events $A_i,i\in I_1$ given co-occurrence of events $B_j,j\in I_2$. When $P[B_j,j\in I_2]=0$, $P[A_i,i\in I_1|B_j,j\in I_2]$ is naturally defined as zero.
\end{defn}

\begin{comm}\label{comment2.2.1} (i) If both $I_1$ and $I_2$ are a singleton, then $P[A_i,i\in I_1|B_j,j\in I_2]=P[A_i|B_j]$, which is the well-known conditional probability of $A_i$ given $B_j$ (see \cite{Kailai:2000}).\\

(ii) If $I_1$ is a singleton, then $P[A_i,i\in I_1|B_j,j\in I_2]=P[A_i|B_j,j\in I_2]$, which is the conditional probability of $A_i$ given co-occurrence of events $B_j,j\in I_2$.\\

(iii) If $I_2$ is a singleton, then $P[A_i,i\in I_1|B_j,j\in I_2]=P[A_i,i\in I_1|B_j]$, which is the conditional probability of co-occurrence of events $A_i,i\in I_1$ given $B_j$.
\end{comm}

\begin{defn}\label{definition2.3} Let $ A_i\in \mathscr{F}_1, i\in I_1$, $ B_j\in \mathscr{F}_2, j\in I_2$, where $I_1, I_2\subset \mathbb{N}$. Then\\

1. $P[X\in A_i,i\in I_1;Y\in B_j,j\in I_2]:=P[X^{-1}(A_i),i\in I_1;Y^{-1}(B_j),j\in I_2]$ is called the probability of co-occurrence of events $X\in A_i,i\in I_1$ and $Y\in B_j,j\in I_2$.\\

2. Given $P[Y\in B_j,j\in I_2]>0$,
$$P[X\in A_i,i\in I_1|Y\in B_j,j\in I_2]:=\frac{P[X\in A_i,i\in I_1;Y\in B_j,j\in I_2]}{P[Y\in B_j,j\in I_2]}$$
is called the conditional probability of co-occurrence of events $X\in A_i, i\in I_1$ given co-occurrence of events $Y\in B_j,j\in I_2$. When $P[Y\in B_j,j\in I_2]=0$, $P[X\in A_i,i\in I_1|Y\in B_j,j\in I_2]$ is naturally defined as zero.
\end{defn}

\begin{comm}\label{comment2.3.1} Let $\mathscr{G}_1$ and $\mathscr{G}_2$ be sub $\sigma$-fields of $\mathscr{F}$. Then\\

(i) For identical random objects $I_{\mathscr{G}_1}$ and $I_{\mathscr{G}_2}$, it is evident that
$$P[I_{\mathscr{G}_1}\in A_i,i\in I_1;I_{\mathscr{G}_2}\in B_j,j\in I_2]=P[A_i,i\in I_1;B_j,j\in I_2],$$
and
$$P[I_{\mathscr{G}_1}\in A_i,i\in I_1|I_{\mathscr{G}_2}\in B_j,j\in I_2]=P[A_i,i\in I_1|B_j,j\in I_2],$$
where $ A_i\in \mathscr{G}_1, i\in I_1$, $ B_j\in \mathscr{G}_2, j\in I_2$ and $I_1, I_2\subset \mathbb{N}$.\\

(ii) If $X$ and $Y$ are independent, then for $ A_i\in \mathscr{F}_1, i\in I_1$, $ B_j\in \mathscr{F}_2, j\in I_2$, where $I_1, I_2\subset \mathbb{N}$, we have $$P[X\in A_i,i\in I_1;Y\in B_j,j\in I_2]=P[X\in A_i,i\in I_1]\cdot P[Y\in B_j,j\in I_2],$$ and $$P[X\in A_i,i\in I_1|Y\in B_j,j\in I_2]=P[X\in A_i,i\in I_1]$$
provided that $P[Y\in B_j,j\in I_2]>0$.
\end{comm}

\begin{defn}\label{definition2.4}Let $(\Omega_i,\mathscr{F}_i), i\in I\, (I\subset \mathbb{N})$ be measurable spaces and $X_i:(\Omega,\mathscr{F}) \rightarrow(\Omega_i,\mathscr{F}_i),i\in I$ random objects. Given $\{I_1, I_2\}$ is a partition of $I$:  $I_1,I_2\subset I$, $I_1\cap I_2=\emptyset$ and $I=I_1+I_2$. Then for $A_{ij}\in \mathscr{F}_{i},j\in J_{i}\subset \mathbb{N},i\in I$, we take
$$P[X_{i}\in A_{ij},j\in J_{i},i\in I_1;X_{i}\in A_{ij},j\in J_{i},i\in I_2]:=P( (\bigcap\limits_{j\in J_{i} \atop i\in I_1}X_{i}\in A_{ij})\bigcap (\bigcap\limits_{j\in J_{i} \atop i\in I_2}X_{i}\in A_{ij})),$$
which is called the probability of co-occurrence of events $X_{i}\in A_{ij}, j\in J_{i}, i\in I_1$ and $X_{i}\in A_{ij}, j\in J_{i}, i\in I_2$.
And take
\begin{eqnarray*}
&&P[X_{i}\in A_{ij},j\in J_{i},i\in I_1|X_{i}\in A_{ij},j\in J_{i},i\in I_2]\\
&:=&\frac {P[X_{i}\in A_{ij},j\in J_{i},i\in I_1;X_{i}\in A_{ij},j\in J_{i},i\in I_2]}{P[X_{i}\in A_{ij},j\in J_{i},i\in I_2]},\\
\end{eqnarray*}
which is called the conditional probability of co-occurrence of events $X_{i}\in A_{ij},j\in J_{i},i\in I_1$ given co-occurrence of events $X_{i}\in A_{ij},j\in J_{i},i\in I_2$ provided that $P[X_{i}\in A_{ij},j\in J_{i},i\in I_2]>0$ . When $P[X_{i}\in A_{ij},j\in J_{i},i\in I_2]=0$, $$P[X_{i}\in A_{ij},j\in J_{i},i\in I_1|X_{i}\in A_{ij},j\in J_{i},i\in I_2]$$ is naturally defined as zero.
\end{defn}

\begin{comm}\label{comment2.4.1}In Definition \ref{definition2.4}, let $X_{ij}=X_{i},j\in J_{i},i\in I$, then
\begin{eqnarray*}
&&P[X_{i}\in A_{ij},j\in J_{i},i\in I_1;X_{i}\in A_{ij},j\in J_{i},i\in I_2]\\
&=&P[X_{ij}\in A_{ij},j\in J_{i},i\in I_1;X_{ij}\in A_{ij},j\in J_{i},i\in I_2],\\
&& P[X_{i}\in A_{ij},j\in J_{i},i\in I_1|X_{i}\in A_{ij},j\in J_{i},i\in I_2] \\
&=& P[X_{ij}\in A_{ij},j\in J_{i},i\in I_1|X_{ij}\in A_{ij},j\in J_{i},i\in I_2].
\end{eqnarray*}

\end{comm}

\begin{defn}\label{definition2.5} Let $(\Omega_1,\mathscr{F}_1)$ and $(\Omega_2,\mathscr{F}_2)$ be measurable spaces. Let $X:(\Omega,\mathscr{F})\rightarrow(\Omega_1,\mathscr{F}_1)$ and $Y:(\Omega,\mathscr{F})\rightarrow(\Omega_2,\mathscr{F}_2)$ be random objects.
Given $B\in \mathscr{F}_2$. Define $\nu(A)=P[X\in A,Y\in B]$ for each $A\in \mathscr{F}_1$, then $\nu$ is a finite measure on $(\Omega_1,\mathscr{F}_1)$ and is denoted by $P[X,Y\in B]$, i.e., $$P[X,Y\in B](A)=P[X\in A,Y\in B],\, \forall A\in \mathscr{F}_1. $$
$P[X,Y\in B]$ is called the probability of co-occurrence of $X$ and $Y\in B$.
Let
$$ \mu=\left\{\begin{array}{cc}\displaystyle\frac{P[X,Y\in B]}{P[Y\in B]},&\ \ {\rm if}\ \ P[Y\in B]>0,\\ \displaystyle 0,& \ \ {\rm if}\ \ P[Y\in B]=0,\end{array}\right.$$
then $\mu$ is a probability measure on $(\Omega_1,\mathscr{F}_1)$ provided that $P[Y\in B]>0$ and is denoted by $P[X|Y\in B]$. $P[X|Y\in B]$ is called the conditional probability of $X$ given $Y\in B$. $P[X|Y\in B]$ is defined as zero measure provided that $P[Y\in B]=0$.
\end{defn}

\begin{comm}\label{comment2.5.1} (i) For identity random object $I_{\mathscr{G}}$, it is evident that
$$P[X,I_{\mathscr{G}}\in B]=P[X,B],$$
and
$$P[X|I_{\mathscr{G}}\in B]=P[X|B],$$
where $\mathscr{G}$ is a sub $\sigma$-field of $\mathscr{F}$, and $B\in \mathscr{G}$.

(ii) For identity random objects $I_{\mathscr{G}_1}$ and $I_{\mathscr{G}_2}$ , we have
$$P[I_{\mathscr{G}_1},I_{\mathscr{G}_2}\in B]=P[I_{\mathscr{G}_1},B]=P[I_0,B],$$
and
$$P[I_{\mathscr{G}_1}|I_{\mathscr{G}_2}\in B]=P[I_{\mathscr{G}_1}|B]=P[I_0|B],$$
where $\mathscr{G}_1$ and $\mathscr{G}_2$ are sub $\sigma$-fields of $\mathscr{F}$, and $B\in \mathscr{G}_2$.

(iii) If $X$ and $Y$ are independent, then we have
$$P[X|Y\in B]=P[X],$$
where $\forall B\in \mathscr{F}_2$ provided that $P[Y\in B]>0$.

(iv) Generally, $P[X,Y\in B]$ is not a probability measure on $(\Omega_1,\mathscr{F}_1)$.

\end{comm}

\begin{defn}\label{definition2.6}Let $(\mathbb{E},\mathscr{E})$ be a measurable space and $Z:(\Omega,\mathscr{F})\rightarrow(\mathbb{E},\mathscr{E})$ a random object. Let $(\Omega_i,\mathscr{F}_i), i\in  I\subset \mathbb{N}$ be measurable spaces and $X_i:(\Omega,\mathscr{F})\rightarrow(\Omega_i,\mathscr{F}_i),i\in I$ random objects. Given $I_1,I_2\subset I$, $I_1\cap I_2=\emptyset$ and $I=I_1+ I_2$. Then for $A_{ij}\in \mathscr{F}_{i},j\in J_{i}\subset \mathbb{N},i\in I$, We define
$$\nu(A)=P[Z\in A,X_{i}\in A_{ij},j\in J_{i},i\in I],\ \  \forall A\in \mathscr{E},$$
then $\nu$ is a finite measure on $(\mathbb{E},\mathscr{E})$ and denoted by $P[Z; X_{i}\in A_{ij}, j\in J_{i}, i\in I]$, i.e.,
$$P[Z;X_{i}\in A_{ij}, j\in J_{i}, i\in I](A)=P[Z\in A,X_{i}\in A_{ij},j\in J_{i},i\in I],\ \  \forall A\in \mathscr{E}.$$
$P[Z;X_{i}\in A_{ij}, j\in J_{i}, i\in I]$ is called the probability of co-occurrence of $Z$ and events $X_{i}\in A_{ij}, j\in J_{i}, i\in I$.
And we take
\begin{eqnarray*}
&&P[Z;X_{i}\in A_{ij},j\in J_{i},i\in I_1|X_{i}\in A_{ij},j\in J_{i},i\in I_2] \\
&:=&\frac {P[Z;X_{i}\in A_{ij},j\in J_{i},i\in I_1;X_{i}\in A_{ij},j\in J_{i},i\in I_2]}{P[X_{i}\in A_{ij},j\in J_{i},i\in I_2]}\\
&=&\frac {P[Z;X_{i}\in A_{ij},j\in J_{i},i\in I]}{P[X_{i}\in A_{ij},j\in J_{i},i\in I_2]},
\end{eqnarray*}
which is called the conditional probability of co-occurrence of $Z$ and events $X_{i}\in A_{ij},j\in J_{i},i\in I_1$ given co-occurrence of events $X_{i}\in A_{ij},j\in J_{i},i\in I_2$ provided that $P[X_{i}\in A_{ij},j\in J_{i},i\in I_2]>0$. If $P[X_{i}\in A_{ij},j\in J_{i},i\in I_2]=0$, then $P[Z;X_{i}\in A_{ij},j\in J_{i},i\in I_1|X_{i}\in A_{ij},j\in J_{i},i\in I_2]$ is defined as zero.
\end{defn}

\begin{comm}\label{comment2.6.1} (i) Let $A_i=\bigcap_{j\in J_i}A_{ij}, \,i\in I$, then by the definition above, we have
\begin{eqnarray*}
P[Z;X_{i}\in A_{ij}, j\in J_{i}, i\in I]&=&P[Z;X_{i}\in A_i, i\in I],\\
P[Z;X_{i}\in A_{ij},j\in J_{i},i\in I_1|X_{i}\in A_{ij},j\in J_{i},i\in I_2]&=&P[Z;X_{i}\in A_{i}, i\in I_1|X_{i}\in A_{i},i\in I_2].
\end{eqnarray*}

(ii) Let $I_j\subset I,j=1,2,3,4$, which are pairwise disjoint. Let $A_{i}\in \mathscr{F}_{i},i\in I_j,j=1,2,3,4$, then by the definition above, we also have
$$\begin{array}{l}\displaystyle P[Z;X_{i}\in \Omega_{i}, i\in I_1;X_{i}\in A_{i},i\in I_2]=P[Z;X_{i}\in A_{i},i\in I_2],\\
\displaystyle P[Z;X_{i}\in \Omega_{i}, i\in I_1|X_{i}\in A_{i},i\in I_2]=P[Z|X_{i}\in A_{i},i\in I_2],\\
\displaystyle P[Z|X_{i}\in \Omega_{i}, i\in I_1;X_{i}\in A_{i},i\in I_2]=P[Z|X_{i}\in A_{i},i\in I_2],\\
\displaystyle P[Z;X_{i}\in A_{i}, i\in I_1;X_{i}\in \Omega_{i},i\in I_3|X_{i}\in A_{i},i\in I_2]\\
\displaystyle =P[Z;X_{i}\in A_{i}, i\in I_1|X_{i}\in A_{i},i\in I_2],\\
\displaystyle P[Z;X_{i}\in A_{i}, i\in I_1;X_{i}\in \Omega_{i},i\in I_3|X_{i}\in A_{i},i\in I_2;X_{i}\in \Omega_{i},i\in I_4]\\
\displaystyle =P[Z;X_{i}\in A_{i}, i\in I_1|X_{i}\in A_{i},i\in I_2].
\end{array}$$

(iii) Generally, the probability of co-occurrence $P[Z;X_i\in A_i,i\in I]$ and the conditional probability of co-occurrence $P[Z;X_i\in A_i,i\in I_1|X_i\in A_i,i\in I_2]$ are not necessarily probability measures on $(\mathbb{E},\mathscr{E})$. But the probability measure $P[Z]$ induced by $Z$ on $(\mathbb{E},\mathscr{E})$ is a special case of them since
$$P[Z]=P[Z;X_i\in \Omega_i,i\in I]=P[Z;X_i\in \Omega_i,i\in I_1|X_i\in \Omega_i,i\in I_2].$$
\end{comm}

\begin{thm}\label{theorem2.6.1} Let $(\mathbb{E},\mathscr{E})$ be measurable space and $Z:(\Omega,\mathscr{F}) \rightarrow(\mathbb{E},\mathscr{E})$ a random object. Let $(\Omega_i,\mathscr{F}_i), i\in I\, (I\subset \mathbb{N})$ be measurable spaces and $X_i:(\Omega,\mathscr{F})\rightarrow(\Omega_i,\mathscr{F}_i),i\in I$ random objects. Given $I_1,I_2\subset I$, $I_1\cap I_2=\emptyset$ and $I=I_1+ I_2$. If $A_{i}\in \mathscr{F}_{i},i\in I$, then
$$P[Z,X_I\in \prod_{i\in I}A_i]=P[Z;X_{I_1}\in \prod_{i\in I_1}A_i,X_{I_2}\in\prod_{i\in I_2}A_i]=P[Z;X_i\in A_i,i\in I],$$
and
$$P[Z,X_{I_1}\in \prod_{i\in I_1}A_i|X_{I_2}\in \prod_{i\in I_2}A_i]=P[Z;X_i\in A_i,i\in I_1|X_i\in A_i,i\in I_2].$$
\end{thm}

\begin{comm}\label{comment2.6.2}Let $(\Omega,\mathscr{F},P)$ be a given probability space. Let $(\Omega_i,\mathscr{F}_i), i\in I \subset \mathbb{N}$, be measurable spaces and $X_i:(\Omega,\mathscr{F}) \rightarrow(\Omega_i,\mathscr{F}_i),i\in I$, random objects. Given $I_i\subset I,i=1,2,3$, which are pairwise disjoint. For $A_{i}\in \mathscr{F}_{i},i\in I_1+I_2$, we take
$$P[X_i,i\in I_3;X_i\in A_i,i\in I_1+I_2]:=P[X_{I_3};X_i\in A_i,i\in I_1+I_2],$$
which is called the probability of co-occurrence of $X_i,i\in I_3$ and $X_i\in A_i,i\in I_1+I_2$;
and
$$P[X_i,i\in I_3;X_i\in A_i,i\in I_1|X_i\in A_i,i\in I_2]:=P[X_{I_3};X_i\in A_i,i\in I_1|X_i\in A_i,i\in I_2],$$
which is called the conditional probability of co-occurrence of $X_i,i\in I_3$ and $X_i\in A_i,i\in I_1$ given co-occurrence of $X_i\in A_i,i\in I_2$.
\end{comm}

\begin{defn}\label{definition2.7} Given $ B\in \mathscr{F}_2$. Then if there exists a nonnegative $\mathscr{F}_1$-measurable function $\varphi:\Omega_1\rightarrow \mathbb{R}$ such that
$$ \int_{A}\varphi(\omega_1)P[X](d\omega_1)=P[X\in A,Y\in B],\, \forall A\in \mathscr{F}_1,$$
then the function $\varphi(\omega_1)$ is called the conditional probability of event $Y\in B$ given $X$ (or $X=\omega_1$) and denoted by $P[Y\in B|X](\omega_1)$(or $P[Y\in B|X=\omega_1]$, see \cite{Robert:2000}), i.e.,
$$ \int_{A}P[Y\in B|X](\omega_1)P[X](d\omega_1)=P[X\in A,Y\in B],\, \forall A\in \mathscr{F}_1.$$
\end{defn}

As to the existence and uniqueness of the conditional probability of event $Y\in B$ given $X$, i.e., $P[Y\in B|X](\omega_1)$, we have the following Theorem $\ref{theorem2.7.1}$.

\begin{thm}\label{theorem2.7.1} Let $B\in \mathscr{F}_2$, then  $P[Y\in B|X](\omega_1)$ exists and is unique a.e. $\{P[X]\}$.
\end{thm}

\begin{comm}\label{comment2.7.1} (i) For identical random object $I_{\mathscr{G}}$, where $\mathscr{G}$ is a sub $\sigma$-field of $\mathscr{F}$,  $P[Y\in B|I_{\mathscr{G}}](\omega)$ is also written as $P[Y\in B|\mathscr{G}](\omega)$.

(ii) For any identical random objects $I_{\mathscr{G}_1}$ and $I_{\mathscr{G}_2}$ , where $\mathscr{G}_i,i=1,2$ are sub $\sigma$-fields of $\mathscr{F}$, we have
$$P[I_{\mathscr{G}_2}\in B|I_{\mathscr{G}_1}](\omega)=P[B|\mathscr{G}_1]\, (or P[B|\mathscr{G}_1](\omega)),$$
which is well known as the conditional probability of $B\in \mathscr{G}_2$ given the sub $\sigma$-field $\mathscr{G}_1$ of $\mathscr{F}$ (see \cite{Robert:2000}).

(iii) By Definition \ref{definition2.7} and Theorem \ref{theorem2.7.1}, $\varphi(\omega_1)=P[Y\in B|X](\omega_1)$ means that
$$\varphi(\omega_1)=P[Y\in B|X](\omega_1), \ \ a.e. \{P[X]\},$$
where $\varphi(\omega_1)$ is a nonnegative $\mathscr{F}_1$-measurable real function on $\Omega_1$.
\end{comm}

\begin{thm}\label{theorem2.7.2} Let $(\Omega,\mathscr{F},P)$, $(\Omega_1,\mathscr{F}_1)$, $(\Omega_2,\mathscr{F}_2)$, $X$ and $Y$ be the same as above.

1. Let $B_1\subset B_2,\,\,\forall B_1,B_2\in \mathscr{F}_2$, then
$$P[Y\in B_1|X](\omega_1)\leq P[Y\in B_2|X](\omega_1), \,a.e. \{P[X]\}.$$

2. $P[Y\in \Omega_2|X](\omega_1)=1,\,a.e. \{P[X]\}.$

\end{thm}

Let  $(\Omega_i,\mathscr{F}_i),\,i=1,2,3$ be measurable spaces. Let $X_i:(\Omega,\mathscr{F})\rightarrow(\Omega_i,\mathscr{F}_i),\,i=1,2,3$ be random objects. For fixed $A_2\in\mathscr{F}_2$, $P[X_1,X_2\in A_2]$ is the probability of co-occurrence of $X_1$ and $X_2\in A_2$ (see Definition \ref{definition2.5}), which is a positive finite measure on $(\Omega_1,\mathscr{F}_1)$ when $P[X_2\in A_2]>0$.

\begin{defn}\label{definition2.8} Given $A_2\in \mathscr{F}_2$, $A_3\in\mathscr{F}_3$ and $P[X_1,X_2\in A_2]>0$. Then if there exists a nonnegative $\mathscr{F}_1$-measurable function $\varphi:\Omega_1\rightarrow \mathbb{R}$ such that
$$ \int_{A_1}\varphi(\omega_1)P[X_1,X_2\in A_2](d\omega_1)=P[X_1\in A_1,X_2\in A_2,X_3\in A_3],\, \forall A_1\in \mathscr{F}_1,$$
then the function $\varphi(\omega_1)$ is called the conditional probability of event $X_3\in A_3$ given co-occurrence of $X_1$ and $X_2\in A_2$, which is denoted by $P[X_3\in A_3|X_1,X_2\in A_2](\omega_1)$, i.e.,
$$\begin{array}{l}\displaystyle \int_{A_1}P[X_3\in A_3|X_1,X_2\in A_2](\omega_1)P[X_1,X_2\in A_2](d\omega_1)\\\\
\displaystyle=P[X_1\in A_1,X_2\in A_2,X_3\in A_3],\, \forall A_1 \in \mathscr{F}_1.\end{array}
$$
If $P[X_1,X_2\in A_2]$ is zero measure, then $P[X_3\in A_3|X_1,X_2\in A_2](\omega_1)$ is defined as zero.
\end{defn}

About the existence and uniqueness of the conditional probability of event $X_3\in A_3$ given co-occurrence of $X_1$ and $X_2\in A_2$, i.e., $P[X_3\in A_3|X_1,X_2\in A_2](\omega_1)$, we have the following theorem.

\begin{thm}\label{theorem2.8.1} Let $(\Omega,\mathscr{F},P)$, $(\Omega_i,\mathscr{F}_i),\,i=1,2,3$ and $X_i(\omega),\,i=1,2,3$ be the same as above.
Given $A_2\in \mathscr{F}_2$, $A_3\in\mathscr{F}_3$ and $P[X_1,X_2\in A_2]>0$. Then  $P[X_3\in A_3|X_1,X_2\in A_2](\omega_1)$ exists and is unique  a.e. $\{P[X_1,X_2\in A_2]\}$.
\end{thm}

\begin{comm}\label{comment2.8.1}(i) $P[X_3\in A_3|X_1,X_2\in\Omega_2](\omega_1)=P[X_3\in A_3|X_1](\omega_1)$.

(ii) $P[X_3\in \Omega_3|X_1,X_2\in A_2](\omega_1)=1,\ a.e.\{P[X_1,X_2\in A_2]\}$ provided that $P[X_1,X_2\in A_2]>0$.

(iii) For $A_{31},A_{32}\in\mathscr{F}_3$ and $A_{31}\subset A_{32}$,
$$P[X_3\in A_{31}|X_1,X_2\in A_2](\omega_1)\leq P[X_3\in A_{32}|X_1,X_2\in A_2](\omega_1),\ a.e.\{P[X_1,X_2\in A_2]\}$$
provided that $P[X_1,X_2\in A_2]>0$.

(iv) By Definition \ref{definition2.8} and Theorem \ref{theorem2.8.1}, $\varphi(\omega_1)=P[X_3\in A_3|X_1,X_2\in A_2](\omega_1)$ means that
$$\varphi(\omega_1)=P[X_3\in A_3|X_1,X_2\in A_2](\omega_1), \ \ a.e.\{P[X_1,X_2\in A_2]\},$$
where $\varphi(\omega_1)$ is a nonnegative $\mathscr{F}_1$-measurable real function on $\Omega_1$ and $P[X_2\in A_2]>0$.

(v) Without any ambiguity, $P[X_3\in A_3|X_1,X_2\in A_2](\omega_1)$ is abbreviated to $P[X_3\in A_3|X_1,X_2\in A_2]$.
\end{comm}

By Theorem \ref{theorem2.8.1} we have the following corollary.

\begin{col}\label{corollary2.8.1}Let $(\Omega,\mathscr{F},P)$ be a given probability space. Let $(\Omega_i,\mathscr{F}_i), i\in \Lambda$ be measurable spaces and $X_i:(\Omega,\mathscr{F}) \rightarrow(\Omega_i,\mathscr{F}_i),i\in \Lambda$ random objects, where $\Lambda$ is a nonempty index set. Given $I_j\subset \Lambda,j=1,2,3$, which are pairwise disjoint. For fixed $A_{I_j}\in \mathscr{F}_{I_j}, j=2,3$ with $P[X_{I_2}\in A_{I_2}]>0$, then $P[X_{I_3}\in A_{I_3}|X_{I_1},X_{I_2}\in A_{I_2}](\omega_{I_1})$ exists and is unique a.e.$\{P[X_{I_1},X_{I_2}\in A_{I_2}]\}$.
\end{col}

\begin{defn}\label{definition2.9}Let $(\Omega,\mathscr{F},P)$ be a given probability space. Let $(\Omega_i,\mathscr{F}_i), i\in I\, (I\subset \mathbb{N})$ be measurable spaces and $X_i:(\Omega,\mathscr{F})\rightarrow(\Omega_i,\mathscr{F}_i),i\in I$ random objects. Given $I_j\subset I,j=1,2,3$, which are pairwise disjoint. Given fixed $A_i\in \mathscr{F}_i, i\in I_2+I_3$ with $P[X_i\in A_i,i\in I_2]>0$. If there exists a nonnegative $\mathscr{F}_{I_1}$-measurable function $\varphi:\Omega_{I_1}\rightarrow \mathbb{R}$ such that
$$ \int_{A_{I_1}}\varphi(\omega_{I_1})P[X_{I_1};X_i\in A_i,i\in I_2](d\omega_{I_1})=P[X_{I_1}\in A_{I_1};X_i\in A_i,i\in I_2+I_3],\, \forall A_{I_1}\in \mathscr{F}_{I_1},$$
then the function $\varphi(\omega_{I_1})$ is called the conditional probability of co-occurrence of events $X_i\in A_i,i\in I_3$ given co-occurrence of $X_{I_1}$ and $X_i\in A_i,i\in I_2$, which is denoted by $P[X_i\in A_i,i\in I_3|X_{I_1};X_i\in A_i,i\in I_2](\omega_{I_1})$, i.e.,
$$\begin{array}{l}\displaystyle \int_{A_{I_1}}P[X_i\in A_i,i\in I_3|X_{I_1};X_i\in A_i,i\in I_2](\omega_{I_1})P[X_{I_1};X_i\in A_i,i\in I_2](d\omega_{I_1})\\\\
\displaystyle=P[X_{I_1}\in A_{I_1};X_i\in A_i,i\in I_2+I_3],\, \forall A_{I_1}\in \mathscr{F}_{I_1}.\end{array}
$$
If $P[X_i\in A_i,i\in I_2]=0$, then $P[X_i\in A_i,i\in I_3|X_{I_1};X_i\in A_i,i\in I_2](\omega_{I_1})$ is defined as zero.
\end{defn}

\begin{thm}\label{theorem2.9.1} Let $(\Omega_i,\mathscr{F}_i), i\in I\subset \mathbb{N}$ be measurable spaces and $X_i:(\Omega,\mathscr{F}) \rightarrow(\Omega_i,\mathscr{F}_i),i\in I$ random objects. Given $I_j\subset I,j=1,2,3$, which are pairwise disjoint. For fixed $A_i\in \mathscr{F}_i, i\in I_2+I_3$, there exists $P[X_i\in A_i,i\in I_3|X_{I_1};X_i\in A_i,i\in I_2](\omega_{I_1})$ and it is unique a.e. $P[X_{I_1};X_i\in A_i,i\in I_2]$ provided that $P[X_i\in A_i,i\in I_2]>0$. And we have
\begin{equation}\label{eqn2.9.1.1}\begin{array}{l}\displaystyle P[X_i\in A_i,i\in I_3|X_{I_1};X_i\in A_i,i\in I_2](\omega_{I_1})\\
\displaystyle =P[X_{I_3}\in \prod_{i\in I_3}A_i|X_{I_1},X_{I_2}\in \prod_{i\in I_2}A_i](\omega_{I_1}).\end{array}\end{equation}
\end{thm}

\begin{comm}\label{comment2.9.1} For fixed $A_i\in \mathscr{F}_i,i\in I_2+I_3$, we take
$$\begin{array}{l}P[X_i\in A_i,i\in I_3|X_i,i\in I_1;X_i\in A_i,i\in I_2](\omega_i,i\in I_1)\\
:=P[X_i\in A_i,i\in I_3|X_{I_1};X_i\in A_i,i\in I_2](\omega_{I_1}),\end{array}$$
which is called the conditional probability of co-occurrence of $X_i\in A_i,i\in I_3$ given co-occurrence of $X_i,i\in I_1$ and $X_i\in A_i,i\in I_2$.
\end{comm}

\begin{comm}\label{comment2.9.2} Given $I_j\subset I,j=1,2,3,4$, which are pairwise disjoint. For fixed $A_i\in \mathscr{F}_i,i\in I_2+I_3$, we have

(i) $$\begin{array}{l}P[X_i\in A_i,i\in I_3;X_i\in \Omega_i,i\in I_4|X_i,i\in I_1;X_i\in A_i,i\in I_2](\omega_{I_1})\\
=P[X_i\in A_i,i\in I_3|X_i,i\in I_1;X_i\in A_i,i\in I_2](\omega_{I_1}).\end{array}$$

(ii)$$\begin{array}{l}P[X_i\in A_i,i\in I_3|X_i,i\in I_1;X_i\in A_i,i\in I_2;X_i\in \Omega_i,i\in I_4](\omega_{I_1})\\
=P[X_i\in A_i,i\in I_3|X_i,i\in I_1;X_i\in A_i,i\in I_2](\omega_{I_1}).\end{array}$$
\end{comm}

\begin{thm}\label{theorem2.9.2} Let $(\Omega,\mathscr{F},P)$ be a given probability space and $(\Omega_i,\mathscr{F}_i),\,i=1,2,3,4$ be measurable spaces. Let $X_i:(\Omega,\mathscr{F}) \rightarrow(\Omega_i,\mathscr{F}_i),\,i=1,2,3,4$ be random objects. Given $A_i\in \mathscr{F}_i,\,i=2,4$ and $A_{31},A_{32}\in\mathscr{F}_3$. If $A_{31}\subset A_{32}$, then
$$P[X_3\in A_{31},X_4\in A_4|X_1,X_2\in A_2](\omega_1)\leq P[X_3\in A_{32},X_4\in A_4|X_1,X_2\in A_2](\omega_1),$$
$a.e.\{P[X_1,X_2\in A_2]\}$ provided that $P[X_1,X_2\in A_2]>0$.
\end{thm}

The proof of Theorem \ref{theorem2.9.2} is similar to Theorem \ref{theorem2.7.2}.

\begin{defn}\label{definition2.10} Let $(\Omega,\mathscr{F},P)$ be a given probability space, $(\Omega_1,\mathscr{F}_1)$ and  $(\Omega_2,\mathscr{F}_2)$ measurable spaces. Let $X:(\Omega,\mathscr{F}) \rightarrow(\Omega_1,\mathscr{F}_1)$ and $Y:(\Omega,\mathscr{F}) \rightarrow(\Omega_2,\mathscr{F}_2)$ be random objects. Let $K(\omega_1,B):\Omega_1\times\mathscr{F}_2\rightarrow[0,+\infty)$ be a finite kernel from $(\Omega_1,\mathscr{F}_1)$ to $(\Omega_2,\mathscr{F}_2)$, i.e., $K(\omega_1,B)$ satisfies the following conditions (see Definition 4.3.1 in \cite{yan:2004}):

(i) For $\forall B\in \mathscr{F}_2$, $K(\cdot,B)$ is a $\mathscr{F}_1$-measurable function on $(\Omega_1,\mathscr{F}_1)$.

(ii) For $\forall \omega_1\in \Omega_1$, $K(\omega_1,\cdot)$ is a finite measure on $(\Omega_2,\mathscr{F}_2)$.

If for $\forall A\in \mathscr{F}_1$ and $\forall B\in \mathscr{F}_2$, it holds that
$$\int_AK(\omega_1,B)P[X](d\omega_1)=P[X\in A,Y\in B],$$
then $K(\omega_1,B)$ is defined as the conditional probability of $Y$ given $X$(or $X=\omega_1$) and denoted by $P[Y|X](\omega_1,B)$\,(or $P[Y|X=\omega_1]$), i.e.,
$$\int_AP[Y|X](\omega_1,B)P[X](d\omega_1)=P[X\in A,Y\in B], \, \forall A\in \mathscr{F}_1,\,\forall B\in \mathscr{F}_2.$$
\end{defn}

\begin{comm}\label{comment2.10.1} (i) For fixed $B\in \mathscr{F}_2$, by Definition \ref{definition2.7} we have $P[Y|X](\omega_1,B)=P[Y\in B|X](\omega_1)$.

(ii) Similar to the proof of Theorem \ref{theorem2.7.2}, we obtain
$$P[Y|X](\omega_1,\Omega_2)=1,\,a.e. \{P[X]\}.$$

(iii) For an identical random object $I_{\mathscr{G}}$, where $\mathscr{G}$ is a sub $\sigma$-field of $\mathscr{F}$, we have
$$P[Y|I_{\mathscr{G}}](\omega,B)=P[Y|\mathscr{G}](\omega,B), $$
which is well known as the regular conditional probability of $Y$ given $\mathscr{G}$ (see Definition 5.6.3 in \cite{Robert:2000}, or Definition 7.3.3 in \cite{yan:2004}).

(iv) For an identical random object $I_{\mathscr{G}}$, where $\mathscr{G}$ is a sub $\sigma$-field of $\mathscr{F}$, we have
$$P[I_{\mathscr{G}}|X](\omega_1,B)=P^{\omega_1}_{\mathscr{G}}, $$
which has been defined in Definition 10.4.2 in \cite{Bogachev:2007}.

(v) For identical random object $I_{\mathscr{G}_1}$ and $I_{\mathscr{G}_2}$, where $\mathscr{G}_1$ and $\mathscr{G}_2$ are sub $\sigma$-fields of $\mathscr{F}$, we have
$$P[I_{\mathscr{G}_2}|I_{\mathscr{G}_1}](\omega,B)=P^{\mathscr{G}_1}_{\mathscr{G}_2}, $$
which is well known as the regular conditional probability of $\mathscr{G}_2$ given $\mathscr{G}_1$ (see Definition 10.4.2 in \cite{Bogachev:2007}).
Especially for a sub $\sigma$-fields $\mathscr{G}$ of $\mathscr{F}$, $P[I_{\mathscr{F}}|I_{\mathscr{G}}](\omega,B)=P(\omega,B)$ which is the regular conditional probability of $P$ with respect to $\mathscr{G}$ (see Definition 7.3.1 in \cite{yan:2004}).

(vi) Let $\mathds{1}_B(\omega_1)$ be the indicator of $B\in \mathscr{F}_1$. Set $K(\omega_1,B)=\mathds{1}_B(\omega_1), \forall \omega_1\in \Omega_1, \forall B\in \mathscr{F}_1$. Then it is easily shown that $P[X|X](\omega_1,B)=K(\omega_1,B)=\mathds{1}_B(\omega_1)$.

(vii) If $X$ and $Y$ are independent, then $P[Y|X](\omega_1,B)=P[Y](B)$.

(viii) Similar to the proof of Theorem \ref{theorem2.7.2}, for $B_1\subset B_2$, $B_1,B_2\in\mathscr{F}_2$, we obtain
$$P[Y|X](\omega_1,B_1)\leq P[Y|X](\omega_1,B_2),\,a.e.\{P[X]\}.$$
\end{comm}

\begin{thm}\label{theorem2.10.1} Let $(\Omega,\mathscr{F},P)$, $(\Omega_1,\mathscr{F}_1)$, $(\Omega_2,\mathscr{F}_2)$, $X$ and $Y$ be the same as above. If there exists $P[Y|X](\omega_1,B)$, then $P[Y|X](\omega_1,B)$ is unique under a.e. $\{P[X]\}$ and $\forall B\in \mathscr{F}_2$.
\end{thm}

The following Theorem \ref{theorem2.10.2} shows the existence of $P[Y|X](\omega_1,B)$ under certain conditions, and its argument is similar to that of Theorem 7.3.9 in \cite{yan:2004}.

\begin{thm}\label{theorem2.10.2} Let $(\Omega,\mathscr{F},P)$ be a given probability space, $(\Omega_1,\mathscr{F}_1)$ a measurable space, and $(\Omega_2,\mathscr{F}_2)$ a separable measurable space, i.e., that is countably generated. Let $X:(\Omega,\mathscr{F})\rightarrow(\Omega_1,\mathscr{F}_1)$ and $Y:(\Omega,\mathscr{F})\rightarrow(\Omega_2,\mathscr{F}_2)$ be random objects. $P[X]$ and $P[Y]$ be the probability measures induced by $X$ on $(\Omega_1,\mathscr{F}_1)$ and $Y$ on $(\Omega_2,\mathscr{F}_2)$, respectively. If $P[Y]$ is a compact measure(see Definition 7.3.6 in \cite{yan:2004}), then there exists the conditional probability of $Y$ given $X$, i.e., $P[Y|X](\omega_1,B)$.
\end{thm}

If $X=I_{\mathscr{G}}$ in Theorem \ref{theorem2.10.2}, then we have the following Corollary \ref{corollary2.10.1} which is Theorem 7.3.9 in \cite{yan:2004}.

\begin{col}\label{corollary2.10.1} Let $(\Omega,\mathscr{F},P)$ be a given probability space, and $(\Omega_2,\mathscr{F}_2)$ a separable measurable space, i.e., that is countably generated. Let $Y:(\Omega,\mathscr{F})\rightarrow(\Omega_2,\mathscr{F}_2)$ be a random object and $P[Y]$ the probability measure induced by $Y$ on $(\Omega_2,\mathscr{F}_2)$. If $P[Y]$ is a compact measure, then for any sub $\sigma$-field $\mathscr{G}$ of $\mathscr{F}$, there exists the regular conditional probability of $Y$ given $\mathscr{G}$, i.e., $P[Y|\mathscr{G}](\omega,B)$.
\end{col}

\begin{comm}\label{comment2.10.2}There exists such an example (see Example 10.4.19 in \cite{Bogachev:2007}) that shows nonexistence of $P[X_2|X_1](\omega,B)$.
\end{comm}

Let $I\subset \mathbb{N}$, $\mathbb{R}^I$ the product space of $\mathbb{R}_i=\mathbb{R},\ i\in I$ with the product topology. Then it is well known that $\mathbb{R}^I$ is a complete separable metric space and $\prod_{i\in I}\mathscr{B}(\mathbb{R}_i)=\prod_{i\in I}\mathscr{B}(\mathbb{R})=\mathscr{B}(\mathbb{R}^I)$.

\begin{thm}\label{theorem2.10.3} Let $(\Omega,\mathscr{F},P)$ be a given probability space, and $X_i(\omega),\ i\in I\subset \mathbb{N}$ random variables on $(\Omega,\mathscr{F},P)$. Let $I_1,I_2\subset I$, and $I_1\cap I_2=\emptyset$. Then there exists the conditional probability of $X_{I_2}$ given $X_{I_1}$, i.e., $P[X_{I_2}|X_{I_1}](x_{I_1},B_{I_2})$, where $x_{I_1}\in \mathbb{R}^{I_1}$ and $B_{I_2}\in\mathscr{B}(\mathbb{R}^{I_2})$.
\end{thm}

\begin{defn}\label{definition2.11} Let $(\Omega,\mathscr{F},P)$ be a given probability space, $(\Omega_i,\mathscr{F}_i),\,i=1,2,3,4$ measurable spaces. Let $X_i:(\Omega,\mathscr{F})\rightarrow(\Omega_i,\mathscr{F}_i),\ i=1,2,3,4$ be random objects. Let $K(\omega_1,A_3):\Omega_1\times\mathscr{F}_3\rightarrow[0,+\infty)$ be a finite kernel from $(\Omega_1,\mathscr{F}_1)$ to $(\Omega_3,\mathscr{F}_3)$, i.e., $K(\omega_1,A_3)$ satisfies the following conditions:

(i) For $\forall A_3\in \mathscr{F}_3$, $K(\cdot,A_3)$ is a $\mathscr{F}_1$-measurable function on $(\Omega_1,\mathscr{F}_1)$.

(ii) For $\forall \omega_1\in \Omega_1$, $K(\omega_1,\cdot)$ is a finite measure on $(\Omega_3,\mathscr{F}_3)$.

Given $A_2\in\mathscr{F}_2$, $A_4\in\mathscr{F}_4$ and $P[X_1,X_2\in A_2]>0$. If
$$\int_{A_1}K(\omega_1,A_3)P[X_1,X_2\in A_2](d\omega_1)=P[X_1\in A_1,X_2\in A_2,X_3\in A_3,X_4\in A_4]$$
holds for $\forall A_1\in \mathscr{F}_1$ and $\forall A_3\in \mathscr{F}_3$, then $K(\omega_1,A_3)$ is defined as the conditional probability of co-occurrence of $X_3$ and $X_4\in A_4$ given co-occurrence of $X_1$ and $X_2\in A_2$, which is denoted by $P[X_3,X_4\in A_4|X_1,X_2\in A_2](\omega_1,A_3)$, i.e.,
$$\begin{array}{l}\displaystyle\int_{A_1}P[X_3,X_4\in A_4|X_1,X_2\in A_2](\omega_1,A_3)P[X_1,X_2\in A_2](d\omega_1)\\\\
\displaystyle =P[X_1\in A_1,X_2\in A_2,X_3\in A_3,X_4\in A_4], \, \forall A_1\in \mathscr{F}_1,\,\forall A_3\in \mathscr{F}_3.
\end{array}$$
If $P[X_1,X_2\in A_2]=0$, then $P[X_3,X_4\in A_4|X_1,X_2\in A_2](\omega_1,A_3)$ is defined as zero.
\end{defn}

Similar to the proof of Theorem \ref{theorem2.10.1}, we have the following theorem.

\begin{thm}\label{theorem2.11.1} Let $(\Omega,\mathscr{F},P)$, $(\Omega_i,\mathscr{F}_i),\,i=1,2,3,4$ and $X_i(\omega),\,i=1,2,3,4$ be the same as above.
Given $A_i\in \mathscr{F}_i,\,i=2,4$ and $P[X_1,X_2\in A_2]>0$. If $P[X_3,X_4\in A_4|X_1,X_2\in A_2](\omega_1,A_3)$ exists, then  $P[X_3,X_4\in A_4|X_1,X_2\in A_2](\omega_1,A_3)$ is unique under a.e. $P[X_1,X_2\in A_2]$ and $\forall A_3\in\mathscr{F}_3$.
\end{thm}

\begin{comm}\label{comment2.11.1}(i) for fixed $A_3\in\mathscr{F}_3$,
$$P[X_3\in A_3,X_4\in A_4|X_1,X_2\in A_2](\omega_1)=P[X_3,X_4\in A_4|X_1,X_2\in A_2](\omega_1,A_3).$$

(ii) $P[X_3,X_4\in\Omega_4|X_1,X_2\in\Omega_2](\omega_1,A_3)=P[X_3|X_1](\omega_1,A_3)$.

(iii) $P[X_3,X_4\in\Omega_4|X_1,X_2\in A_2](\omega_1,A_3)=P[X_3|X_1,X_2\in A_2](\omega_1,A_3)$.

(iv) $P[X_3,X_4\in A_4|X_1,X_2\in\Omega_2](\omega_1,A_3)=P[X_3,X_4\in A_4|X_1](\omega_1,A_3)$.

(v) $P[X_3,X_4\in A_4|X_1,X_2\in A_2](\omega_1,\Omega_3)=P[X_4\in A_4|X_1,X_2\in A_2](\omega_1)$.

(vi) For $A_{31}\subset A_{32}$, $A_{31},\ A_{32}\in\mathscr{F}_3$,
$$P[X_3,X_4\in A_4|X_1,X_2\in A_2](\omega_1,A_{31})\leq P[X_3,X_4\in A_4|X_1,X_2\in A_2](\omega_1,A_{32}),$$
$a.e. \{P[X_1,X_2\in A_2]\}$ provided that $P[X_1,X_2\in A_2]>0$.

(vii) By definition it is evident that $P[X_3,X_4\in A_4|X_1,X_2\in A_2](\omega_1,A_3)=0$ provided that $P[X_2\in A_2,X_4\in A_4]=0$.

(viii) Without any ambiguity, $P[X_3,X_4\in A_4|X_1,X_2\in A_2](\omega_1,A_3)$ is abbreviated to $P[X_3,X_4\in A_4|X_1,X_2\in A_2]$.

\end{comm}

Theorem \ref{theorem2.11.2} below shows the existence of $P[X_3,X_4\in A_4|X_1,X_2\in A_2](\omega_1,A_3)$. Its proof is similar to Theorem \ref{theorem2.10.2}.

\begin{thm}\label{theorem2.11.2} Let $(\Omega,\mathscr{F},P)$ be a given probability space, $(\Omega_i,\mathscr{F}_i),\ i=1,2,4$ measurable spaces, and $(\Omega_3,\mathscr{F}_3)$ a separable measurable space, i.e., that is countably generated. Let $X_i:(\Omega,\mathscr{F})\rightarrow(\Omega_i,\mathscr{F}_i),\ i=1,2,3,4$ be random objects. If $P[X_3]$ is a compact measure, then for each fixed $A_2\in\mathscr{F}_2$ and each fixed $A_4\in\mathscr{F}_4$ there exists the conditional probability of co-occurrence of $X_3$ and $X_4\in A_4$ given co-occurrence of $X_1$ and $X_2\in A_2$, i.e., $P[X_3,X_4\in A_4|X_1,X_2\in A_2](\omega_1,A_3)$ provided that $P[X_1,X_2\in A_2]>0$.
\end{thm}

\begin{thm}\label{theorem2.11.3} Let $(\Omega,\mathscr{F},P)$ be a given probability space, and $X_i(\omega),\ i\in I\subset \mathbb{N}$ random variables on $(\Omega,\mathscr{F},P)$. Let $I_i\subset I,\ i=1,2,3,4$ be pairwise disjoint, and $X_{I_i},\ i=1,2,3,4$ defined the same as in Theorem \ref{theorem2.10.3}. Then there exists the conditional probability of co-occurrence of $X_{I_3}$ and $X_{I_4}\in A_{I_4}$ given co-occurrence of $X_{I_1}$ and $X_{I_2}\in A_{I_2}$, i.e., $P[X_{I_3},X_{I_4}\in A_{I_4}|X_{I_1},X_{I_2}\in A_{I_2}](x_{I_1},A_{I_3})$,  where $x_{I_1}\in \mathbb{R}^{I_1}$, $A_{I_2}\in\mathscr{B}(\mathbb{R}^{I_2})$, $A_{I_3}\in\mathscr{B}(\mathbb{R}^{I_3})$ and $A_{I_4}\in\mathscr{B}(\mathbb{R}^{I_4})$ provided that $P[X_{I_4}\in A_{I_4},X_{I_2}\in A_{I_2}]>0$.
\end{thm}

\begin{defn}\label{definition2.12} Let $(\Omega,\mathscr{F},P)$ be a given probability space, $(\Omega_{i},\mathscr{F}_{i}),\ i\in \mathbb{N}$ measurable spaces. Let $X_{i}:(\Omega,\mathscr{F})\rightarrow(\Omega_{i},\mathscr{F}_{i}),\ i\in \mathbb{N}$ be random objects. Given $I_i\subset \mathbb{N},\ i=1,2,3,4$ which are pairwise disjoint. For fixed $A_i\in \mathscr{F}_i,\ i\in I_2+I_4$, if $P[X_{I_3},X_{I_4}\in \prod_{i\in I_4}A_i|X_{I_1},X_{I_2}\in \prod_{i\in I_2}A_i](\omega_{I_1},A_{I_3})$ exists, then the conditional probability of co-occurrence of $X_i,\ i\in I_3$ and $X_i\in A_i,\ i\in I_4$ given co-occurrence of $X_i,\ i\in I_1$ and $X_i\in A_i,\ i\in I_2$ is defined as $P[X_{I_3},X_{I_4}\in \prod_{i\in I_4}A_i|X_{I_1},X_{I_2}\in \prod_{i\in I_2}A_i](\omega_{I_1},A_{I_3})$, and denoted by $$P[X_i,i\in I_3;X_i\in A_i,i\in I_4|X_i,i\in I_1;X_i\in A_i,i\in I_2](\omega_i,\ i\in I_1;A_{I_3}),$$ i.e.,
$$\begin{array}{l}\displaystyle P[X_i,i\in I_3;X_i\in A_i,i\in I_4|X_i,i\in I_1;X_i\in A_i,i\in I_2](\omega_i,\ i\in I_1;A_{I_3})\\
\displaystyle =P[X_{I_3},X_{I_4}\in \prod_{i\in I_4}A_i|X_{I_1},X_{I_2}\in \prod_{i\in I_2}A_i](\omega_{I_1},A_{I_3})\\
\displaystyle =P[X_{I_3},\bigcap_{i\in I_4}(X_i\in A_i)|X_{I_1},\bigcap_{i\in I_2}(X_i\in A_i)](\omega_{I_1},A_{I_3}),
\end{array}$$
where $A_{I_3}\in\mathscr{F}_{I_3}$. Furthermore
$$P[X_{I_3},X_{I_4}\in \prod_{i\in I_4}A_i|X_{I_1},X_{I_2}\in \prod_{i\in I_2}A_i](\omega_{I_1},A_{I_3})$$
is written as
$$P[X_i,i\in I_3;X_i\in A_i,i\in I_4|X_i,i\in I_1;X_i\in A_i,i\in I_2](\omega_i,\ i\in I_1;A_i,i\in I_3),$$
where $A_i\in\mathscr{F}_i,\ i\in I_3$, that we shall explain below.
\end{defn}

\begin{comm}\label{comment2.12.1}(i) In Definition \ref{definition2.12}, let
$$\mathscr{C}_3=\{\prod_{i\in s}A_i\times\prod_{i\in I_3\backslash s}\Omega_i:A_i\in\mathscr{F}_i,i\in s,s\in S_{I_3}\},$$
where $S_{I_3}$ stands for all of finite subsets of $I_3$, and $\mathscr{C}'_3=\{\prod_{i\in I_3}A_i:A_i\in\mathscr{F}_i,i\in I_3\}$, then it is well known that $\mathscr{C}_3$ is a half algebra on $\Omega_{I_3}$, $\mathscr{C}_3\subset\mathscr{C}'_3$, and that $\sigma(\mathscr{C}_3)=\sigma(\mathscr{C}'_3)=\mathscr{F}_{I_3}$. Therefore if $\widetilde K(\omega_{I_1},A_{I_3}):\Omega_{I_1}\times\mathscr{C}_3\rightarrow [0,+\infty)$ satisfies the conditions:

(a) For $\forall \omega_{I_1}\in \Omega_{I_1}$, $\widetilde K(\omega_{I_1},\cdot)$ is  a countably additive finite set function on $\mathscr{C}_3$.

(b) For $\forall A_{I_3}\in \mathscr{C}_3$, $\widetilde K(\cdot,A_{I_3})$ is a $\mathscr{F}_{I_1}$-measurable function on $(\Omega_{I_1},\mathscr{F}_{I_1})$ and
$$\begin{array}{l}\displaystyle\int_{A_{I_1}}\widetilde K(\omega_{I_1},A_{I_3})P[X_{I_1},X_{I_2}\in \prod_{i\in I_2}A_i](d\omega_{I_1})\\
\displaystyle=P[X_{I_1}\in A_{I_1},X_{I_2}\in \prod_{i\in I_2}A_i,X_{I_3}\in A_{I_3},X_{I_4}\in \prod_{i\in I_4}A_i],\ \forall A_{I_1}\in\mathscr{F}_{I_1}.\end{array}$$
Then there exists $P[X_{I_3},X_{I_4}\in \prod_{i\in I_4}A_i|X_{I_1},X_{I_2}\in \prod_{i\in I_2}A_i](\omega_{I_1},A_{I_3})$ such that
$$P[X_{I_3},X_{I_4}\in \prod_{i\in I_4}A_i|X_{I_1},X_{I_2}\in \prod_{i\in I_2}A_i](\omega_{I_1},A_{I_3})=\widetilde K(\omega_{I_1},A_{I_3}),\ \forall A_{I_3}\in \mathscr{C}_3.
$$
Conversely, if there exists $P[X_{I_3},X_{I_4}\in \prod_{i\in I_4}A_i|X_{I_1},X_{I_2}\in \prod_{i\in I_2}A_i](\omega_{I_1},A_{I_3})$, then its restriction to $\Omega_{I_1}\times\mathscr{C}_3$ is $\widetilde K(\omega_{I_1},A_{I_3})$ that satisfies the conditions (a) and (b). Therefore $P[X_{I_3},X_{I_4}\in \prod_{i\in I_4}A_i|X_{I_1},X_{I_2}\in \prod_{i\in I_2}A_i](\omega_{I_1},A_{I_3})$ is also denoted by
$$P[X_{I_3},X_{I_4}\in \prod_{i\in I_4}A_i|X_{I_1},X_{I_2}\in \prod_{i\in I_2}A_i](\omega_{I_1},\prod_{i\in I_3}A_i),$$
where $A_i\in\mathscr{F}_i,\ i\in I_3$. So for convenience's sake,
$$P[X_{I_3},X_{I_4}\in \prod_{i\in I_4}A_i|X_{I_1},X_{I_2}\in \prod_{i\in I_2}A_i](\omega_{I_1},A_{I_3})$$
is written as
$$P[X_i,i\in I_3;X_i\in A_i,i\in I_4|X_i,i\in I_1;X_i\in A_i,i\in I_2](\omega_i,\ i\in I_1;A_i,i\in I_3).$$
Hence we have
$$\begin{array}{l}\displaystyle P[X_i,i\in I_3;X_i\in A_i,i\in I_4|X_i,i\in I_1;X_i\in A_i,i\in I_2](\omega_i,\ i\in I_1;A_i,i\in I_3)\\
\displaystyle =P[X_{I_3},X_{I_4}\in \prod_{i\in I_4}A_i|X_{I_1},X_{I_2}\in \prod_{i\in I_2}A_i](\omega_{I_1},\prod_{i\in I_3}A_i),\\
\end{array}$$
where $\forall A_i\in\mathscr{F}_i,\ i\in I_3$, which means $P[X_{I_3},X_{I_4}\in \prod_{i\in I_4}A_i|X_{I_1},X_{I_2}\in \prod_{i\in I_2}A_i](\omega_{I_1},A_{I_3})$.

(ii) If $K(\omega_{I_1},A_{I_3})$ is a finite kernel from $(\Omega_{I_1},\mathscr{F}_{I_1})$ to $(\Omega_{I_3},\mathscr{F}_{I_3})$ such that, for $\forall A_{I_3}\in \mathscr{C}_3$,
$$\begin{array}{l}\displaystyle \int_{A_{I_1}}K(\omega_{I_1},A_{I_3})P[X_{I_1},X_{I_2}\in A_{I_2}](d\omega_{I_1})\\
\displaystyle =P[X_{I_1}\in A_{I_1},X_{I_2}\in A_{I_2},X_{I_3}\in A_{I_3},X_{I_4}\in A_{I_4}], \ \forall A_{I_1}\in\mathscr{F}_{I_1},\end{array}$$
where fixed $A_{I_j}\in\mathscr{F}_{I_j},j=2,4$, then there exists $P[X_{I_3},X_{I_4}\in A_{I_4}|X_{I_1},X_{I_2}\in A_{I_2}](\omega_{I_1},A_{I_3})$, and
$$P[X_{I_3},X_{I_4}\in A_{I_4}|X_{I_1},X_{I_2}\in A_{I_2}](\omega_{I_1},A_{I_3})=K(\omega_{I_1},A_{I_3}).$$

(iii) Let
$$\mathscr{C}_1=\left\{\prod_{i\in s}A_i\times\prod_{i\in I_1\backslash s}\Omega_i:A_i\in\mathscr{F}_i,i\in s,s\in S_{I_1}\right\},$$
where $S_{I_1}$ stands for all of finite subsets of $I_1$. If $K(\omega_{I_1},A_{I_3})$ is a finite kernel from $(\Omega_{I_1},\mathscr{F}_{I_1})$ to $(\Omega_{I_3},\mathscr{F}_{I_3})$ such that, for $\forall A_{I_1}\in \mathscr{C}_1$ and $\forall A_{I_3}\in \mathscr{C}_3$,
$$\begin{array}{l}\displaystyle \int_{A_{I_1}}K(\omega_{I_1},A_{I_3})P[X_{I_1},X_{I_2}\in A_{I_2}](d\omega_{I_1})\\
\displaystyle =P[X_{I_1}\in A_{I_1},X_{I_2}\in A_{I_2},X_{I_3}\in A_{I_3},X_{I_4}\in A_{I_4}], \end{array}$$
where fixed $A_{I_j}\in\mathscr{F}_{I_j},j=2,4$, then there exists $P[X_{I_3},X_{I_4}\in A_{I_4}|X_{I_1},X_{I_2}\in A_{I_2}](\omega_{I_1},A_{I_3})$, and
$$P[X_{I_3},X_{I_4}\in A_{I_4}|X_{I_1},X_{I_2}\in A_{I_2}](\omega_{I_1},A_{I_3})=K(\omega_{I_1},A_{I_3}).$$
\end{comm}

\begin{comm}\label{comment2.12.2}Let $(\Omega,\mathscr{F},P)$ be a given probability space, $(\Omega_{i},\mathscr{F}_{i}),\ i\in \mathbb{N}$ measurable spaces. Let $X_{i}:(\Omega,\mathscr{F})\rightarrow(\Omega_{i},\mathscr{F}_{i}),\ i\in \mathbb{N}$ be random objects. Let $I_i\subset \mathbb{N},\ i=1,2,3,\cdots,6$ which are pairwise disjoint. And let $A_i\in \mathscr{F}_i,\ i\in I_3+I_4+I_5+I_6$ with $P[X_i\in A_i,\ i\in I_3+I_4+ I_5+I_6]>0$.
\begin{enumerate}
\renewcommand\theenumi{\roman{enumi}}
\renewcommand{\labelenumi}{(\theenumi)}
\item If $$P[X_i,i\in I_2;X_i\in A_i,i\in I_4|X_i,i\in I_1;X_i\in A_i,i\in I_3](\omega_i,\ i\in I_1;A_i,i\in I_2)$$ exists, then for fixed    $A_i\in\mathscr{F}_i,\ i\in I_2$,
$$\begin{array}{l}\displaystyle P[X_i,i\in I_2;X_i\in A_i,i\in I_4|X_i,i\in I_1;X_i\in A_i,i\in I_3](\omega_i,\ i\in I_1;A_i,i\in I_2)\\
\displaystyle =P[X_i\in A_i,i\in I_2+I_4|X_i,i\in I_1;X_i\in A_i,i\in I_3](\omega_i,\ i\in I_1);\\
\end{array}$$
\item If $$P[X_{I_2};X_i\in A_i,i\in I_4;X_i\in A_i,i\in I_6|X_{I_1};X_i\in A_i,i\in I_3;X_i\in A_i,i\in I_5](\omega_{I_1};A_i,i\in I_2)$$ exists, then
$$\begin{array}{l}\displaystyle P[X_{I_2};X_i\in A_i,i\in I_4;X_i\in A_i,i\in I_6|X_{I_1};X_i\in A_i,i\in I_3;X_i\in A_i,i\in I_5](\omega_{I_1};A_i,i\in I_2)\\
\displaystyle =P[X_i,i\in I_2;X_i\in A_i,i\in I_4+I_6|X_i,i\in I_1;X_i\in A_i,i\in I_3+I_5](\omega_i,\ i\in I_1;A_i,i\in I_2);
\end{array}$$
and vice versa.
\item If $$P[X_i,i\in I_2;X_i\in A_i,i\in I_4|X_i,i\in I_1;X_i\in A_i,i\in I_3;X_i\in \Omega_i,i\in I_5](\omega_{I_1};A_i,i\in I_2)$$ exists, then
$$\begin{array}{l}\displaystyle P[X_i,i\in I_2;X_i\in A_i,i\in I_4|X_i,i\in I_1;X_i\in A_i,i\in I_3;X_i\in \Omega_i,i\in I_5](\omega_{I_1};A_i,i\in I_2)\\
\displaystyle =P[X_i,i\in I_2;X_i\in A_i,i\in I_4|X_i,i\in I_1;X_i\in A_i,i\in I_3](\omega_i,\ i\in I_1;A_i,i\in I_2);
\end{array}$$
and vice versa.
\item If $$P[X_i,i\in I_2;X_i\in A_i,i\in I_4|X_i,i\in I_1;X_i\in \Omega_i,i\in I_3](\omega_i,\ i\in I_1;A_i,i\in I_2)$$ exists, then
$$\begin{array}{l}\displaystyle P[X_i,i\in I_2;X_i\in A_i,i\in I_4|X_i,i\in I_1;X_i\in \Omega_i,i\in I_3](\omega_i,\ i\in I_1;A_i,i\in I_2)\\
\displaystyle =P[X_i,i\in I_2;X_i\in A_i,i\in I_4|X_i,i\in I_1](\omega_i,\ i\in I_1;A_i,i\in I_2);
\end{array}$$
and vice versa.
\item If $$P[X_i,i\in I_2;X_i\in A_i,i\in I_4;X_i\in \Omega_i,i\in I_6|X_i,i\in I_1;X_i\in A_i,i\in I_3](\omega_{I_1};A_i,i\in I_2)$$ exists, then
$$\begin{array}{l}\displaystyle P[X_i,i\in I_2;X_i\in A_i,i\in I_4;X_i\in \Omega_i,i\in I_6|X_i,i\in I_1;X_i\in A_i,i\in I_3](\omega_{I_1};A_i,i\in I_2)\\
\displaystyle =P[X_i,i\in I_2;X_i\in A_i,i\in I_4|X_i,i\in I_1;X_i\in A_i,i\in I_3](\omega_i,\ i\in I_1;A_i,i\in I_2);
\end{array}$$
and vice versa.
\item If $$P[X_i,i\in I_2;X_i\in \Omega_i,i\in I_4|X_i,i\in I_1;X_i\in A_i,i\in I_3](\omega_i,\ i\in I_1;A_i,i\in I_2)$$ exists, then
$$\begin{array}{l}\displaystyle P[X_i,i\in I_2;X_i\in \Omega_i,i\in I_4|X_i,i\in I_1;X_i\in A_i,i\in I_3](\omega_i,\ i\in I_1;A_i,i\in I_2)\\
\displaystyle =P[X_i,i\in I_2|X_i,i\in I_1;X_i\in A_i,i\in I_3](\omega_i,\ i\in I_1;A_i,i\in I_2);
\end{array}$$
and vice versa.
\item If $$P[X_i,i\in I_2;X_i\in A_i,i\in I_4|X_i,i\in I_1;X_i\in A_i,i\in I_3](\omega_i,\ i\in I_1;A_i,i\in I_2)$$ exists, then
$$\begin{array}{l}\displaystyle P[X_i,i\in I_2;X_i\in A_i,i\in I_4|X_i,i\in I_1;X_i\in A_i,i\in I_3](\omega_i,\ i\in I_1;\Omega_i,i\in I_2)\\
\displaystyle =P[X_i\in A_i,i\in I_4|X_i,i\in I_1;X_i\in A_i,i\in I_3](\omega_i,\ i\in I_1).
\end{array}$$
\end{enumerate}
\end{comm}

Let $(\Omega,\mathscr{F},P)$ be a given probability space, $(\Omega_i,\mathscr{F}_i),\,i=1,2,3,\cdots,8$ measurable spaces. Let $X_i:(\Omega,\mathscr{F})\rightarrow(\Omega_i,\mathscr{F}_i),\ i=1,2,3,\cdots,8$ be random objects. Given $A_i\in\mathscr{F}_i,\ i=5,6,7,8$ with $P[X_i\in A_i,\ i=5,6,7,8]>0$, let
$$P[X_3,X_4;X_7\in A_7,X_8\in A_8|X_1,X_2;X_5\in A_5,X_6\in A_6](\omega_1,\omega_2;A_3,A_4)$$
be the conditional probability of co-occurrence of $X_i,\ i=3,4$ and $X_i\in A_i,\ i=7,8$ given co-occurrence of $X_i,\ i=1,2$ and $X_i\in A_i,\ i=5,6$, where $A_i\in \mathscr{F}_i,\ i=3,4$. Similar to the proof of Theorem \ref{theorem2.7.2}, we easily have the following Theorem \ref{theorem2.12.2}.

\begin{thm}\label{theorem2.12.2}Let $(\Omega,\mathscr{F},P)$, $(\Omega_i,\mathscr{F}_i),\,i=1,2,3,\cdots,8$ and $X_i:\Omega\rightarrow\Omega_i,\ i=1,2,3,\cdots,8$ be the same as above and $A_i\in\mathscr{F}_i,\ i=5,6,7,8$. If $A_{41}\subset A_{42}$, where $A_{41},\ A_{42}\in\mathscr{F}_4$, then
$$\begin{array}{l}\displaystyle P[X_3,X_4;X_7\in A_7,X_8\in A_8|X_1,X_2;X_5\in A_5,X_6\in A_6](\omega_1,\omega_2;A_3,A_{41})\\
\displaystyle \leq P[X_3,X_4;X_7\in A_7,X_8\in A_8|X_1,X_2;X_5\in A_5,X_6\in A_6](\omega_1,\omega_2;A_3,A_{42}),
\end{array}
$$
a.e.$\{P[X_1,X_2;X_5\in A_5,X_6\in A_6]\}$ provided that $P[X_1,X_2;X_5\in A_5,X_6\in A_6]>0$.
\end{thm}

\begin{comm}\label{comment2.12.3} In view of the definitions above we note that, probability of co-occurrence and conditional probability of co-occurrence are not necessarily probability measures while probability measure is a special case of them.
\end{comm}

The next section is devoted to the transformation of $P[X_i,i\in I_3;X_i\in A_i,i\in I_4|X_i,i\in I_1;X_i\in A_i,i\in I_2](\omega_i,\ i\in I_1;A_i,i\in I_3)$ in Definition \ref{definition2.12}.

\section {Transformation of Conditional Probability of Co-occurrence}\label{section3}

Now we study the transformation of conditional probability of co-occurrence.

\begin{thm}\label{theorem3.1}Let $(\Omega,\mathscr{F},P)$ be a given probability space, $(\Omega_i,\mathscr{F}_i),\,i=1,2,3,4,5$ measurable spaces and $X_i:(\Omega,\mathscr{F})\rightarrow(\Omega_i,\mathscr{F}_i),\ i=1,2,3,4,5$ random objects. Given fixed $A_i\in\mathscr{F}_i,\ i=4,5$. If there exists $P[X_2,X_3;X_4\in A_4|X_1,X_5\in A_5](\omega_1;A_2,A_3)$, then so does $P[X_2;X_3\in A_3,X_4\in A_4|X_1,X_5\in A_5](\omega_1,A_2)$ for each fixed $A_3\in\mathscr{F}_3$, and
$$\begin{array}{l}\displaystyle P[X_2;X_3\in A_3,X_4\in A_4|X_1,X_5\in A_5](\omega_1,A_2)\\\\
\displaystyle=P[X_2,X_3;X_4\in A_4|X_1,X_5\in A_5](\omega_1;A_2,A_3).
\end{array}$$
\end{thm}

\begin{col}\label{corollary3.1.1}Let $(\Omega,\mathscr{F},P)$ be a given probability space, $(\Omega_i,\mathscr{F}_i),\,i=1,2,3,4$ measurable spaces and $X_i:(\Omega,\mathscr{F})\rightarrow(\Omega_i,\mathscr{F}_i),\ i=1,2,3,4$ random objects. Given fixed $A_4\in\mathscr{F}_4$.
\begin{enumerate}
\item If there exists $P[X_2,X_3|X_1](\omega_1;A_2,A_3)$, then so does $P[X_2,X_3\in A_3|X_1](\omega_1,A_2)$ for each fixed $A_3\in\mathscr{F}_3$, and
$$P[X_2,X_3\in A_3|X_1](\omega_1,A_2)=P[X_2,X_3|X_1](\omega_1;A_2,A_3).$$
\item If there exists $P[X_2,X_3;X_4\in A_4|X_1](\omega_1;A_2,A_3)$, then so does $P[X_2;X_3\in A_3,X_4\in A_4|X_1](\omega_1,A_2)$ for each fixed $A_3\in\mathscr{F}_3$, and
$$P[X_2;X_3\in A_3,X_4\in A_4|X_1](\omega_1,A_2)=P[X_2,X_3;X_4\in A_4|X_1](\omega_1;A_2,A_3).$$
\item If there exists $P[X_2,X_3|X_1,X_4\in A_4](\omega_1;A_2,A_3)$, then so does $P[X_2,X_3\in A_3|X_1,X_4\in A_4](\omega_1,A_2)$ for each fixed $A_3\in\mathscr{F}_3$, and
$$P[X_2,X_3\in A_3|X_1,X_4\in A_4](\omega_1,A_2)=P[X_2,X_3|X_1,X_4\in A_4](\omega_1;A_2,A_3).$$
\end{enumerate}
\end{col}

By Theorem \ref{theorem3.1} and Corollary \ref{corollary3.1.1}, we have the following corollary.

\begin{col}\label{corollary3.1.2}Let $(\Omega,\mathscr{F},P)$ be a given probability space, $(\Omega_{i},\mathscr{F}_{i}),\ i\in \mathbb{N}$ measurable spaces. Let $X_i:(\Omega,\mathscr{F})\rightarrow(\Omega_i,\mathscr{F}_i),\ i\in \mathbb{N}$ be random objects. Given $I_i\subset \mathbb{N},\ i=1,2,3,4,5$ and $I_i\cap I_j=\emptyset,\ i\neq j$. And given fixed $A_i\in \mathscr{F}_i,\ i\in I_4+I_5$.
\begin{enumerate}
\item If there exists $P[X_i,i\in I_2+I_3|X_i,i\in I_1](\omega_i,i\in I_1;A_i,i\in I_2+I_3)$, then so does
$$P[X_i,i\in I_2;X_i\in A_i,i\in I_3|X_i,i\in I_1](\omega_i,i\in I_1;A_i,i\in I_2)$$
for each fixed $A_i\in\mathscr{F}_i,\ i\in I_3$ and
$$\begin{array}{l}\displaystyle P[X_i,i\in I_2;X_i\in A_i,i\in I_3|X_i,i\in I_1](\omega_i,i\in I_1;A_i,i\in I_2)\\
\displaystyle=P[X_i,i\in I_2+I_3|X_i,i\in I_1](\omega_i,i\in I_1;A_i,i\in I_2;A_i,i\in I_3).
\end{array}$$
\item If there exists $P[X_i,i\in I_2+I_3;X_i\in A_i,i\in I_4|X_i,i\in I_1](\omega_i,i\in I_1;A_i,i\in I_2+I_3)$, then so does
$$P[X_i,i\in I_2;X_{I_3}\in \prod_{i\in I_3}A_i,X_{I_4}\in \prod_{i\in I_4}A_i|X_i,i\in I_1](\omega_i,i\in I_1;A_i,i\in I_2)$$
for each fixed $A_i\in\mathscr{F}_i,\ i\in I_3$, and
$$\begin{array}{l}\displaystyle P[X_i,i\in I_2;X_{I_3}\in \prod_{i\in I_3}A_i,X_{I_4}\in \prod_{i\in I_4}A_i|X_i,i\in I_1](\omega_i,i\in I_1;A_i,i\in I_2)\\
\displaystyle=P[X_i,i\in I_2+I_3;X_i\in A_i,i\in I_4|X_i,i\in I_1](\omega_i,i\in I_1;A_i,i\in I_2;A_i,i\in I_3).
\end{array}$$
\item If there exists $P[X_i,i\in I_2+I_3|X_i,i\in I_1;X_i\in A_i,i\in I_4](\omega_i,i\in I_1;A_i,i\in I_2+I_3)$, then so does
$$P[X_i,i\in I_2;X_{I_3}\in \prod_{i\in I_3}A_i|X_i,i\in I_1;X_{I_4}\in \prod_{i\in I_4}A_i](\omega_i,i\in I_1;A_i,i\in I_2)$$
for each fixed $A_i\in\mathscr{F}_i,\ i\in I_3$, and
$$\begin{array}{l}\displaystyle P[X_i,i\in I_2;X_{I_3}\in \prod_{i\in I_3}A_i|X_i,i\in I_1,X_{I_4}\in \prod_{i\in I_4}A_i](\omega_i,i\in I_1;A_i,i\in I_2)\\
\displaystyle=P[X_i,i\in I_2+I_3|X_i,i\in I_1;X_i\in A_i,i\in I_4](\omega_i,i\in I_1;A_i,i\in I_2;A_i,i\in I_3).
\end{array}$$
\item If there exists
$$P[X_i,i\in I_2+I_3;X_i\in A_i,i\in I_4|X_i,i\in I_1;X_i\in A_i,i\in I_5](\omega_i,i\in I_1;A_i,i\in I_2+I_3),$$
then so does
$$P[X_i,i\in I_2;X_{I_3}\in \prod_{i\in I_3}A_i,X_{I_4}\in \prod_{i\in I_4}A_i|X_i,i\in I_1;X_{I_5}\in \prod_{i\in I_5}A_i](\omega_i,i\in I_1;A_i,i\in I_2)$$
for each fixed $A_i\in\mathscr{F}_i,\ i\in I_3$, and
$$\begin{array}{l}\displaystyle P[X_i,i\in I_2;X_{I_3}\in \prod_{i\in I_3}A_i,X_{I_4}\in \prod_{i\in I_4}A_i|X_i,i\in I_1,X_{I_5}\in \prod_{i\in I_5}A_i](\omega_i,i\in I_1;A_i,i\in I_2)\\
\displaystyle=P[X_i,i\in I_2+I_3;X_i\in A_i,i\in I_4|X_i,i\in I_1;X_{I_5}\in \prod_{i\in I_5}A_i](\omega_{I_1},\prod_{i\in I_2+I_3}A_i).
\end{array}$$
\end{enumerate}
\end{col}

From Corollary \ref{corollary3.1.2} and Theorem \ref{theorem2.11.3} it follows that

\begin{col}\label{corollary3.1.3}  Let $(\Omega,\mathscr{F},P)$ be a given probability space. Let $I_i\subset \mathbb{N},\ i=1,2,3,4,5$ and $I_i\cap I_j=\emptyset,\ i\neq j$. Let $X_i,\ i\in \mathbb{N}$ be random variables on $(\Omega,\mathscr{F},P)$. Given $A_i\in \mathscr{B}(\mathbb{R}),\ i\in I_4+I_5$.
\begin{enumerate}
\item For each fixed $A_i\in\mathscr{B}(\mathbb{R}),\ i\in I_3$,
\begin{eqnarray*}
&&P[X_i,i\in I_2;X_i\in A_i,i\in I_3|X_i,i\in I_1](x_i,i\in I_1;A_i,i\in I_2)\\
&=&P[X_i,i\in I_2+I_3|X_i,i\in I_1](x_i,i\in I_1;A_i,i\in I_2;A_i,i\in I_3).
\end{eqnarray*}
\item For each fixed $A_i\in\mathscr{B}(\mathbb{R}),\ i\in I_3$,
\begin{eqnarray*}&& P[X_i,i\in I_2;X_{I_3}\in \prod_{i\in I_3}A_i,X_{I_4}\in \prod_{i\in I_4}A_i|X_i,i\in I_1](x_i,i\in I_1;A_i,i\in I_2)\\
&=&P[X_i,i\in I_2+I_3;X_i\in A_i,i\in I_4|X_i,i\in I_1](x_i,i\in I_1;A_i,i\in I_2;A_i,i\in I_3).
\end{eqnarray*}
\item For each fixed $A_i\in\mathscr{F}_i,\ i\in I_3$,
\begin{eqnarray*}
& &P[X_i,i\in I_2;X_{I_3}\in \prod_{i\in I_3}A_i|X_i,i\in I_1,X_{I_4}\in \prod_{i\in I_4}A_i](x_i,i\in I_1;A_i,i\in I_2)\\
&=&P[X_i,i\in I_2+I_3|X_i,i\in I_1;X_i\in A_i,i\in I_4](x_i,i\in I_1;A_i,i\in I_2;A_i,i\in I_3).
\end{eqnarray*}
\item For each fixed $A_i\in\mathscr{F}_i,\ i\in I_3$,
\begin{eqnarray*}
&& P[X_i,i\in I_2;X_{I_3}\in \prod_{i\in I_3}A_i,X_{I_4}\in \prod_{i\in I_4}A_i|X_i,i\in I_1,X_{I_5}\in \prod_{i\in I_5}A_i](x_i,i\in I_1;A_i,i\in I_2)\\
&=&P[X_i,i\in I_2+I_3;X_i\in A_i,i\in I_4|X_i,i\in I_1;X_{I_5}\in \prod_{i\in I_5}A_i](x_{I_1},\prod_{i\in I_2+I_3}A_i).
\end{eqnarray*}
\end{enumerate}
\end{col}

\begin{thm}\label{theorem3.2}Let $(\Omega,\mathscr{F},P)$ be a given probability space, $(\Omega_i,\mathscr{F}_i),\,i=1,2,3,4,5$ measurable spaces and $X_i:(\Omega,\mathscr{F})\rightarrow(\Omega_i,\mathscr{F}_i),\ i=1,2,3,4,5$ random objects. Given fixed $A_i\in\mathscr{F}_i,\ i=3,4,5$. If there exists $P[X_2,X_4\in A_4|X_1;X_3\in A_3,X_5\in A_5](\omega_1,A_2)$, then so does $P[X_2;X_3\in A_3,X_4\in A_4|X_1,X_5\in A_5](\omega_1,A_2)$, and
\begin{eqnarray*}
&& P[X_2;X_3\in A_3,X_4\in A_4|X_1,X_5\in A_5](\omega_1,A_2)\\
&=&P[X_2,X_4\in A_4|X_1;X_3\in A_3,X_5\in A_5](\omega_1,A_2)\cdot P[X_3\in A_3|X_1,X_5\in A_5](\omega_1).
\end{eqnarray*}
Conversely, if there exists $P[X_2;X_3\in A_3,X_4\in A_4|X_1,X_5\in A_5](\omega_1,A_2)$, then so does $P[X_2,X_4\in A_4|X_1;X_3\in A_3,X_5\in A_5](\omega_1,A_2)$, and
\begin{eqnarray*}
&& P[X_2,X_4\in A_4|X_1;X_3\in A_3,X_5\in A_5](\omega_1,A_2)\\
&=&\left\{\begin{array}{l}\displaystyle \frac{P[X_2;X_3\in A_3,X_4\in A_4|X_1,X_5\in A_5](\omega_1,A_2)}{P[X_3\in A_3|X_1,X_5\in A_5](\omega_1)},\ \ \omega_1\in E,\\\\
\displaystyle 0,\ \ \ \ \ \ \ \ \ \omega_1\in E^c,\end{array}\right.
\end{eqnarray*}
where $E=\{\omega_1:P[X_3\in A_3|X_1,X_5\in A_5](\omega_1)\neq 0\}$ and $E^c=\Omega_1\backslash E$.
\end{thm}

We have the following corollary.

\begin{col}\label{corollary3.2.1}Let $(\Omega,\mathscr{F},P)$ be a given probability space, $(\Omega_{i},\mathscr{F}_{i}),\ i\in \mathbb{N}$ measurable spaces. Let $X_i:(\Omega,\mathscr{F})\rightarrow(\Omega_i,\mathscr{F}_i),\ i\in \mathbb{N}$ be random objects. Given $I_i\subset \mathbb{N},\ i=1,2,3,4,5$ and $I_i\cap I_j=\emptyset,\ i\neq j$. And given fixed $A_i\in \mathscr{F}_i,\ i\in I_3+I_4+I_5$. If there exists $$P[X_i,i\in I_2;X_i\in A_i,i\in I_4|X_i,i\in I_1;X_i\in A_i,i\in I_3+I_5](\omega_{I_1};A_i,i\in I_2),$$ then so does $$P[X_i,i\in I_2;X_i\in A_i,i\in I_3+I_4|X_i,i\in I_1;X_i\in A_i,i\in I_5](\omega_{I_1};A_i,i\in I_2),$$ and
$$\begin{array}{l}\displaystyle P[X_i,i\in I_2;X_i\in A_i,i\in I_3+I_4|X_i,i\in I_1;X_i\in A_i,i\in I_5](\omega_{I_1};A_i,i\in I_2)\\
\displaystyle =P[X_i,i\in I_2;X_i\in A_i,i\in I_4|X_i,i\in I_1;X_i\in A_i,i\in I_3+I_5](\omega_{I_1};A_i,i\in I_2)\\
\displaystyle \ \ \ \cdot P[X_i\in A_i,i\in I_3|X_i,i\in I_1;X_i\in A_i,i\in I_5](\omega_{I_1}).\end{array}$$
Conversely, if there exists $$P[X_i,i\in I_2;X_i\in A_i,i\in I_3+I_4|X_i,i\in I_1;X_i\in A_i,i\in I_5](\omega_{I_1};A_i,i\in I_2),$$ then so does $$P[X_i,i\in I_2;X_i\in A_i,i\in I_4|X_i,i\in I_1;X_i\in A_i,i\in I_3+I_5](\omega_{I_1};A_i,i\in I_2),$$ and
$$\begin{array}{l}\displaystyle P[X_i,i\in I_2;X_i\in A_i,i\in I_4|X_i,i\in I_1;X_i\in A_i,i\in I_3+I_5](\omega_{I_1};A_i,i\in I_2)\\\\
\displaystyle =\left\{\begin{array}{l}\displaystyle \frac{P[X_i,i\in I_2;X_i\in A_i,i\in I_3+I_4|X_i,i\in I_1;X_i\in A_i,i\in I_5](\omega_{I_1};A_i,i\in I_2)}{P[X_i\in A_i,i\in I_3|X_i,i\in I_1;X_i\in A_i,i\in I_5](\omega_{I_1})},\ \omega_{I_1}\in E,\\\\
\displaystyle 0,\ \ \ \ \ \ \ \ \ \omega_{I_1}\in E^c,\end{array}\right.\end{array}$$
where $E=\{\omega_{I_1}:P[X_i\in A_i,i\in I_3|X_i,i\in I_1;X_i\in A_i,i\in I_5](\omega_{I_1})\neq 0\}$ and $E^c=\Omega_{I_1}\setminus E$.
\end{col}

According to Theorem \ref{theorem3.1} and Theorem \ref{theorem3.2}, we obtain the following theorem.

\begin{thm}\label{theorem3.3}Let $(\Omega,\mathscr{F},P)$ be a given probability space, $(\Omega_i,\mathscr{F}_i),\,i=1,2,3,4,5$ measurable spaces and $X_i:(\Omega,\mathscr{F})\rightarrow(\Omega_i,\mathscr{F}_i),\ i=1,2,3,4,5$ random objects. Given fixed $A_i\in\mathscr{F}_i,\ i=4,5$. If there exists $P[X_2,X_3;X_4\in A_4|X_1,X_5\in A_5](\omega_1;A_2,A_3)$, then so do $P[X_2;X_3\in A_3,X_4\in A_4|X_1,X_5\in A_5](\omega_1,A_2)$ and $P[X_2,X_4\in A_4|X_1;X_3\in A_3,X_5\in A_5](\omega_1,A_2)$ for each fixed $A_3\in\mathscr{F}_3$, and
$$\begin{array}{l}\displaystyle P[X_2;X_3\in A_3,X_4\in A_4|X_1,X_5\in A_5](\omega_1,A_2)\\
\displaystyle =P[X_2,X_3;X_4\in A_4|X_1,X_5\in A_5](\omega_1;A_2,A_3)\end{array}$$
and
$$\begin{array}{l}\displaystyle P[X_2,X_4\in A_4|X_1;X_3\in A_3,X_5\in A_5](\omega_1,A_2)\\\\
\displaystyle =\left\{\begin{array}{l}\displaystyle \frac{P[X_2,X_3;X_4\in A_4|X_1,X_5\in A_5](\omega_1;A_2,A_3)}{P[X_3\in A_3|X_1,X_5\in A_5](\omega_1)},\ \omega_1\in E,\\\\
\displaystyle 0,\ \ \ \ \ \ \ \ \ \omega_1\in E^c,\end{array}\right.\end{array}$$
where $E=\{\omega_1:P[X_3\in A_3|X_1,X_5\in A_5](\omega_1)\neq 0\}$ and $E^c=\Omega_1\setminus E$.
\end{thm}

\begin{col}\label{corollary3.3.1}Let $(\Omega,\mathscr{F},P)$ be a given probability space, $(\Omega_i,\mathscr{F}_i),\,i=1,2,3,4$ measurable spaces and $X_i:(\Omega,\mathscr{F})\rightarrow(\Omega_i,\mathscr{F}_i),\ i=1,2,3,4$ random objects.
\begin{enumerate}
\item If there exists $P[X_2,X_3|X_1](\omega_1;A_2,A_3)$, then so does $P[X_2|X_1](\omega_1,A_2)$, and
$$P[X_2|X_1](\omega_1,A_2)=P[X_2,X_3|X_1](\omega_1;A_2,\Omega_3).$$
\item If there exists $P[X_2,X_3|X_1](\omega_1;A_2,A_3)$, then so do $P[X_2,X_3\in A_3|X_1](\omega_1,A_2)$ and $P[X_2|X_1,X_3\in A_3](\omega_1,A_2)$ for each fixed $A_3\in\mathscr{F}_3$, and
$$P[X_2,X_3\in A_3|X_1](\omega_1,A_2)=P[X_2,X_3|X_1](\omega_1;A_2,A_3)$$
and
$$P[X_2|X_1,X_3\in A_3](\omega_1,A_2)=\left\{\begin{array}{l}\displaystyle \frac{P[X_2,X_3|X_1](\omega_1;A_2,A_3)}{P[X_3\in A_3|X_1](\omega_1)},\ \omega_1\in E,\\\\
\displaystyle 0,\ \ \ \ \ \ \ \ \ \omega_1\in E^c,\end{array}\right.$$
where $E=\{\omega_1:P[X_3\in A_3|X_1](\omega_1)\neq 0\}$ and $E^c=\Omega_1\setminus E$.
\item If there exists $P[X_2,X_3;X_4\in A_4|X_1](\omega_1;A_2,A_3)$ for fixed $A_4\in\mathscr{F}_4$, then so do $P[X_2;X_3\in A_3,X_4\in A_4|X_1](\omega_1,A_2)$ and $P[X_2,X_4\in A_4|X_1,X_3\in A_3](\omega_1,A_2)$ for each fixed $A_3\in\mathscr{F}_3$, and
$$P[X_2;X_3\in A_3,X_4\in A_4|X_1](\omega_1,A_2)=P[X_2,X_3;X_4\in A_4|X_1](\omega_1;A_2,A_3)$$
and
$$\begin{array}{l}\displaystyle P[X_2,X_4\in A_4|X_1,X_3\in A_3](\omega_1,A_2)\\\\
\displaystyle =\left\{\begin{array}{l}\displaystyle \frac{P[X_2,X_3;X_4\in A_4|X_1](\omega_1;A_2,A_3)}{P[X_3\in A_3|X_1](\omega_1)},\ \omega_1\in E,\\\\
\displaystyle 0,\ \ \ \ \ \ \ \ \ \omega_1\in E^c,\end{array}\right.\end{array}$$
where $E=\{\omega_1:P[X_3\in A_3|X_1](\omega_1)\neq 0\}$ and $E^c=\Omega_1\setminus E$.
\item If there exists $P[X_2,X_3|X_1,X_4\in A_4](\omega_1;A_2,A_3)$ for fixed $A_4\in\mathscr{F}_4$, then so do $P[X_2,X_3\in A_3|X_1,X_4\in A_4](\omega_1,A_2)$ and $P[X_2|X_1;X_3\in A_3,X_4\in A_4](\omega_1,A_2)$ for each fixed $A_3\in\mathscr{F}_3$, and
$$P[X_2,X_3\in A_3|X_1,X_4\in A_4](\omega_1,A_2)=P[X_2,X_3|X_1,X_4\in A_4](\omega_1; A_2,A_3)$$
and
$$\begin{array}{l}\displaystyle P[X_2|X_1;X_3\in A_3,X_4\in A_4](\omega_1,A_2)\\\\
\displaystyle =\left\{\begin{array}{l}\displaystyle \frac{P[X_2,X_3|X_1,X_4\in A_4](\omega_1;A_2,A_3)}{P[X_3\in A_3|X_1,X_4\in A_4](\omega_1)},\ \omega_1\in E,\\\\
\displaystyle 0,\ \ \ \ \ \ \ \ \ \omega_1\in E^c,\end{array}\right.\end{array}$$
where $E=\{\omega_1:P[X_3\in A_3|X_1,X_4\in A_4](\omega_1)\neq 0\}$ and $E^c=\Omega_1\setminus E$.
\end{enumerate}
\end{col}

We also have the following corollary.

\begin{col}\label{corollary3.3.2}Let $(\Omega,\mathscr{F},P)$ be a given probability space, $(\Omega_{i},\mathscr{F}_{i}),\ i\in \mathbb{N}$ measurable spaces. Let $X_i:(\Omega,\mathscr{F})\rightarrow(\Omega_i,\mathscr{F}_i),\ i\in \mathbb{N}$ be random objects. Given $I_i\subset \mathbb{N},\ i=1,2,3,4,5$ and $I_i\cap I_j=\emptyset,\ i\neq j$. And given fixed $A_i\in \mathscr{F}_i,\ i\in I_4+I_5$.
\begin{enumerate}
\item If there exists $P[X_i,i\in I_2+I_3|X_i,i\in I_1](\omega_i,i\in I_1;A_i,i\in I_2+I_3)$, then so does
$$P[X_i,i\in I_2|X_i,i\in I_1](\omega_i,i\in I_1;A_i,i\in I_2),$$
and
$$\begin{array}{l}\displaystyle P[X_i,i\in I_2|X_i,i\in I_1](\omega_i,i\in I_1;A_i,i\in I_2)\\
\displaystyle=P[X_i,i\in I_2+I_3|X_i,i\in I_1](\omega_i,i\in I_1;A_i,i\in I_2;\Omega_i,i\in I_3).
\end{array}$$
\item If there exists $P[X_i,i\in I_2+I_3|X_i,i\in I_1](\omega_i,i\in I_1;A_i,i\in I_2+I_3)$, then so do
$$P[X_i,i\in I_2;X_i\in A_i,i\in I_3|X_i,i\in I_1](\omega_i,i\in I_1;A_i,i\in I_2)$$
and
$$P[X_i,i\in I_2|X_i,i\in I_1;X_i\in A_i,i\in I_3](\omega_i,i\in I_1;A_i,i\in I_2)$$
for fixed $A_i\in\mathscr{F}_i,\ i\in I_3$, and
$$\begin{array}{l}\displaystyle P[X_i,i\in I_2;X_i\in A_i,i\in I_3|X_i,i\in I_1](\omega_i,i\in I_1;A_i,i\in I_2)\\
\displaystyle=P[X_i,i\in I_2+I_3|X_i,i\in I_1](\omega_i,i\in I_1;A_i,i\in I_2;A_i,i\in I_3)
\end{array}$$
and
$$\begin{array}{l}\displaystyle P[X_i,i\in I_2|X_i,i\in I_1;X_i\in A_i,i\in I_3](\omega_i,i\in I_1;A_i,i\in I_2)\\\\
\displaystyle =\left\{\begin{array}{l}\displaystyle \frac{P[X_i,i\in I_2+I_3|X_i,i\in I_1](\omega_i,i\in I_1;A_i,i\in I_2;A_i,i\in I_3)}{P[X_i\in A_i,i\in I_3|X_i,i\in I_1](\omega_i,i\in I_1)},\ \omega_{I_1}\in E,\\\\
\displaystyle 0,\ \ \ \ \ \ \ \ \ \omega_{I_1}\in E^c,\end{array}\right.\end{array}$$
where $E=\{\omega_{I_1}:P[X_i\in A_i,i\in I_3|X_i,i\in I_1](\omega_{I_1})\neq 0\}$ and $E^c=\Omega_{I_1}\setminus E$.\\
\item If there exists $P[X_i,i\in I_2+I_3;X_i\in A_i,i\in I_4|X_i,i\in I_1](\omega_i,i\in I_1;A_i,i\in I_2+I_3)$, then so do
$$P[X_i,i\in I_2;X_{I_3}\in \prod_{i\in I_3}A_i,X_{I_4}\in \prod_{i\in I_4}A_i|X_i,i\in I_1](\omega_i,i\in I_1;A_i,i\in I_2)$$
and
$$P[X_i,i\in I_2;X_{I_4}\in \prod_{i\in I_4}A_i|X_i,i\in I_1;X_{I_3}\in \prod_{i\in I_3}A_i](\omega_i,i\in I_1;A_i,i\in I_2)$$
for fixed $A_i\in\mathscr{F}_i,\ i\in I_3$, and
$$\begin{array}{l}\displaystyle P[X_i,i\in I_2;X_{I_3}\in \prod_{i\in I_3}A_i,X_{I_4}\in \prod_{i\in I_4}A_i|X_i,i\in I_1](\omega_i,i\in I_1;A_i,i\in I_2)\\
\displaystyle=P[X_i,i\in I_2+I_3;X_i\in A_i,i\in I_4|X_i,i\in I_1](\omega_i,i\in I_1;A_i,i\in I_2;A_i,i\in I_3)
\end{array}$$
and
$$\begin{array}{l}\displaystyle P[X_i,i\in I_2;X_{I_4}\in \prod_{i\in I_4}A_i|X_i,i\in I_1;X_{I_3}\in \prod_{i\in I_3}A_i](\omega_i,i\in I_1;A_i,i\in I_2)\\\\
\displaystyle =\left\{\begin{array}{l}\displaystyle \frac{P[X_i,i\in I_2+I_3;X_i\in A_i,i\in I_4|X_i,i\in I_1](\omega_{I_1},\prod\limits_{i\in I_2+I_3}A_i)}{P[X_i\in A_i,i\in I_3|X_i,i\in I_1](\omega_i,i\in I_1)},\ \omega_{I_1}\in E,\\\\
\displaystyle 0,\ \ \ \ \ \ \ \ \ \omega_{I_1}\in E^c,\end{array}\right.\end{array}$$
where $E=\{\omega_{I_1}:P[X_i\in A_i,i\in I_3|X_i,i\in I_1](\omega_{I_1})\neq 0\}$ and $E^c=\Omega_{I_1}\setminus E$.\\
\item If there exists $P[X_i,i\in I_2+I_3|X_i,i\in I_1;X_i\in A_i,i\in I_4](\omega_i,i\in I_1;A_i,i\in I_2+I_3)$, then so do
$$P[X_i,i\in I_2;X_{I_3}\in \prod_{i\in I_3}A_i|X_i,i\in I_1;X_{I_4}\in \prod_{i\in I_4}A_i](\omega_i,i\in I_1;A_i,i\in I_2)$$
and
$$P[X_i,i\in I_2|X_i,i\in I_1;X_{I_3}\in \prod_{i\in I_3}A_i,X_{I_4}\in \prod_{i\in I_4}A_i](\omega_i,i\in I_1;A_i,i\in I_2)$$
for fixed $A_i\in\mathscr{F}_i,\ i\in I_3$, and
$$\begin{array}{l}\displaystyle P[X_i,i\in I_2;X_{I_3}\in \prod_{i\in I_3}A_i|X_i,i\in I_1,X_{I_4}\in \prod_{i\in I_4}A_i](\omega_i,i\in I_1;A_i,i\in I_2)\\
\displaystyle=P[X_i,i\in I_2+I_3|X_i,i\in I_1;X_i\in A_i,i\in I_4](\omega_i,i\in I_1;A_i,i\in I_2;A_i,i\in I_3)
\end{array}$$
and
$$\begin{array}{l}\displaystyle P[X_i,i\in I_2|X_i,i\in I_1;X_{I_3}\in \prod_{i\in I_3}A_i,X_{I_4}\in \prod_{i\in I_4}A_i](\omega_i,i\in I_1;A_i,i\in I_2)\\\\
\displaystyle =\left\{\begin{array}{l}\displaystyle \frac{P[X_i,i\in I_2+I_3|X_i,i\in I_1;X_i\in A_i,i\in I_4](\omega_{I_1},\prod\limits_{i\in I_2+I_3}A_i)}{P[X_i\in A_i,i\in I_3|X_i,i\in I_1;X_i\in A_i,i\in I_4](\omega_i,i\in I_1)},\ \omega_{I_1}\in E,\\\\
\displaystyle 0,\ \ \ \ \ \ \ \ \ \omega_{I_1}\in E^c,\end{array}\right.\end{array}$$
where $E=\{\omega_{I_1}:P[X_i\in A_i,i\in I_3|X_i,i\in I_1;X_i\in A_i,i\in I_4](\omega_{I_1})\neq 0\}$ and $E^c=\Omega_{I_1}\setminus E$.\\
\item If there exists $P[X_i,i\in I_2+I_3;X_i\in A_i,i\in I_4|X_i,i\in I_1;X_i\in A_i,i\in I_5](\omega_i,i\in I_1;A_i,i\in I_2+I_3)$, then so do
$$P[X_i,i\in I_2;X_{I_3}\in \prod_{i\in I_3}A_i,X_{I_4}\in \prod_{i\in I_4}A_i|X_i,i\in I_1;X_{I_5}\in \prod_{i\in I_5}A_i](\omega_i,i\in I_1;A_i,i\in I_2)$$
and
$$P[X_i,i\in I_2;X_{I_4}\in \prod_{i\in I_4}A_i|X_i,i\in I_1;X_{I_3}\in \prod_{i\in I_3}A_i,X_{I_5}\in \prod_{i\in I_5}A_i](\omega_i,i\in I_1;A_i,i\in I_2)$$
for fixed $A_i\in\mathscr{F}_i,\ i\in I_3$, and
$$\begin{array}{l}\displaystyle P[X_i,i\in I_2;X_{I_3}\in \prod_{i\in I_3}A_i,X_{I_4}\in \prod_{i\in I_4}A_i|X_i,i\in I_1;X_{I_5}\in \prod_{i\in I_5}A_i](\omega_i,i\in I_1;A_i,i\in I_2)\\
\displaystyle=P[X_i,i\in I_2+I_3;X_i\in A_i,i\in I_4|X_i,i\in I_1;X_{I_5}\in \prod_{i\in I_5}A_i](\omega_{I_1},\prod_{i\in I_2+I_3}A_i)
\end{array}$$
and
$$\begin{array}{l}\displaystyle P[X_i,i\in I_2;X_{I_4}\in \prod_{i\in I_4}A_i|X_i,i\in I_1;X_{I_3}\in \prod_{i\in I_3}A_i,X_{I_5}\in \prod_{i\in I_5}A_i](\omega_i,i\in I_1;A_i,i\in I_2)\\\\
\displaystyle =\left\{\begin{array}{l}\displaystyle \frac{P[X_i,i\in I_2+I_3;X_{I_4}\in \prod\limits_{i\in I_4}A_i|X_i,i\in I_1;X_{I_5}\in \prod\limits_{i\in I_5}A_i](\omega_{I_1},\prod\limits_{i\in I_2+I_3}A_i)}{P[X_i\in A_i,i\in I_3|X_i,i\in I_1;X_i\in A_i,i\in I_5](\omega_i,i\in I_1)},\ \omega_{I_1}\in E,\\\\
\displaystyle 0,\ \ \ \ \ \ \ \ \ \omega_{I_1}\in E^c,\end{array}\right.\end{array}$$
where $E=\{\omega_{I_1}:P[X_i\in A_i,i\in I_3|X_i,i\in I_1;X_i\in A_i,i\in I_5](\omega_{I_1})\neq 0\}$ and $E^c=\Omega_{I_1}\setminus E$.
\end{enumerate}
\end{col}

From Corollary \ref{corollary3.3.2} and Theorem \ref{theorem2.11.3} it follows that

\begin{col}\label{corollary3.3.3}  Let $(\Omega,\mathscr{F},P)$ be a given probability space. Let $I_i\subset \mathbb{N},\ i=1,2,3,4,5$ and $I_i\cap I_j=\emptyset,\ i\neq j$. Let $X_i,\ i\in \mathbb{N}$ be random variables on $(\Omega,\mathscr{F},P)$. Given fixed $A_i\in \mathscr{B}(\mathbb{R}),\ i\in I_4+I_5$.
\begin{enumerate}
\item
$$\begin{array}{l}\displaystyle P[X_i,i\in I_2|X_i,i\in I_1](x_i,i\in I_1;A_i,i\in I_2)\\
\displaystyle=P[X_i,i\in I_2+I_3|X_i,i\in I_1](x_i,i\in I_1;A_i,i\in I_2,\mathbb{R}^{I_3}).
\end{array}$$
\item For fixed $A_i\in\mathscr{B}(\mathbb{R}),\ i\in I_3$,
$$\begin{array}{l}\displaystyle P[X_i,i\in I_2;X_i\in A_i,i\in I_3|X_i,i\in I_1](x_i,i\in I_1;A_i,i\in I_2)\\
\displaystyle=P[X_i,i\in I_2+I_3|X_i,i\in I_1](x_i,i\in I_1;A_i,i\in I_2;A_i,i\in I_3),
\end{array}$$
and
$$\begin{array}{l}\displaystyle P[X_i,i\in I_2|X_i,i\in I_1;X_i\in A_i,i\in I_3](x_i,i\in I_1;A_i,i\in I_2)\\\\
\displaystyle =\left\{\begin{array}{l}\displaystyle \frac{P[X_i,i\in I_2+I_3|X_i,i\in I_1](x_i,i\in I_1;A_i,i\in I_2;A_i,i\in I_3)}{P[X_i\in A_i,i\in I_3|X_i,i\in I_1](x_i,i\in I_1)},\ x_{I_1}\in E,\\\\
\displaystyle 0,\ \ \ \ \ \ \ \ \ x_{I_1}\in E^c,\end{array}\right.\end{array}$$
where $E=\{x_{I_1}:P[X_i\in A_i,i\in I_3|X_i,i\in I_1](x_{I_1})\neq 0\}$ and $E^c=\mathbb{R}^{I_1}\setminus E$.\\
\item For fixed $A_i\in\mathscr{B}(\mathbb{R}),\ i\in I_3$,
$$\begin{array}{l}\displaystyle P[X_i,i\in I_2;X_{I_3}\in \prod_{i\in I_3}A_i,X_{I_4}\in \prod_{i\in I_4}A_i|X_i,i\in I_1](x_i,i\in I_1;A_i,i\in I_2)\\
\displaystyle=P[X_i,i\in I_2+I_3;X_i\in A_i,i\in I_4|X_i,i\in I_1](x_i,i\in I_1;A_i,i\in I_2;A_i,i\in I_3),
\end{array}$$
and
$$\begin{array}{l}\displaystyle P[X_i,i\in I_2;X_{I_4}\in \prod_{i\in I_4}A_i|X_i,i\in I_1;X_{I_3}\in \prod_{i\in I_3}A_i](x_i,i\in I_1;A_i,i\in I_2)\\\\
\displaystyle =\left\{\begin{array}{l}\displaystyle \frac{P[X_i,i\in I_2+I_3;X_i\in A_i,i\in I_4|X_i,i\in I_1](x_{I_1},\prod\limits_{i\in I_2+ I_3}A_i)}{P[X_i\in A_i,i\in I_3|X_i,i\in I_1](x_i,i\in I_1)},\ x_{I_1}\in E,\\\\
\displaystyle 0,\ \ \ \ \ \ \ \ \  x_{I_1}\in E^c,\end{array}\right.\end{array}$$
where $E=\{x_{I_1}:P[X_i\in A_i,i\in I_3|X_i,i\in I_1](x_{I_i})\neq 0\}$ and $E^c=\mathbb{R}^{I_1}\setminus E$.\\
\item For fixed $A_i\in\mathscr{F}_i,\ i\in I_3$,
$$\begin{array}{l}\displaystyle P[X_i,i\in I_2;X_{I_3}\in \prod_{i\in I_3}A_i|X_i,i\in I_1,X_{I_4}\in \prod_{i\in I_4}A_i](x_i,i\in I_1;A_i,i\in I_2)\\
\displaystyle=P[X_i,i\in I_2+I_3|X_i,i\in I_1;X_i\in A_i,i\in I_4](x_i,i\in I_1;A_i,i\in I_2;A_i,i\in I_3),
\end{array}$$
and
$$\begin{array}{l}\displaystyle P[X_i,i\in I_2|X_i,i\in I_1;X_{I_3}\in \prod_{i\in I_3}A_i,X_{I_4}\in \prod_{i\in I_4}A_i](x_i,i\in I_1;A_i,i\in I_2)\\\\
\displaystyle =\left\{\begin{array}{l}\displaystyle \frac{P[X_i,i\in I_2+I_3|X_i,i\in I_1;X_i\in A_i,i\in I_4](x_{I_1},\prod\limits_{i\in I_2+ I_3}A_i)}{P[X_i\in A_i,i\in I_3|X_i,i\in I_1;X_i\in A_i,i\in I_4](x_i,i\in I_1)},\ x_{I_1}\in E,\\\\
\displaystyle 0,\ \ \ \ \ \ \ \ \  x_{I_1}\in E^c,\end{array}\right.\end{array}$$
where $E=\{x_{I_1}:P[X_i\in A_i,i\in I_3|X_i,i\in I_1;X_i\in A_i,i\in I_4](x_{I_i})\neq 0\}$ and $E^c=\mathbb{R}^{I_1}\setminus E$.\\
\item For fixed $A_i\in\mathscr{F}_i,\ i\in I_3$,
$$\begin{array}{l}\displaystyle P[X_i,i\in I_2;X_{I_3}\in \prod_{i\in I_3}A_i,X_{I_4}\in \prod_{i\in I_4}A_i|X_i,i\in I_1;X_{I_5}\in \prod_{i\in I_5}A_i](x_i,i\in I_1;A_i,i\in I_2)\\
\displaystyle=P[X_i,i\in I_2+I_3;X_i\in A_i,i\in I_4|X_i,i\in I_1;X_{I_5}\in \prod_{i\in I_5}A_i](x_{I_1},\prod_{i\in I_2+I_3}A_i),
\end{array}$$
and
$$\begin{array}{l}\displaystyle P[X_i,i\in I_2;X_{I_4}\in \prod_{i\in I_4}A_i|X_i,i\in I_1;X_{I_3}\in \prod_{i\in I_3}A_i,X_{I_5}\in \prod_{i\in I_5}A_i](x_i,i\in I_1;A_i,i\in I_2)\\\\
\displaystyle =\left\{\begin{array}{l}\displaystyle \frac{P[X_i,i\in I_2+I_3;X_{I_4}\in \prod\limits_{i\in I_4}A_i|X_i,i\in I_1;X_{I_5}\in \prod\limits_{i\in I_5}A_i](x_{I_1},\prod\limits_{i\in I_2+I_3}A_i)}{P[X_i\in A_i,i\in I_3|X_i,i\in I_1;X_i\in A_i,i\in I_5](x_i,i\in I_1)},\ x_{I_1}\in E,\\\\
\displaystyle 0,\ \ \ \ \ \ \ \ \ x_{I_1}\in E^c,\end{array}\right.\end{array}$$
where $E=\{x_{I_1}:P[X_i\in A_i,i\in I_3|X_i,i\in I_1;X_i\in A_i,i\in I_5](x_{I_i})\neq 0\}$ and $E^c=\mathbb{R}^{I_1}\setminus E$.
\end{enumerate}
\end{col}

\begin{thm}\label{theorem3.4}Let $(\Omega,\mathscr{F},P)$ be a given probability space, $(\Omega_i,\mathscr{F}_i),\,i=1,2,3,4,5,6$ measurable spaces and $X_i:(\Omega,\mathscr{F})\rightarrow(\Omega_i,\mathscr{F}_i),\ i=1,2,3,4,5,6$ random objects. Given fixed $A_i\in\mathscr{F}_i,\ i=5,6$. If there exists $P[X_2,X_3,X_4;X_5\in A_5|X_1,X_6\in A_6](\omega_1;A_2,A_3,A_4)$, then so does $P[X_2;X_3\in A_3,X_5\in A_5|X_1;X_4\in A_4,X_6\in A_6](\omega_1,A_2)$ for fixed $A_i\in\mathscr{F}_i,\ i=3,4$, and
$$\begin{array}{l}\displaystyle P[X_2;X_3\in A_3,X_5\in A_5|X_1;X_4\in A_4,X_6\in A_6](\omega_1,A_2)\\
\displaystyle =\left\{\begin{array}{l}\displaystyle \frac{P[X_2,X_3,X_4;X_5\in A_5|X_1,X_6\in A_6](\omega_1;A_2,A_3,A_4)}{P[X_4\in A_4|X_1,X_6\in A_6](\omega_1)},\ \omega_1\in E,\\\\
\displaystyle 0,\ \ \ \ \ \ \ \ \ \omega_1\in E^c,\end{array}\right.\end{array}$$
where $E=\{\omega_1:P[X_4\in A_4|X_1,X_6\in A_6](\omega_1)\neq 0\}$ and $E^c=\Omega_1 \setminus E$.
\end{thm}

Then we have the following corollary.

\begin{col}\label{corollary3.4.1}Let $(\Omega,\mathscr{F},P)$ be a given probability space, $(\Omega_i,\mathscr{F}_i),\,i=1,2,3,4,5$ measurable spaces and $X_i:(\Omega,\mathscr{F})\rightarrow(\Omega_i,\mathscr{F}_i),\ i=1,2,3,4,5$ random objects. Given fixed $A_5\in\mathscr{F}_5$.
\begin{enumerate}
\item If there exists $P[X_2,X_3,X_4|X_1](\omega_1;A_2,A_3,A_4)$, then so does $P[X_2,X_3\in A_3|X_1,X_4\in A_4](\omega_1,A_2)$ for each fixed $A_3\in\mathscr{F}_3$ and each fixed $A_4\in\mathscr{F}_4$, and
$$\begin{array}{l}\displaystyle P[X_2,X_3\in A_3|X_1,X_4\in A_4](\omega_1,A_2)
\displaystyle =\left\{\begin{array}{l}\displaystyle \frac{P[X_2,X_3,X_4|X_1](\omega_1;A_2,A_3,A_4)}{P[X_4\in A_4|X_1](\omega_1)},\ \omega_1\in E,\\\\
\displaystyle 0,\ \ \ \ \ \ \ \ \ \omega_1\in E^c,\end{array}\right.\end{array}$$
where $E=\{\omega_1:P[X_4\in A_4|X_1](\omega_1)\neq 0\}$ and $E^c=\Omega_1\setminus E$.\\
\item If there exists $P[X_2,X_3,X_4;X_5\in A_5|X_1](\omega_1;A_2,A_3,A_4)$, then so does $P[X_2;X_3\in A_3,X_5\in A_5|X_1,X_4\in A_4](\omega_1,A_2)$ for each fixed $A_3\in\mathscr{F}_3$ and each fixed $A_4\in\mathscr{F}_4$, and
$$\begin{array}{l}\displaystyle P[X_2;X_3\in A_3,X_5\in A_5|X_1,X_4\in A_4](\omega_1,A_2)\\\\
\displaystyle =\left\{\begin{array}{l}\displaystyle \frac{P[X_2,X_3,X_4;X_5\in A_5|X_1](\omega_1;A_2,A_3,A_4)}{P[X_4\in A_4|X_1](\omega_1)},\ \omega_1\in E,\\\\
\displaystyle 0,\ \ \ \ \ \ \ \ \ \omega_1\in E^c,\end{array}\right.\end{array}$$
where $E=\{\omega_1:P[X_4\in A_4|X_1](\omega_1)\neq 0\}$ and $E^c=\Omega_1\setminus E$.\\
\item If there exists $P[X_2,X_3,X_4|X_1,X_5\in A_5](\omega_1;A_2,A_3,A_4)$, then so does $P[X_2,X_3\in A_3|X_1;X_4\in A_4,X_5\in A_5](\omega_1,A_2)$ for each fixed $A_3\in\mathscr{F}_3$ and each fixed $A_4\in\mathscr{F}_4$, and
$$\begin{array}{l}\displaystyle P[X_2,X_3\in A_3|X_1;X_4\in A_4,X_5\in A_5](\omega_1,A_2)\\\\
\displaystyle=\left\{\begin{array}{l}\displaystyle \frac{P[X_2,X_3,X_4|X_1,X_5\in A_5](\omega_1;A_2,A_3,A_4)}{P[X_4\in A_4|X_1,X_5\in A_5](\omega_1)},\ \omega_1\in E,\\
\displaystyle 0,\ \ \ \ \ \ \ \ \ \omega_1\in E^c,\end{array}\right.\end{array}$$
where $E=\{\omega_1:P[X_4\in A_4|X_1,X_5\in A_5](\omega_1)\neq 0\}$ and $E^c=\Omega_1\setminus E$.
\end{enumerate}
\end{col}

By Theorem \ref{theorem3.4} and Corollary \ref{corollary3.4.1} we have the following corollary.

\begin{col}\label{corollary3.4.2}Let $(\Omega,\mathscr{F},P)$ be a given probability space, $(\Omega_{i},\mathscr{F}_{i}),\ i\in \mathbb{N}$ measurable spaces. Let $X_i:(\Omega,\mathscr{F})\rightarrow(\Omega_i,\mathscr{F}_i),\ i\in \mathbb{N}$ be random objects. Given $I_i\subset \mathbb{N},\ i=1,2,3,4,5,6$ and $I_i\cap I_j=\emptyset,\ i\neq j$. And given fixed $A_i\in \mathscr{F}_i,\ i\in I_5+I_6$.
\begin{enumerate}
\item If there exists $P[X_i,i\in I_2+I_3+I_4|X_i,i\in I_1](\omega_i,i\in I_1;A_i,i\in I_2+I_3+I_4)$, then so does
$$P[X_i,i\in I_2;X_{I_3}\in \prod_{i\in I_3}A_i|X_i,i\in I_1;X_{I_4}\in \prod_{i\in I_4}A_i](\omega_i,i\in I_1;A_i,i\in I_2)$$
for fixed $A_i\in\mathscr{F}_i,\ i\in I_3+I_4$, and
$$\begin{array}{l}\displaystyle P[X_i,i\in I_2;X_{I_3}\in \prod_{i\in I_3}A_i|X_i,i\in I_1;X_{I_4}\in \prod_{i\in I_4}A_i](\omega_i,i\in I_1;A_i,i\in I_2)\\\\
\displaystyle =\left\{\begin{array}{l}\displaystyle \frac{P[X_i,i\in I_{234}|X_i,i\in I_1](\omega_{I_1},\prod\limits_{i\in I_{234}}A_i)}{P[X_i\in A_i,i\in I_4|X_i,i\in I_1](\omega_i,i\in I_1)},\ \omega_{I_1}\in E,\\\\
\displaystyle 0,\ \ \ \ \ \ \ \ \ \omega_{I_1}\in E^c,\end{array}\right.\end{array}$$
where $E=\{\omega_{I_1}:P[X_i\in A_i,i\in I_4|X_i,i\in I_1](\omega_{I_1})\neq 0\}$, $E^c=\Omega_{I_1}\setminus E$ and $I_{234}=I_2+I_3+I_4$.\\
\item If there exists $P[X_i,i\in I_2+I_3+I_4|X_i,i\in I_1;X_i\in A_i,i\in I_5](\omega_i,i\in I_1;A_i,i\in I_2+I_3+I_4)$, then so does
$$P[X_i,i\in I_2;X_{I_3}\in \prod_{i\in I_3}A_i|X_i,i\in I_1;X_{I_4}\in \prod_{i\in I_4}A_i,X_{I_5}\in \prod_{i\in I_5}A_i](\omega_i,i\in I_1;A_i,i\in I_2)$$
for fixed  $A_i\in\mathscr{F}_i,\ i\in I_3+I_4$, and
$$\begin{array}{l}\displaystyle P[X_i,i\in I_2;X_{I_3}\in \prod_{i\in I_3}A_i|X_i,i\in I_1;X_{I_4}\in \prod_{i\in I_4}A_i,X_{I_5}\in \prod_{i\in I_5}A_i](\omega_i,i\in I_1;A_i,i\in I_2)\\\\
\displaystyle =\left\{\begin{array}{l}\displaystyle \frac{P[X_i,i\in I_{234}|X_i,i\in I_1;X_i\in A_i,i\in I_5](\omega_{I_1},\prod\limits_{i\in I_{234}}A_i)}{P[X_i\in A_i,i\in I_4|X_i,i\in I_1;X_i\in A_i,i\in I_5](\omega_i,i\in I_1)},\ \omega_{I_1}\in E,\\\\
\displaystyle 0,\ \ \ \ \ \ \ \ \  \omega_{I_1}\in E^c,\end{array}\right.\end{array}$$
where $E=\{\omega_{I_1}:P[X_i\in A_i,i\in I_4|X_i,i\in I_1;X_i\in A_i,i\in I_5](\omega_{I_1})\neq 0\}$, $E^c=\Omega_{I_1}\setminus E$ and $I_{234}=I_2+ I_3+I_4$.\\
\item If there exists $P[X_i,i\in I_2+I_3+I_4;X_i\in A_i,i\in I_5|X_i,i\in I_1](\omega_i,i\in I_1;A_i,i\in I_2+I_3+I_4)$, then so does
$$P[X_i,i\in I_2;X_{I_3}\in \prod_{i\in I_3}A_i, X_{I_5}\in \prod_{i\in I_5}A_i|X_i,i\in I_1;X_{I_4}\in \prod_{i\in I_4}A_i](\omega_i,i\in I_1;A_i,i\in I_2)$$
for fixed $A_i\in\mathscr{F}_i,\ i\in I_3+I_4$, and
$$\begin{array}{l}\displaystyle P[X_i,i\in I_2;X_{I_3}\in \prod_{i\in I_3}A_i, X_{I_5}\in \prod_{i\in I_5}A_i|X_i,i\in I_1;X_{I_4}\in \prod_{i\in I_4}A_i](\omega_i,i\in I_1;A_i,i\in I_2)\\\\
\displaystyle =\left\{\begin{array}{l}\displaystyle \frac{P[X_i,i\in I_{234};X_{I_5}\in \prod\limits_{i\in I_5}A_i|X_i,i\in I_1](\omega_{I_1},\prod\limits_{i\in I_{234}}A_i)}{P[X_i\in A_i,i\in I_4|X_i,i\in I_1](\omega_i,i\in I_1)},\ \omega_{I_1}\in E,\\\\
\displaystyle 0,\ \ \ \ \ \ \ \ \ \omega_{I_1}\in E^c,\end{array}\right.\end{array}$$
where $E=\{\omega_{I_1}:P[X_i\in A_i,i\in I_4|X_i,i\in I_1](\omega_{I_1})\neq 0\}$, $E^c=\Omega_{I_1}\setminus E$ and $I_{234}=I_2+I_3+I_4$.\\
\item If there exists $P[X_i,i\in I_2+I_3+I_4;X_i\in A_i,i\in I_5|X_i,i\in I_1;X_i\in A_i,i\in I_6](\omega_i,i\in I_1;A_i,i\in I_2+I_3+I_4)$, then so does
$$\begin{array}{l}\displaystyle P[X_i,i\in I_2;X_{I_3}\in \prod_{i\in I_3}A_i, X_{I_5}\in \prod_{i\in I_5}A_i|X_i,i\in I_1;X_{I_4}\in \prod_{i\in I_4}A_i,X_{I_6}\in \prod_{i\in I_6}A_i]\\
\displaystyle (\omega_i,i\in I_1;A_i,i\in I_2)\end{array}$$
for fixed $A_i\in\mathscr{F}_i,\ i\in I_3+I_4$, and
$$\begin{array}{l}$$\begin{array}{l}\displaystyle P[X_i,i\in I_2;X_{I_3}\in \prod_{i\in I_3}A_i, X_{I_5}\in \prod_{i\in I_5}A_i|X_i,i\in I_1;X_{I_4}\in \prod_{i\in I_4}A_i,X_{I_6}\in \prod_{i\in I_6}A_i]\\
\displaystyle (\omega_i,i\in I_1;A_i,i\in I_2)\end{array}$$\\\\
\displaystyle =\left\{\begin{array}{l}\displaystyle \frac{P[X_i,i\in I_{234};X_{I_5}\in \prod\limits_{i\in I_5}A_i|X_i,i\in I_1;X_{I_6}\in \prod\limits_{i\in I_6}A_i](\omega_{I_1},\prod\limits_{i\in I_{234}}A_i)}{P[X_i\in A_i,i\in I_4|X_i,i\in I_1;X_{I_6}\in \prod\limits_{i\in I_6}A_i](\omega_i,i\in I_1)},\ \omega_{I_1}\in E,\\\\
\displaystyle 0,\ \ \ \ \ \ \ \ \ \omega_{I_1}\in E^c,\end{array}\right.\end{array}$$
where $E=\{\omega_{I_1}:P[X_i\in A_i,i\in I_4|X_i,i\in I_1;X_{I_6}\in \prod_{i\in I_6}A_i](\omega_{I_1})\neq 0\}$, $E^c=\Omega_{I_1}\setminus E$ and $I_{234}=I_2+I_3+I_4$.
\end{enumerate}
\end{col}

From Corollary \ref{corollary3.4.2} and Theorem \ref{theorem2.11.3} the corollary below follows.

\begin{col}\label{corollary3.4.3}  Let $(\Omega,\mathscr{F},P)$ be a given probability space. Let $I_i\subset \mathbb{N},\ i=1,2,3,4,5,6$ and $I_i\cap I_j=\emptyset,\ i\neq j$. Let $X_i,\ i\in \mathbb{N}$ be random variables on $(\Omega,\mathscr{F},P)$. Given $A_i\in \mathscr{B}(\mathbb{R}),\ i\in I_5+I_6$.
\begin{enumerate}
\item For fixed $A_i\in\mathscr{B}(\mathbb{R}),\ i\in I_3+I_4$,
$$\begin{array}{l}\displaystyle P[X_i,i\in I_2;X_{I_3}\in \prod_{i\in I_3}A_i|X_i,i\in I_1;X_{I_4}\in \prod_{i\in I_4}A_i](x_i,i\in I_1;A_i,i\in I_2)\\\\
\displaystyle =\left\{\begin{array}{l}\displaystyle \frac{P[X_i,i\in I_{234}|X_i,i\in I_1](x_{I_1},\prod\limits_{i\in I_{234}}A_i)}{P[X_i\in A_i,i\in I_4|X_i,i\in I_1](x_i,i\in I_1)},\ x_{I_1}\in E,\\\\
\displaystyle 0,\ \ \ \ \ \ \ \ \ x_{I_1}\in E^c,\end{array}\right.\end{array}$$
where $E=\{x_{I_1}:P[X_i\in A_i,i\in I_4|X_i,i\in I_1](x_{I_1})\neq 0\}$, $E^c=\mathbb{R}^{I_1}\setminus E$ and $I_{234}=I_2+I_3+I_4$.
\item For fixed  $A_i\in\mathscr{B}(\mathbb{R}),\ i\in I_3+I_4$,
$$\begin{array}{l}\displaystyle P[X_i,i\in I_2;X_{I_3}\in \prod_{i\in I_3}A_i|X_i,i\in I_1;X_{I_4}\in \prod_{i\in I_4}A_i,X_{I_5}\in \prod_{i\in I_5}A_i](x_i,i\in I_1;A_i,i\in I_2)\\\\
\displaystyle =\left\{\begin{array}{l}\displaystyle \frac{P[X_i,i\in I_{234}|X_i,i\in I_1;X_i\in A_i,i\in I_5](x_{I_1},\prod\limits_{i\in I_{234}}A_i)}{P[X_i\in A_i,i\in I_4|X_i,i\in I_1;X_i\in A_i,i\in I_5](x_i,i\in I_1)},\ x_{I_1}\in E,\\\\
\displaystyle 0,\ \ \ \ \ \ \ \ \ x_{I_1}\in E^c,\end{array}\right.\end{array}$$
where $E=\{x_{I_1}:P[X_i\in A_i,i\in I_4|X_i,i\in I_1;X_i\in A_i,i\in I_5](x_{I_1})\neq 0\}$, $E^c=\mathbb{R}^{I_1}\setminus E$ and $I_{234}=I_2+I_3+I_4$.
\item For fixed $A_i\in\mathscr{B}(\mathbb{R}),\ i\in I_3+I_4$,
$$\begin{array}{l}\displaystyle P[X_i,i\in I_2;X_{I_3}\in \prod_{i\in I_3}A_i, X_{I_5}\in \prod_{i\in I_5}A_i|X_i,i\in I_1;X_{I_4}\in \prod_{i\in I_4}A_i](x_i,i\in I_1;A_i,i\in I_2)\\\\
\displaystyle =\left\{\begin{array}{l}\displaystyle \frac{P[X_i,i\in I_{234};X_{I_5}\in \prod\limits_{i\in I_5}A_i|X_i,i\in I_1](x_{I_1},\prod\limits_{i\in I_{234}}A_i)}{P[X_i\in A_i,i\in I_4|X_i,i\in I_1](x_i,i\in I_1)},\ x_{I_1}\in E,\\\\
\displaystyle 0,\ \ \ \ \ \ \ \ \  x_{I_1}\in E^c,\end{array}\right.\end{array}$$
where $E=\{x_{I_1}:P[X_i\in A_i,i\in I_4|X_i,i\in I_1](x_{I_1})\neq 0\}$, $E^c=\mathbb{R}^{I_1}\setminus E$ and $I_{234}=I_2+I_3+I_4$.
\item For fixed $A_i\in\mathscr{B}(\mathbb{R}),\ i\in I_3+I_4$,
$$\begin{array}{l}$$\begin{array}{l}\displaystyle P[X_i,i\in I_2;X_{I_3}\in \prod_{i\in I_3}A_i, X_{I_5}\in \prod_{i\in I_5}A_i|X_i,i\in I_1;X_{I_4}\in \prod_{i\in I_4}A_i,X_{I_6}\in \prod_{i\in I_6}A_i]\\
\displaystyle (x_i,i\in I_1;A_i,i\in I_2)\end{array}$$\\\\
\displaystyle =\left\{\begin{array}{l}\displaystyle \frac{P[X_i,i\in I_{234};X_{I_5}\in \prod\limits_{i\in I_5}A_i|X_i,i\in I_1;X_{I_6}\in \prod\limits_{i\in I_6}A_i](x_{I_1},\prod\limits_{i\in I_{234}}A_i)}{P[X_i\in A_i,i\in I_4|X_i,i\in I_1;X_{I_6}\in \prod\limits_{i\in I_6}A_i](x_i,i\in I_1)},\ \omega_{I_1}\in E,\\\\
\displaystyle 0,\ \ \ \ \ \ \ \ \  x_{I_1}\in E^c,\end{array}\right.\end{array}$$
where $E=\{x_{I_1}:P[X_i\in A_i,i\in I_4|X_i,i\in I_1;X_{I_6}\in \prod_{i\in I_6}A_i](x_{I_1})\neq 0\}$, $E^c=\mathbb{R}^{I_1}\setminus E$ and $I_{234}=I_2+ I_3+ I_4$.
\end{enumerate}
\end{col}

\begin{thm}\label{theorem3.5}Let $(\Omega,\mathscr{F},P)$ be a given probability space, $(\Omega_{i},\mathscr{F}_{i}),\ i\in \mathbb{N}$ measurable spaces. Let $X_i:(\Omega,\mathscr{F})\rightarrow(\Omega_i,\mathscr{F}_i),\ i\in \mathbb{N}$ be random objects. Given $I_i\subset \mathbb{N},\ i=1,2,3,4$ and $I_i\cap I_j=\emptyset,\ i\neq j$. And given fixed $A_i\in \mathscr{F}_i,\ i\in I_3+I_4$, if there exists $P[X_i,i\in I_2;X_i\in A_i,i\in I_3|X_i,i\in I_1;X_i\in A_i,i\in I_4](\omega_i,i\in I_1;A_{I_2})$, then $P[X_i,i\in I_1+I_2;X_i\in A_i,i\in I_3+I_4]$ is also the measure induced by $P[X_i,i\in I_2;X_i\in A_i,i\in I_3|X_i,i\in I_1;X_i\in A_i,i\in I_4](\omega_i,i\in I_1;A_{I_2})$ and $P[X_i,i\in I_1;X_i\in A_i,i\in I_4]$, and
$$\begin{array}{l}\displaystyle P[X_i,i\in I_1+I_2;X_i\in A_i,i\in I_3+I_4](A_{I_1}\times A_{I_2})\\\\
\displaystyle =\int_{A_{I_1}}P[X_i,i\in I_2;X_i\in A_i,i\in I_3|X_i,i\in I_1;X_i\in A_i,i\in I_4](\omega_i,i\in I_1;A_{I_2})\\
\displaystyle \cdot P[X_i,i\in I_1;X_i\in A_i,i\in I_4](d(\omega_i,i\in I_1)),\ \forall A_{I_j}\in\mathscr{F}_{I_j},j=1,2.\\
\end{array}$$
Especially,
$$\begin{array}{l}\displaystyle \int_{\Omega_{I_1}}P[X_i,i\in I_2;X_i\in A_i,i\in I_3|X_i,i\in I_1;X_i\in A_i,i\in I_4](\omega_i,i\in I_1;A_{I_2})\\
\displaystyle \cdot P[X_i,i\in I_1;X_i\in A_i,i\in I_4](d(\omega_i,i\in I_1))\\\\
\displaystyle =P[X_i,i\in I_2;X_i\in A_i,i\in I_3+I_4](A_{I_2}),\ \forall A_{I_2}\in\mathscr{F}_{I_2},
\end{array}$$
which is the probability of co-occurrence of $X_i,i\in I_2$ and $X_i\in A_i,\ i\in I_3+I_4$ on $(\Omega_{I_2},\mathscr{F}_{I_2})$; And
$$\begin{array}{l}\displaystyle \int_{A_{I_1}}P[X_i,i\in I_2;X_i\in A_i,i\in I_3|X_i,i\in I_1;X_i\in A_i,i\in I_4](\omega_i,i\in I_1;\Omega_{I_2})\\
\displaystyle \cdot P[X_i,i\in I_1;X_i\in A_i,i\in I_4](d(\omega_i,i\in I_1))\\\\
\displaystyle =P[X_i,i\in I_1;X_i\in A_i,i\in I_3+I_4](A_{I_1}),\ \forall A_{I_1}\in\mathscr{F}_{I_1},
\end{array}$$
which is the probability of co-occurrence of $X_i,i\in I_1$ and $X_i\in A_i,\ i\in I_3+I_4$ on $(\Omega_{I_1},\mathscr{F}_{I_1})$.
\end{thm}

From Theorem \ref{theorem3.5} and Theorem \ref{theorem2.11.3} the corollary below follows.

\begin{col}\label{corollary3.5.1}  Let $(\Omega,\mathscr{F},P)$ be a given probability space. Let $X_i,\ i\in \mathbb{N}$ be random variables on $(\Omega,\mathscr{F},P)$. Let $I_i\subset \mathbb{N},\ i=1,2,3,4$ and $I_i\cap I_j=\emptyset,\ i\neq j$. Given fixed $A_i\in \mathscr{B}(\mathbb{R}),\ i\in I_3+I_4 $,
then $P[X_i,i\in I_1+I_2;X_i\in A_i,i\in I_3+I_4]$ is also the measure induced by $P[X_i,i\in I_2;X_i\in A_i,i\in I_3|X_i,i\in I_1;X_i\in A_i,i\in I_4](x_i,i\in I_1;A_{I_2})$ and $P[X_i,i\in I_1;X_i\in A_i,i\in I_4]$, and
$$\begin{array}{l}\displaystyle P[X_i,i\in I_1+I_2;X_i\in A_i,i\in I_3+I_4](A_{I_1}\times A_{I_2})\\\\
\displaystyle =\int_{A_{I_1}}P[X_i,i\in I_2;X_i\in A_i,i\in I_3|X_i,i\in I_1;X_i\in A_i,i\in I_4](x_i,i\in I_1;A_{I_2})\\
\displaystyle \cdot P[X_i,i\in I_1;X_i\in A_i,i\in I_4](d(x_i,i\in I_1)),\ \forall A_{I_j}\in\mathscr{B}(\mathbb{R}^{I_j}),j=1,2.\\
\end{array}$$
Especially,
$$\begin{array}{l}\displaystyle \int_{\mathbb{R}^{I_1}}P[X_i,i\in I_2;X_i\in A_i,i\in I_3|X_i,i\in I_1;X_i\in A_i,i\in I_4](x_i,i\in I_1;A_{I_2})\\
\displaystyle \cdot P[X_i,i\in I_1;X_i\in A_i,i\in I_4](d(x_i,i\in I_1))\\\\
\displaystyle =P[X_i,i\in I_2;X_i\in A_i,i\in I_3+I_4](A_{I_2}),\ \forall A_{I_2}\in\mathscr{B}(\mathbb{R}^{I_2}),
\end{array}$$
which is the probability of co-occurrence of $X_i,i\in I_2$ and $X_i\in A_i,\ i\in I_3+I_4$ on $(\mathbb{R}^{I_2},\mathscr{B}(\mathbb{R}^{I_2}))$; And
$$\begin{array}{l}\displaystyle \int_{A_{I_1}}P[X_i,i\in I_2;X_i\in A_i,i\in I_3|X_i,i\in I_1;X_i\in A_i,i\in I_4](x_i,i\in I_1;\mathbb{R}^{I_2})\\
\displaystyle \cdot P[X_i,i\in I_1;X_i\in A_i,i\in I_4](d(x_i,i\in I_1))\\\\
\displaystyle =P[X_i,i\in I_1;X_i\in A_i,i\in I_3+I_4](A_{I_1}),\ \forall A_{I_1}\in\mathscr{B}(\mathbb{R}^{I_1}),
\end{array}$$
which is the probability of co-occurrence of $X_i,i\in I_1$ and $X_i\in A_i,\ i\in I_3+I_4$ on $(\mathbb{R}^{I_1},\mathscr{B}(\mathbb{R}^{I_1}))$.
\end{col}

\begin{thm}\label{theorem3.6}Let $(\Omega,\mathscr{F},P)$ be a given probability space, $(\Omega_i,\mathscr{F}_i),\,i=1,2,3,4$ measurable spaces and $X_i:(\Omega,\mathscr{F})\rightarrow(\Omega_i,\mathscr{F}_i),\ i=1,2,3,4$ random objects. Given fixed $A_i\in\mathscr{F}_i,\ i=3,4$. If there exists $P[X_2|X_1,X_4\in A_4](\omega_1,A_2)$ $(or\  P[X_2,X_4\in A_4|X_1](\omega_1,A_2))$, then for fixed $A_2\in\mathscr{F}_2$,
\begin{equation}\label{eqn3.6.1}\begin{array}{l}\displaystyle P[X_2\in A_2,X_3\in A_3|X_1,X_4\in A_4](\omega_1)\\\\
\displaystyle =\int_{A_2}P[X_3\in A_3|X_1,X_2;X_4\in A_4](\omega_1,\omega_2)P[X_2|X_1,X_4\in A_4](\omega_1,d\omega_2),\\
\ \ \ a.e.\{P[X_1,X_4\in A_4]\},\end{array}\end{equation}
provided that $P[X_1,X_4\in A_4]>0$; or
\begin{equation}\label{eqn3.6.2}\begin{array}{l}\displaystyle P[X_i\in A_i,i=2,3,4|X_1](\omega_1)\\\\
\displaystyle =\int_{A_2}P[X_3\in A_3|X_1,X_2;X_4\in A_4](\omega_1,\omega_2)P[X_2,X_4\in A_4|X_1](\omega_1,d\omega_2),\\
\ \ \  a.e.\{P[X_1]\}.\end{array}\end{equation}
\end{thm}

We have the following corollary.

\begin{col}\label{corollary3.6.1}Let $(\Omega,\mathscr{F},P)$ be a given probability space, $(\Omega_{i},\mathscr{F}_{i}),\ i\in \mathbb{N}$ measurable spaces. Let $X_i:(\Omega,\mathscr{F})\rightarrow(\Omega_i,\mathscr{F}_i),\ i\in \mathbb{N}$ be random objects. Given $I_i\subset \mathbb{N},\ i=1,2,3,4$ and $I_i\cap I_j=\emptyset,\ i\neq j$. And given fixed $A_i\in \mathscr{F}_i,\ i\in I_3+I_4$. If there exists $P[X_i,i\in I_2|X_i,i\in I_1;X_i\in A_i,i\in I_4](\omega_i,i\in I_1;A_i,i\in I_2)$ $(or\ P[X_i,i\in I_2;X_i\in A_i,i\in I_4|X_i,i\in I_1](\omega_i,i\in I_1;A_i,i\in I_2))$, then for fixed $A_i\in\mathscr{F}_i,i\in I_2$, provided that $P[X_i,i\in I_1;X_i\in A_i,i\in I_4]>0$,
$$\begin{array}{l}\displaystyle P[X_i\in A_i,i\in I_2+I_3|X_i,i\in I_1;X_i\in A_i,i\in I_4](\omega_i,i\in I_1)\\\\
\displaystyle =\int_{\prod_{i\in I_2}A_i}P[X_i\in A_i,i\in I_3|X_i,i\in I_1+I_2;X_i\in A_i,i\in I_4](\omega_i,i\in I_1;\omega_i,i\in I_2)\\
\displaystyle \ \ \ \cdot P[X_i,i\in I_2|X_i,i\in I_1;X_i\in A_i,i\in I_4](\omega_i,i\in I_1;d(\omega_i,i\in I_2)),\\
\quad a.e.\{P[X_i,i\in I_1;X_i\in A_i,i\in I_4]\};\end{array}$$
or
$$\begin{array}{l}\displaystyle P[X_i\in A_i,i\in I_2+I_3+I_4|X_i,i\in I_1](\omega_i,i\in I_1)\\\\
\displaystyle =\int_{\prod_{i\in I_2}A_i}P[X_i\in A_i,i\in I_3|X_i,i\in I_1+I_2;X_i\in A_i,i\in I_4](\omega_i,i\in I_1;\omega_i,i\in I_2)\\
\displaystyle \ \ \ \cdot P[X_i,i\in I_2;X_i\in A_i,i\in I_4|X_i,i\in I_1](\omega_i,i\in I_1;d(\omega_i,i\in I_2)), \quad a.e.\{P[X_i,i\in I_1]\}.
\end{array}$$
\end{col}

From Corollary \ref{corollary3.6.1} and Theorem \ref{theorem2.11.3} it follows that

\begin{col}\label{corollary3.6.2}Let $(\Omega,\mathscr{F},P)$ be a given probability space. Let $X_i,\ i\in \mathbb{N}$ be random variables on $(\Omega,\mathscr{F},P)$. Let $I_i\subset \mathbb{N},\ i=1,2,3,4$ and $I_i\cap I_j=\emptyset,\ i\neq j$. Given fixed $A_i\in \mathscr{B}(\mathbb{R}),\ i\in I_3+I_4 $, then for fixed $A_i\in\mathscr{B}(\mathbb{R}),i\in I_2$,
$$\begin{array}{l}\displaystyle P[X_i\in A_i,i\in I_2+I_3|X_i,i\in I_1;X_i\in A_i,i\in I_4](x_i,i\in I_1)\\\\
\displaystyle =\int_{\prod_{i\in I_2}A_i}P[X_i\in A_i,i\in I_3|X_i,i\in I_1+I_2;X_i\in A_i,i\in I_4](x_i,i\in I_1;x_i,i\in I_2)\\
\displaystyle \ \ \ \cdot P[X_i,i\in I_2|X_i,i\in I_1;X_i\in A_i,i\in I_4](x_i,i\in I_1;d(x_i,i\in I_2)),\\
\quad  a.e.\{P[X_i,i\in I_1;X_i\in A_i,i\in I_4]\},\end{array}$$
provided that $P[X_i,i\in I_1;X_i\in A_i,i\in I_4]>0$; or
$$\begin{array}{l}\displaystyle P[X_i\in A_i,i\in I_2+I_3+I_4|X_i,i\in I_1](x_i,i\in I_1)\\\\
\displaystyle =\int_{\prod_{i\in I_2}A_i}P[X_i\in A_i,i\in I_3|X_i,i\in I_1+I_2;X_i\in A_i,i\in I_4](x_i,i\in I_1;x_i,i\in I_2)\\
\displaystyle \ \ \ \cdot P[X_i,i\in I_2;X_i\in A_i,i\in I_4|X_i,i\in I_1](x_i,i\in I_1;d(x_i,i\in I_2)), \quad a.e.\{P[X_i,i\in I_1]\}.\end{array}$$
\end{col}

\begin{thm}\label{theorem3.7}Let $(\Omega,\mathscr{F},P)$ be a given probability space, $(\Omega_i,\mathscr{F}_i),\,i=1,2,\cdots,6$ measurable spaces and $X_i:(\Omega,\mathscr{F})\rightarrow(\Omega_i,\mathscr{F}_i),\ i=1,2,\cdots,6$ random objects. Given fixed $A_i\in\mathscr{F}_i,\ i=4,5,6$. If there exist $P[X_3,X_4\in A_4|X_1,X_2;X_5\in A_5,X_6\in A_6](\omega_1,\omega_2;A_3)$ and $P[X_2,X_6\in A_6|X_1,X_5\in A_5](\omega_1,A_2)$, then so does $P[X_2,X_3;X_4\in A_4,X_6\in A_6|X_1,X_5\in A_5](\omega_1;A_2,A_3)$, and  for $\forall A_i\in\mathscr{F}_i,i=2,3$,
\begin{equation}\label{eqn3.7.1}\begin{array}{l}\displaystyle P[X_2,X_3;X_4\in A_4,X_6\in A_6|X_1,X_5\in A_5](\omega_1;A_2,A_3)\\\\
\displaystyle =\int_{A_2}P[X_3,X_4\in A_4|X_1,X_2;X_5\in A_5,X_6\in A_6](\omega_1,\omega_2;A_3)\\
\displaystyle \ \ \ \cdot P[X_2,X_6\in A_6|X_1,X_5\in A_5](\omega_1,d\omega_2),\ \ a.e.\{P[X_1,X_5\in A_5]\}.\end{array}\end{equation}
\end{thm}

By Theorem \ref{theorem3.7} and Theorem \ref{theorem3.3} we have the following corollary.

\begin{col}\label{corollary3.7.1}Let $(\Omega,\mathscr{F},P)$ be a given probability space, $(\Omega_i,\mathscr{F}_i),\,i=1,2,\cdots,6$ measurable spaces and $X_i:(\Omega,\mathscr{F})\rightarrow(\Omega_i,\mathscr{F}_i),\ i=1,2,\cdots,6$ random objects. Given fixed $A_i\in\mathscr{F}_i,\ i=4,5,6$. If there exist $P[X_3,X_4\in A_4|X_1,X_2;X_5\in A_5,X_6\in A_6](\omega_1,\omega_2;A_3)$ and $P[X_2,X_6\in A_6|X_1,X_5\in A_5](\omega_1,A_2)$, then so do $P[X_3;X_2\in A_2,X_4\in A_4,X_6\in A_6|X_1,X_5\in A_5](\omega_1,A_3)$ for each fixed $A_2\in\mathscr{F}_2$ and $P[X_2;X_3\in A_3,X_4\in A_4,X_6\in A_6|X_1,X_5\in A_5](\omega_1,A_2)$ for each fixed $A_3\in\mathscr{F}_3$. For each fixed  $A_2\in\mathscr{F}_2$,
$$\begin{array}{l}\displaystyle P[X_3;X_2\in A_2,X_4\in A_4,X_6\in A_6|X_1,X_5\in A_5](\omega_1,A_3)\\\\
\displaystyle =\int_{A_2}P[X_3,X_4\in A_4|X_1,X_2;X_5\in A_5,X_6\in A_6](\omega_1,\omega_2;A_3)\\
\displaystyle \ \ \ \cdot P[X_2,X_6\in A_6|X_1,X_5\in A_5](\omega_1,d\omega_2),\ \ a.e.\{P[X_1,X_5\in A_5]\}.\end{array}$$
For each fixed $A_3\in\mathscr{F}_3$,
$$\begin{array}{l}\displaystyle P[X_2;X_3\in A_3,X_4\in A_4,X_6\in A_6|X_1,X_5\in A_5](\omega_1,A_2)\\\\
\displaystyle =\int_{A_2}P[X_3,X_4\in A_4|X_1,X_2;X_5\in A_5,X_6\in A_6](\omega_1,\omega_2;A_3)\\
\displaystyle \ \ \ \cdot P[X_2,X_6\in A_6|X_1,X_5\in A_5](\omega_1,d\omega_2),\ \ a.e.\{P[X_1,X_5\in A_5]\}.\end{array}$$
\end{col}

The following corollary comes from Corollary \ref{corollary3.7.1}.

\begin{col}\label{corollary3.7.2}Let $(\Omega,\mathscr{F},P)$ be a given probability space, $(\Omega_i,\mathscr{F}_i),\,i=1,2,\cdots,6$ measurable spaces and $X_i:(\Omega,\mathscr{F})\rightarrow(\Omega_i,\mathscr{F}_i),\ i=1,2,\cdots,6$ random objects. Given fixed $A_i\in\mathscr{F}_i,\ i=4,5$.
\begin{enumerate}
\item If there exist $P[X_3|X_1,X_2](\omega_1,\omega_2;A_3)$ and $P[X_2|X_1](\omega_1,A_2)$,  then so do $P[X_3,X_2\in A_2|X_1](\omega_1,A_3)$ for each fixed $A_2\in\mathscr{F}_2$ and $P[X_2,X_3\in A_3|X_1](\omega_1,A_2)$ for each fixed $A_3\in\mathscr{F}_3$. For each fixed  $A_2\in\mathscr{F}_2$,
$$\begin{array}{l}\displaystyle P[X_3,X_2\in A_2|X_1](\omega_1,A_3)\\\\
\displaystyle =\int_{A_2}P[X_3|X_1,X_2](\omega_1,\omega_2;A_3)P[X_2|X_1](\omega_1,d\omega_2),\ \ a.e.\{P[X_1]\}.\end{array}$$
For each fixed $A_3\in\mathscr{F}_3$,
$$\begin{array}{l}\displaystyle P[X_2,X_3\in A_3|X_1](\omega_1,A_2)\\\\
\displaystyle =\int_{A_2}P[X_3|X_1,X_2](\omega_1,\omega_2;A_3)P[X_2|X_1](\omega_1,d\omega_2),\ \ a.e.\{P[X_1]\}.\end{array}$$
\item If there exist $P[X_3,X_4\in A_4|X_1,X_2](\omega_1,\omega_2;A_3)$ and $P[X_2|X_1](\omega_1,A_2)$, then so do $P[X_3;X_2\in A_2,X_4\in A_4|X_1](\omega_1,A_3)$ for each fixed $A_2\in\mathscr{F}_2$ and $P[X_2;X_3\in A_3,X_4\in A_4|X_1](\omega_1,A_2)$ for each fixed $A_3\in\mathscr{F}_3$. For each fixed  $A_2\in\mathscr{F}_2$,
$$\begin{array}{l}\displaystyle P[X_3;X_2\in A_2,X_4\in A_4|X_1](\omega_1,A_3)\\\\
\displaystyle =\int_{A_2}P[X_3,X_4\in A_4|X_1,X_2](\omega_1,\omega_2;A_3)P[X_2|X_1](\omega_1,d\omega_2),\ \ a.e.\{P[X_1]\}.\end{array}$$
For each fixed $A_3\in\mathscr{F}_3$,
$$\begin{array}{l}\displaystyle P[X_2;X_3\in A_3,X_4\in A_4|X_1](\omega_1,A_2)\\\\
\displaystyle =\int_{A_2}P[X_3,X_4\in A_4|X_1,X_2](\omega_1,\omega_2;A_3)P[X_2|X_1](\omega_1,d\omega_2),\ \ a.e.\{P[X_1]\}.\end{array}$$
\item If there exist $P[X_3|X_1,X_2;X_5\in A_5](\omega_1,\omega_2;A_3)$ and $P[X_2|X_1,X_5\in A_5](\omega_1,A_2)$, then so do $P[X_3,X_2\in A_2|X_1,X_5\in A_5](\omega_1,A_3)$ for each fixed $A_2\in\mathscr{F}_2$ and $P[X_2,X_3\in A_3|X_1,X_5\in A_5](\omega_1,A_2)$ for each fixed $A_3\in\mathscr{F}_3$. For each fixed  $A_2\in\mathscr{F}_2$,
$$\begin{array}{l}\displaystyle P[X_3,X_2\in A_2|X_1,X_5\in A_5](\omega_1,A_3)\\\\
\displaystyle =\int_{A_2}P[X_3|X_1,X_2;X_5\in A_5](\omega_1,\omega_2;A_3)\\
\displaystyle\ \ \ \cdot P[X_2|X_1,X_5\in A_5](\omega_1,d\omega_2),\quad \ a.e.\{P[X_1,X_5\in A_5]\}.\end{array}$$
For each fixed $A_3\in\mathscr{F}_3$,
$$\begin{array}{l}\displaystyle P[X_2,X_3\in A_3|X_1,X_5\in A_5](\omega_1,A_2)\\\\
\displaystyle =\int_{A_2}P[X_3|X_1,X_2;X_5\in A_5](\omega_1,\omega_2;A_3)\\
\displaystyle\ \ \ \cdot P[X_2|X_1,X_5\in A_5](\omega_1,d\omega_2), \quad   a.e.\{P[X_1,X_5\in A_5]\}.\end{array}$$
\item If there exist $P[X_3|X_1,X_2;X_5\in A_5,X_6\in A_6](\omega_1,\omega_2;A_3)$ and $P[X_2,X_6\in A_6|X_1,X_5\in A_5](\omega_1,A_2)$, then so do $P[X_3,X_2\in A_2,X_6\in A_6|X_1,X_5\in A_5](\omega_1,A_3)$ for each fixed $A_2\in\mathscr{F}_2$ and $P[X_2,X_3\in A_3,X_6\in A_6|X_1,X_5\in A_5](\omega_1,A_2)$ for each fixed $A_3\in\mathscr{F}_3$. For each fixed  $A_2\in\mathscr{F}_2$,
$$\begin{array}{l}\displaystyle P[X_3,X_2\in A_2,X_6\in A_6|X_1,X_5\in A_5](\omega_1,A_3)\\\\
\displaystyle =\int_{A_2}P[X_3|X_1,X_2;X_5\in A_5,X_6\in A_6](\omega_1,\omega_2;A_3)\\
\displaystyle\ \ \ \cdot P[X_2,X_6\in A_6|X_1,X_5\in A_5](\omega_1,d\omega_2),\quad a.e.\{P[X_1,X_5\in A_5]\}.\end{array}$$
For each fixed $A_3\in\mathscr{F}_3$,
$$\begin{array}{l}\displaystyle P[X_2,X_3\in A_3,X_6\in A_6|X_1,X_5\in A_5](\omega_1,A_2)\\\\
\displaystyle =\int_{A_2}P[X_3|X_1,X_2;X_5\in A_5,X_6\in A_6](\omega_1,\omega_2;A_3)\\
\displaystyle \ \ \ \cdot P[X_2,X_6\in A_6|X_1,X_5\in A_5](\omega_1,d\omega_2),\quad a.e.\{P[X_1,X_5\in A_5]\}.\end{array}$$
\end{enumerate}
\end{col}

By Theorem \ref{theorem3.7}, Corollary \ref{corollary3.7.1} and Corollary \ref{corollary3.7.2}, we obtain the following corollary.

\begin{col}\label{corollary3.7.3}Let $\Lambda$ be a nonempty index set. Let $(\Omega,\mathscr{F},P)$ be a given probability space, $(\Omega_{i},\mathscr{F}_{i}),\ i\in \Lambda$ measurable spaces. Let $X_i:(\Omega,\mathscr{F})\rightarrow(\Omega_i,\mathscr{F}_i),\ i\in \Lambda$ be random objects. Given $I_i\subset \Lambda,\ i=1,2,\cdots,6$ and $I_i\cap I_j=\emptyset,\ i\neq j$. Let $I_{12}=I_1+I_2$ and fixed $A_{I_i}\in \mathscr{F}_{I_i},\ i=4,5,6$.
\begin{enumerate}
\item If there exist $P[X_{I_3},X_{I_4}\in A_{I_4}|X_{I_{12}};X_{I_5}\in A_{I_5},X_{I_6}\in A_{I_6}](\omega_{I_1},\omega_{I_2};A_{I_3})$ and $P[X_{I_2},X_{I_6}\in A_{I_6}|X_{I_1},X_{I_5}\in A_{I_5}](\omega_{I_1},A_{I_2})$, then so does $P[X_{I_2},X_{I_3};X_{I_4}\in A_{I_4},X_{I_6}\in A_{I_6}|X_{I_1},X_{I_5}\in A_{I_5}](\omega_{I_1};A_{I_2},A_{I_3})$, and  for $\forall A_{I_i}\in\mathscr{F}_{I_i},i=2,3$,
$$\begin{array}{l}\displaystyle P[X_{I_2},X_{I_3};X_{I_4}\in A_{I_4},X_{I_6}\in A_{I_6}|X_{I_1},X_{I_5}\in A_{I_5}](\omega_{I_1};A_{I_2},A_{I_3})\\\\
\displaystyle =\int_{A_{I_2}}P[X_{I_3},X_{I_4}\in A_{I_4}|X_{I_1},X_{I_2};X_{I_5}\in A_{I_5},X_{I_6}\in A_{I_6}](\omega_{I_1},\omega_{I_2};A_{I_3})\\
\displaystyle \ \ \ \cdot P[X_{I_2},X_{I_6}\in A_{I_6}|X_{I_1},X_{I_5}\in A_{I_5}](\omega_{I_1},d\omega_{I_2}),\quad a.e.\{P[X_{I_1},X_{I_5}\in A_{I_5}]\}.\end{array}$$
\item If there exist $P[X_{I_3},X_{I_4}\in A_{I_4}|X_{I_{12}};X_{I_5}\in A_{I_5},X_{I_6}\in A_{I_6}](\omega_{I_1},\omega_{I_2};A_{I_3})$ and $P[X_{I_2},X_{I_6}\in A_{I_6}|X_{I_1},X_{I_5}\in A_{I_5}](\omega_{I_1},A_{I_2})$, then so do $P[X_{I_3};X_{I_2}\in A_{I_2},X_{I_4}\in A_{I_4},X_{I_6}\in A_{I_6}|X_{I_1},X_{I_5}\in A_{I_5}](\omega_{I_1},A_{I_3})$ for each fixed $A_{I_2}\in\mathscr{F}_{I_2}$ and $P[X_{I_2};X_{I_3}\in A_{I_3},X_{I_4}\in A_{I_4},X_{I_6}\in A_{I_6}|X_{I_1},X_{I_5}\in A_{I_5}](\omega_{I_1},A_{I_2})$ for each fixed $A_{I_3}\in\mathscr{F}_{I_3}$. For each fixed $A_{I_2}\in\mathscr{F}_{I_2}$,
$$\begin{array}{l}\displaystyle P[X_{I_3};X_{I_2}\in A_{I_2},X_{I_4}\in A_{I_4},X_{I_6}\in A_{I_6}|X_{I_1},X_{I_5}\in A_{I_5}](\omega_{I_1},A_{I_3})\\\\
\displaystyle =\int_{A_{I_2}}P[X_{I_3},X_{I_4}\in A_{I_4}|X_{I_{12}},X_{I_5}\in A_{I_5},X_{I_6}\in A_{I_6}](\omega_{I_1},\omega_{I_2};A_{I_3})\\
\displaystyle \ \ \cdot P[X_{I_2},X_{I_6}\in A_{I_6}|X_{I_1},X_{I_5}\in A_{I_5}](\omega_{I_1},d\omega_{I_2}),\quad a.e.\{P[X_{I_1},X_{I_5}\in A_{I_5}]\}.\end{array}$$
For each fixed $A_{I_3}\in\mathscr{F}_{I_3}$,
$$\begin{array}{l}\displaystyle P[X_{I_2};X_{I_3}\in A_{I_3},X_{I_4}\in A_{I_4},X_{I_6}\in A_{I_6}|X_{I_1},X_{I_5}\in A_{I_5}](\omega_{I_1},A_{I_2})\\\\
\displaystyle =\int_{A_{I_2}}P[X_{I_3},X_{I_4}\in A_{I_4}|X_{I_{12}},X_{I_5}\in A_{I_5},X_{I_6}\in A_{I_6}](\omega_{I_1},\omega_{I_2};A_{I_3})\\
\displaystyle \ \ \cdot P[X_{I_2},X_{I_6}\in A_{I_6}|X_{I_1},X_{I_5}\in A_{I_5}](\omega_{I_1},d\omega_{I_2}),\quad a.e.\{P[X_{I_1},X_{I_5}\in A_{I_5}]\}.\end{array}$$
\item If there exist $P[X_{I_3},X_{I_4}\in A_{I_4}|X_{I_{12}},X_{I_5}\in A_{I_5}](\omega_{I_1},\omega_{I_2};A_{I_3})$ and $P[X_{I_2}|X_{I_1},X_{I_5}\in A_{I_5}](\omega_{I_1},A_{I_2})$, then so do $P[X_{I_3};X_{I_2}\in A_{I_2},X_{I_4}\in A_{I_4}|X_{I_1},X_{I_5}\in A_{I_5}](\omega_{I_1},A_{I_3})$ for each fixed $A_{I_2}\in\mathscr{F}_{I_2}$ and $P[X_{I_2};X_{I_3}\in A_{I_3},X_{I_4}\in A_{I_4}|X_{I_1},X_{I_5}\in A_{I_5}](\omega_{I_1},A_{I_2})$ for each fixed $A_{I_3}\in\mathscr{F}_{I_3}$. For each fixed $A_{I_2}\in\mathscr{F}_{I_2}$,
$$\begin{array}{l}\displaystyle P[X_{I_3};X_{I_2}\in A_{I_2},X_{I_4}\in A_{I_4}|X_{I_1},X_{I_5}\in A_{I_5}](\omega_{I_1},A_{I_3})\\\\
\displaystyle =\int_{A_{I_2}}P[X_{I_3},X_{I_4}\in A_{I_4}|X_{I_{12}},X_{I_5}\in A_{I_5}](\omega_{I_1},\omega_{I_2};A_{I_3})\\
\displaystyle \ \ \cdot P[X_{I_2}|X_{I_1},X_{I_5}\in A_{I_5}](\omega_{I_1},d\omega_{I_2}),\quad a.e.\{P[X_{I_1},X_{I_5}\in A_{I_5}]\}.\end{array}$$
For each fixed $A_{I_3}\in\mathscr{F}_{I_3}$,
$$\begin{array}{l}\displaystyle P[X_{I_2};X_{I_3}\in A_{I_3},X_{I_4}\in A_{I_4}|X_{I_1},X_{I_5}\in A_{I_5}](\omega_{I_1},A_{I_2})\\\\
\displaystyle =\int_{A_{I_2}}P[X_{I_3},X_{I_4}\in A_{I_4}|X_{I_{12}},X_{I_5}\in A_{I_5}](\omega_{I_1},\omega_{I_2};A_{I_3})\\
\displaystyle \ \ \cdot P[X_{I_2}|X_{I_1},X_{I_5}\in A_{I_5}](\omega_{I_1},d\omega_{I_2}),\quad a.e.\{P[X_{I_1},X_{I_5}\in A_{I_5}]\}.\end{array}$$
\item If there exist $P[X_{I_3}|X_{I_{12}},X_{I_5}\in A_{I_5}](\omega_{I_1},\omega_{I_2};A_{I_3})$ and $P[X_{I_2}|X_{I_1},X_{I_5}\in A_{I_5}](\omega_{I_1},A_{I_2})$, then so do $P[X_{I_3},X_{I_2}\in A_{I_2}|X_{I_1},X_{I_5}\in A_{I_5}](\omega_{I_1},A_{I_3})$ for each fixed $A_{I_2}\in\mathscr{F}_{I_2}$ and $P[X_{I_2},X_{I_3}\in A_{I_3}|X_{I_1},X_{I_5}\in A_{I_5}](\omega_{I_1},A_{I_2})$ for each fixed $A_{I_3}\in\mathscr{F}_{I_3}$. For each fixed $A_{I_2}\in\mathscr{F}_{I_2}$,
$$\begin{array}{l}\displaystyle P[X_{I_3},X_{I_2}\in A_{I_2}|X_{I_1},X_{I_5}\in A_{I_5}](\omega_{I_1},A_{I_3})\\\\
\displaystyle =\int_{A_{I_2}}P[X_{I_3}|X_{I_{12}},X_{I_5}\in A_{I_5}](\omega_{I_1},\omega_{I_2};A_{I_3})\\
\displaystyle \ \ \cdot P[X_{I_2}|X_{I_1},X_{I_5}\in A_{I_5}](\omega_{I_1},d\omega_{I_2}),\ \ a.e.\{P[X_{I_1},X_{I_5}\in A_{I_5}]\}.\end{array}$$
For each fixed $A_{I_3}\in\mathscr{F}_{I_3}$,
$$\begin{array}{l}\displaystyle P[X_{I_2},X_{I_3}\in A_{I_3}|X_{I_1},X_{I_5}\in A_{I_5}](\omega_{I_1},A_{I_2})\\\\
\displaystyle =\int_{A_{I_2}}P[X_{I_3}|X_{I_{12}},X_{I_5}\in A_{I_5}](\omega_{I_1},\omega_{I_2};A_{I_3})\\
\displaystyle \ \ \cdot P[X_{I_2}|X_{I_1},X_{I_5}\in A_{I_5}](\omega_{I_1},d\omega_{I_2}),\ \  a.e.\{P[X_{I_1},X_{I_5}\in A_{I_5}]\}.\end{array}$$
\item If there exist $P[X_{I_3},X_{I_4}\in A_{I_4}|X_{I_{12}}](\omega_{I_1},\omega_{I_2};A_{I_3})$ and $P[X_{I_2}|X_{I_1}](\omega_{I_1},A_{I_2})$, then so do $P[X_{I_3};X_{I_2}\in A_{I_2},X_{I_4}\in A_{I_4}|X_{I_1}](\omega_{I_1},A_{I_3})$ for each fixed $A_{I_2}\in\mathscr{F}_{I_2}$ and $P[X_{I_2};X_{I_3}\in A_{I_3},X_{I_4}\in A_{I_4}|X_{I_1}](\omega_{I_1},A_{I_2})$ for each fixed $A_{I_3}\in\mathscr{F}_{I_3}$. For each fixed $A_{I_2}\in\mathscr{F}_{I_2}$,
$$\begin{array}{l}\displaystyle P[X_{I_3};X_{I_2}\in A_{I_2},X_{I_4}\in A_{I_4}|X_{I_1}](\omega_{I_1},A_{I_3})\\\\
\displaystyle=\int_{A_{I_2}}P[X_{I_3},X_{I_4}\in A_{I_4}|X_{I_{12}}](\omega_{I_1},\omega_{I_2};A_{I_3})\\
\displaystyle \ \ \cdot P[X_{I_2}|X_{I_1}](\omega_{I_1},d\omega_{I_2}),\quad a.e.\{P[X_{I_1}]\}.
\end{array}$$
For each fixed $A_{I_3}\in\mathscr{F}_{I_3}$,
$$\begin{array}{l}\displaystyle P[X_{I_2};X_{I_3}\in A_{I_3},X_{I_4}\in A_{I_4}|X_{I_1}](\omega_{I_1},A_{I_2})\\\\
\displaystyle =\int_{A_{I_2}}P[X_{I_3},X_{I_4}\in A_{I_4}|X_{I_{12}}](\omega_{I_1},\omega_{I_2};A_{I_3})\\
\displaystyle \ \ \cdot P[X_{I_2}|X_{I_1}](\omega_{I_1},d\omega_{I_2}),\quad a.e.\{P[X_{I_1}]\}.
\end{array}$$
\item If there exist $P[X_{I_3}|X_{I_{12}}](\omega_{I_1},\omega_{I_2};A_{I_3})$ and $P[X_{I_2}|X_{I_1}](\omega_{I_1},A_{I_2})$, then so do $P[X_{I_3},X_{I_2}\in A_{I_2}|X_{I_1}](\omega_{I_1},A_{I_3})$ for each fixed $A_{I_2}\in\mathscr{F}_{I_2}$ and $P[X_{I_2},X_{I_3}\in A_{I_3}|X_{I_1}](\omega_{I_1},A_{I_2})$ for each fixed $A_{I_3}\in\mathscr{F}_{I_3}$. For each fixed $A_{I_2}\in\mathscr{F}_{I_2}$,
$$\begin{array}{l}\displaystyle P[X_{I_3},X_{I_2}\in A_{I_2}|X_{I_1}](\omega_{I_1},A_{I_3})\\\\
\displaystyle=\int_{A_{I_2}}P[X_{I_3}|X_{I_{12}}](\omega_{I_1},\omega_{I_2};A_{I_3})P[X_{I_2}|X_{I_1}](\omega_{I_1},d\omega_{I_2}),\quad a.e.\{P[X_{I_1}]\}.
\end{array}$$
For each fixed $A_{I_3}\in\mathscr{F}_{I_3}$,
$$\begin{array}{l}\displaystyle P[X_{I_2},X_{I_3}\in A_{I_3}|X_{I_1}](\omega_{I_1},A_{I_2})\\\\
\displaystyle=\int_{A_{I_2}}P[X_{I_3}|X_{I_{12}}](\omega_{I_1},\omega_{I_2};A_{I_3})P[X_{I_2}|X_{I_1}](\omega_{I_1},d\omega_{I_2}),\quad a.e.\{P[X_{I_1}]\}.
\end{array}$$
\end{enumerate}
\end{col}

By Corollary \ref{corollary3.7.3} and Theorem \ref{theorem2.11.3}, we have the following corollary.

\begin{col}\label{corollary3.7.4}Let $(\Omega,\mathscr{F},P)$ be a given probability space. Let $X_i,\ i\in \mathbb{N}$ be random variables on $(\Omega,\mathscr{F})$. For $I\subset \mathbb{N}$, let $X_I(\omega)=(X_i(\omega),i\in I)$, then $X_I$ is a random object from $(\Omega,\mathscr{F})$ to $(\mathbb{R}^I,\mathscr{B}(\mathbb{R}^I))$. Given $I_i\subset \mathbb{N},\ i=1,2,\cdots,6$ and $I_i\cap I_j=\emptyset,\ i\neq j$. Let $I_{12}=I_1+I_2$ and fixed $A_{I_i}\in \mathscr{B}(\mathbb{R}^{I_i}),\ i=4,5,6$.
\begin{enumerate}
\item For $\forall A_{I_i}\in\mathscr{B}(\mathbb{R}^{I_i}),\ i=2,3$,
$$\begin{array}{l}\displaystyle P[X_{I_2},X_{I_3};X_{I_4}\in A_{I_4},X_{I_6}\in A_{I_6}|X_{I_1},X_{I_5}\in A_{I_5}](x_{I_1};A_{I_2},A_{I_3})\\\\
\displaystyle =\int_{A_{I_2}}P[X_{I_3},X_{I_4}\in A_{I_4}|X_{I_{12}},X_{I_5}\in A_{I_5},X_{I_6}\in A_{I_6}](x_{I_1},x_{I_2};A_{I_3})\\
\displaystyle \ \ \cdot P[X_{I_2},X_{I_6}\in A_{I_6}|X_{I_1},X_{I_5}\in A_{I_5}](x_{I_1},dx_{I_2}),\quad a.e.\{P[X_{I_1},X_{I_5}\in A_{I_5}]\}.\end{array}$$
\item For each fixed $A_{I_2}\in\mathscr{B}(\mathbb{R}^{I_2})$,
$$\begin{array}{l}\displaystyle P[X_{I_3};X_{I_2}\in A_{I_2},X_{I_4}\in A_{I_4},X_{I_6}\in A_{I_6}|X_{I_1},X_{I_5}\in A_{I_5}](x_{I_1},A_{I_3})\\\\
\displaystyle =\int_{A_{I_2}}P[X_{I_3},X_{I_4}\in A_{I_4}|X_{I_{12}},X_{I_5}\in A_{I_5},X_{I_6}\in A_{I_6}](x_{I_1},x_{I_2};A_{I_3})\\
\displaystyle \ \ \cdot P[X_{I_2},X_{I_6}\in A_{I_6}|X_{I_1},X_{I_5}\in A_{I_5}](x_{I_1},dx_{I_2}),\quad a.e.\{P[X_{I_1},X_{I_5}\in A_{I_5}]\}.\end{array}$$
For each fixed $A_{I_3}\in\mathscr{B}(\mathbb{R}^{I_3})$,
$$\begin{array}{l}\displaystyle P[X_{I_2};X_{I_3}\in A_{I_3},X_{I_4}\in A_{I_4},X_{I_6}\in A_{I_6}|X_{I_1},X_{I_5}\in A_{I_5}](x_{I_1},A_{I_2})\\\\
\displaystyle =\int_{A_{I_2}}P[X_{I_3},X_{I_4}\in A_{I_4}|X_{I_{12}},X_{I_5}\in A_{I_5},X_{I_6}\in A_{I_6}](x_{I_1},x_{I_2};A_{I_3})\\
\displaystyle \ \ \cdot P[X_{I_2},X_{I_6}\in A_{I_6}|X_{I_1},X_{I_5}\in A_{I_5}](x_{I_1},dx_{I_2}),\quad a.e.\{P[X_{I_1},X_{I_5}\in A_{I_5}]\}.\end{array}$$
\item For each fixed $A_{I_2}\in\mathscr{B}(\mathbb{R}^{I_2})$,
$$\begin{array}{l}\displaystyle P[X_{I_3};X_{I_2}\in A_{I_2},X_{I_4}\in A_{I_4}|X_{I_1},X_{I_5}\in A_{I_5}](x_{I_1},A_{I_3})\\\\
\displaystyle =\int_{A_{I_2}}P[X_{I_3},X_{I_4}\in A_{I_4}|X_{I_{12}},X_{I_5}\in A_{I_5}](x_{I_1},x_{I_2};A_{I_3})\\
\displaystyle \ \ \cdot P[X_{I_2}|X_{I_1},X_{I_5}\in A_{I_5}](x_{I_1},dx_{I_2}),\ \ a.e.\{P[X_{I_1},X_{I_5}\in A_{I_5}]\}.\end{array}$$
For each fixed $A_{I_3}\in\mathscr{B}(\mathbb{R}^{I_3})$,
$$\begin{array}{l}\displaystyle P[X_{I_2};X_{I_3}\in A_{I_3},X_{I_4}\in A_{I_4}|X_{I_1},X_{I_5}\in A_{I_5}](x_{I_1},A_{I_2})\\\\
\displaystyle =\int_{A_{I_2}}P[X_{I_3},X_{I_4}\in A_{I_4}|X_{I_{12}},X_{I_5}\in A_{I_5}](x_{I_1},x_{I_2};A_{I_3})\\
\displaystyle \ \ \cdot P[X_{I_2}|X_{I_1},X_{I_5}\in A_{I_5}](x_{I_1},dx_{I_2}),\ \ a.e.\{P[X_{I_1},X_{I_5}\in A_{I_5}]\}.\end{array}$$
\item For each fixed $A_{I_2}\in\mathscr{B}(\mathbb{R}^{I_2})$,
$$\begin{array}{l}\displaystyle P[X_{I_3},X_{I_2}\in A_{I_2}|X_{I_1},X_{I_5}\in A_{I_5}](x_{I_1},A_{I_3})\\\\
\displaystyle =\int_{A_{I_2}}P[X_{I_3}|X_{I_{12}},X_{I_5}\in A_{I_5}](x_{I_1},x_{I_2};A_{I_3})\\
\displaystyle \ \ \cdot P[X_{I_2}|X_{I_1},X_{I_5}\in A_{I_5}](x_{I_1},dx_{I_2}),\ \  a.e.\{P[X_{I_1},X_{I_5}\in A_{I_5}]\}.\end{array}$$
For each fixed $A_{I_3}\in\mathscr{B}(\mathbb{R}^{I_3})$,
$$\begin{array}{l}\displaystyle P[X_{I_2},X_{I_3}\in A_{I_3}|X_{I_1},X_{I_5}\in A_{I_5}](x_{I_1},A_{I_2})\\\\
\displaystyle =\int_{A_{I_2}}P[X_{I_3}|X_{I_{12}},X_{I_5}\in A_{I_5}](x_{I_1},x_{I_2};A_{I_3})\\
\displaystyle \ \ \cdot P[X_{I_2}|X_{I_1},X_{I_5}\in A_{I_5}](x_{I_1},dx_{I_2}),\ \ a.e.\{P[X_{I_1},X_{I_5}\in A_{I_5}]\}.\end{array}$$
\item For each fixed $A_{I_2}\in\mathscr{B}(\mathbb{R}^{I_2})$,
$$\begin{array}{l}\displaystyle P[X_{I_3};X_{I_2}\in A_{I_2},X_{I_4}\in A_{I_4}|X_{I_1}](x_{I_1},A_{I_3})\\\\
\displaystyle=\int_{A_{I_2}}P[X_{I_3},X_{I_4}\in A_{I_4}|X_{I_{12}}](x_{I_1},x_{I_2};A_{I_3})P[X_{I_2}|X_{I_1}](x_{I_1},dx_{I_2}),\ \ a.e.\{P[X_{I_1}]\}.
\end{array}$$
For each fixed $A_{I_3}\in\mathscr{B}(\mathbb{R}^{I_3})$,
$$\begin{array}{l}\displaystyle P[X_{I_2};X_{I_3}\in A_{I_3},X_{I_4}\in A_{I_4}|X_{I_1}](x_{I_1},A_{I_2})\\\\
\displaystyle=\int_{A_{I_2}}P[X_{I_3},X_{I_4}\in A_{I_4}|X_{I_{12}}](x_{I_1},x_{I_2};A_{I_3})P[X_{I_2}|X_{I_1}](x_{I_1},dx_{I_2}),\ \ a.e.\{P[X_{I_1}]\}.
\end{array}$$
\item For each fixed $A_{I_2}\in\mathscr{B}(\mathbb{R}^{I_2})$,
$$\begin{array}{l}\displaystyle P[X_{I_3},X_{I_2}\in A_{I_2}|X_{I_1}](x_{I_1},A_{I_3})\\\\
\displaystyle=\int_{A_{I_2}}P[X_{I_3}|X_{I_{12}}](x_{I_1},x_{I_2};A_{I_3})P[X_{I_2}|X_{I_1}](x_{I_1},dx_{I_2}),\ \ a.e.\{P[X_{I_1}]\}.
\end{array}$$
For each fixed $A_{I_3}\in\mathscr{B}(\mathbb{R}^{I_3})$,
$$\begin{array}{l}\displaystyle P[X_{I_2},X_{I_3}\in A_{I_3}|X_{I_1}](x_{I_1},A_{I_2})\\\\
\displaystyle=\int_{A_{I_2}}P[X_{I_3}|X_{I_{12}}](x_{I_1},x_{I_2};A_{I_3})P[X_{I_2}|X_{I_1}](x_{I_1},dx_{I_2}),\ \ a.e.\{P[X_{I_1}]\}.
\end{array}$$
\end{enumerate}
\end{col}

\begin{lem}\label{lemma3.9}Let $(\Omega_i,\mathscr{F}_i),i=1,2,3$ be measurable spaces. If $K_2(\omega_1,A_2)$ is a finite kernel from $(\Omega_1,\mathscr{F}_1)$ to $(\Omega_2,\mathscr{F}_2)$, and $K_3(\omega_1,A_3)$ is also a finite kernel from $(\Omega_1,\mathscr{F}_1)$ to $(\Omega_3,\mathscr{F}_3)$, then there exists a finite kernel $\widehat{K}(\omega_1,A_{23})$ from $(\Omega_1,\mathscr{F}_1)$ to $(\Omega_2\times\Omega_3,\mathscr{F}_2\times \mathscr{F}_3)$ such that
$$\widehat{K}(\omega_1,A_2\times A_3)=K_2(\omega_1,A_2)\cdot K_3(\omega_1,A_3),\ \ \forall A_i\in\mathscr{F}_i,\ i=2,3.$$
$\widehat{K}(\omega_1,A_{23})$ is denoted by $K_2(\omega_1,\cdot)\times K_3(\omega_1,\cdot)$ or $K_2(\omega_1,\cdot)\times K_3(\omega_1,\cdot)(A_{23})$ where $A_{23}\in \mathscr{F}_2\times \mathscr{F}_3$.
\end{lem}

\begin{col}\label{corollarylem3.9.1}Let $(\Omega_i,\mathscr{F}_i),i=1,2,3$ be measurable spaces. If $K(\omega_1,A_2)$ is a finite kernel from $(\Omega_1,\mathscr{F}_1)$ to $(\Omega_2,\mathscr{F}_2)$, and $\mu$ is a finite measure on $(\Omega_3,\mathscr{F}_3)$, then there exists a finite kernel $\widehat{K}(\omega_1,A_{23})$ from $(\Omega_1,\mathscr{F}_1)$ to $(\Omega_2\times\Omega_3,\mathscr{F}_2\times \mathscr{F}_3)$ such that
$$\widehat{K}(\omega_1,A_2\times A_3)=K(\omega_1,A_2)\cdot \mu(A_3),\ \ \forall A_i\in\mathscr{F}_i,\ i=2,3.$$
$\widehat{K}(\omega_1,A_{23})$ is denoted by $K(\omega_1,\cdot)\times \mu$ or $K(\omega_1,\cdot)\times \mu(A_{23})$ where $A_{23}\in \mathscr{F}_2\times \mathscr{F}_3$.
\end{col}

Let $K_3(\omega_1,A_3)=\mu(A_3)$, then Corollary \ref{corollarylem3.9.1} follows from Lemma \ref{lemma3.9}.

For the sake of convenience, $P[X_2\in A_2|X_1,X_3\in A_3](\omega_1)$ and $P[X_2,X_4\in A_4|X_1,X_3\in A_3](\omega_1,A_2)$ are respectively abbreviated to $P[X_2\in A_2|X_1,X_3\in A_3]$ and $P[X_2,X_4\in A_4|X_1,X_3\in A_3]$ without any ambiguity. The others are similar.

\begin{thm}\label{theorem3.8}Let $(\Omega,\mathscr{F},P)$ be a given probability space, and let $(\Omega_i,\mathscr{F}_i),\ i=1,2,\cdots,6$ be measurable spaces. Let $X_i:(\Omega,\mathscr{F})\rightarrow(\Omega_i,\mathscr{F}_i),\ i=1,2,\cdots,6$ be random objects, and $P[X_i]$ the probability on $(\Omega_i,\mathscr{F}_i)$ induced by $X_i$, $i=1,2,\cdots,6$. Then $(\Omega_i,\mathscr{F}_i,P[X_i]),\ i=1,2,\cdots,6$ are probability spaces.

1. If $(X_1,X_2)$ and $X_3$ are independent, then for fixed $A_i\in\mathscr{F}_i,i=2,3$,
$$P[X_1;X_2\in A_2,X_3\in A_3]=P[X_3\in A_3]\cdot P[X_1,X_2\in A_2],$$
$$P[X_1,X_2\in A_2|X_3\in A_3]=P[X_1,X_2\in A_2]\ \ \text{provided that}\ \  P[X_3\in A_3]>0,$$
$$P[X_3\in A_3|X_1,X_2\in A_2]=P[X_3\in A_3]\ \ \text{provided that} \ \ P[X_2\in A_2]>0.$$

2. If $(X_1,X_3)$ and $(X_2,X_4)$ are independent, then for fixed $A_i\in\mathscr{F}_i,i=3,4$,
$$P[X_1,X_2;X_3\in A_3,X_4\in A_4]=P[X_1,X_3\in A_3]\times P[X_2,X_4\in A_4],$$
and
$$P[X_2,X_4\in A_4|X_1,X_3\in A_3](\omega_1,A_2)=P[X_2,X_4\in A_4](A_2)$$
provided that $P[X_3\in A_3]>0$.

3. If $(X_1,X_3,X_4,X_6)$ and $(X_2,X_5)$ are independent, and $P[X_3,X_6\in A_6|X_1,X_4\in A_4]$ exists for fixed $A_i\in\mathscr{F}_i,i=4,6$ with $P[X_4\in A_4]>0$, then for each fixed $A_5\in\mathscr{F}_5$, there exists $P[X_3,X_2;X_5\in A_5,X_6\in A_6|X_1,X_4\in A_4]$, and
$$P[X_3,X_2;X_5\in A_5,X_6\in A_6|X_1,X_4\in A_4]=P[X_3,X_6\in A_6|X_1,X_4\in A_4]\times P[X_2,X_5\in A_5],$$
which is defined in Corollary \ref{corollarylem3.9.1}.
\end{thm}

\begin{col}\label{corollary3.8.1}Let $\Lambda$ be a nonempty index set. Let $(\Omega,\mathscr{F},P)$ be a given probability space, and $(\Omega_{i},\mathscr{F}_{i}),\ i\in \Lambda$ measurable spaces. Let $X_i:(\Omega,\mathscr{F})\rightarrow(\Omega_i,\mathscr{F}_i),\ i\in \Lambda$ be random objects. Given $I_i\subset \Lambda,\ i=1,2,\cdots,6$ that are pairwise disjoint.

1. If $(X_{I_1},X_{I_2})$ and $X_{I_3}$ are independent, then for fixed $A_{I_i}\in\mathscr{F}_{I_i},i=2,3$,
$$P[X_{I_1};X_{I_2}\in A_{I_2},X_{I_3}\in A_{I_3}]=P[X_{I_3}\in A_{I_3}]\cdot P[X_{I_1},X_{I_2}\in A_{I_2}],$$
$$P[X_{I_1},X_{I_2}\in A_{I_2}|X_{I_3}\in A_{I_3}]=P[X_{I_1},X_{I_2}\in A_{I_2}]\ \ \text{provided that}\ \  P[X_{I_3}\in A_{I_3}]>0,$$
$$P[X_{I_3}\in A_{I_3}|X_{I_1},X_{I_2}\in A_{I_2}]=P[X_{I_3}\in A_{I_3}]\ \ \text{provided that} \ \ P[X_{I_2}\in A_{I_2}]>0.$$

2. If $(X_{I_1},X_{I_3})$ and $(X_{I_2},X_{I_4})$ are independent, then for fixed $A_{I_i}\in\mathscr{F}_{I_i},i=3,4$,
$$P[X_{I_1},X_{I_2};X_{I_3}\in A_{I_3},X_{I_4}\in A_{I_4}]=P[X_{I_1},X_{I_3}\in A_{I_3}]\times P[X_{I_2},X_{I_4}\in A_{I_4}],$$
and
$$P[X_{I_2},X_{I_4}\in A_{I_4}|X_{I_1},X_{I_3}\in A_{I_3}](\omega_{I_1},A_{I_2})=P[X_{I_2},X_{I_4}\in A_{I_4}](A_{I_2})$$
provided that $P[X_{I_3}\in A_{I_3}]>0$.

3. If $(X_{I_1},X_{I_3},X_{I_4},X_{I_6})$ and $(X_{I_2},X_{I_5})$ are independent, and $P[X_{I_3},X_{I_6}\in A_{I_6}|X_{I_1},X_{I_4}\in A_{I_4}]$ exists for fixed $A_{I_i}\in\mathscr{F}_{I_i},i=4,6$ with $P[X_{I_4}\in A_{I_4}]>0$, then for each fixed $A_{I_5}\in\mathscr{F}_{I_5}$, there exists $P[X_{I_3},X_{I_2};X_{I_5}\in A_{I_5},X_{I_6}\in A_{I_6}|X_{I_1},X_{I_4}\in A_{I_4}]$, and
$$\begin{array}{l}P[X_{I_3},X_{I_2};X_{I_5}\in A_{I_5},X_{I_6}\in A_{I_6}|X_{I_1},X_{I_4}\in A_{I_4}]\\
=P[X_{I_3},X_{I_6}\in A_{I_6}|X_{I_1},X_{I_4}\in A_{I_4}]\times P[X_{I_2},X_{I_5}\in A_{I_5}].\end{array}$$
\end{col}

\begin{defn}\label{definition3.9}Let $(\Omega,\mathscr{F},P)$ be a given probability space, and let $(\Omega_i,\mathscr{F}_i),\ i=1,2,\cdots,6$ be measurable spaces. Let $X_i:(\Omega,\mathscr{F})\rightarrow(\Omega_i,\mathscr{F}_i),\ i=1,2,\cdots,6$ be random objects, and $P[X_i]$ the probability on $(\Omega_i,\mathscr{F}_i)$ induced by $X_i$, $i=1,2,\cdots,6$. Then $(\Omega_i,\mathscr{F}_i,P[X_i]),\ i=1,2,\cdots,6$ are probability spaces.

1. For fixed $A_i\in\mathscr{F}_i,i=1,2,3$ with $P[X_1\in A_1]>0$, if
$$P[X_2\in A_2,X_3\in A_3|X_1\in A_1]=P[X_2\in A_2|X_1\in A_1]\cdot P[X_3\in A_3|X_1\in A_1],$$
then event $X_2\in A_2$ and event $X_3\in A_3$ are called conditionally independent given $X_1\in A_1$.

2. For fixed $A_i\in\mathscr{F}_i,i=1,3,4$ with $P[X_1\in A_1]>0$, if
$$P[X_2;X_3\in A_3,X_4\in A_4|X_1\in A_1]=P[X_3\in A_3|X_1\in A_1]\cdot P[X_2,X_4\in A_4|X_1\in A_1],$$
then co-occurrence of $X_2$ and $X_4\in A_4$ and event $X_3\in A_3$ are called conditionally independent given $X_1\in A_1$.

3. For fixed $A_i\in\mathscr{F}_i,i=1,4,5$ with $P[X_1\in A_1]>0$, if
$$P[X_2,X_3;X_4\in A_4,X_5\in A_5|X_1\in A_1]=P[X_2,X_4\in A_4|X_1\in A_1]\times P[X_3,X_5\in A_5|X_1\in A_1],$$
then co-occurrence of $X_2$ and $X_4\in A_4$ and co-occurrence of $X_3$ and $X_5\in A_5$ are called conditionally independent given $X_1\in A_1$.

4. For fixed $A_i\in\mathscr{F}_i,i=3,4,5$ with $P[X_3\in A_3]>0$, if
$$\begin{array}{l}P[X_4\in A_4,X_5\in A_5|X_1,X_3\in A_3]\\
=P[X_4\in A_4|X_1,X_3\in A_3]\cdot P[X_5\in A_5|X_1,X_3\in A_3],\end{array}$$
then event $X_4\in A_4$ and event $X_5\in A_5$ are called conditionally independent given co-occurrence of $X_1$ and $X_3\in A_3$.

5. For fixed $A_i\in\mathscr{F}_i,i=3,4,5$ with $P[X_3\in A_3]>0$, if
$$\begin{array}{l}P[X_2;X_4\in A_4,X_5\in A_5|X_1,X_3\in A_3]\\
=P[X_5\in A_5|X_1,X_3\in A_3]\cdot P[X_2,X_4\in A_4|X_1,X_3\in A_3],\end{array}$$
then event $X_5\in A_5$ and co-occurrence of $X_2$ and $X_4\in A_4$ are called conditionally independent given co-occurrence of $X_1$ and $X_3\in A_3$.

6. For fixed $A_i\in\mathscr{F}_i,i=4,5,6$ with $P[X_4\in A_4]>0$, if
$$\begin{array}{l}P[X_2,X_3;X_5\in A_5,X_6\in A_6|X_1,X_4\in A_4]\\
=P[X_2,X_5\in A_5|X_1,X_4\in A_4]\times P[X_3,X_6\in A_6|X_1,X_4\in A_4],\end{array}$$
which is defined in Lemma \ref{lemma3.9}, then co-occurrence of $X_2$ and $X_5\in A_5$ and co-occurrence of $X_3$ and $X_6\in A_6$ are called conditionally independent given co-occurrence of $X_1$ and $X_4\in A_4$.
\end{defn}

\begin{comm}\label{commentdef3.9.1} There are still other similar concepts as the special cases of the above definitions which are not enumerated. For example, in Definition \ref{definition3.9} 6., if $A_6=\Omega_6$, then we have
$$\begin{array}{l}P[X_2,X_3;X_5\in A_5|X_1,X_4\in A_4]\\
=P[X_2,X_5\in A_5|X_1,X_4\in A_4]\times P[X_3|X_1,X_4\in A_4],\end{array}$$
which means that co-occurrence of $X_2$ and $X_5\in A_5$ and random object $X_3$ are conditionally independent given co-occurrence of $X_1$ and $X_4\in A_4$.
\end{comm}

\begin{thm}\label{theorem3.9}Let $(\Omega,\mathscr{F},P)$ be a given probability space, and let $(\Omega_i,\mathscr{F}_i),\ i=1,2,\cdots,6$ be measurable spaces. Let $X_i:(\Omega,\mathscr{F})\rightarrow(\Omega_i,\mathscr{F}_i),\ i=1,2,\cdots,6$ be random objects on $(\Omega,\mathscr{F})$, and $P[X_i]$ the probability on $(\Omega_i,\mathscr{F}_i)$ induced by $X_i$, $i=1,2,\cdots,6$. Then $(\Omega_i,\mathscr{F}_i,P[X_i]),\ i=1,2,\cdots,6$ are probability spaces.

1. For fixed $A_i\in\mathscr{F}_i,i=1,2,3$ with $P[X_1\in A_1,X_3\in A_3]>0$, then
$$P[X_2\in A_2|X_1\in A_1,X_3\in A_3]=P[X_2\in A_2|X_1\in A_1]$$
if and only if
$$P[X_2\in A_2,X_3\in A_3|X_1\in A_1]=P[X_2\in A_2|X_1\in A_1]\cdot P[X_3\in A_3|X_1\in A_1].$$

2. For fixed $A_i\in\mathscr{F}_i,i=1,3,4$ with $P[X_1\in A_1,X_3\in A_3,X_4\in A_4]>0$, then
$$P[X_2,X_4\in A_4|X_1\in A_1,X_3\in A_3]=P[X_2,X_4\in A_4|X_1\in A_1]$$
or
$$P[X_3\in A_3|X_2;X_1\in A_1,X_4\in A_4]=P[X_3\in A_3|X_1\in A_1]$$
if and only if
$$P[X_2;X_3\in A_3,X_4\in A_4|X_1\in A_1]=P[X_3\in A_3|X_1\in A_1]\cdot P[X_2,X_4\in A_4|X_1\in A_1].$$

3. For fixed $A_i\in\mathscr{F}_i,i=1,4,5$ with $P[X_1\in A_1,X_5\in A_5]>0$, then
$$P[X_2,X_4\in A_4|X_3;X_1\in A_1,X_5\in A_5]=P[X_2,X_4\in A_4|X_1\in A_1]$$
if and only if
$$P[X_2,X_3;X_4\in A_4,X_5\in A_5|X_1\in A_1]=P[X_2,X_4\in A_4|X_1\in A_1]\times P[X_3,X_5\in A_5|X_1\in A_1].$$

4. For fixed $A_i\in\mathscr{F}_i,i=2,3,4$ with $P[X_3\in A_3,X_4\in A_4]>0$, then
$$P[X_2\in A_2|X_1;X_3\in A_3,X_4\in A_4]=P[X_2\in A_2|X_1,X_3\in A_3]$$
if and only if
$$P[X_2\in A_2,X_4\in A_4|X_1,X_3\in A_3]=P[X_2\in A_2|X_1,X_3\in A_3]\cdot P[X_4\in A_4|X_1,X_3\in A_3].$$

5. For fixed $A_i\in\mathscr{F}_i,i=3,4,5$ with $P[X_3\in A_3,X_4\in A_4,X_5\in A_5]>0$, then
$$P[X_2,X_4\in A_4|X_1;X_3\in A_3,X_5\in A_5]=P[X_2,X_4\in A_4|X_1,X_3\in A_3]$$
or
$$P[X_5\in A_5|X_1,X_2;X_3\in A_3,X_4\in A_4]=P[X_5\in A_5|X_1,X_3\in A_3]$$
if and only if
$$\begin{array}{l}P[X_2;X_4\in A_4,X_5\in A_5|X_1,X_3\in A_3]\\
=P[X_2,X_4\in A_4|X_1,X_3\in A_3]\cdot P[X_5\in A_5|X_1,X_3\in A_3]\end{array}$$
given the existence of $P[X_2,X_4\in A_4|X_1,X_3\in A_3]$.

6. For fixed $A_i\in\mathscr{F}_i,i=4,5,6$ with $P[X_4\in A_4,X_6\in A_6]>0$, then
$$P[X_2,X_5\in A_5|X_1,X_3;X_4\in A_4,X_6\in A_6]=P[X_2,X_5\in A_5|X_1,X_4\in A_4]$$
if and only if
$$\begin{array}{l}P[X_2,X_3;X_5\in A_5,X_6\in A_6|X_1,X_4\in A_4]\\
=P[X_2,X_5\in A_5|X_1,X_4\in A_4]\times P[X_3,X_6\in A_6|X_1,X_4\in A_4]\end{array}$$
given the existences of $P[X_2,X_5\in A_5|X_1,X_4\in A_4]$ and $P[X_3,X_6\in A_6|X_1,X_4\in A_4]$.
\end{thm}

\begin{comm}\label{comment3.9.1}In 6. of Theorem \ref{theorem3.9},
$$P[X_2,X_5\in A_5|X_1,X_3;X_4\in A_4,X_6\in A_6]=P[X_2,X_5\in A_5|X_1,X_4\in A_4]$$
means that
$$\begin{array}{l}P[X_2,X_5\in A_5|X_1,X_3;X_4\in A_4,X_6\in A_6](\omega_1,\omega_2;A_2)\\
=P[X_2,X_5\in A_5|X_1,X_4\in A_4](\omega_1,A_2), \ \ a.e.\{P[X_1,X_3;X_4\in A_4,X_6\in A_6]\},\ \forall A_2\in\mathscr{F}_2.\end{array}$$
we also have
$$\begin{array}{l}\ \ \ \ P[X_2,X_5\in A_5|X_1,X_3;X_4\in A_4,X_6\in A_6]=P[X_2,X_5\in A_5|X_1,X_4\in A_4]\\
\Leftrightarrow P[X_2,X_5\in A_5|X_1;X_3\in A_3,X_4\in A_4,X_6\in A_6]=P[X_2,X_5\in A_5|X_1,X_4\in A_4],\\
\ \ \ \forall A_3\in\mathscr{F}_3\  with\  P[X_i\in A_i,i=3,4,6]>0.
\end{array}$$
\end{comm}

\begin{col}\label{corollary3.9.1}Let $(\Omega,\mathscr{F},P)$ be a given probability space, and let $(\Omega_i,\mathscr{F}_i),\ i=1,2,\cdots,6$ be measurable spaces. Let $X_i:(\Omega,\mathscr{F})\rightarrow(\Omega_i,\mathscr{F}_i),\ i=1,2,\cdots,6$ be random objects, and $P[X_i]$ the probability on $(\Omega_i,\mathscr{F}_i)$ induced by $X_i$, $i=1,2,\cdots,6$. Then $(\Omega_i,\mathscr{F}_i,P[X_i]),\ i=1,2,\cdots,6$ are probability spaces.

1. If $(X_1,X_2,X_3,X_4)$ and $X_5$ are independent, then for fixed $A_i\in\mathscr{F}_i,\ i=3,4,5$ with $P[X_3\in A_3,X_4\in A_4,X_5\in A_5]>0$,
$$P[X_2,X_4\in A_4|X_1,X_3\in A_3,X_5\in A_5]=P[X_2,X_4\in A_4|X_1,X_3\in A_3]$$
and
$$P[X_5\in A_5|X_1,X_2;X_3\in A_3,X_4\in A_4]=P[X_5\in A_5]$$
given the existence of $P[X_2,X_4\in A_4|X_1,X_3\in A_3]$.

2. If $(X_1,X_3,X_4,X_6)$ and $(X_2,X_5)$ are independent, then for fixed $A_i\in\mathscr{F}_i,\ i=4,5,6$ with $P[X_4\in A_4,X_5\in A_5]>0$,
$$P[X_3,X_6\in A_6|X_1,X_2;X_4\in A_4,X_5\in A_5]=P[X_3,X_6\in A_6|X_1,X_4\in A_4]$$
given the existence of $P[X_3,X_6\in A_6|X_1,X_4\in A_4]$.
\end{col}

\begin{col}\label{corollary3.9.2}Let $\Lambda$ be a nonempty index set. Let $(\Omega,\mathscr{F},P)$ be a given probability space, and $(\Omega_{i},\mathscr{F}_{i}),\ i\in \Lambda$ measurable spaces. Let $X_i:(\Omega,\mathscr{F})\rightarrow(\Omega_i,\mathscr{F}_i),\ i\in \Lambda$ be random objects. Given $I_i\subset \Lambda,\ i=1,2,\cdots,6$ that are pairwise disjoint.

1. For fixed $A_{I_i}\in\mathscr{F}_{I_i},i=1,2,3$ with $P[X_{I_1}\in A_{I_1},X_{I_3}\in A_{I_3}]>0$, then
$$P[X_{I_2}\in A_{I_2}|X_{I_1}\in A_{I_1},X_{I_3}\in A_{I_3}]=P[X_{I_2}\in A_{I_2}|X_{I_1}\in A_{I_1}]$$
if and only if
$$P[X_{I_2}\in A_{I_2},X_{I_3}\in A_{I_3}|X_{I_1}\in A_{I_1}]=P[X_{I_2}\in A_{I_2}|X_{I_1}\in A_{I_1}]\cdot P[X_{I_3}\in A_{I_3}|X_{I_1}\in A_{I_1}].$$

2. For fixed $A_{I_i}\in\mathscr{F}_{I_i},i=1,3,4$ with $P[X_{I_1}\in A_{I_1},X_{I_3}\in A_{I_3},X_{I_4}\in A_{I_4}]>0$, then
$$P[X_{I_2},X_{I_4}\in A_{I_4}|X_{I_1}\in A_{I_1},X_{I_3}\in A_{I_3}]=P[X_{I_2},X_{I_4}\in A_{I_4}|X_{I_1}\in A_{I_1}]$$
or
$$P[X_{I_3}\in A_{I_3}|X_{I_2};X_{I_1}\in A_{I_1},X_{I_4}\in A_{I_4}]=P[X_{I_3}\in A_{I_3}|X_{I_1}\in A_{I_1}]$$
if and only if
$$\begin{array}{l}P[X_{I_2};X_{I_3}\in A_{I_3},X_{I_4}\in A_{I_4}|X_{I_1}\in A_{I_1}]\\
=P[X_{I_3}\in A_{I_3}|X_{I_1}\in A_{I_1}]\cdot P[X_{I_2},X_{I_4}\in A_{I_4}|X_{I_1}\in A_{I_1}].\end{array}$$

3. For fixed $A_{I_i}\in\mathscr{F}_{I_i},i=1,4,5$ with $P[X_{I_1}\in A_{I_1},X_{I_5}\in A_{I_5}]>0$, then
$$P[X_{I_2},X_{I_4}\in A_{I_4}|X_{I_3};X_{I_1}\in A_{I_1},X_{I_5}\in A_{I_5}]=P[X_{I_2},X_{I_4}\in A_{I_4}|X_{I_1}\in A_{I_1}]$$
if and only if
$$\begin{array}{l}P[X_{I_2},X_{I_3};X_{I_4}\in A_{I_4},X_{I_5}\in A_{I_5}|X_{I_1}\in A_{I_1}]\\
=P[X_{I_2},X_{I_4}\in A_{I_4}|X_{I_1}\in A_{I_1}]\times P[X_{I_3},X_{I_5}\in A_{I_5}|X_{I_1}\in A_{I_1}].\end{array}$$

4. For fixed $A_{I_i}\in\mathscr{F}_{I_i},i=2,3,4$ with $P[X_{I_3}\in A_{I_3},X_{I_4}\in A_{I_4}]>0$, then
$$P[X_{I_2}\in A_{I_2}|X_{I_1};X_{I_3}\in A_{I_3},X_{I_4}\in A_{I_4}]=P[X_{I_2}\in A_{I_2}|X_{I_1},X_{I_3}\in A_{I_3}]$$
if and only if
$$\begin{array}{l}P[X_{I_2}\in A_{I_2},X_{I_4}\in A_{I_4}|X_{I_1},X_{I_3}\in A_{I_3}]\\
=P[X_{I_2}\in A_{I_2}|X_{I_1},X_{I_3}\in A_{I_3}]\cdot P[X_{I_4}\in A_{I_4}|X_{I_1},X_{I_3}\in A_{I_3}].\end{array}$$

5. For fixed $A_{I_i}\in\mathscr{F}_{I_i},i=3,4,5$ with $P[X_{I_3}\in A_{I_3},X_{I_4}\in A_{I_4},X_{I_5}\in A_{I_5}]>0$, then
$$P[X_{I_2},X_{I_4}\in A_{I_4}|X_{I_1};X_{I_3}\in A_{I_3},X_{I_5}\in A_{I_5}]=P[X_{I_2},X_{I_4}\in A_{I_4}|X_{I_1},X_{I_3}\in A_{I_3}]$$
or
$$P[X_{I_5}\in A_{I_5}|X_{I_1},X_{I_2};X_{I_3}\in A_{I_3},X_{I_4}\in A_{I_4}]=P[X_{I_5}\in A_{I_5}|X_{I_1},X_{I_3}\in A_{I_3}]$$
if and only if
$$\begin{array}{l}P[X_{I_2};X_{I_4}\in A_{I_4},X_{I_5}\in A_{I_5}|X_{I_1},X_{I_3}\in A_{I_3}]\\
=P[X_{I_2},X_{I_4}\in A_{I_4}|X_{I_1},X_{I_3}\in A_{I_3}]\cdot P[X_{I_5}\in A_{I_5}|X_{I_1},X_{I_3}\in A_{I_3}]\end{array}$$
given the existence of $P[X_{I_2},X_{I_4}\in A_{I_4}|X_{I_1},X_{I_3}\in A_{I_3}]$.

6. For fixed $A_{I_i}\in\mathscr{F}_{I_i},i=4,5,6$ with $P[X_{I_4}\in A_{I_4},X_{I_6}\in A_{I_6}]>0$, then
$$P[X_{I_2},X_{I_5}\in A_{I_5}|X_{I_1},X_{I_3};X_{I_4}\in A_{I_4},X_{I_6}\in A_{I_6}]=P[X_{I_2},X_{I_5}\in A_{I_5}|X_{I_1},X_{I_4}\in A_{I_4}]$$
if and only if
$$\begin{array}{l}P[X_{I_2},X_{I_3};X_{I_5}\in A_{I_5},X_{I_6}\in A_{I_6}|X_{I_1},X_{I_4}\in A_{I_4}]\\
=P[X_{I_2},X_{I_5}\in A_{I_5}|X_{I_1},X_{I_4}\in A_{I_4}]\times P[X_{I_3},X_{I_6}\in A_{I_6}|X_{I_1},X_{I_4}\in A_{I_4}]\end{array}$$
given the existences of $P[X_{I_2},X_{I_5}\in A_{I_5}|X_{I_1},X_{I_4}\in A_{I_4}]$ and $P[X_{I_3},X_{I_6}\in A_{I_6}|X_{I_1},X_{I_4}\in A_{I_4}]$.
\end{col}

From Corollary \ref{corollary3.9.2} and Theorem \ref{theorem2.11.3} the corollary below follows.

\begin{col}\label{corollary3.9.3}  Let $(\Omega,\mathscr{F},P)$ be a given probability space. Let $X_i,\ i\in \mathbb{N}$ be random variables on $(\Omega,\mathscr{F},P)$. Given $I_i\subset \mathbb{N},\ i=1,2,\cdots,6$ that are pairwise disjoint.

1. For fixed $A_i\in\mathscr{B}(\mathbb{R}),i\in I_1+I_2+I_3$ with $P[X_i\in A_i,i\in I_1+I_3]>0$, then
$$P[X_i\in A_i,i\in I_2|X_i\in A_i,i\in I_1+I_3]=P[X_i\in A_i,i\in I_2|X_i\in A_i,i\in I_1]$$
if and only if
$$\begin{array}{l}P[X_i\in A_i,i\in I_2+I_3|X_i\in A_i,i\in I_1]\\
=P[X_i\in A_i,i\in I_2|X_i\in A_i,i\in I_1]\cdot P[X_i\in A_i,i\in I_3|X_i\in A_i,i\in I_1].\end{array}$$

2. For fixed $A_i\in\mathscr{B}(\mathbb{R}),i\in I_1+I_3+I_4$ with $P[X_i\in A_i,i\in I_1+I_3+I_4]>0$, then
$$P[X_i,i\in I_2;X_i\in A_i,i\in I_4|X_i\in A_i,i\in I_1+I_3]=P[X_i,i\in I_2;X_i\in A_i,i\in I_4|X_i\in A_i,i\in I_1]$$
or
$$P[X_i\in A_i,i\in I_3|X_i,i\in I_2;X_i\in A_i,i\in I_1+I_4]=P[X_i\in A_i,i\in I_3|X_i\in A_i,i\in I_1]$$
if and only if
$$\begin{array}{l}P[X_i,i\in I_2;X_i\in A_i,i\in I_3+I_4|X_i\in A_i,i\in I_1]\\
=P[X_i\in A_i,i\in I_3|X_i\in A_i,i\in I_1]\cdot P[X_i,i\in I_2;X_i\in A_i,i\in I_4|X_i\in A_i,i\in I_1].\end{array}$$

3. For fixed $A_i\in\mathscr{B}(\mathbb{R}),i\in I_1+I_4+I_5$ with $P[X_i\in A_i,i\in I_1+I_5]>0$, then
$$\begin{array}{l}P[X_i,i\in I_2;X_i\in A_i,i\in I_4|X_i,i\in I_3;X_i\in A_i,i\in I_1+I_5]\\
=P[X_i,i\in I_2;X_i\in A_i,i\in I_4|X_i\in A_i,i\in I_1]\end{array}$$
if and only if
$$\begin{array}{l}P[X_i,i\in I_2+I_3;X_i\in A_i,i\in I_4+I_5|X_i\in A_i,i\in I_1]\\
=P[X_i,i\in I_2;X_i\in A_i,i\in I_4|X_i\in A_i,i\in I_1]\times P[X_i,i\in I_3;X_i\in A_i,i\in I_5|X_i\in A_i,i\in I_1].\end{array}$$

4. For fixed $A_i\in\mathscr{B}(\mathbb{R}),i\in I_2+I_3+I_4$ with $P[X_i\in A_i,i\in I_3+I_4]>0$, then
$$P[X_i\in A_i,i\in I_2|X_i,i\in I_1;X_i\in A_i,i\in I_3+I_4]=P[X_i\in A_i,i\in I_2|X_i,i\in I_1;X_i\in A_i,i\in I_3]$$
if and only if
$$\begin{array}{l}P[X_i\in A_i,i\in I_2+I_4|X_i,i\in I_1;X_i\in A_i,i\in I_3]\\
=P[X_i\in A_i,i\in I_2|X_i,i\in I_1;X_i\in A_i,i\in I_3]\cdot P[X_i\in A_i,i\in I_4|X_i,i\in I_1;X_i\in A_i,i\in I_3].\end{array}$$

5. For fixed $A_i\in\mathscr{B}(\mathbb{R}),i\in I_3+I_4+I_5$ with $P[X_i\in A_i,i\in I_3+I_4+I_5]>0$, then
$$\begin{array}{l}P[X_i,i\in I_2;X_i\in A_i,i\in I_4|X_i,i\in I_1;X_i\in A_i,i\in I_3+I_5]\\
=P[X_i,i\in I_2;X_i\in A_i,i\in I_4|X_i,i\in I_1;X_i\in A_i,i\in I_3]\end{array}$$
or
$$\begin{array}{l}P[X_i\in A_i,i\in I_5|X_i,i\in I_1+I_2;X_i\in A_i,i\in I_3+I_4]\\
=P[X_i\in A_i,i\in I_5|X_i,i\in I_1;X_i\in A_i,i\in I_3]\end{array}$$
if and only if
$$\begin{array}{l}P[X_i,i\in I_2;X_i\in A_i,i\in I_4+I_5|X_i,i\in I_1;X_i\in A_i,i\in I_3]\\
=P[X_i,i\in I_2;X_i\in A_i,i\in I_4|X_i,i\in I_1;X_i\in A_i,i\in I_3]\\
\ \ \ \cdot P[X_i\in A_i,i\in I_5|X_i,i\in I_1;X_i\in A_i,i\in I_3].\end{array}$$

6. For fixed $A_i\in\mathscr{B}(\mathbb{R}),i\in I_4+I_5+I_6$ with $P[X_i\in A_i,i\in I_4+I_6]>0$, then
$$\begin{array}{l}P[X_i,i\in I_2;X_i\in A_i,i\in I_5|X_i,i\in I_1+I_3;X_i\in A_i,i\in I_4+I_6]\\
=P[X_i,i\in I_2;X_i\in A_i,i\in I_5|X_i,i\in I_1;X_i\in A_i,i\in I_4]\end{array}$$
if and only if
$$\begin{array}{l}P[X_i,i\in I_2+I_3;X_i\in A_i,i\in I_5+I_6|X_i,i\in I_1;X_i\in A_i,i\in I_4]\\
=P[X_i,i\in I_2;X_i\in A_i,i\in I_5|X_i,i\in I_1;X_i\in A_i,i\in I_4]\\
\ \ \ \times P[X_i,i\in I_3;X_i\in A_i,i\in I_6|X_i,i\in I_1;X_i\in A_i,i\in I_4].\end{array}$$
\end{col}

We also obtain the following corollary.

\begin{col}\label{corollary3.9.4}Let $\Lambda$ be a nonempty index set. Let $(\Omega,\mathscr{F},P)$ be a given probability space, and $(\Omega_{i},\mathscr{F}_{i}),\ i\in \Lambda$ measurable spaces. Let $X_i:(\Omega,\mathscr{F})\rightarrow(\Omega_i,\mathscr{F}_i),\ i\in \Lambda$ be random objects. Given $I_i\subset \Lambda,\ i=1,2,\cdots,6$ that are pairwise disjoint.

1. If $(X_{I_1},X_{I_2},X_{I_3},X_{I_4})$ and $X_{I_5}$ are independent, then for fixed $A_{I_i}\in\mathscr{F}_{I_i},\ i=3,4,5$ with $P[X_{I_3}\in A_{I_3},X_{I_4}\in A_{I_4},X_{I_5}\in A_{I_5}]>0$,
$$P[X_{I_2},X_{I_4}\in A_{I_4}|X_{I_1},X_{I_3}\in A_{I_3},X_{I_5}\in A_{I_5}]=P[X_{I_2},X_{I_4}\in A_{I_4}|X_{I_1},X_{I_3}\in A_{I_3}]$$
and
$$P[X_{I_5}\in A_{I_5}|X_{I_1},X_{I_2};X_{I_3}\in A_{I_3},X_{I_4}\in A_{I_4}]=P[X_{I_5}\in A_{I_5}]$$
given the existence of $P[X_{I_2},X_{I_4}\in A_{I_4}|X_{I_1},X_{I_3}\in A_{I_3}]$.

2. If $(X_{I_1},X_{I_3},X_{I_4},X_{I_6})$ and $(X_{I_2},X_{I_5})$ are independent, then for fixed $A_{I_i}\in\mathscr{F}_{I_i},\ i=4,5,6$ with $P[X_{I_4}\in A_{I_4},X_{I_5}\in A_{I_5}]>0$,
$$P[X_{I_3},X_{I_6}\in A_{I_6}|X_{I_1},X_{I_2};X_{I_4}\in A_{I_4},X_{I_5}\in A_{I_5}]=P[X_{I_3},X_{I_6}\in A_{I_6}|X_{I_1},X_{I_4}\in A_{I_4}]$$
given the existence of $P[X_{I_3},X_{I_6}\in A_{I_6}|X_{I_1},X_{I_4}\in A_{I_4}]$.
\end{col}

From Corollary \ref{corollary3.9.4} and Theorem \ref{theorem2.11.3} the corollary below follows.

\begin{col}\label{corollary3.9.5}  Let $(\Omega,\mathscr{F},P)$ be a given probability space. Let $X_i,\ i\in \mathbb{N}$ be random variables on $(\Omega,\mathscr{F},P)$. Given $I_i\subset \mathbb{N},\ i=1,2,\cdots,6$ that are pairwise disjoint.

1. If $(X_i,i\in I_1+I_2+I_3+I_4)$ and $(X_i,i\in I_5)$ are independent, then for fixed $A_i\in\mathscr{B}(\mathbb{R}),i\in I_3+I_4+I_5$ with $P[X_i\in A_i,i\in I_3+I_4+I_5]>0$,
$$\begin{array}{l}P[X_i,i\in I_2;X_i\in A_i,i\in I_4|X_i,i\in I_1;X_i\in A_i,i\in I_3+I_5]\\
=P[X_i,i\in I_2;X_i\in A_i,i\in I_4|X_i,i\in I_1;X_i\in A_i,i\in I_3]\end{array}$$
and
$$\begin{array}{l}P[X_i\in A_i,i\in I_5|X_i,i\in I_1+I_2;X_i\in A_i,i\in I_3+I_4]\\
=P[X_i\in A_i,i\in I_5].\end{array}$$

2. If $(X_i,i\in I_1+I_3+I_4+I_5)$ and $(X_i,i\in I_2+I_5)$ are independent, then for fixed $A_i\in\mathscr{B}(\mathbb{R}),i\in I_4+I_5+I_6$ with $P[X_i\in A_i,i\in I_4+I_5]>0$,
$$\begin{array}{l}P[X_i,i\in I_3;X_i\in A_i,i\in I_6|X_i,i\in I_1+I_2;X_i\in A_i,i\in I_4+I_5]\\
=P[X_i,i\in I_1;X_i\in A_i,i\in I_3|X_i,i\in I_1;X_i\in A_i,i\in I_4].\end{array}$$
\end{col}

\section{Conditional Probability of Co-occurrence Given The Density of Probability of Co-occurrence with respect to Product Measure}\label{section4}

By Proposition 4.4.3 in \cite{yan:2004} we have the product probability space $(\Omega_I,\mathscr{F}_I,P_I)$ of $(\Omega_i,\mathscr{F}_i,P[X_i]),\ i\in I$, where $P_I=\prod_{i\in I}P[X_i]$ is the product probability measure of $P[X_i],\ i\in I$. In addition, let $P[X_I]$ be the probability measure on $(\Omega_I,\mathscr{F}_I)$ induced by $X_I$, which is also called the probability of co-occurrence of $X_i,i\in I$ provided $I$ is a countable index set. Then the the density of the probability $P[X_I]$ is defined with respect to their product probability measure $P_I$ as follows.

\begin{defn}\label{definition4.1} Let $(\Omega,\mathscr{F},P)$, $(\Omega_{i},\mathscr{F}_{i}),\ i\in I$; $X_i(\omega),\ i\in I$; $P_I$ and $P[X_I]$ be the same as above. If there is a nonnegative $\mathscr{F}_I$-measurable function $f:\Omega_I\rightarrow \mathbb{R}^+$ such that
$$P[X_I](A_I)=\int_{A_I}f(\omega_I)P_I(d\omega_I),\ \ \forall A_I\in\mathscr{F}_I,$$
then $f(\omega_I)$ is defined as the density of the probability $P[X_I]$ with respect to their product probability measure $P_I$, i.e., the density of $P[X_I]$ with respect to $\prod_{i\in I}P[X_i]$.
\end{defn}

The existence and uniqueness of the density of $P[X_I]$ with respect to $\prod_{i\in I}P[X_i]$ is given in Theorem \ref{theorem4.1}.

\begin{thm}\label{theorem4.1}
\begin{enumerate}
\item If the density of $P[X_I]$ with respect to $P_I$ exists, then it is unique a.e.$\{P_I\}$, and $P[X_I]<<P_I$.\\
\item If $P[X_I]<<P_I$, then the density of $P[X_I]$ with respect to $P_I$ exists.
\end{enumerate}
\end{thm}

Generally the density of $P[X_I]$ with respect to $P_I$ need not exist. For example, let $\Omega=[0,1]$, $\mathscr{F}=\mathscr{B}([0,1])=[0,1]\cap\mathscr{B}(\mathbb{R})$ and $P=\lambda$, where $\lambda$ is Lebesgue measure. Then $(\Omega,\mathscr{F},P)=([0,1],\mathscr{B}([0,1]),\lambda)$ is a probability space. Let $\Omega_1=\Omega_2=\Omega=[0,1]$, $\mathscr{F}_1=\mathscr{F}_2=\mathscr{F}=\mathscr{B}([0,1])$. Let $X_1$ and $X_2$ be the identical random object $I_0(x)$ on $([0,1],\mathscr{B}([0,1]))$, i.e., $X_1(x)=X_2(x)=x,\ x\in [0,1]$. Then $P[X_1]=P[X_2]=\lambda$. It is evident that
$$\begin{array}{l}\displaystyle (\Omega_1\times\Omega_2,\mathscr{F}_1\times\mathscr{F}_2,P[X_1]\times P[X_2])=([0,1]\times[0,1],\mathscr{B}([0,1])\times\mathscr{B}([0,1]),\lambda\times\lambda)\\
\displaystyle =([0,1]\times[0,1],\mathscr{B}([0,1]\times[0,1]),\lambda\times\lambda).\end{array}$$
Set $X_{12}(x)=(X_1(x),X_2(x))=(x,x)$, then $X_{12}(x)$ is a random object from $([0,1],\mathscr{B}([0,1]))$ to $([0,1]\times[0,1],\mathscr{B}([0,1])\times\mathscr{B}([0,1]))$. Let $E=\{(x,x):x\in [0,1]\}$, then $E\in \mathscr{B}([0,1])\times\mathscr{B}([0,1])$, or $E\in \mathscr{B}([0,1]\times[0,1])$. Obviously, $P[X_1]\times P[X_2](E)=\lambda\times\lambda(E)=0$.
 But $P[X_{12}](E)=P[X_{12}\in E]=\lambda([0,1])=1\neq 0$. Thus $P[X_{12}]$(or $P[X_1,X_2]$) is not absolutely continuous with respect to $P[X_1]\times P[X_2]$. Therefore the density of $P[X_1,X_2]$ with respect to $P[X_1]\times P[X_2]$ doesn't exist by Theorem \ref{theorem4.1}.

\begin{thm}\label{theorem4.2}Let $(\Omega,\mathscr{F},P)$, $(\Omega_{i},\mathscr{F}_{i}),\ i\in I$; $X_i(\omega),\ i\in I$; $(\Omega_{i},\mathscr{F}_{i},P[X_i]),\ i\in I$; $(\Omega_{I},\mathscr{F}_{I})$ and $P_I$ be the same as above. If the density $f_I(\omega_I)$ of $P[X_I]$ with respect to $P_I$ exists, then so does the density $f_{I_1}(\omega_{I_1})$ of $P[X_{I_1}]$ with respect to $P_{I_1}$ for $\forall I_1\subset I$, and
$$f_{I_1}(\omega_{I_1})=\left\{\begin{array}{l}\displaystyle\int_{\Omega_{I^c_1}}f_I(\omega_{I_1},\omega_{I^c_1})P_{I^c_1}(d\omega_{I^c_1}),\ \ \omega_{I_1}\in E^c,\\\\
\displaystyle 0,\ \ \ \omega_{I_1}\in E,\end{array}\right.$$
where $I^c_1=I\backslash I_1$,
$$E=\{\omega_{I_1}:\int_{\Omega_{I^c_1}}f_I(\omega_{I_1},\omega_{I^c_1})P_{I^c_1}(d\omega_{I^c_1})=+\infty\} \ with\  P_{I_1}(E)=0,$$
and $E^c=\Omega_{I_1}\backslash E$.
\end{thm}

By Theorem \ref{theorem4.2} and Fubini's Theorem we obtain the following corollary.

\begin{col}\label{corollary4.2.1}Let $(\Omega,\mathscr{F},P)$, $(\Omega_{i},\mathscr{F}_{i}),\ i\in I$; $X_i(\omega),\ i\in I$; $(\Omega_{i},\mathscr{F}_{i},P[X_i]),\ i\in I$; $(\Omega_{I},\mathscr{F}_{I})$ and $P_I$ be the same as above. If $X_i(\omega),\ i\in I$  are independent, and the density $f_I(\omega_I)$ of $P[X_I]$ with respect to $P_I$ exists, then $f_I(\omega_I)=\prod_{j\in J}f_{I_j}(\omega_{I_j})$, where $J$ a finite index set; $\forall I_j\subset I,\ j\in J$; $I_j,\ j\in J$ pairwise disjoint, $I=\sum_{j\in J}I_j$; and $f_{I_j}(\omega_{I_j})$ the density of $P[X_{I_j}]$ with respect to $P_{I_j}$, $j\in J$.
\end{col}

\begin{thm}\label{theorem4.3}Let $(\Omega,\mathscr{F},P)$, $(\Omega_{i},\mathscr{F}_{i}),\ i\in I$; $X_i(\omega),\ i\in I$; $(\Omega_{i},\mathscr{F}_{i},P[X_i]),\ i\in I$; $(\Omega_{I},\mathscr{F}_{I})$ and $P_I$ be the same as above. If the density $f_I(\omega_I)$ of $P[X_I]$ with respect to $P_I$ exists, then $P[X_{I_2}|X_{I_1}](\omega_{I_1},A_{I_2})$ exists for $\forall I_1,I_2\subset I$, where $I=I_1+I_2$, $I_1\cap I_2=\emptyset$, and
$$P[X_{I_2}|X_{I_1}](\omega_{I_1},A_{I_2})=\left\{\begin{array}{l}\displaystyle\frac{\int_{A_{I_2}}f_I(\omega_{I_1},\omega_{I_2})P_{I_2}(d\omega_{I_2})}
{\int_{\Omega_{I_2}}f_I(\omega_{I_1},\omega_{I_2})P_{I_2}(d\omega_{I_2})},\ \ \omega_{I_1}\in E^c,\\\\
\displaystyle 0,\ \ \omega_{I_1}\in E,\end{array}\right.$$
where $E=E_1\cup E_2$,
$$E_1=\{\omega_{I_1}:\int_{\Omega_{I_2}}f_I(\omega_{I_1},\omega_{I_2})P_{I_2}(d\omega_{I_2})=+\infty\}\ with\ P_{I_1}(E_1)=0,$$
$$E_2=\{\omega_{I_1}:\int_{\Omega_{I_2}}f_I(\omega_{I_1},\omega_{I_2})P_{I_2}(d\omega_{I_2})=0\},$$
and $E^c=\Omega_{I_1}\backslash E$.
\end{thm}

By Corollary \ref{corollary3.3.1} and Theorem \ref{theorem4.3}, we have the following corollary.

\begin{col}\label{corollary4.3.1}Let $(\Omega,\mathscr{F},P)$, $(\Omega_{i},\mathscr{F}_{i}),\ i\in I$; $X_i(\omega),\ i\in I$; $(\Omega_{i},\mathscr{F}_{i},P[X_i]),\ i\in I$; $(\Omega_{I},\mathscr{F}_{I})$ and $P_I$ be the same as above. If the density $f_I(\omega_I)$ of $P[X_I]$ with respect to $P_I$ exists, then so does $P[X_{I_2}|X_{I_1}](\omega_{I_1},A_{I_2})$ for $\forall I_1,\ I_2\subset I,\ I_1\cap I_2=\emptyset$ , and
$$P[X_{I_2}|X_{I_1}](\omega_{I_1},A_{I_2})=\left\{\begin{array}{l}\displaystyle\frac{\int_{A_{I_2}\times\Omega_{I_3}}f_I(\omega_{I_1},\omega_{I^c_1})
P_{I^c_1}(d\omega_{I^c_1})}{\int_{\Omega_{I^c_1}}f_I(\omega_{I_1},\omega_{I^c_1})
P_{I^c_1}(d\omega_{I^c_1})},\ \ \omega_{I_1}\in E^c,\\\\
\displaystyle 0,\ \ \omega_{I_1}\in E,\end{array}\right.$$
where $E=E_1\cup E_2$,
$$E_1=\{\omega_{I_1}:\int_{\Omega_{I^c_1}}f_I(\omega_{I_1},\omega_{I^c_1})P_{I^c_1}(d\omega_{I^c_1})=+\infty\}\ with\ P_{I_1}(E_1)=0,$$
$$E_2=\{\omega_{I_1}:\int_{\Omega_{I^c_1}}f_I(\omega_{I_1},\omega_{I^c_1})P_{I^c_1}(d\omega_{I^c_1})=0\},$$
$E^c=\Omega_{I_1}\backslash E$, $I_3=I\backslash (I_1+I_2)$ and $I^c_1=I\backslash I_1$.
\end{col}

Corollary \ref{corollary4.3.2} below comes from Corollary \ref{corollary3.4.1} 1., Corollary \ref{corollary4.3.1}.

\begin{col}\label{corollary4.3.2}Let $(\Omega,\mathscr{F},P)$, $(\Omega_{i},\mathscr{F}_{i}),\ i\in I$; $X_i(\omega),\ i\in I$; $(\Omega_{i},\mathscr{F}_{i},P[X_i]),\ i\in I$; $(\Omega_{I},\mathscr{F}_{I})$ and $P_I$ be the same as above. If the density $f_I(\omega_I)$ of $P[X_I]$ with respect to $P_I$ exists, then so does $P[X_{I_2},X_{I_4}\in A_{I_4}|X_{I_1},X_{I_3}\in A_{I_3}](\omega_{I_1},A_{I_2})$ for $I_j\subset I,\ j=1,2,3,4$ and fixed $A_{I_j}\in\mathscr{F}_{I_j},\ j=3,4$, where $I_j,j=1,2,3,4$ are pairwise disjoint. And
$$\begin{array}{l}\displaystyle P[X_{I_2},X_{I_3}\in A_{I_3}|X_{I_1},X_{I_4}\in A_{I_4}](\omega_{I_1},A_{I_2})\\
\displaystyle =\left\{\begin{array}{l}\displaystyle \frac{\int_{(\prod_{j=2,3,4}A_{I_j})\times \Omega_{I_5}}f_I(\omega_{I_1},\omega_{I^c_1})P_{I^c_1}(d\omega_{I^c_1})}{\int_{A_{I_4}\times(\prod_{j=2,3,5}\Omega_{I_j})}f_I(\omega_{I_1},\omega_{I^c_1})
P_{I^c_1}(d\omega_{I^c_1})},\ \omega_{I_1}\in E^c,\\\\
\displaystyle 0,\ \ \ \ \ \ \ \ \ \omega_{I_1}\in E,\end{array}\right.
\end{array}$$
where $E=E_1\cup E_2$,
$$E_1=\{\omega_{I_1}:\int_{\Omega_{I^c_1}}f_I(\omega_{I_1},\omega_{I^c_1})P_{I^c_1}(d\omega_{I^c_1})=+\infty\}\ with\ P_{I_1}(E_1)=0,$$
$$E_2=\{\omega_{I_1}:\int_{A_{I_4}\times(\prod_{j=2,3,5}\Omega_{I_j})}f_I(\omega_{I_1},\omega_{I^c_1})P_{I^c_1}(d\omega_{I^c_1})=0\},$$
$E^c=\Omega_{I_1}\backslash E$, $I_5=I\backslash (I_1+I_2+I_3+I_4)$ and $I^c_1=I\backslash I_1$.
\end{col}

Let $(\Omega,\mathscr{F},P)$ be a fixed probability space, $(\Omega_{i},\mathscr{F}_{i},\mu_i),\ i\in I$ $\sigma$-finite measure spaces, where $I$ is a nonempty finite index set. Let $X_i:(\Omega,\mathscr{F})\rightarrow(\Omega_i,\mathscr{F}_i),\ i\in I$ be random objects. By Proposition 4.3.7 in \cite{yan:2004} we have the product measure space $(\Omega_I,\mathscr{F}_I,\mu_I)$ of $(\Omega_i,\mathscr{F}_i,\mu_i),\ i\in I$, where $\mu_I=\prod_{i\in I}\mu_i$ is the product measure of $\mu_i,\ i\in I$. In addition, let $P[X_I]=P[X_i,i\in I]$ be the probability measure on $(\Omega_I,\mathscr{F}_I)$ induced by $X_I$. Then we give the definition of the density of the probability of co-occurrence of $X_i,i\in I$ with respect to the product measure $\mu_I$.

\begin{defn}\label{definition4.2} Let $(\Omega,\mathscr{F},P)$, $(\Omega_{i},\mathscr{F}_{i},\mu_i),\ i\in I$; $X_i(\omega),\ i\in I$; $(\Omega_{I},\mathscr{F}_{I})$, $\mu_I$ and
$P[X_I]=P[X_i,i\in I]$ be the same as above. If there is a nonnegative $\mathscr{F}_I$-measurable function $f:\Omega_I\rightarrow \mathbb{R}^+$ such that
$$\int_{A_I}f(\omega_I)\mu_I(d\omega_I)=P[X_I](A_I),\ \ \forall A_I\in\mathscr{F}_I,$$
then $f(\omega_I)$ is defined as the density of the probability of co-occurrence of $X_i,i\in I$ with respect to the product measure $\mu_I$, i.e., the density of $P[X_I]$ with respect to $\prod_{i\in I}\mu_i$.
\end{defn}

\begin{comm}\label{comment4.2.1}Let $I$ be a finite index set with the cardinal number $|I|$. Let $X_i(x),\ i\in I$ be random variables on $(\Omega,\mathscr{F},P)$ and $(\mathbb{R}^I,\mathscr{B}(\mathbb{R}^I),\lambda_I)$ the $|I|$-dimensional Euclidean Borel measurable space with Lebesgue measure $\lambda_I$. Then the density $f_I(x_I)$ of $P[X_I]$ with respect to $\lambda_I$ is the joint density of  the random vector $X_I$.
\end{comm}

Regarding the existence and uniqueness of the density of $P[X_I]$ with respect to $\prod_{i\in I}\mu_i$ we have the following Theorem \ref{theorem4.4}.

\begin{thm}\label{theorem4.4}
1. If the density of $P[X_I]$ with respect to $\mu_I$ exists, then it is unique a.e.$\{\mu_I\}$, and $P[X_I]<<\mu_I$.

2. If $P[X_I]<<\mu_I$, then the density of $P[X_I]$ with respect to $\mu_I$ exists.
\end{thm}

The proof of Theorem \ref{theorem4.4} is similar to that of Theorem \ref{theorem4.1}.

\begin{thm}\label{theorem4.5}Let $(\Omega,\mathscr{F},P)$, $(\Omega_{i},\mathscr{F}_{i},\mu_i),\ i\in I$; $X_i(\omega),\ i\in I$; $(\Omega_{I},\mathscr{F}_{I})$ and $\mu_I$ be the same as above. If the density $f_I(\omega_I)$ of $P[X_I]$ with respect to $\mu_I$ exists, then so does the density $f_{I_1}(\omega_{I_1})$ of $P[X_{I_1}]$ with respect to $\mu_{I_1}$ for $\forall I_1\subset I$, and
$$f_{I_1}(\omega_{I_1})=\left\{\begin{array}{l}\displaystyle\int_{\Omega_{I^c_1}}f_I(\omega_{I_1},\omega_{I^c_1})\mu_{I^c_1}(d\omega_{I^c_1}),\ \ \omega_{I_1}\in E^c,\\\\
\displaystyle 0,\ \ \ \omega_{I_1}\in E,\end{array}\right.$$
where $I^c_1=I\backslash I_1$,
$$E=\{\omega_{I_1}:\int_{\Omega_{I^c_1}}f_I(\omega_{I_1},\omega_{I^c_1})\mu_{I^c_1}(d\omega_{I^c_1})=+\infty\} \ with\  \mu_{I_1}(E)=0,$$
and $E^c=\Omega_{I_1}\backslash E$.
\end{thm}

Furthermore, by Theorem \ref{theorem4.5} we obtain the following corollary.

\begin{col}\label{corollary4.5.1}Let $(\Omega,\mathscr{F},P)$, $(\Omega_{i},\mathscr{F}_{i},\mu_i),\ i\in I$; $X_i(\omega),\ i\in I$; $(\Omega_{I},\mathscr{F}_{I})$ and $\mu_I$ be the same as above. If $X_i(\omega),\ i\in I$  are independent, and the density $f_I(\omega_I)$ of $P[X_I]$ with respect to $\mu_I$ exists, then $f_I(\omega_I)=\prod_{j\in J}f_{I_j}(\omega_{I_j})$, where $J$ a finite index set; $\forall I_j\subset I,\ j\in J$; $I_j,\ j\in J$ pairwise disjoint, $I=\sum_{j\in J}I_j$; and $f_{I_j}(\omega_{I_j})$ the density of $P[X_{I_j}]$ with respect to $\mu_{I_j}$, $j\in J$.
\end{col}

\begin{thm}\label{theorem4.6}Let $(\Omega,\mathscr{F},P)$, $(\Omega_{i},\mathscr{F}_{i},\mu_i),\ i\in I$; $X_i(\omega),\ i\in I$; $(\Omega_{I},\mathscr{F}_{I})$ and $\mu_I$ be the same as above. If the density $f_I(\omega_I)$ of $P[X_I]$ with respect to $\mu_I$ exists, then $P[X_{I_2}|X_{I_1}](\omega_{I_1},A_{I_2})$ exists for $\forall I_1,I_2\subset I$, where $I=I_1+I_2$, $I_1\cap I_2=\emptyset$, and
$$P[X_{I_2}|X_{I_1}](\omega_{I_1},A_{I_2})=\left\{\begin{array}{l}\displaystyle\frac{\int_{A_{I_2}}f_I(\omega_{I_1},\omega_{I_2})\mu_{I_2}(d\omega_{I_2})}
{\int_{\Omega_{I_2}}f_I(\omega_{I_1},\omega_{I_2})\mu_{I_2}(d\omega_{I_2})},\ \ \omega_{I_1}\in E^c,\\\\
\displaystyle 0,\ \ \omega_{I_1}\in E,\end{array}\right.$$
where $E=E_1\cup E_2$,
$$E_1=\{\omega_{I_1}:\int_{\Omega_{I_2}}f_I(\omega_{I_1},\omega_{I_2})\mu_{I_2}(d\omega_{I_2})=+\infty\}\ with\ \mu_{I_1}(E_1)=0,$$
$$E_2=\{\omega_{I_1}:\int_{\Omega_{I_2}}f_I(\omega_{I_1},\omega_{I_2})\mu_{I_2}(d\omega_{I_2})=0\},$$
and $E^c=\Omega_{I_1}\backslash E$.
\end{thm}

The proof of Theorem \ref{theorem4.6} is similar to that of Theorem \ref{theorem4.3}. In addition, by Corollary \ref{corollary3.3.1} and Theorem \ref{theorem4.6}, we have the following corollary.

\begin{col}\label{corollary4.6.1}Let $(\Omega,\mathscr{F},P)$, $(\Omega_{i},\mathscr{F}_{i},\mu_i),\ i\in I$; $X_i(\omega),\ i\in I$; $(\Omega_{I},\mathscr{F}_{I})$ and $\mu_I$ be the same as above. If the density $f_I(\omega_I)$ of $P[X_I]$ with respect to $\mu_I$ exists, then so does $P[X_{I_2}|X_{I_1}](\omega_{I_1},A_{I_2})$ for $\forall I_1,\ I_2\subset I,\ I_1\cap I_2=\emptyset$ , and
$$P[X_{I_2}|X_{I_1}](\omega_{I_1},A_{I_2})=\left\{\begin{array}{l}\displaystyle\frac{\int_{A_{I_2}\times\Omega_{I_3}}f_I(\omega_{I_1},\omega_{I^c_1})
\mu_{I^c_1}(d\omega_{I^c_1})}{\int_{\Omega_{I^c_1}}f_I(\omega_{I_1},\omega_{I^c_1})\mu_{I^c_1}(d\omega_{I^c_1})},\ \ \omega_{I_1}\in E^c,\\
\displaystyle 0,\ \ \omega_{I_1}\in E,\end{array}\right.$$
where $E=E_1\cup E_2$,
$$E_1=\{\omega_{I_1}:\int_{\Omega_{I^c_1}}f_I(\omega_{I_1},\omega_{I^c_1})\mu_{I^c_1}(d\omega_{I^c_1})=+\infty\}\ with\ \mu_{I_1}(E_1)=0,$$
$$E_2=\{\omega_{I_1}:\int_{\Omega_{I^c_1}}f_I(\omega_{I_1},\omega_{I^c_1})\mu_{I^c_1}(d\omega_{I^c_1})=0\},$$
$E^c=\Omega_{I_1}\backslash E$, $I_3=I\backslash (I_1+I_2)$ and $I^c_1=I\backslash I_1$.
\end{col}

Corollary \ref{corollary4.6.2} follows from Corollary \ref{corollary3.4.1} and Corollary \ref{corollary4.6.1}.

\begin{col}\label{corollary4.6.2}Let $(\Omega,\mathscr{F},P)$, $(\Omega_{i},\mathscr{F}_{i},\mu_i),\ i\in I$; $X_i(\omega),\ i\in I$; $(\Omega_{I},\mathscr{F}_{I})$ and $\mu_I$ be the same as above. If the density $f_I(\omega_I)$ of $P[X_I]$ with respect to $\mu_I$ exists, then so does $P[X_{I_2},X_{I_4}\in A_{I_4}|X_{I_1},X_{I_3}\in A_{I_3}](\omega_{I_1},A_{I_2})$ for $I_j\subset I,\ j=1,2,3,4$ and fixed $A_{I_j}\in\mathscr{F}_{I_j},\ j=3,4$, where $I_j,j=1,2,3,4$ are pairwise disjoint. And
$$\begin{array}{l}\displaystyle P[X_{I_2},X_{I_3}\in A_{I_3}|X_{I_1},X_{I_4}\in A_{I_4}](\omega_{I_1},A_{I_2})\\
\displaystyle =\left\{\begin{array}{l}\displaystyle \frac{\int_{(\prod_{j=2,3,4}A_{I_j})\times \Omega_{I_5}}f_I(\omega_{I_1},\omega_{I^c_1})\mu_{I^c_1}(d\omega_{I^c_1})}{\int_{A_{I_4}\times(\prod_{j=2,3,5}\Omega_{I_j})}f_I(\omega_{I_1},\omega_{I^c_1})
\mu_{I^c_1}(d\omega_{I^c_1})},\ \omega_{I_1}\in E^c,\\
\displaystyle 0,\ \ \ \ \ \ \ \ \ \omega_{I_1}\in E,\end{array}\right.
\end{array}$$
where $E=E_1\cup E_2$,
$$E_1=\{\omega_{I_1}:\int_{\Omega_{I^c_1}}f_I(\omega_{I_1},\omega_{I^c_1})\mu_{I^c_1}(d\omega_{I^c_1})=+\infty\}\ with\ \mu_{I_1}(E_1)=0,$$
$$E_2=\{\omega_{I_1}:\int_{A_{I_4}\times(\prod_{j=2,3,5}\Omega_{I_j})}f_I(\omega_{I_1},\omega_{I^c_1})\mu_{I^c_1}(d\omega_{I^c_1})=0\},$$
$E^c=\Omega_{I_1}\backslash E$, $I_5=I\backslash (I_1+I_2+I_3+I_4)$ and $I^c_1=I\backslash I_1$.
\end{col}

\begin{col}\label{corollary4.6.3}Let $(\Omega,\mathscr{F},P)$ be a given probability space. Let $I$ be a finite index set and $X_i(\omega),\ i\in I$ random variables on $(\Omega,\mathscr{F},P)$. If the density $f_I(x_I)$ of $P[X_I]$ with respect to $\lambda_I$ exists, where $\lambda_I$ is Lebesgue measure on $(\mathbb{R}^I,\mathscr{B}(\mathbb{R}^I))$, then so does $P[X_{I_2},X_{I_4}\in A_{I_4}|X_{I_1},X_{I_3}\in A_{I_3}](x_{I_1},A_{I_2})$ for $I_j\subset I,\ j=1,2,3,4$ and fixed $A_{I_j}\in\mathscr{B}(\mathbb{R}^{I_j}),\ j=3,4$, where $I_j,j=1,2,3,4$ are pairwise disjoint. And
$$\begin{array}{l}\displaystyle P[X_{I_2},X_{I_3}\in A_{I_3}|X_{I_1},X_{I_4}\in A_{I_4}](x_{I_1},A_{I_2})\\
\displaystyle =\left\{\begin{array}{l}\displaystyle \frac{\int_{(\prod_{j=2,3,4}A_{I_j})\times \mathbb{R}^{I_5}}f_I(x_{I_1},x_{I^c_1})\lambda_{I^c_1}(dx_{I^c_1})}{\int_{A_{I_4}\times(\prod_{j=2,3,5}\mathbb{R}^{I_j})}f_I(x_{I_1},x_{I^c_1})
\lambda_{I^c_1}(dx_{I^c_1})},\ x_{I_1}\in E^c,\\\\
\displaystyle 0,\ \ \ \ \ \ \ \ \ x_{I_1}\in E,\end{array}\right.
\end{array}$$
where $E=E_1\cup E_2$,
$$E_1=\{x_{I_1}:\int_{\mathbb{R}^{I^c_1}}f_I(x_{I_1},x_{I^c_1})\lambda_{I^c_1}(dx_{I^c_1})=+\infty\}\ with\ \lambda_{I_1}(E_1)=0,$$
$$E_2=\{x_{I_1}:\int_{A_{I_4}\times(\prod_{j=2,3,5}\mathbb{R}^{I_j})}f_I(x_{I_1},x_{I^c_1})\lambda_{I^c_1}(dx_{I^c_1})=0\},$$
$E^c=\mathbb{R}^{I_1}\backslash E$, $I_5=I\backslash (I_1+I_2+I_3+I_4)$ and $I^c_1=I\backslash I_1$.

\end{col}

By Proposition 4.3.7 in \cite{yan:2004} we have the product measure space $(\Omega_I,\mathscr{F}_I,\mu_I)$ of $(\Omega_i,\mathscr{F}_i,\mu_i),\ i\in I$, where $\mu_I=\prod_{i\in I}\mu_i$ is the product measure of $\mu_i,\ i\in I$. In addition, let $P[X_i],\ i\in I$ be the probability measure on $(\Omega_i,\mathscr{F}_i)$ induced by $X_i$, respectively. Let $P[X_I]=P[X_i,i\in I]$ be the probability measure on $(\Omega_I,\mathscr{F}_I)$ induced by $X_I$, and $P_I=\prod_{i\in I}P[X_i]$ the product probability measure of $P[X_i],\ i\in I$ on $(\Omega_I,\mathscr{F}_I)$. Then we give the relation of the density $f_{P_I}(\omega_I)$ with the density $f_{\mu_I}(\omega_I)$, where $f_{P_I}(\omega_I)$ is the density of $P[X_I]$ with respect to the product probability measure $P_I$ while $f_{\mu_I}(\omega_I)$  is the density of $P[X_I]$ with respect to the product measure $\mu_I$.

\begin{lem}\label{lemma4.3}Let $(\Omega,\mathscr{F},P)$, $(\Omega_{i},\mathscr{F}_{i},\mu_i),\ i\in I$; $X_i(\omega),\ i\in I$; $P[X_i],\ i\in I$;
$(\Omega_I,\mathscr{F}_I)$, $P_I=\prod_{i\in I}P[X_i]$ and $\mu_I=\prod_{i\in I}\mu_i$ be the same as above. If there exist the densities $f_i(\omega_i),\ i\in I$ of $P[X_i]$ with respect to $\mu_i$, then so does the density $f_I(\omega_I)$ of $P_I$ with respect to $\mu_I$ such that
$$P_I(A_I)=\int_{A_I}f_I(\omega_I)\mu_I(d\omega_I),\ \forall A_I\in\mathscr{F}_I,$$
and $f_I(\omega_I)=\prod_{i\in I}f_i(\omega_i)$. Conversely if there exists the density $f_I(\omega_I)$ of $P_I$ with respect to $\mu_I$, then so do
the densities $f_i(\omega_i),\ i\in I$ of $P[X_i]$ with respect to $\mu_i$, and there exists $E\in \mathscr{F}_I$ with $\mu_I(E)=0$ such that
$$f_i(\omega_i)=\int_{\Omega_{I^c_i}}\mathds{1}_{E^c}\cdot f_I(\omega_i,\omega_{I^c_i})\mu_{I^c_i}(d\omega_{I^c_i}),\ i\in I,$$
where $I^c_i=I\backslash\{i\}$, $E^c=\Omega_I\backslash E$ and $\mathds{1}_{E^c}$ is the indicator of $E^c$. And
$$f_I(\omega_I)=\prod_{i\in I}f_i(\omega_i),\ \ a.e.\{\mu_I\}.$$
\end{lem}

\begin{thm}\label{theorem4.7}Let $(\Omega,\mathscr{F},P)$, $(\Omega_{i},\mathscr{F}_{i},\mu_i),\ i\in I$; $X_i(\omega),\ i\in I$; $P[X_i],\ i\in I$; $(\Omega_{I},\mathscr{F}_{I})$, $\mu_I$, $P[X_I]$, $P_I$ be the same as above. If there exist the density $f_{P_I}(\omega_I)$ of $P[X_I]$ with respect to $P_I$ and the densities $f_i(\omega_i),\ i\in I$ of $P[X_i]$ with respect to $\mu_i$, then so does the density $f_{\mu_I}(\omega_I)$ of $P[X_I]$ with respect to $\mu_I$, and $f_{\mu_I}(\omega_I)=f_{P_I}(\omega_I)\cdot\prod_{i\in I}f_i(\omega_i)$.
\end{thm}

\section{The Integral with respect to Probability and Conditional Probability of Co-occurrence}\label{section5}

According to  Definitions \ref{definition2.5}, \ref{definition2.6}, \ref{definition2.10}, \ref{definition2.11} and \ref{definition2.12}, the probability of co-occurrence and conditional probability of co-occurrence are not necessarily probability measures while they are finite measures. And the integrals with respect to them have been concerned a little in the above sections. This section further investigates the integrals with respect to them.

Let $(\Omega,\mathscr{F},P)$ be a given probability space, and let $(\Omega_i,\mathscr{F}_i),\ i=1,2$ be measurable spaces. Let $X_i:(\Omega,\mathscr{F})\rightarrow(\Omega_i,\mathscr{F}_i),\ i=1,2$ be random objects, and $P[X_i]$ the probability on $(\Omega_i,\mathscr{F}_i)$ induced by $X_i$, $i=1,2$. Then $(\Omega_i,\mathscr{F}_i,P[X_i]),\ i=1,2$ are probability spaces. Let $Y(\omega_2)$ be a random variable on $(\Omega_2,\mathscr{F}_2,P[X_2])$ and $A_1\in\mathscr{F}_1$. Note that, we only consider such a random variable $Y(\omega_2)$ with $E(Y)\in\mathbb{R}$. So $Y(\omega_2)$ may be an extended random variable with $E(Y)\in\mathbb{R}$. Hence if $Y(\omega_2)$ is finite a.e.$\{P[X_2]\}$, then $Y(\omega_2)$ is also regarded as a random variable.

\begin{defn}\label{definition5.1}Let $(\Omega,\mathscr{F},P)$, $(\Omega_i,\mathscr{F}_i),\ i=1,2$; $X_i(\omega),\ i=1,2$; $P[X_i],\ i=1,2$; $Y(\omega_2)$ and $A_1$ be the same as above. Then the integral of $Y(\omega_2)$ with respect to $P[X_2,X_1\in A_1]$ is denoted by $E_{[X_2,X_1\in A_1]}(Y)$, i.e.,
$$E_{[X_2,X_1\in A_1]}(Y)=\int_{\Omega_2}Y(\omega_2)P[X_2,X_1\in A_1](d\omega_2).$$
\end{defn}

\begin{comm}\label{comment5.1.1} (i) We take
$$E_{[X_2]}(Y):=\int_{\Omega_2}Y(\omega_2)P[X_2](d\omega_2),$$
which is the expectation of $Y(\omega_2)$.

(ii) In Definition \ref{definition5.1}, let $A_1=\Omega_1$, then
$$E_{[X_2,X_1\in \Omega_1]}(Y)=E_{[X_2]}(Y).$$

(iii) Since $P[X_2,X_1\in A_1]\leq P[X_2]$ for each fixed $A_1\in\mathscr{F}_1$, it is true that $$E_{[X_2]}(Y)\in \mathbb{R}\Rightarrow E_{[X_2,X_1\in A_1]}(Y)\in \mathbb{R}.$$

(iv) Let $I_0$ be the identity random object $I_{\mathscr{F}}$ from $(\Omega,\mathscr{F})$ to $(\Omega,\mathscr{F})$, $Y(\omega)$ a random variable on $(\Omega,\mathscr{F},P)$ and $A\in\mathscr{F}$, then $E_{[I_0]}(Y)=E(Y)$, and we take $$E_{[A]}(Y):=E_{[I_0,A]}(Y)=E_{[I_0,I_0\in A]}(Y)=\int_{\Omega}Y(\omega)P[I_0,A](d\omega).$$
\end{comm}

\begin{thm}\label{theorem5.1}Let $(\Omega,\mathscr{F},P)$ be a given probability space, let $Y(\omega)$ be a random variable on $(\Omega,\mathscr{F},P)$ with its expectation $E(Y)\in \mathbb{R}$, and let $I_0$ be the identical random object $I_{\mathscr{F}}$ from $(\Omega,\mathscr{F})$ to $(\Omega,\mathscr{F})$. Then
$$\begin{array}{ll}\displaystyle E_{[I_0,A]}(Y)&\displaystyle=\int_{\Omega}Y(\omega)P[I_0,A](d\omega)=\int_A Y(\omega)P(d\omega)\\
&\displaystyle =E(Y\cdot \mathds{1}_A),\ \ \forall A\in \mathscr{F},\end{array}$$
where $\mathds{1}_A$ is the indicator of $A$.
\end{thm}

\begin{defn}\label{definition5.2}Let $(\Omega,\mathscr{F},P)$, $(\Omega_i,\mathscr{F}_i),\ i=1,2$; $X_i(\omega),\ i=1,2$; $P[X_i],\ i=1,2$; $Y(\omega_2)$ and $A_1$ be the same as above. Then the integral of $Y(\omega_2)$ with respect to $P[X_2|X_1\in A_1]$ is denoted by $E_{[X_2]}(Y|X_1\in A_1)$, i.e.,
$$E_{[X_2]}(Y|X_1\in A_1)=\int_{\Omega_2}Y(\omega_2)P[X_2|X_1\in A_1](d\omega_2),$$
which is called the conditional expectation of $Y$ given $X_1\in A_1$.
\end{defn}

\begin{comm}\label{commentdef5.2.1} (i) By Definition \ref{definition2.5} and Definition \ref{definition5.1}, we have
$$E_{[X_2]}(Y|X_1\in A_1)=\left\{\begin{array}{l}
\displaystyle \frac{1}{P[X_1\in A_1]}E_{[X_2,X_1\in A_1]}(Y),\ \ \ \ P[X_1\in A_1]>0,\\\\
0,\ \ \ \ P[X_1\in A_1]=0,
\end{array}\right.$$
or
$$ E_{[X_2]}(Y|X_1\in A_1)\cdot P[X_1\in A_1]=E_{[X_2,X_1\in A_1]}(Y).$$

(ii) Let $I_0$ be the identical random object $I_{\mathscr{F}}$ from $(\Omega,\mathscr{F})$ to $(\Omega,\mathscr{F})$, $Y(\omega)$ a random variable on $(\Omega,\mathscr{F},P)$ and $P(A)>0,\ A\in\mathscr{F}$, then by (i) and Theorem \ref{theorem5.1}, we have
$$\begin{array}{ll}\displaystyle E(Y|A)&\displaystyle=E_{[I_0]}(Y|I_0\in A)=\frac{1}{P[I_0\in A]}E_{[I_0,I_0\in A]}(Y)\\\\
&\displaystyle =\frac{1}{P(A)}E_{[I_0,A]}(Y)=\frac{1}{P(A)}\int_{A}Y(\omega)P(d\omega),\end{array}$$
which is the conditional expectation of $Y$ given $A$ (see 5.3.5, \cite{Robert:2000}). In addition, we have
$$E(Y|A)\cdot P(A)=E_{[I_0]}(Y|I_0\in A)\cdot P[I_0\in A]=E_{[A]}(Y)=E(Y\cdot \mathds{1}_A).$$

\end{comm}

\begin{thm}\label{theorem5.2}Let $(\Omega,\mathscr{F},P)$ be a given probability space, and let $(\Omega_i,\mathscr{F}_i),\ i=1,2$ be measurable spaces. Let $X_i:(\Omega,\mathscr{F})\rightarrow(\Omega_i,\mathscr{F}_i),\ i=1,2$ be random objects, and $P[X_i]$ the probability on $(\Omega_i,\mathscr{F}_i)$ induced by $X_i$, $i=1,2$. Let $Y(\omega_2)$ be a random variable on $(\Omega_2,\mathscr{F}_2,P[X_2])$ with its expectation $E_{[X_2]}(Y)\in \mathbb{R}$. Then

1. for $\forall A_1\in \mathscr{F}_1$,
$$E_{[X_2,X_1\in A_1]}(Y)=\int_{X_1\in A_1} Y\circ X_2(\omega)P(d\omega)=E(Y\circ X_2\cdot \mathds{1}_{[X_1\in A_1]}),$$
where $Y\circ X_2(\omega)$ is the composite mapping of $Y$ and $X_2$, and $\mathds{1}_{[X_1\in A_1]}$ is the indicator of $X_1\in A_1$.

2. for $\forall A_1\in \mathscr{F}_1$ with $P[X_1\in A_1]>0$,
$$E_{[X_2]}(Y|X_1\in A_1)=\frac{1}{P[X_1\in A_1]}\int_{X_1\in A_1} Y\circ X_2(\omega)P(d\omega)=\frac{E_{[X_2,X_1\in A_1]}(Y)}{P[X_1\in A_1]},$$
where $Y\circ X_2(\omega)$ is the composite mapping of $Y$ and $X_2$. In addition, we have
$$E_{[X_2]}(Y|X_1\in A_1)\cdot P[X_1\in A_1]=E_{[X_2,X_1\in A_1]}(Y).$$
\end{thm}

\begin{comm}\label{comment5.2.1}Let $Y(\omega_2)$ be a random variable on $(\Omega_2,\mathscr{F}_2,P[X_2])$ with its expectation $E_{[X_2]}(Y)\in \mathbb{R}$.

(i) For $A_1,A_2\in\mathscr{F}_1$, $A_1\cap A_2=\emptyset$, by Theorem \ref{theorem5.2} we have
$$E_{[X_2,X_1\in A_1]}(Y)+E_{[X_2,X_1\in A_2]}(Y)=E_{[X_2,X_1\in A_1+A_2]}(Y).$$

(ii) From Theorem \ref{theorem5.2} it follows that
$$E_{[X_2,X_2\in A_2]}(Y)=E_{[X_2]}(Y\cdot \mathds{1}_{A_2}),\ \ \forall A_2\in\mathscr{F}_2,$$
where $\mathds{1}_{A_2}$ is the indicator of $A_2$.
\end{comm}

\begin{col}\label{corollary5.2.1}Let $(\Omega,\mathscr{F},P)$, $(\Omega_{i},\mathscr{F}_{i}),\ i\in \Lambda$ and $X_i(\omega),\ i\in \Lambda$  be the same as above. Let $I_1,I_2\subset\Lambda$ with $I_1\cap I_2=\emptyset$. Let $Y(\omega_{I_2})$ be a random variable on $(\Omega_{I_2},\mathscr{F}_{I_2},P[X_{I_2}])$ with its expectation $E_{[X_{I_2}]}(Y)\in \mathbb{R}$. Then

1. for $\forall A_{I_1}\in \mathscr{F}_{I_1}$,
$$E_{[X_{I_2},X_{I_1}\in A_{I_1}]}(Y)=E(Y\circ X_{I_2}\cdot \mathds{1}_{[X_{I_1}\in A_{I_1}]}),$$
where $Y\circ X_{I_2}(\omega)$ is the composite mapping of $Y$ and $X_{I_2}$.

2. for $\forall A_{I_1}\in \mathscr{F}_{I_1}$,
$$E_{[X_{I_2}]}(Y|X_{I_1}\in A_{I_1}])\cdot P[X_{I_1}\in A_{I_1}]=E_{[X_{I_2},X_{I_1}\in A_{I_1}]}(Y).$$
\end{col}

\begin{comm}\label{comment5.2.2} If $\Lambda$ is a countable index set, then by Comment \ref{comment2.6.2}, for $A_i\in \mathscr{F}_{i},\ i\in I_1$,  $E_{[X_{I_2},X_{I_1}\in \prod_{i\in I_1}A_i]}(Y)$ is also written as $E_{[X_i,i\in I_2;X_i\in A_i,i\in I_1]}(Y)$, and $E_{[X_{I_2}]}(Y|X_{I_1}\in \prod_{i\in I_1}A_i)$ as $E_{[X_i,i\in I_2]}(Y|X_i\in A_i,\ i\in I_1)$, i.e.,
$$E_{[X_i,i\in I_2;X_i\in A_i,i\in I_1]}(Y)=E_{[X_{I_2},X_{I_1}\in \prod_{i\in I_1}A_i]}(Y)=E_{[X_{I_2},\bigcap_{i\in I_1} X_i\in A_i]}(Y),$$
and
$$E_{[X_i,i\in I_2]}(Y|X_i\in A_i,\ i\in I_1)=E_{[X_{I_2}]}(Y|X_{I_1}\in \prod_{i\in I_1}A_i)=E_{[X_{I_2}]}(Y|\bigcap_{i\in I_1} X_i\in A_i),$$
which is also called the conditional expectation of $Y$ given co-occurrence of $X_i\in A_i,\ i\in I_1$.

Especially, let $Y(\omega)$ a random variable on $(\Omega,\mathscr{F},P)$ and $A_i\in\mathscr{F},i\in I\subset\Lambda$, then we take
$$E_{[A_i,i\in I]}(Y):=E_{[I_0;I_0\in A_i,i\in I]}(Y)=E_{[I_0,\bigcap_{i\in I}A_i]}(Y),$$
and
$$E(Y|A_i,\ i\in I):=E(Y|I_0\in A_i,\ i\in I)=E_{[I_0]}(Y|\bigcap_{i\in I}A_i),$$
which is also called the conditional expectation of $Y$ given co-occurrence of $A_i,\ i\in I$.
\end{comm}

By Comment \ref{comment5.2.2} and Corollary \ref{corollary5.2.1} we obtain the following corollary.

\begin{col}\label{corollary5.2.2}Let $(\Omega,\mathscr{F},P)$, $(\Omega_{i},\mathscr{F}_{i}),\ i\in \Lambda$ and $X_i(\omega),\ i\in \Lambda$  be the same as above, where $\Lambda$ is a countable index set. Let $I_1,I_2\subset\Lambda$ with $I_1\cap I_2=\emptyset$. Let $Y(\omega_{I_2})$ be a random variable on $(\Omega_{I_2},\mathscr{F}_{I_2},P[X_{I_2}])$ with its expectation $E_{[X_{I_2}]}(Y)\in \mathbb{R}$. Then for $A_i\in\mathscr{F}_i,i\in I_1$,
$$E_{[X_i,i\in I_2;X_i\in A_i,i\in I_1]}(Y)=E(Y\circ X_{I_2}\cdot \mathds{1}_{[X_i\in A_i,i\in I_1]});$$
and for $A_i\in\mathscr{F}_i,i\in I_1$,
$$E_{[X_i,i\in I_2]}(Y|X_i\in A_i,\ i\in I_1)\cdot P[X_i\in A_i,\ i\in I_1]=E_{[X_i,i\in I_2;X_i\in A_i,i\in I_1]}(Y).$$
\end{col}

As a special case of Corollary \ref{corollary5.2.2}, the following corollary is held.

\begin{col}\label{corollary5.2.3}Let $I_0$ be the identical random object $I_{\mathscr{F}}$ from $(\Omega,\mathscr{F})$ to $(\Omega,\mathscr{F})$, $Y(\omega)$ a random variable on $(\Omega,\mathscr{F},P)$ with its expectation $E(Y)\in \mathbb{R}$, and $A_i\in\mathscr{F},i\in I$, where $I$ is a countable index set. Then we have
$$E_{[A_i,i\in I]}(Y)=E_{[I_0;A_i,i\in I]}(Y)=E(Y\cdot \mathds{1}_{[A_i,i\in I]})$$
and
$$E(Y|A_i,i\in I)\cdot P[A_i,i\in I]=E_{[A_i,i\in I]}(Y).$$
\end{col}

\begin{defn}\label{definition5.3}Let $(\Omega,\mathscr{F},P)$ be a given probability space, and let $(\Omega_i,\mathscr{F}_i),\ i=1,2,3$ be measurable spaces. Let $X_i:(\Omega,\mathscr{F})\rightarrow(\Omega_i,\mathscr{F}_i),\ i=1,2,3$ be random objects, and $P[X_i]$ the probability on $(\Omega_i,\mathscr{F}_i)$ induced by $X_i$, $i=1,2,3$. Then $(\Omega_i,\mathscr{F}_i,P[X_i]),\ i=1,2,3$ are probability spaces. Let $Y(\omega_2)$ be a random variable on $(\Omega_2,\mathscr{F}_2,P[X_2])$ and $A_i\in\mathscr{F}_i,\ i=1,3$. Then the integral of $Y(\omega_2)$ with respect to $P[X_2,X_3\in A_3|X_1\in A_1]$ is denoted by $E_{[X_2,X_3\in A_3]}(Y|X_1\in A_1)$, i.e.,
$$E_{[X_2,X_3\in A_3]}(Y|X_1\in A_1)=\int_{\Omega_2}Y(\omega_2)P[X_2,X_3\in A_3|X_1\in A_1](d\omega_2),$$
which is called the integral of $Y$ with respect to $P[X_2,X_3\in A_3]$ given $X_1\in A_1$.
\end{defn}

\begin{comm}\label{commentdef5.3.1}
\begin{enumerate}
\renewcommand\theenumi{\roman{enumi}}
\renewcommand{\labelenumi}{(\theenumi)}
\item Since
$$P[X_2,X_3\in A_3|X_1\in A_1]=P[X_2|X_1\in A_1,X_3\in A_3]\cdot P[X_3\in A_3|X_1\in A_1],$$
we have
$$E_{[X_2,X_3\in A_3]}(Y|X_1\in A_1)=E_{[X_2]}(Y|X_1\in A_1,X_3\in A_3)\cdot P[X_3\in A_3|X_1\in A_1].$$
\item $E_{[X_2,X_3\in \Omega_3]}(Y|X_1\in A_1)=E_{[X_2]}(Y|X_1\in A_1)$,\\\\
$E_{[X_2,X_3\in A_3]}(Y|X_1\in \Omega_1)=E_{[X_2,X_3\in A_3]}(Y)$,\\\\
$E_{[X_2,X_3\in \Omega_3]}(Y|X_1\in \Omega_1)=E_{[X_2]}(Y)$.\\
\item As a special case of (i), we have
$$E_{[A_3]}(Y|A_1)=E_{[I_0,A_3]}(Y|A_1)=E(Y|A_1,A_3)\cdot P[A_3|A_1].$$
\end{enumerate}
\end{comm}

\begin{comm}\label{commentdef5.3.2}(i) Let $I_i\subset\Lambda,\ i=1,2,3$, which are pairwise disjoint. Let $Y(\omega_{I_2})$ be a random variable on $(\Omega_{I_2},\mathscr{F}_{I_2},P[X_{I_2}])$ with its expectation $E_{[X_{I_2}]}(Y)\in \mathbb{R}$. Then for $A_{I_i}\in \mathscr{F}_{I_i}, i=1,3$, we have
$$E_{[X_{I_2},X_{I_3}\in A_{I_3}]}(Y|X_{I_1}\in A_{I_1})=E_{[X_{I_2}]}(Y|X_{I_1}\in A_{I_1},X_{I_3}\in A_{I_3})\cdot P[X_{I_3}\in A_{I_3}|X_{I_1}\in A_{I_1}].$$
(ii) If $\Lambda$ is a countable index set, then by Comment \ref{comment2.6.2}, for $A_i\in \mathscr{F}_{i},\ i\in I_1+I_3$, we take
$$E_{[X_i,i\in I_2;X_i\in A_i,i\in I_3]}(Y|X_i\in A_i,i\in I_1):=E_{[X_{I_2},X_{I_3}\in \prod_{i\in I_3}A_i]}(Y|X_{I_1}\in \prod_{i\in I_1}A_i),$$
which is called the integral of $Y$ with respect to $P[X_i,i\in I_2;X_i\in A_i,i\in I_3]$ given co-occurrence of $X_i\in A_i,i\in I_1$. Since
$$E_{[X_{I_2},X_{I_3}\in \prod_{i\in I_3}A_i]}(Y|X_{I_1}\in \prod_{i\in I_1}A_i)=E_{[X_{I_2},\bigcap_{i\in I_3} X_i\in A_i]}(Y|\bigcap_{i\in I_1} X_i\in A_i),$$ we also have
$$E_{[X_i,i\in I_2;X_i\in A_i,i\in I_3]}(Y|X_i\in A_i,i\in I_1)=E_{[X_{I_2},\bigcap_{i\in I_3} X_i\in A_i]}(Y|\bigcap_{i\in I_1} X_i\in A_i).$$\\\\
(iii) Let $I_0$ be the identical random object $I_{\mathscr{F}}$ from $(\Omega,\mathscr{F})$ to $(\Omega,\mathscr{F})$, $Y(\omega)$ a random variable on $(\Omega,\mathscr{F},P)$ with its expectation $E(Y)\in \mathbb{R}$. Let $\Lambda$ be a countable index set, $I_1,I_2\subset \Lambda$ with $I_1\bigcap I_2=\emptyset$, and $A_i\in\mathscr{F},i\in I_1+I_2$. Then we take
$$E_{[A_i,i\in I_2]}(Y|A_i,i\in I_1):=E_{[I_0;A_i,i\in I_2]}(Y|I_0\in A_i,i\in I_1)=E_{[I_0,\bigcap_{i\in I_2}A_i]}(Y|\bigcap_{i\in I_1}A_i),$$
which is called the integral of $Y$ with respect to $P[I_0;A_i,i\in I_2]$ given co-occurrence of $A_i,i\in I_1$.
So it follows from (i) that
$$E_{[A_i,i\in I_2]}(Y|A_i,i\in I_1)=E(Y|A_i,i\in I_1+I_2)\cdot P[A_i,i\in I_2|A_i,i\in I_1].$$
\end{comm}

By Corollary \ref{corollary5.2.1}, Comment \ref{comment5.2.2}, Corollary \ref{corollary5.2.2}, Corollary \ref{corollary5.2.3} and Comment \ref{commentdef5.3.2} we obtain the following corollary.

\begin{col}\label{corollary5.2.4}Let $(\Omega,\mathscr{F},P)$ be a given probability space, $(\Omega_{i},\mathscr{F}_{i}),\ i\in \Lambda$ measurable spaces, where $\Lambda$ is a nonempty index set. Let $X_i:(\Omega,\mathscr{F})\rightarrow(\Omega_i,\mathscr{F}_i),\ i\in \Lambda$ be random objects. Let $I_i\subset\Lambda,\ i=1,2,3$, which are pairwise disjoint.

1. If $Y(\omega_{I_2})$ is a random variable on $(\Omega_{I_2},\mathscr{F}_{I_2},P[X_{I_2}])$ with its expectation $E_{[X_{I_2}]}(Y)\in \mathbb{R}$, then for $A_{I_i}\in \mathscr{F}_{I_i}, i=1,3$,
$$E_{[X_{I_2},X_{I_3}\in A_{I_3}]}(Y|X_{I_1}\in A_{I_1})\cdot P[X_{I_1}\in A_{I_1}]=E_{[X_{I_2};X_{I_i}\in A_{I_i},i=1,3]}(Y).$$
Especially, if $\Lambda$ is a countable index set, then for $A_i\in \mathscr{F}_{i},\ i\in I_1+I_3$,
$$E_{[X_i,i\in I_2;X_i\in A_i,i\in I_3]}(Y|X_i\in A_i,i\in I_1)\cdot P[X_i\in A_i,i\in I_1]=E_{[X_i,i\in I_2;X_i\in A_i,i\in I_1+I_3]}(Y).$$

2. If $\Lambda$ is a countable index set, and $Y(\omega)$ is a random variable on $(\Omega,\mathscr{F},P)$ with its expectation $E(Y)\in \mathbb{R}$, then for $A_i\in\mathscr{F},i\in I_1+I_2$,
$$E_{[A_i,i\in I_2]}(Y|A_i,i\in I_1)\cdot P[A_i,i\in I_1]=E_{[A_i,i\in I_1+I_2]}(Y).$$
\end{col}

\begin{defn}\label{definition5.4}Let $(\Omega,\mathscr{F},P)$ be a given probability space, and $(\Omega_i,\mathscr{F}_i),\ i=1,2$ measurable spaces. Let $X_i:(\Omega,\mathscr{F})\rightarrow(\Omega_i,\mathscr{F}_i),\ i=1,2$ be random objects, and $P[X_i]$ the probability on $(\Omega_i,\mathscr{F}_i)$ induced by $X_i$, $i=1,2$. Let $Y(\omega_2)$ be a random variable on $(\Omega_2,\mathscr{F}_2,P[X_2])$ with $E_{[X_2]}(Y)\in \mathbb{R}$. If there exists a $\mathscr{F}_1$-measurable function $\varphi(\omega_1)$ such that
$$\int_{A_1}\varphi(\omega_1)P[X_1](d\omega_1)=E_{[X_2,X_1\in A_1]}(Y),\ \ \forall A_1\in\mathscr{F}_1,$$
then $\varphi(\omega_1)$ is called the conditional expectation of $Y(\omega_2)$ given $X_1$ and denoted by $E_{[X_2]}(Y|X_1)(\omega_1)$, i.e.,
$$\int_{A_1}E_{[X_2]}(Y|X_1)(\omega_1)P[X_1](d\omega_1)=E_{[X_2,X_1\in A_1]}(Y),\ \ \forall A_1\in\mathscr{F}_1.$$
\end{defn}

\begin{comm}\label{commentdef5.4.1}\begin{enumerate}
\renewcommand\theenumi{\roman{enumi}}
\renewcommand{\labelenumi}{(\theenumi)}
\item Since
$$\int_{\Omega_1}E_{[X_2]}(Y|X_1)(\omega_1)P[X_1](d\omega_1)=E_{[X_2]}(Y)\in \mathbb{R},$$
$|E_{[X_2]}(Y|X_1)(\omega_1)|<+\infty$, $a.e.\{P[X_1]\}$.\\
\item If $X_1=I_{\mathscr{G}}$, where $\mathscr{G}$ is a sub $\sigma$-field of $\mathscr{F}$, then $E_{[X_2]}(Y|I_{\mathscr{G}})(\omega)$ is also denoted by $E_{[X_2]}(Y|\mathscr{G})(\omega)$, i.e., $E_{[X_2]}(Y|\mathscr{G})(\omega)=E_{[X_2]}(Y|I_{\mathscr{G}})(\omega)$.\\
\item If $X_1=I_{\mathscr{G}}$ and $X_2=I_0=I_{\mathscr{F}}$, where $\mathscr{G}$ is a sub $\sigma$-field of $\mathscr{F}$, then $E_{[I_0]}(Y|I_{\mathscr{G}})(\omega)$ is also denoted by $E(Y|\mathscr{G})(\omega)$, i.e., $E(Y|\mathscr{G})(\omega)=E_{[I_0]}(Y|I_{\mathscr{G}})(\omega)$, which has been defined in Section 7.2, \cite{yan:2004}.\\
\item By Comment \ref{comment5.1.1} and Definition \ref{definition5.4} we have
$$E_{[X_1]}(E_{[X_2]}(Y|X_1)\cdot \mathds{1}_{A_1})=E_{[X_2,X_1\in A_1]}(Y),$$
where $A_1\in\mathscr{F}_1$ and $\mathds{1}_{A_1}$ is the indicator of $A_1$. Especially, we have
$$E_{[X_1]}(E_{[X_2]}(Y|X_1))=E_{[X_2]}(Y).$$
\end{enumerate}
\end{comm}

\begin{defn}\label{definition5.5}Let $(\Omega,\mathscr{F},P)$ be a given probability space, and $(\Omega_i,\mathscr{F}_i),\ i=1,2,3$ measurable spaces. Let $X_i:(\Omega,\mathscr{F})\rightarrow(\Omega_i,\mathscr{F}_i),\ i=1,2,3$ be random objects, and $P[X_i]$ the probability on $(\Omega_i,\mathscr{F}_i)$ induced by $X_i$, $i=1,2,3$. Let $Y(\omega_2)$ be a random variable on $(\Omega_2,\mathscr{F}_2,P[X_2])$ with $E_{[X_2]}(Y)\in \mathbb{R}$. For fixed $A_3\in\mathscr{F}_3$ with $P[X_3\in A_3]>0$, if there exists a $\mathscr{F}_1$-measurable function $\varphi(\omega_1)$ such that, for $\forall A_1\in\mathscr{F}_1$,
$$\int_{A_1}\varphi(\omega_1)P[X_1,X_3\in A_3](d\omega_1)=E_{[X_2;X_1\in A_1,X_3\in A_3]}(Y),$$
then $\varphi(\omega_1)$ is called the conditional expectation of $Y(\omega_2)$ given co-occurrence of $X_1$ and $X_3\in A_3$, and denoted by $E_{[X_2]}(Y|X_1,X_3\in A_3)(\omega_1)$, i.e.,
$$\begin{array}{l}\displaystyle\int_{A_1}E_{[X_2]}(Y|X_1,X_3\in A_3)(\omega_1)P[X_1,X_3\in A_3](d\omega_1)\\\\
=E_{[X_2;X_1\in A_1,X_3\in A_3]}(Y),\ \ \forall A_1\in\mathscr{F}_1.\end{array}$$
For fixed $A_3\in\mathscr{F}_3$ with $P[X_3\in A_3]=0$, $E_{[X_2]}(Y|X_1,X_3\in A_3)(\omega_1)$ is defined as zero.
\end{defn}

\begin{comm}\label{commentdef5.5.1}\begin{enumerate}
\renewcommand\theenumi{\roman{enumi}}
\renewcommand{\labelenumi}{(\theenumi)}
\item By definition and Comment \ref{comment2.6.1}, we have
$$E_{[X_2]}(Y|X_1,X_3\in \Omega_3)(\omega_1)=E_{[X_2]}(Y|X_1)(\omega_1),$$
which has been defined in Definition \ref{definition5.4}.
\item By Definition \ref{definition5.1} and Definition \ref{definition5.5} we have
$$E_{[X_1,X_3\in A_3]}(E_{[X_2]}(Y|X_1,X_3\in A_3)\cdot \mathds{1}_{A_1})=E_{[X_2;X_1\in A_1,X_3\in A_3]}(Y),$$
where $A_1\in\mathscr{F}_1$ and $\mathds{1}_{A_1}$ is the indicator of $A_1$. Especially, we have
$$E_{[X_1,X_3\in A_3]}(E_{[X_2]}(Y|X_1,X_3\in A_3))=E_{[X_2,X_3\in A_3]}(Y).$$
\end{enumerate}
\end{comm}

\begin{defn}\label{definition5.6}Let $(\Omega,\mathscr{F},P)$ be a given probability space, and $(\Omega_i,\mathscr{F}_i),\ i=1,2,4$ measurable spaces. Let $X_i:(\Omega,\mathscr{F})\rightarrow(\Omega_i,\mathscr{F}_i),\ i=1,2,4$ be random objects, and $P[X_i]$ the probability on $(\Omega_i,\mathscr{F}_i)$ induced by $X_i$, $i=1,2,4$. Let $Y(\omega_2)$ be a random variable on $(\Omega_2,\mathscr{F}_2,P[X_2])$ with $E_{[X_2]}(Y)\in \mathbb{R}$. For fixed $A_4\in\mathscr{F}_4$, if there exists a $\mathscr{F}_1$-measurable function $\varphi(\omega_1)$ such that, for $\forall A_1\in\mathscr{F}_1$,
$$\int_{A_1}\varphi(\omega_1)P[X_1](d\omega_1)=E_{[X_2;X_1\in A_1,X_4\in A_4]}(Y),$$
then $\varphi(\omega_1)$ is called the integral of $Y$ with respect to $P[X_2,X_4\in A_4]$ given $X_1$, and denoted by $E_{[X_2,X_4\in A_4]}(Y|X_1)(\omega_1)$, i.e.,
$$\begin{array}{l}\displaystyle\int_{A_1}E_{[X_2,X_4\in A_4]}(Y|X_1)(\omega_1)P[X_1](d\omega_1)=E_{[X_2;X_1\in A_1,X_4\in A_4]}(Y),\ \ \forall A_1\in\mathscr{F}_1.\end{array}$$
\end{defn}

\begin{comm}\label{commentdef5.6.1}
\begin{enumerate}
\renewcommand\theenumi{\roman{enumi}}
\renewcommand{\labelenumi}{(\theenumi)}
\item By definition and Comment \ref{comment2.6.1}, we have
$$E_{[X_2,X_4\in \Omega_4]}(Y|X_1)(\omega_1)=E_{[X_2]}(Y|X_1)(\omega_1),$$
which has been defined in Definition \ref{definition5.4}.
\item By Definition \ref{definition5.1} and Definition \ref{definition5.6} we have
$$E_{[X_1]}(E_{[X_2,X_4\in A_4]}(Y|X_1)\cdot \mathds{1}_{A_1})=E_{[X_2;X_1\in A_1,X_4\in A_4]}(Y),$$
where $A_1\in\mathscr{F}_1$ and $\mathds{1}_{A_1}$ is the indicator of $A_1$. Especially, we have
$$E_{[X_1]}(E_{[X_2,X_4\in A_4]}(Y|X_1))=E_{[X_2,X_4\in A_4]}(Y).$$
\end{enumerate}
\end{comm}

\begin{defn}\label{definition5.7}Let $(\Omega,\mathscr{F},P)$ be a given probability space, and $(\Omega_i,\mathscr{F}_i),\ i=1,2,3,4$ measurable spaces. Let $X_i:(\Omega,\mathscr{F})\rightarrow(\Omega_i,\mathscr{F}_i),\ i=1,2,3,4$ be random objects, and $P[X_i]$ the probability on $(\Omega_i,\mathscr{F}_i)$ induced by $X_i$, $i=1,2,3,4$. Let $Y(\omega_2)$ be a random variable on $(\Omega_2,\mathscr{F}_2,P[X_2])$ with $E_{[X_2]}(Y)\in \mathbb{R}$. For fixed $A_i\in\mathscr{F}_i,\ i=3,4$ with $P[X_3\in A_3]>0$, if there exists a $\mathscr{F}_1$-measurable function $\varphi(\omega_1)$ such that, for $\forall A_1\in\mathscr{F}_1$,
$$\int_{A_1}\varphi(\omega_1)P[X_1,X_3\in A_3](d\omega_1)=E_{[X_2;X_i\in A_i,i=1,3,4]}(Y),$$
then $\varphi(\omega_1)$ is called the integral of $Y$ with respect to $P[X_2,X_4\in A_4]$ given co-occurrence of $X_1$ and $X_3\in A_3$, and denoted by $E_{[X_2,X_4\in A_4]}(Y|X_1,X_3\in A_3)(\omega_1)$, i.e.,
$$\begin{array}{l}\displaystyle\int_{A_1}E_{[X_2,X_4\in A_4]}(Y|X_1,X_3\in A_3)(\omega_1)P[X_1,X_3\in A_3](d\omega_1)\\\\
=E_{[X_2;X_i\in A_i,i=1,3,4]}(Y),\ \ \forall A_1\in\mathscr{F}_1.\end{array}$$
For $P[X_3\in A_3]=0$, $E_{[X_2,X_4\in A_4]}(Y|X_1,X_3\in A_3)(\omega_1)$ is defined as zero.
\end{defn}

\begin{comm}\label{commentdef5.7.1}\begin{enumerate}
\renewcommand\theenumi{\roman{enumi}}
\renewcommand{\labelenumi}{(\theenumi)}
\item By Comment \ref{commentdef5.3.1}, Comment \ref{commentdef5.3.2} and Definition \ref{definition5.5} we have
$$E_{[X_2,X_4\in \Omega_4]}(Y|X_1,X_3\in A_3)(\omega_1)=E_{[X_2]}(Y|X_1,X_3\in A_3)(\omega_1).$$
\item By Comment \ref{commentdef5.3.2} and Definition \ref{definition5.6} we have
$$E_{[X_2,X_4\in A_4]}(Y|X_1,X_3\in \Omega_3)(\omega_1)=E_{[X_2,X_4\in A_4]}(Y|X_1)(\omega_1).$$
\item By Definition \ref{definition5.1} and Definition \ref{definition5.7} we have
$$E_{[X_1,X_3\in A_3]}(E_{[X_2,X_4\in A_4]}(Y|X_1,X_3\in A_3)\cdot \mathds{1}_{A_1})=E_{[X_2;X_i\in A_i,i=1,3,4]}(Y),$$
where $A_1\in\mathscr{F}_1$ and $\mathds{1}_{A_1}$ is the indicator of $A_1$. Especially, we have
$$E_{[X_1,X_3\in A_3]}(E_{[X_2,X_4\in A_4]}(Y|X_1,X_3\in A_3))=E_{[X_2;X_i\in A_i,i=3,4]}(Y).$$
\end{enumerate}
\end{comm}

About the existence and uniqueness of $E_{[X_2,X_4\in A_4]}(Y|X_1,X_3\in A_3)(\omega_1)$ we have the following theorem.

\begin{thm}\label{theorem5.3}Let $(\Omega,\mathscr{F},P)$, $(\Omega_i,\mathscr{F}_i),\ i=1,2,3,4$; $X_i(\omega),\ i=1,2,3,4$ and $P[X_i],\ i=1,2,3,4$ be the same as in Definition \ref{definition5.7}. If $Y(\omega_2)$ is a random variable on $(\Omega_2,\mathscr{F}_2,P[X_2])$ with $E_{[X_2]}(Y)\in \mathbb{R}$, then for $\forall A_i\in\mathscr{F}_i,i=3,4$, $E_{[X_2,X_4\in A_4]}(Y|X_1,X_3\in A_3)(\omega_1)$ exists and it is unique a.e.$\{P[X_1,X_3\in A_3]\}$ provided $P[X_3\in A_3]>0$.
\end{thm}

In view of Comment \ref{commentdef5.5.1}, Comment \ref{commentdef5.6.1} and Comment \ref{commentdef5.7.1}, we obtain the following corollary.

\begin{col}\label{corollary5.3.1}Let $(\Omega,\mathscr{F},P)$, $(\Omega_i,\mathscr{F}_i),\ i=1,2,3,4$; $X_i(\omega),\ i=1,2,3,4$ and $P[X_i],\ i=1,2,3,4$ be the same as above. If $Y(\omega_2)$ is a random variable on $(\Omega_2,\mathscr{F}_2,P[X_2])$ with $E_{[X_2]}(Y)\in \mathbb{R}$, then for $\forall A_i\in\mathscr{F}_i,i=3,4$,
\begin{enumerate}
\item $E_{[X_2]}(Y|X_1,X_3\in A_3)(\omega_1)$ exists and it is unique a.e.$\{P[X_1,X_3\in A_3]\}$ provided $P[X_3\in A_3]>0$;\\
\item $E_{[X_2,X_4\in A_4]}(Y|X_1)(\omega_1)$ exists and it is unique a.e.$\{P[X_1]\}$;\\
\item $E_{[X_2]}(Y|X_1)(\omega_1)$ exists and it is unique a.e.$\{P[X_1]\}$.
\end{enumerate}
\end{col}

\begin{comm}\label{comment5.3.1}Let $(\Omega,\mathscr{F},P)$ be a given probability space, and let $I_{\mathscr{G}}$ and $I_0$ be identical random objects on $(\Omega,\mathscr{F})$, where $\mathscr{G}$ is a sub $\sigma$-field of $\mathscr{F}$.

(i) For fixed $B\in\mathscr{F}$, by Corollary \ref{corollary5.3.1} there exists $E(\mathds{1}_B|\mathscr{G})=E_{[I_0]}(\mathds{1}_B|I_{\mathscr{G}})$. And by definition we have
$$E(\mathds{1}_B|\mathscr{G})=E_{[I_0]}(\mathds{1}_B|I_{\mathscr{G}})=P[B|I_{\mathscr{G}}]=P[B|\mathscr{G}],$$
where $\mathds{1}_B$ is the indicator of $B$.

(ii) Let $Y$ be a random variable on $(\Omega,\mathscr{F},P)$ with $E(Y)\in \mathbb{R}$.  By Corollary \ref{corollary5.3.1} there exists $E(Y|\mathscr{G})=E_{[I_0]}(Y|I_{\mathscr{G}})$. And by definition we have
$$E_{[I_{\mathscr{G}}]}(E(Y|\mathscr{G}))=E_{[I_{\mathscr{G}}]}(E_{[I_0]}(Y|I_{\mathscr{G}}))=E_{[I_0]}(Y)=E(Y).$$

(iii) Let $Y$ be a random variable on $(\Omega,\mathscr{F},P)$ with $E(Y)\in \mathbb{R}$, let $\mathscr{G}=\{\emptyset,\Omega\}$. Then $E(Y|\mathscr{G})$ is a constant and $$E(Y|\mathscr{G})=E_{[I_0]}(Y|I_{\mathscr{G}})=E(Y).$$

(iv) By Theorem \ref{theorem5.3} and Corollary \ref{corollary5.3.1}, if a $\mathscr{F}_1$-measurable function $\varphi(\omega_1)$ is equal to $E_{[X_2,X_4\in A_4]}(Y|X_1,X_3\in A_3)$ a.e.$\{P[X_1,X_3\in A_3]\}$, then it is also denoted by
$$\varphi=E_{[X_2,X_4\in A_4]}(Y|X_1,X_3\in A_3)$$
without any ambiguity.
\end{comm}

\begin{lem}\label{lemma5.4.1}Let $(\Omega,\mathscr{F},P)$, $(\Omega_i,\mathscr{F}_i),\ i=1,2,3,4$; $X_i(\omega),\ i=1,2,3,4$ and $P[X_i],\ i=1,2,3,4$ be the same as in Definition \ref{definition5.7}. If $Y(\omega_2)$ be a random variable on $(\Omega_2,\mathscr{F}_2,P[X_2])$ with $E_{[X_2]}(Y)\in \mathbb{R}$, then for fixed $A_i\in \mathscr{F}_i,i=3,4$,
\begin{equation}\label{eqnlem5.4.1.1}\begin{array}{l}\displaystyle\int_{A_1\times\Omega_2}Y(\omega_2)P[X_1,X_2;X_i\in A_i,i=3,4](d(\omega_1,\omega_2))\\
\displaystyle=\int_{\Omega_2}Y(\omega_2)P[X_2;X_i\in A_i,i=1,3,4](d\omega_2),\ \ \ \  \forall A_1\in\mathscr{F}_1.\end{array}\end{equation}
\end{lem}

\begin{comm}\label{commentlem5.4.1}(i) The result of Lemma \ref{lemma5.4.1} can be extended the general case, i.e.,
\begin{equation}\label{eqncom5.3.1.1}\begin{array}{l}\displaystyle\int_{A_1\times A_2}Y(\omega_2)P[X_1,X_2;X_i\in A_i,i=3,4](d(\omega_1,\omega_2))\\
\displaystyle =\int_{A_2}Y(\omega_2)P[X_2;X_i\in A_i,i=1,3,4](d\omega_2),\ \ \ \ \forall A_i\in\mathscr{F}_i,\ i=1,2,3,4.\end{array}\end{equation}

(ii) Since $\mathds{1}_{[A_1\times A_2]}=\mathds{1}_{A_1}\cdot \mathds{1}_{A_2}$, Equation (\ref{eqncom5.3.1.1}) above is rewritten as
$$E_{[X_1,X_2;X_i\in A_i,i=3,4]}(Y(\omega_2)\cdot \mathds{1}_{A_1}\cdot \mathds{1}_{A_2})=E_{[X_2;X_i\in A_i,i=1,3,4]}(Y\cdot \mathds{1}_{A_2}),$$
$\forall A_i\in\mathscr{F}_i,\ i=1,2,3,4$. Especially we have
$$E_{[X_1,X_2;X_i\in A_i,i=3,4]}(Y(\omega_2)\cdot \mathds{1}_{A_1})=E_{[X_2;X_i\in A_i,i=1,3,4]}(Y),\ \ \forall A_i\in \mathscr{F}_i,i=1,3,4,$$
which is Equation (\ref{eqnlem5.4.1.1}).
\end{comm}

\begin{thm}\label{theorem5.4}Let $(\Omega,\mathscr{F},P)$, $(\Omega_i,\mathscr{F}_i),\ i=1,2,3,4$; $X_i(\omega),\ i=1,2,3,4$ and $P[X_i],\ i=1,2,3,4$ be the same as in Definition \ref{definition5.7}. Let $Y(\omega_2)$ be a random variable on $(\Omega_2,\mathscr{F}_2,P[X_2])$ with $E_{[X_2]}(Y)\in \mathbb{R}$. If there exists $P[X_2,X_4\in A_4|X_1,X_3\in A_3](\omega_1,A_2)$ for fixed $A_i\in\mathscr{F}_i,i=3,4$ with $P[X_3\in A_3]>0$, then
$$\begin{array}{l}E_{[X_2,X_4\in A_4]}(Y|X_1,X_3\in A_3)(\omega_1)\\
\displaystyle=\int_{\Omega_2}Y(\omega_2)P[X_2,X_4\in A_4|X_1,X_3\in A_3](\omega_1,d\omega_2),\ \ a.e.\{P[X_1,X_3\in A_3]\}\end{array}$$.
\end{thm}

\begin{comm}\label{comment5.4.1}(i) In Theorem \ref{theorem5.4}, ``$a.e.\{P[X_1,X_3\in A_3]\}$'' means that $P[X_1,X_3\in A_3](E)=0$, where
$$E=\{\omega_1:\int_{\Omega_2}Y(\omega_2)P[X_2,X_4\in A_4|X_1,X_3\in A_3](\omega_1,d\omega_2)=\infty \ or\  \infty-\infty\}.$$

(ii)For any random variable $Y$ on $(\Omega_2,\mathscr{F}_2,P[X_2])$, if there exists $P[X_2,X_4\in A_4|X_1,X_3\in A_3]$, then we always take
$$E_{[X_2,X_4\in A_4]}(Y|X_1,X_3\in A_3)(\omega_1)=\int_{\Omega_2}Y(\omega_2)P[X_2,X_4\in A_4|X_1,X_3\in A_3](\omega_1,d\omega_2)$$
without any ambiguity.
\end{comm}

By Comment \ref{comment2.11.1} we have the following corollary.

\begin{col}\label{corollary5.4.1}Let $(\Omega,\mathscr{F},P)$, $(\Omega_i,\mathscr{F}_i),\ i=1,2,3,4$; $X_i(\omega),\ i=1,2,3,4$ and $P[X_i],\ i=1,2,3,4$ be the same as above. Let $Y(\omega_2)$ be a random variable on $(\Omega_2,\mathscr{F}_2,P[X_2])$ with $E_{[X_2]}(Y)\in \mathbb{R}$.
\begin{enumerate}
\item If there exists $P[X_2|X_1,X_3\in A_3](\omega_1,A_2)$ for fixed $A_3\in\mathscr{F}_3$ with $P[X_3\in A_3]>0$, then
$$\begin{array}{l}E_{[X_2]}(Y|X_1,X_3\in A_3)(\omega_1)\\
\displaystyle=\int_{\Omega_2}Y(\omega_2)P[X_2|X_1,X_3\in A_3](\omega_1,d\omega_2),\ \ a.e.\{P[X_1,X_3\in A_3]\}.\end{array}$$
\item If there exists $P[X_2,X_4\in A_4|X_1](\omega_1,A_2)$ for fixed $A_4\in\mathscr{F}_4$, then
$$E_{[X_2,X_4\in A_4]}(Y|X_1)(\omega_1)=\int_{\Omega_2}Y(\omega_2)P[X_2,X_4\in A_4|X_1](\omega_1,d\omega_2),\ \ a.e.\{P[X_1]\}.$$
\item If there exists $P[X_2|X_1](\omega_1,A_2)$, then
$$E_{[X_2]}(Y|X_1)(\omega_1)=\int_{\Omega_2}Y(\omega_2)P[X_2|X_1](\omega_1,d\omega_2),\  \ a.e.\{P[X_1]\}.$$
\end{enumerate}
\end{col}

Noting that $P[I_0|I_{\mathscr{G}}](\omega,A)=P(\omega,A)$ (see Comment \ref{comment2.10.1} (v)), by Comment \ref{commentdef5.4.1} (iii) and Corollary \ref{corollary5.4.1} 3. we obtain the following corollary which is Theorem 7.3.2 in \cite{yan:2004}.

\begin{col}\label{corollary5.4.2}Let $(\Omega,\mathscr{F},P)$ be a given probability space and $\mathscr{G}$ a sub $\sigma$-field of $\mathscr{F}$. Let $Y(\omega)$ be a random variable on $(\Omega,\mathscr{F},P)$ with $E(Y)\in \mathbb{R}$. If there exists $P(\omega,A)$, then
$$E(Y|\mathscr{G})(\omega)=\int_{\Omega}Y(\omega')P(\omega,d\omega').$$
\end{col}

By Theorem \ref{theorem5.3} and Theorem \ref{theorem5.4}, we have the following corollary.

\begin{col}\label{corollary5.4.3} Let $(\Omega,\mathscr{F},P)$, $(\Omega_{i},\mathscr{F}_{i}),\ i\in \Lambda$ and  $X_i(\omega),\ i\in \Lambda$ be the same as above. Let $I_i\subset\Lambda,\ i=1,2,3,4$, which are pairwise disjoint. Let $Y(\omega_{I_2})$ be a random variable on $(\Omega_{I_2},\mathscr{F}_{I_2},P[X_{I_2}])$ with $E_{[X_{I_2}]}(Y)\in \mathbb{R}$. Then for fixed $A_{I_i}\in\mathscr{F}_{I_i},\ i=3,4$, there exists $E_{[X_{I_2},X_{I_4}\in A_{I_4}]}(Y|X_{I_1},X_{I_3}\in A_{I_3})(\omega_{I_1})$. Furthermore if there exists $P[X_{I_2},X_{I_4}\in A_{I_4}|X_{I_1},X_{I_3}\in A_{I_3}](\omega_{I_1},A_{I_2})$ for fixed $A_{I_i}\in\mathscr{F}_{I_i},\ i=3,4$ with $P[X_{I_3}\in A_{I_3}]>0$, then
$$\begin{array}{l}\displaystyle E_{[X_{I_2},X_{I_4}\in A_{I_4}]}(Y|X_{I_1},X_{I_3}\in A_{I_3})(\omega_{I_1})\\
\displaystyle =\int_{\Omega_{I_2}}Y(\omega_{I_2})P[X_{I_2},X_{I_4}\in A_{I_4}|X_{I_1},X_{I_3}\in A_{I_3}](\omega_{I_1},d\omega_{I_2}),\ a.e.\{P[X_{I_1},X_{I_3}\in A_{I_3}]\}.\end{array}$$
\end{col}

\begin{comm}\label{comment5.4.2}If $\Lambda$ is a countable index set, then for $A_i\in \mathscr{F}_{i},\ i\in I_3+I_4$, we take
$$\begin{array}{l}\displaystyle E_{[X_i,i\in I_2;X_i\in A_i,i\in I_4]}(Y|X_i,i\in I_1;X_i\in A_i,i\in I_3)(\omega_i,i\in I_1)\\
\displaystyle :=E_{[X_{I_2},\bigcap_{i\in I_4} X_i\in A_i]}(Y|X_{I_1},\bigcap_{i\in I_3} X_i\in A_i)(\omega_i,i\in I_1)\\
\displaystyle =E_{[X_{I_2},X_{I_4}\in \prod_{i\in I_4}A_i]}(Y|X_{I_1},X_{I_3}\in \prod_{i\in I_3}A_i)(\omega_{I_1}),\end{array}$$
which is called the integral of $Y$ with respect to $P[X_i,i\in I_2;X_i\in A_i,i\in I_4]$ given co-occurrence of $X_i,i\in I_1$ and $X_i\in A_i,i\in I_3$. By Definition \ref{definition2.12} and Comment \ref{comment2.12.1}, if there exists $P[X_i,i\in I_2;X_i\in A_i,i\in I_4|X_i,i\in I_1;X_i\in A_i,i\in I_3](\omega_{I_1},A_{I_2})$, then
$$\begin{array}{l}\displaystyle E_{[X_i,i\in I_2;X_i\in A_i,i\in I_4]}(Y|X_i,i\in I_1;X_i\in A_i,i\in I_3)(\omega_i,i\in I_1)\\
\displaystyle =\int_{\Omega_{I_2}}Y(\omega_{I_2})P[X_i,i\in I_2;X_i\in A_i,i\in I_4|X_i,i\in I_1;X_i\in A_i,i\in I_3](\omega_{I_1},d\omega_{I_2}),\\
a.e.\{P[X_i,i\in I_1;X_i\in A_i,i\in I_3]\}.\end{array}$$
\end{comm}

According to Theorem \ref{theorem2.11.3}, we have the following theorem.

\begin{thm}\label{theorem5.5} Let $(\Omega,\mathscr{F},P)$ be a given probability space, and $X_i(\omega),\ i\in I\subset \mathbb{N}$ random variables on $(\Omega,\mathscr{F},P)$. Let $I_i\subset I,\ i=1,2,3,4$ be pairwise disjoint. Let $Y(x_{I_2})$ be a random variable on $(\mathbb{R}^{I_2},\mathscr{B}(\mathbb{R}^{I_2}),P[X_{I_2}])$ with $E_{[X_{I_2}]}(Y)\in \mathbb{R}$. Then
$$\begin{array}{l}\displaystyle E_{[X_{I_2},X_{I_4}\in A_{I_4}]}(Y|X_{I_1},X_{I_3}\in A_{I_3})(x_{I_1})\\\\
\displaystyle =\int_{\mathbb{R}^{I_2}}Y(x_{I_2})P[X_{I_2},X_{I_4}\in A_{I_4}|X_{I_1},X_{I_3}\in A_{I_3}](x_{I_1},dx_{I_2}),\ a.e.\{P[X_{I_1},X_{I_3}\in A_{I_3}]\},\end{array}$$
where $x_{I_i}\in \mathbb{R}^{I_i},\ i=1,2$ and $A_{I_i}\in\mathscr{B}(\mathbb{R}^{I_i}),\ i=3,4$ with $P[X_{I_3}\in A_{I_3}]>0$.
\end{thm}

\section{The Definition of $E$-integral and Its Properties}\label{section6}

\begin{defn}\label{definition6.1}Let $(\Omega,\mathscr{F},P)$ be a given probability space, $(\Omega_{i},\mathscr{F}_{i}),\ i\in \Lambda$ measurable spaces, where $\Lambda$ is a nonempty index set. Let $X_i:(\Omega,\mathscr{F})\rightarrow(\Omega_i,\mathscr{F}_i),\ i\in \Lambda$ be random objects. For $I\subset \Lambda$, let $P[X_I]$ be the probability measure on $(\Omega_I,\mathscr{F}_I)$ induced by $X_I$. Let $f(\omega_\Lambda)$ be a $\mathscr{F}_\Lambda$-measurable function on $(\Omega_\Lambda,\mathscr{F}_\Lambda)$. Let $I_j\subset\Lambda,j=1,2,3,4$, which are pairwise disjoint, and $A_{I_j}\in\mathscr{F}_{I_j},j=3,4$. Then the integrals of $f(\omega_\Lambda)$ with respect to $P[X_{I_2},X_{I_4}\in A_{I_4}]$, $P[X_{I_2},X_{I_4}\in A_{I_4}|X_{I_3}\in A_{I_3}]$ or $P[X_{I_2},X_{I_4}\in A_{I_4}|X_{I_1},X_{I_3}\in A_{I_3}]$ (provided its existence) are called $E$-integral. We denote
$$E_{[X_{I_2},X_{I_4}\in A_{I_4}]}(f):=\int_{\Omega_{I_2}}f(\omega_{I_2^c},\omega_{I_2})P[X_{I_2},X_{I_4}\in A_{I_4}](d\omega_{I_2}),$$
$$E_{[X_{I_2},X_{I_4}\in A_{I_4}]}(f|X_{I_3}\in A_{I_3}):=\int_{\Omega_{I_2}}f(\omega_{I_2^c},\omega_{I_2})P[X_{I_2},X_{I_4}\in A_{I_4}|X_{I_3}\in A_{I_3}](d\omega_{I_2}),$$
and
$$\begin{array}{l}E_{[X_{I_2},X_{I_4}\in A_{I_4}]}(f|X_{I_1},X_{I_3}\in A_{I_3})\\\\
\displaystyle :=\int_{\Omega_{I_2}}f(\omega_{I_{12}^c},\omega_{I_1},\omega_{I_2})P[X_{I_2},X_{I_4}\in A_{I_4}|X_{I_1},X_{I_3}\in A_{I_3}](\omega_{I_1},d\omega_{I_2}),\end{array}$$
where $I_2^c=\Lambda\backslash I_2$, $I_{12}^c=\Lambda\backslash (I_1+I_2)$.
\end{defn}

\begin{comm}\label{commentdef6.1.1}(i) If $f(\omega_{I_2})$ is a measurable function on $(\Omega_{I_2},\mathscr{F}_{I_2})$, then so is it on $(\Omega_\Lambda,\mathscr{F}_\Lambda)$. Thus the integrals given in Definition \ref{definition5.1}-\ref{definition5.7} above are $E$-integrals.

(ii) Generally, the integrability of $f(\omega_\Lambda)$ with respect to $P[X_\Lambda]$ can't lead to the integrability of $f(\omega_\Lambda)$ with respect to $P[X_{I_2}]$ for $a.e.\{P[X_{I_2^c}]\}\ \ \omega_{I_2^c}\in\Omega_{I_2^c}$, i.e.,
$$E_{[X_\Lambda]}(f)\in\mathbb{R}\nRightarrow E_{[X_{I_2}]}(f)\in\mathbb{R},\ \ a.e.\{P[X_{I_2^c}]\}.$$
See the following example with $\Lambda=\{1,2\}$.

Let $(\Omega,\mathscr{F},P)=([0,1],\mathscr{B}([0,1]),\lambda)$ with the Lebesgue measure $\lambda$, $$(\Omega_i,\mathscr{F}_i)=([0,1],\mathscr{B}([0,1])),i=1,2,$$ and $X_i=I_0,i=1,2$, where $I_0$ is the identical random object from $([0,1],\mathscr{B}([0,1]))$ to $([0,1],\mathscr{B}([0,1]))$. Then $P[X_1]=P[X_2]=\lambda$.

Let
$$f(x_1,x_2)=\left\{\begin{array}{lll}\displaystyle 0,&\ \ \ \ &(x_1,x_2)\in D_1,\\
\displaystyle \frac{1}{x_2},&\ \ \ \ &(x_1,x_2)\in D_2,
\end{array}\right.$$
where $D_1=\{(x_1,x_2):x_1=x_2,x_1\in (0,1]\}\cup\{(x_1,x_2):x_2=0,x_1\in [0,1]\}$ and $D_2=[0,1]\times[0,1]-D_1$, then $f(x_1,x_2)$ is a $\mathscr{B}([0,1]\times[0,1])$-measurable function. It is evident that $D_1,D_2\in \mathscr{B}([0,1]\times[0,1])$ and $P[I_0,I_0](D_2)=0$. Thus
$$\begin{array}{l}\displaystyle \int_{[0,1]\times[0,1]}f(x_1,x_2)P[I_0,I_0](d(x_1,x_2))=\int_{D_1+D_2}f(x_1,x_2)P[I_0,I_0](d(x_1,x_2))\\\\
\displaystyle =\int_{D_1}f(x_1,x_2)P[I_0,I_0](d(x_1,x_2))=0.\end{array}$$
On the other hand,
$$\int_{[0,1]}f(x_1,x_2)P[I_0](dx_2)=\int_{[0,1]}\frac{1}{x_2}\lambda (dx_2)=+\infty,\ \ \forall x_1\in[0,1].$$
So the claim above is true.

(iii) If $P[X_{I_2}|X_{I_2^c}](\omega_{I_2^c},A_{I_2})$ exists, then
$$E_{[X_\Lambda]}(f)\in\mathbb{R}\Rightarrow E_{[X_{I_2}]}(f|X_{I_2^c})\in\mathbb{R},\ \ a.e.\{P[X_{I_2^c}]\}.$$
\end{comm}

\begin{thm}\label{theorem6.1}Let $(\Omega,\mathscr{F},P)$ be a given probability space, and let $(\Omega_i,\mathscr{F}_i),\ i=1,2,\cdots,n,n+1,\cdots,2n$ be measurable spaces. Let $X_i:(\Omega,\mathscr{F})\rightarrow(\Omega_i,\mathscr{F}_i),\ i=1,2,\cdots,2n$ be random objects, and $P[X_i]$ the probability on $(\Omega_i,\mathscr{F}_i)$ induced by $X_i$, $i=1,2,\cdots,2n$. Then $(\Omega_i,\mathscr{F}_i,P[X_i]),\ i=1,2,\cdots,2n$ are probability spaces. Given fixed $A_i\in\mathscr{F}_i,i=n+1,n+2,\cdots,2n$ with $P[X_i\in A_i,i=n+1,n+2,\cdots,2n]>0$. If there exist $P[X_i,X_{n+i}\in A_{n+i}|X_j,j=1,2,\cdots,i-1;X_{n+j}\in A_{n+j},j=1,2,\cdots,i-1]$, $i=2,3,\cdots,n$, then

1. For $2\leq i\leq n$,
$$P[X_j,j=1,2,\cdots,i;X_{n+j}\in A_{n+j},j=1,2,\cdots,i]$$
is the measure on $(\prod_{j=1}^i\Omega_j,\prod_{j=1}^i\mathscr{F}_j)$ which is induced by $P[X_1,X_{n+1}\in A_{n+1}]$ and
$$P[X_j,X_{n+j}\in A_{n+j}|X_k,k=1,2,\cdots,j-1;X_{n+k}\in A_{n+k},k=1,2,\cdots,j-1],$$
$j=2,3,\cdots,i$.

2. For $1\leq i\leq n-2$, there exists
$$\begin{array}{l}P[X_j,j=i+1,i+2,\cdots,n;X_{n+j}\in A_{n+j},j=i+1,i+2,\cdots,n\\
\ \ \ \ |X_j,j=1,2,\cdots,i;X_{n+j}\in A_{n+j},j=1,2,\cdots,i]\end{array}$$
such that, for $a.e.\{P[X_j,j=1,2,\cdots,i;X_{n+j}\in A_{n+j},j=1,2,\cdots,i]\}$ $(\omega_1,\omega_2,\cdots,\omega_i)\in\prod_{j=1}^{i}\Omega_j$,
$$\begin{array}{l}P[X_j,j=i+1,i+2,\cdots,n;X_{n+j}\in A_{n+j},j=i+1,i+2,\cdots,n\\
\ \ \ \ |X_j,j=1,2,\cdots,i;X_{n+j}\in A_{n+j},j=1,2,\cdots,i]\end{array}$$
is the measure on $(\prod_{j=i+1}\Omega_j,\prod_{j=i+1}\mathscr{F}_j)$ induced by $$P[X_{i+1},X_{n+i+1}\in A_{n+i+1}|X_j,j=1,2,\cdots,i;X_{n+j}\in A_{n+j},j=1,2,\cdots,i]$$ and $$P[X_k,X_{n+k}\in A_{n+k}|X_j,j=1,2,\cdots,k-1;X_{n+j}\in A_{n+j},j=1,2,\cdots,k-1],$$ $k=i+2,i+3,\cdots,n$.

3. Let $Y(\omega_1,\omega_2,\cdots,\omega_n)$ be a random variable on $(\prod_{i=1}^n\Omega_i,\prod_{i=1}^n\mathscr{F}_i,P[X_i,i=1,2,\cdots,n])$ with $E_{[X_i,i=1,2,\cdots,n;X_{n+i}\in A_{n+i},i=1,2,\cdots,n]}(Y)\in \mathbb{R}$. Then there exists $E\in\prod_{i=1}^{n}\mathscr{F}_i$ with
$$P[X_j,j=1,2,\cdots,n;X_{n+j}\in A_{n+j},j=1,2,\cdots,n](E^c)=0, \ \ E^c=\prod_{i=1}^{n}\Omega_i\backslash E,$$
such that
$$\begin{array}{l}\displaystyle E_{[X_j,j=1,2,\cdots,n;X_{n+j}\in A_{n+j},j=1,2,\cdots,n]}(Y)=E_{[X_j,j=1,2,\cdots,n;X_{n+j}\in A_{n+j},j=1,2,\cdots,n]}(Y\mathds{1}_E)\\\\
\displaystyle =E_{[X_j,j=1,2,\cdots,i;X_{n+j}\in A_{n+j},j=1,2,\cdots,i]}(E_{[X_j,j=i+1,i+2,\cdots,n;X_{n+j}\in A_{n+j},j=i+1,i+2,\cdots,n]}\\
\displaystyle \ \ \ (Y\mathds{1}_E|X_j,j=1,2,\cdots,i;X_{n+j}\in A_{n+j},j=1,2,\cdots,i))\\\\
\displaystyle =E_{[X_1,X_{n+1}\in A_{n+1}]}(E_{[X_2,X_{n+2}\in A_{n+2}]}(E_{[X_3,X_{n+3}\in A_{n+3}]}(\cdots \\
\displaystyle \ \ \ E_{[X_n,X_{2n}\in A_{2n}]}(Y\mathds{1}_E|X_j,j=1,2,\cdots,n-1;X_{n+j}\in A_{n+j},j=1,2,\cdots,n-1)\cdots\\
\displaystyle \ \ \ |X_1,X_2;X_{n+1}\in A_{n+1},X_{n+2}\in A_{n+2})|X_1,X_{n+1}\in A_{n+1})),
\end{array}$$
where $$\begin{array}{l}E_{[X_j,j=i+1,i+2,\cdots,n;X_{n+j}\in A_{n+j},j=i+1,i+2,\cdots,n]}(Y\mathds{1}_E\\
\ \ \ \ |X_j,j=1,2,\cdots,i;X_{n+j}\in A_{n+j},j=1,2,\cdots,i)\\\\
\displaystyle =E_{[X_{i+1},X_{n+i+1}\in A_{n+i+1}]}(E_{[X_{i+2},X_{n+i+2}\in A_{n+i+2}]}(E_{[X_{i+3},X_{n+i+3}\in A_{n+i+3}]}(\cdots \\
\displaystyle \ \ \ E_{[X_n,X_{2n}\in A_{2n}]}(Y\mathds{1}_E|X_j,j=1,2,\cdots,n-1;X_{n+j}\in A_{n+j},j=1,2,\cdots,n-1)\cdots\\
\displaystyle \ \ \ |X_j,j=1,2,\cdots,i+2;X_{n+j}\in A_{n+j},j=1,2,\cdots,i+2)\\
\displaystyle \ \ \ |X_j,j=1,2,\cdots,i+1;X_{n+j}\in A_{n+j},j=1,2,\cdots,i+1)\\
\displaystyle \ \ \ |X_j,j=1,2,\cdots,i;X_{n+j}\in A_{n+j},j=1,2,\cdots,i),\\
\ \ \ a.e.\{P[X_j,j=1,2,\cdots,i;X_{n+j}\in A_{n+j},j=1,2,\cdots,i]\},\ \ i=1,2,\cdots,n-1\end{array}$$
which is a measurable real function on $(\prod_{j=1}^{i}\Omega_j,\prod_{j=1}^{i}\mathscr{F}_j)$, respectively; and $\mathds{1}_E$ is the indicator of $E$.
\end{thm}

\begin{comm}\label{comment6.1.1}(i) In view of  Theorem \ref{theorem6.1}, if $E_{[X_i,i=1,2,\cdots,n]}(Y)\in \mathbb{R}$, then $Y$ always means $Y\mathds{1}_E$. In fact, if the value $\infty$ or $\infty-\infty$ of
$$\begin{array}{l}E_{[X_j,j=i+1,i+2,\cdots,n;X_{n+j}\in A_{n+j},j=i+1,i+2,\cdots,n]}(Y\\
\ \ \ \ |X_j,j=1,2,\cdots,i;X_{n+j}\in A_{n+j},j=1,2,\cdots,i),\end{array}$$
$i=n-1,n-2,\cdots,2,1$, is redefined as a finite real number, e.g., zero, then
$$\begin{array}{l}E_{[X_j,j=1,2,\cdots,n;X_{n+j}\in A_{n+j},j=1,2,\cdots,n]}(Y)\\\\
\displaystyle =E_{[X_1,X_{n+1}\in A_{n+1}]}(E_{[X_2,X_{n+2}\in A_{n+2}]}(E_{[X_3,X_{n+3}\in A_{n+3}]}(\cdots \\
\displaystyle \ \ \ E_{[X_n,X_{2n}\in A_{2n}]}(Y|X_j,j=1,2,\cdots,n-1;X_{n+j}\in A_{n+j},j=1,2,\cdots,n-1)\cdots\\
\displaystyle \ \ \ |X_1,X_2;X_{n+1}\in A_{n+1},X_{n+2}\in A_{n+2})|X_1,X_{n+1}\in A_{n+1})).\end{array}$$

(ii) In Theorem \ref{theorem6.1} 2., when $n=3$ and $i=1$, it is Theorem \ref{theorem3.7}.
\end{comm}

By Corollary \ref{corollary3.8.1}, we obtain the following corollary.

\begin{col}\label{corollary6.1.1}Let $(\Omega,\mathscr{F},P)$, $(\Omega_i,\mathscr{F}_i),\ i=1,2,\cdots,n,n+1,\cdots,2n$; $X_i(\omega),\ i=1,2,\cdots,2n$; $A_i\in\mathscr{F}_i,i=n+1,n+2,\cdots,2n$ and $Y(\omega_1,\omega_2,\cdots,\omega_n)$ are the same as in Theorem \ref{theorem6.1}. If $(X_i,X_{n+i}),j=1,2,\cdots,n$ are independent, then there exists $E\in\prod_{i=1}^{n}\mathscr{F}_i$ with $P[X_j,j=1,2,\cdots,n;X_{n+j}\in A_{n+j},j=1,2,\cdots,n](E^c)=0$ such that
$$\begin{array}{l}\displaystyle E_{[X_j,j=1,2,\cdots,n;X_{n+j}\in A_{n+j},j=1,2,\cdots,n]}(\mathds{1}_E\cdot Y)\\\\
\displaystyle =E_{[X_j,j=1,2,\cdots,i;X_{n+j}\in A_{n+j},j=1,2,\cdots,i]}(E_{[X_j,j=i+1,i+2,\cdots,n;X_{n+j}\in A_{n+j},j=i+1,i+2,\cdots,n]}(\mathds{1}_E\cdot Y))\\\\
\displaystyle =E_{[X_1,X_{n+1}\in A_{n+1}]}(E_{[X_2,X_{n+2}\in A_{n+2}]}(\cdots E_{[X_n,X_{2n}\in A_{2n}]}(\mathds{1}_E\cdot Y)\cdots)),
\end{array}$$
where
$$\begin{array}{l}E_{[X_j,j=i+1,i+2,\cdots,n;X_{n+j}\in A_{n+j},j=i+1,i+2,\cdots,n]}(\mathds{1}_E\cdot Y)\\\\
=E_{[X_{i+1},X_{n+i+1}\in A_{n+i+1}]}(E_{[X_{i+2},X_{n+i+2}\in A_{n+i+2}]}(\cdots E_{[X_{n},X_{2n}\in A_{2n}]}(\mathds{1}_E\cdot Y)\cdots)),
\end{array}$$
$i=1,2,\cdots,n-1$, is a measurable real function on $(\prod_{j=1}^{i}\Omega_j,\prod_{j=1}^{i}\mathscr{F}_j)$, respectively.
\end{col}

\begin{thm}\label{theorem6.2}Let $(\Omega,\mathscr{F},P)$ be a given probability space, and let $(\Omega_i,\mathscr{F}_i),\ i=1,2,\cdots,6$ be measurable spaces. Let $X_i:(\Omega,\mathscr{F})\rightarrow(\Omega_i,\mathscr{F}_i),\ i=1,2,\cdots,6$ be random objects, and $P[X_i]$ the probability on $(\Omega_i,\mathscr{F}_i)$ induced by $X_i$, $i=1,2,\cdots,6$. Then $(\Omega_i,\mathscr{F}_i,P[X_i])$, $i=1,2,\cdots,6$ are probability spaces. Let $Y(\omega_2)$ be a random variable on $(\Omega_2,\mathscr{F}_2,P[X_2])$ with $E_{[X_2]}(Y)\in \mathbb{R}$. Let $\mathds{1}_{A_i}$ be the indicator of $A_i$, where $A_i\in\mathscr{F}_i$, $i=1,2,\cdots,6$.

1. For fixed $A_i\in\mathscr{F}_i,i=3,4,5$ and fixed $A_{4i}\in\mathscr{F}_4,i=1,2$ with $A_{41}\cap A_{42}=\emptyset$,
$$\begin{array}{l}E_{[X_2;X_4\in A_{41}+A_{42},X_5\in A_5]}(Y)=E_{[X_2;X_4\in A_{41},X_5\in A_5]}(Y)+E_{[X_2;X_4\in A_{42},X_5\in A_5]}(Y)\end{array}$$
and
$$\begin{array}{l}E_{[X_2;X_4\in A_{41}+A_{42},X_5\in A_5]}(Y|X_1,X_3\in A_3)\\
=E_{[X_2;X_4\in A_{41},X_5\in A_5]}(Y|X_1,X_3\in A_3)+E_{[X_2;X_4\in A_{42},X_5\in A_5]}(Y|X_1,X_3\in A_3).\end{array}$$

2. For fixed $A_3\in\mathscr{F}_3$,
$$\begin{array}{l}E_{[X_i,i=1,2;X_3\in A_3]}(Y\cdot \mathds{1}_{A_1}\cdot \mathds{1}_{A_2})=E_{[X_2;X_i\in A_i,i=1,3]}(Y\cdot \mathds{1}_{A_2}),\ \ \forall A_i\in\mathscr{F}_i,i=1,2.\end{array}$$
Especially,
$$\begin{array}{l}E_{[X_i,i=1,2;X_3\in A_3]}(Y\cdot \mathds{1}_{A_1})=E_{[X_2;X_i\in A_i,i=1,3]}(Y),\end{array}$$
and
$$\begin{array}{l}E_{[X_i,i=1,2;X_3\in A_3]}(Y)=E_{[X_2,X_3\in A_3]}(Y).\end{array}$$

3. For fixed $A_i\in\mathscr{F}_i,i=3,4,6$,
$$E_{[X_2,X_4;X_6\in A_6]}(Y\cdot \mathds{1}_{A_4}|X_1,X_3\in A_3)=E_{[X_2;X_4\in A_4,X_6\in A_6]}(Y|X_1,X_3\in A_3).$$
Especially,
$$E_{[X_2,X_4;X_6\in A_6]}(Y|X_1,X_3\in A_3)=E_{[X_2,X_6\in A_6]}(Y|X_1,X_3\in A_3).$$

4. For fixed $A_i\in\mathscr{F}_i,i=3,4,5,6$, if there exists $P[X_3,X_5\in A_5|X_1,X_4\in A_4]$, then
$$\begin{array}{l}E_{[X_3,X_5\in A_5]}(E_{[X_2;X_6\in A_6]}(Y|X_1,X_3;X_4\in A_4,X_5\in A_5)\cdot \mathds{1}_{A_3}|X_1,X_4\in A_4)\\
=E_{[X_2;X_i\in A_i,i=3,5,6]}(Y|X_1,X_4\in A_4),\ \ a.e.\{P[X_1,X_4\in A_4]\}.\end{array}$$
Especially,
$$\begin{array}{l}E_{[X_3,X_5\in A_5]}(E_{[X_2;X_6\in A_6]}(Y|X_1,X_3;X_4\in A_4,X_5\in A_5)|X_1,X_4\in A_4)\\
=E_{[X_2;X_i\in A_i,i=5,6]}(Y|X_1,X_4\in A_4),\ \ a.e.\{P[X_1,X_4\in A_4]\}.\end{array}$$
\end{thm}

\begin{col}\label{corollary6.2.1}Let $\Lambda$ be a nonempty index set. Let $(\Omega,\mathscr{F},P)$ be a given probability space, and $(\Omega_{i},\mathscr{F}_{i}),\ i\in \Lambda$ measurable spaces. Let $X_i:(\Omega,\mathscr{F})\rightarrow(\Omega_i,\mathscr{F}_i),\ i\in \Lambda$ be random objects. Given $I_i\subset \Lambda,\ i=1,2,\cdots,6$ that are pairwise disjoint. Let $Y(\omega_{I_2})$ be a random variable on $(\Omega_{I_2},\mathscr{F}_{I_2},P[X_{I_2}])$ with $E_{[X_{I_2}]}(Y)\in \mathbb{R}$. Let $\mathds{1}_{A_{I_i}}$ be the indicator of $A_{I_i}$, where $A_{I_i}\in\mathscr{F}_{I_i}$, $i=1,2,\cdots,6$.

1. For fixed $A_{I_i}\in\mathscr{F}_{I_i},i=3,4,5$ and fixed $A_{I_{4i}}\in\mathscr{F}_{I_4},i=1,2$ with $A_{I_{41}}\cap A_{I_{42}}=\emptyset$,
$$\begin{array}{l}E_{[X_{I_2};X_{I_4}\in A_{I_{41}}+A_{I_{42}},X_{I_5}\in A_{I_5}]}(Y)=E_{[X_{I_2};X_{I_4}\in A_{I_{41}},X_{I_5}\in A_{I_5}]}(Y)+E_{[X_{I_2};X_{I_4}\in A_{I_{42}},X_{I_5}\in A_{I_5}]}(Y)\end{array}$$
and
$$\begin{array}{l}E_{[X_{I_2};X_{I_4}\in A_{I_{41}}+A_{I_{42}},X_{I_5}\in A_{I_5}]}(Y|X_{I_1},X_{I_3}\in A_{I_3})\\
=E_{[X_{I_2};X_{I_4}\in A_{I_{41}},X_{I_5}\in A_{I_5}]}(Y|X_{I_1},X_{I_3}\in A_{I_3})+E_{[X_{I_2};X_{I_4}\in A_{I_{42}},X_{I_5}\in A_{I_5}]}(Y|X_{I_1},X_{I_3}\in A_{I_3}).\end{array}$$

2. For fixed $A_{I_3}\in\mathscr{F}_{I_3}$,
$$\begin{array}{l}E_{[X_{I_i},i=1,2;X_{I_3}\in A_{I_3}]}(Y\cdot \mathds{1}_{A_{I_1}}\cdot \mathds{1}_{A_{I_2}})=E_{[X_{I_2};X_{I_i}\in A_{I_i},i=1,3]}(Y\cdot \mathds{1}_{A_{I_2}}),\ \ \forall A_{I_i}\in\mathscr{F}_{I_i},i=1,2.\end{array}$$
Especially,
$$\begin{array}{l}E_{[X_{I_i},i=1,2;X_{I_3}\in A_{I_3}]}(Y\cdot \mathds{1}_{A_{I_1}})=E_{[X_{I_2};X_{I_i}\in A_{I_i},i=1,3]}(Y),\end{array}$$
and
$$\begin{array}{l}E_{[X_{I_i},i=1,2;X_{I_3}\in A_{I_3}]}(Y)=E_{[X_{I_2},X_{I_3}\in A_{I_3}]}(Y).\end{array}$$

3. For fixed $A_{I_i}\in\mathscr{F}_{I_i},i=3,4,6$,
$$E_{[X_{I_2},X_{I_4};X_{I_6}\in A_{I_6}]}(Y\cdot \mathds{1}_{A_{I_4}}|X_{I_1},X_{I_3}\in A_{I_3})=E_{[X_{I_2};X_{I_4}\in A_{I_4},X_{I_6}\in A_{I_6}]}(Y|X_{I_1},X_{I_3}\in A_{I_3}).$$
Especially,
$$E_{[X_{I_2},X_{I_4};X_{I_6}\in A_{I_6}]}(Y|X_{I_1},X_{I_3}\in A_{I_3})=E_{[X_{I_2},X_{I_6}\in A_{I_6}]}(Y|X_{I_1},X_{I_3}\in A_{I_3}).$$

4. For fixed $A_{I_i}\in\mathscr{F}_{I_i},i=3,4,5,6$, if there exists $P[X_{I_3},X_{I_5}\in A_{I_5}|X_{I_1},X_{I_4}\in A_{I_4}]$, then
$$\begin{array}{l}E_{[X_{I_3},X_{I_5}\in A_{I_5}]}(E_{[X_{I_2};X_{I_6}\in A_{I_6}]}(Y|X_{I_1},X_{I_3};X_{I_4}\in A_{I_4},X_{I_5}\in A_{I_5})\cdot \mathds{1}_{A_{I_3}}|X_{I_1},X_{I_4}\in A_{I_4})\\
=E_{[X_{I_2};X_{I_i}\in A_{I_i},i=3,5,6]}(Y|X_{I_1},X_{I_4}\in A_{I_4}),\ \ a.e.\{P[X_{I_1},X_{I_4}\in A_{I_4}]\}.\end{array}$$
Especially,
$$\begin{array}{l}E_{[X_{I_3},X_{I_5}\in A_{I_5}]}(E_{[X_{I_2};X_{I_6}\in A_{I_6}]}(Y|X_{I_1},X_{I_3};X_{I_4}\in A_{I_4},X_{I_5}\in A_{I_5})|X_{I_1},X_{I_4}\in A_{I_4})\\
=E_{[X_{I_2};X_{I_i}\in A_{I_i},i=5,6]}(Y|X_{I_1},X_{I_4}\in A_{I_4}),\ \ a.e.\{P[X_{I_1},X_{I_4}\in A_{I_4}]\}.\end{array}$$
\end{col}

\begin{thm}\label{theorem6.3}Let $(\Omega,\mathscr{F},P)$ be a given probability space, and let $(\Omega_i,\mathscr{F}_i),\ i=1,2,3,4,5$ be measurable spaces. Let $X_i:(\Omega,\mathscr{F})\rightarrow(\Omega_i,\mathscr{F}_i),\ i=1,2,3,4,5$ be random objects, and $P[X_i]$ the probability on $(\Omega_i,\mathscr{F}_i)$ induced by $X_i$, $i=1,2,3,4,5$. Then $(\Omega_i,\mathscr{F}_i,P[X_i])$, $i=1,2,3,4,5$ are probability spaces. If $Y(\omega_2)$ is a random variable on $(\Omega_2,\mathscr{F}_2,P[X_2])$ with $E_{[X_2]}(Y)\in \mathbb{R}$, then for fixed $A_i\in\mathscr{F}_i,\ i=3,4,5$,
$$\begin{array}{l}E_{[X_2;X_3\in A_3,X_4\in A_4]}(Y|X_1,X_5\in A_5)\\
=E_{[X_2,X_4\in A_4]}(Y|X_1;X_3\in A_3,X_5\in A_5)\cdot P[X_3\in A_3|X_1,X_5\in A_5]\end{array}$$
and
$$\begin{array}{l}\displaystyle E_{[X_2,X_4\in A_4]}(Y|X_1;X_3\in A_3,X_5\in A_5)\\\\
\displaystyle =\left\{\begin{array}{l}\displaystyle \frac{E_{[X_2;X_3\in A_3,X_4\in A_4]}(Y|X_1,X_5\in A_5)}{P[X_3\in A_3|X_1,X_5\in A_5]},\ \omega_1\in D,\\
\displaystyle 0,\ \ \ \ \ \ \ \ \ \omega_1\in D^c,\end{array}\right.\end{array}$$
where $D=\{\omega_1:P[X_3\in A_3|X_1,X_5\in A_5](\omega_1)\neq 0\}$ and $D^c=\Omega_1\backslash D$ with $P[X_1;X_3\in A_3,X_5\in A_5](D^c)=0$.
\end{thm}

\begin{comm}\label{comment6.3.1}In the proof of Theorem \ref{theorem6.3} we don't consider if there exist $P[X_2,X_4\in A_4|X_1;X_3\in A_3,X_5\in A_5]$ and $P[X_2;X_3\in A_3,X_4\in A_4|X_1,X_5\in A_5]$. If one of them exists, then, by Theorem \ref{theorem3.2}, so does the other. Then one can give the proof of Theorem \ref{theorem6.3} by Comment \ref{comment5.4.2}.
\end{comm}

We then have the following corollary.

\begin{col}\label{corollary6.3.1}Let $\Lambda$ be a nonempty index set. Let $(\Omega,\mathscr{F},P)$ be a given probability space, and $(\Omega_{i},\mathscr{F}_{i}),\ i\in \Lambda$ measurable spaces. Let $X_i:(\Omega,\mathscr{F})\rightarrow(\Omega_i,\mathscr{F}_i),\ i\in \Lambda$ be random objects. Given $I_i\subset \Lambda,\ i=1,2,\cdots,5$ that are pairwise disjoint. If $Y(\omega_{I_2})$ is a random variable on $(\Omega_{I_2},\mathscr{F}_{I_2},P[X_{I_2}])$ with $E_{[X_{I_2}]}(Y)\in \mathbb{R}$, then for fixed $A_{I_i}\in\mathscr{F}_{I_i},\ i=3,4,5$,
$$\begin{array}{l}E_{[X_{I_2};X_{I_3}\in A_{I_3},X_{I_4}\in A_{I_4}]}(Y|X_{I_1},X_{I_5}\in A_{I_5})\\\\
=E_{[X_{I_2},X_{I_4}\in A_{I_4}]}(Y|X_{I_1};X_{I_3}\in A_{I_3},X_{I_5}\in A_{I_5})\cdot P[X_{I_3}\in A_{I_3}|X_{I_1},X_{I_5}\in A_{I_5}]\end{array}$$
and
$$\begin{array}{l}\displaystyle E_{[X_{I_2},X_{I_4}\in A_{I_4}]}(Y|X_{I_1};X_{I_3}\in A_{I_3},X_{I_5}\in A_{I_5})\\\\
\displaystyle =\left\{\begin{array}{l}\displaystyle \frac{E_{[X_{I_2};X_{I_3}\in A_{I_3},X_{I_4}\in A_{I_4}]}(Y|X_{I_1},X_{I_5}\in A_{I_5})}{P[X_{I_3}\in A_{I_3}|X_{I_1},X_{I_5}\in A_{I_5}]},\ \omega_{I_1}\in D,\\\\
\displaystyle 0,\ \ \ \ \ \ \ \ \ \omega_{I_1}\in D^c,\end{array}\right.\end{array}$$
where $D=\{\omega_{I_1}:P[X_{I_3}\in A_{I_3}|X_{I_1},X_{I_5}\in A_{I_5}](\omega_{I_1})\neq 0\}$ and $D^c=\Omega_{I_1}\backslash D$ with $P[X_{I_1};X_{I_3}\in A_{I_3},X_{I_5}\in A_{I_5}](D^c)=0$.
\end{col}

From Corollary \ref{corollary6.3.1} the corollary below follows.

\begin{col}\label{corollary6.3.2}  Let $(\Omega,\mathscr{F},P)$ be a given probability space. Let $X_i,\ i\in \mathbb{N}$ be random variables on $(\Omega,\mathscr{F},P)$. Given $I_i\subset \mathbb{N},\ i=1,2,\cdots,5$ that are pairwise disjoint. If $Y$ is a random variable on $(\mathbb{R}^{I_2},\mathscr{B}(\mathbb{R}^{I_2}),P[X_{I_2}])$ with $E_{[X_{I_2}]}(Y)\in\mathbb{R}$, then for fixed $A_i\in\mathscr{B}(\mathbb{R}),\  i\in I_3+I_4+I_5$,
$$\begin{array}{l}E_{[X_i,i\in I_2;X_i\in A_i,i\in I_3+I_4]}(Y|X_i,i\in I_1;X_i\in A_i,i\in I_5)\\\\
=E_{[X_i,i\in I_2;X_i\in A_i,i\in I_4]}(Y|X_i,i\in I_1;X_i\in A_i,i\in I_3+I_5)\\
\ \ \ \cdot P[X_i\in A_i,i\in I_3|X_i,i\in I_1;X_i\in A_i,i\in I_5]\end{array}$$
and
$$\begin{array}{l}\displaystyle E_{[X_i,i\in I_2;X_i\in A_i,i\in I_4]}(Y|X_i,i\in I_1;X_i\in A_i,i\in I_3+I_5)\\\\
\displaystyle =\left\{\begin{array}{l}\displaystyle \frac{E_{[X_i,i\in I_2;X_i\in A_i,i\in I_3+I_4]}(Y|X_i,i\in I_1;X_i\in A_i,i\in I_5)}{P[X_i\in A_i,i\in I_3|X_i,i\in I_1;X_i\in A_i,i\in I_5]},\ x_{I_1}\in D,\\\\
\displaystyle 0,\ \ \ \ \ \ \ \ \ x_{I_1}\in D^c,\end{array}\right.\end{array}$$
where $D=\{x_{I_1}:P[X_i\in A_i,i\in I_3|X_i,i\in I_1;X_i\in A_i,i\in I_5](x_{I_1})\neq 0\}$ and $D^c=\mathbb{R}^{I_1}\backslash D$ with $P[X_i,i\in I_1;X_i\in A_i,i\in I_3+I_5](D^c)=0$.
\end{col}

\begin{thm}\label{theorem6.4}Let $(\Omega,\mathscr{F},P)$ be a given probability space, and let $(\Omega_i,\mathscr{F}_i),\ i=1,2,\cdots,6$ be measurable spaces. Let $X_i:(\Omega,\mathscr{F})\rightarrow(\Omega_i,\mathscr{F}_i),\ i=1,2,\cdots,6$ be random objects, and $P[X_i]$ the probability on $(\Omega_i,\mathscr{F}_i)$ induced by $X_i$, $i=1,2,\cdots,6$. Then $(\Omega_i,\mathscr{F}_i,P[X_i]),\ i=1,2,\cdots,6$ are probability spaces.  Let $Y(\omega_2)$ be a random variable on $(\Omega_2,\mathscr{F}_2,P[X_2])$ with $E_{[X_2]}(Y)\in \mathbb{R}$.\\

1. For fixed $A_i\in\mathscr{F}_i,i=4,5,6$ with $P[X_4\in A_4,X_6\in A_6]>0$, if
$$\begin{array}{l}P[X_2,X_3;X_5\in A_5,X_6\in A_6|X_1,X_4\in A_4]\\
=P[X_2,X_5\in A_5|X_1,X_4\in A_4]\times P[X_3,X_6\in A_6|X_1,X_4\in A_4],\end{array}$$
then
$$E_{[X_2,X_5\in A_5]}(Y|X_1,X_3;X_4\in A_4,X_6\in A_6)=E_{[X_2,X_5\in A_5]}(Y|X_1,X_4\in A_4).$$

2. For fixed $A_i\in\mathscr{F}_i,i=4,5,6$ with $P[X_4\in A_4,X_6\in A_6]>0$, if
$$\begin{array}{l}P[X_2;X_5\in A_5,X_6\in A_6|X_1,X_4\in A_4]\\
=P[X_2,X_5\in A_5|X_1,X_4\in A_4]\cdot P[X_6\in A_6|X_1,X_4\in A_4],\end{array}$$
then
$$E_{[X_2,X_5\in A_5]}(Y|X_1;X_4\in A_4,X_6\in A_6)=E_{[X_2,X_5\in A_5]}(Y|X_1,X_4\in A_4).$$

3. For fixed $A_i\in\mathscr{F}_i,i=4,5,6$ with $P[X_4\in A_4,X_6\in A_6]>0$, if
$$\begin{array}{l}P[X_1,X_2;X_4\in A_4,X_5\in A_5|X_6\in A_6]\\
=P[X_2,X_5\in A_5|X_6\in A_6]\times P[X_1,X_4\in A_4|X_6\in A_6],\end{array}$$
then
$$E_{[X_2,X_5\in A_5]}(Y|X_1;X_4\in A_4,X_6\in A_6)=E_{[X_2,X_5\in A_5]}(Y|X_6\in A_6).$$
\end{thm}

\begin{col}\label{corollary6.4.1}Let $\Lambda$ be a nonempty index set. Let $(\Omega,\mathscr{F},P)$ be a given probability space, and $(\Omega_{i},\mathscr{F}_{i}),\ i\in \Lambda$ measurable spaces. Let $X_i:(\Omega,\mathscr{F})\rightarrow(\Omega_i,\mathscr{F}_i),\ i\in \Lambda$ be random objects. Given $I_i\subset \Lambda,\ i=1,2,\cdots,6$ that are pairwise disjoint. Let $Y(\omega_{I_2})$ be a random variable on $(\Omega_{I_2},\mathscr{F}_{I_2},P[X_{I_2}])$ with $E_{[X_{I_2}]}(Y)\in \mathbb{R}$.\\

1. For fixed $A_{I_i}\in\mathscr{F}_{I_i},i=4,5,6$ with $P[X_{I_4}\in A_{I_4},X_{I_6}\in A_{I_6}]>0$, if
$$\begin{array}{l}P[X_{I_2},X_{I_3};X_{I_5}\in A_{I_5},X_{I_6}\in A_{I_6}|X_{I_1},X_{I_4}\in A_{I_4}]\\\\
=P[X_{I_2},X_{I_5}\in A_{I_5}|X_{I_1},X_{I_4}\in A_{I_4}]\times P[X_{I_3},X_{I_6}\in A_{I_6}|X_{I_1},X_{I_4}\in A_{I_4}],\end{array}$$
then
$$E_{[X_{I_2},X_{I_5}\in A_{I_5}]}(Y|X_{I_1},X_{I_3};X_{I_4}\in A_{I_4},X_{I_6}\in A_{I_6})=E_{[X_{I_2},X_{I_5}\in A_{I_5}]}(Y|X_{I_1},X_{I_4}\in A_{I_4}).$$

2. For fixed $A_{I_i}\in\mathscr{F}_{I_i},i=4,5,6$ with $P[X_{I_4}\in A_{I_4},X_{I_6}\in A_{I_6}]>0$, if
$$\begin{array}{l}P[X_{I_2};X_{I_5}\in A_{I_5},X_{I_6}\in A_{I_6}|X_{I_1},X_{I_4}\in A_{I_4}]\\\\
=P[X_{I_2},X_{I_5}\in A_{I_5}|X_{I_1},X_{I_4}\in A_{I_4}]\cdot P[X_{I_6}\in A_{I_6}|X_{I_1},X_{I_4}\in A_{I_4}],\end{array}$$
then
$$E_{[X_{I_2},X_{I_5}\in A_{I_5}]}(Y|X_{I_1};X_{I_4}\in A_{I_4},X_{I_6}\in A_{I_6})=E_{[X_{I_2},X_{I_5}\in A_{I_5}]}(Y|X_{I_1},X_{I_4}\in A_{I_4}).$$

3. For fixed $A_{I_i}\in\mathscr{F}_{I_i},i=4,5,6$ with $P[X_{I_4}\in A_{I_4},X_{I_6}\in A_{I_6}]>0$, if
$$\begin{array}{l}P[X_{I_1},X_{I_2};X_{I_4}\in A_{I_4},X_{I_5}\in A_{I_5}|X_{I_6}\in A_{I_6}]\\\\
=P[X_{I_2},X_{I_5}\in A_{I_5}|X_{I_6}\in A_{I_6}]\times P[X_{I_1},X_{I_4}\in A_{I_4}|X_{I_6}\in A_{I_6}],\end{array}$$
then
$$E_{[X_{I_2},X_{I_5}\in A_{I_5}]}(Y|X_{I_1}; X_{I_4}\in A_{I_4},X_{I_6}\in A_{I_6})=E_{[X_{I_2},X_{I_5}\in A_{I_5}]}(Y|X_{I_6}\in A_{I_6}).$$
\end{col}

From Corollary \ref{corollary6.4.1} the corollary below follows.

\begin{col}\label{corollary6.4.2}  Let $(\Omega,\mathscr{F},P)$ be a given probability space. Let $X_i,\ i\in \mathbb{N}$ be random variables on $(\Omega,\mathscr{F},P)$. Given $I_i\subset \mathbb{N},\ i=1,2,\cdots,5$ that are pairwise disjoint. Let $Y$ be a random variable on $(\mathbb{R}^{I_2},\mathscr{B}(\mathbb{R}^{I_2}),P[X_{I_2}])$ with $E_{[X_{I_2}]}(Y)\in\mathbb{R}$.\\

1. For fixed $A_i\in\mathscr{B}(\mathbb{R}),i\in I_4+I_5+I_6$ with $P[X_i\in A_i,i\in I_4+I_6]>0$, if
$$\begin{array}{l}P[X_i,i\in I_2+I_3;X_i\in A_i,i\in I_5+I_6|X_i,i\in I_1;X_i\in A_i,i\in I_4]\\\\
=P[X_i,i\in I_2;X_i\in A_i,i\in I_5|X_i,i\in I_1;X_i\in A_i,i\in I_4]\\
\times P[X_i,i\in I_3;X_i\in A_i,i\in I_6|X_i,i\in I_1;X_i\in A_i,i\in I_4],\end{array}$$
then
$$\begin{array}{l}E_{[X_i,i\in I_2;X_i\in A_i,i\in I_5]}(Y|X_i,i\in I_1+I_3;X_i\in A_i,i\in I_4+I_6)\\\\
=E_{[X_i,i\in I_2;X_i\in A_i,i\in I_5]}(Y|X_i,i\in I_1;X_i\in A_i,i\in I_4).\end{array}$$

2. For fixed $A_i\in\mathscr{B}(\mathbb{R}),i\in I_4+I_5+I_6$ with $P[X_i\in A_i,i\in I_4+I_6]>0$, if
$$\begin{array}{l}P[X_i,i\in I_2;X_i\in A_i,i\in I_5+I_6|X_i,i\in I_1;X_i\in A_i,i\in I_4]\\\\
=P[X_i,i\in I_2;X_i\in A_i,i\in I_5|X_i,i\in I_1;X_i\in A_i,i\in I_4]\\
\ \ \cdot P[X_i\in A_i,i\in I_6|X_i,i\in I_1;X_i\in A_i,i\in I_4],\end{array}$$
then
$$\begin{array}{l}E_{[X_i,i\in I_2;X_i\in A_i,i\in I_5]}(Y|X_i,i\in I_1;X_i\in A_i,i\in I_4+I_6)\\\\
=E_{[X_i,i\in I_2;X_i\in A_i,i\in I_5]}(Y|X_i,i\in I_1;X_i\in A_i,i\in I_4).\end{array}$$

3. For fixed $A_i\in\mathscr{B}(\mathbb{R}),i\in I_4+I_5+I_6$ with $P[X_i\in A_i,i\in I_4+I_6]>0$, if
$$\begin{array}{l}P[X_i,i\in I_1+I_2;X_i\in A_i,i\in I_4+I_5|X_i\in A_i,i\in I_6]\\\\
=P[X_i,i\in I_2;X_i\in A_i,i\in I_5|X_i\in A_i,i\in I_6]\\
\times P[X_i,i\in I_1;X_i\in A_i,i\in I_4|X_i\in A_i,i\in I_6],\end{array}$$
then
$$\begin{array}{l}E_{[X_i,i\in I_2;X_i\in A_i,i\in I_5]}(Y|X_i,i\in I_1;X_i\in A_i,i\in I_4+I_6)\\\\
=E_{[X_i,i\in I_2;X_i\in A_i,i\in I_5]}(Y|X_i\in A_i,i\in I_6).\end{array}$$
\end{col}

\begin{thm}\label{theorem6.5}Let $(\Omega,\mathscr{F},P)$ be a given probability space, and let $(\Omega_i,\mathscr{F}_i),\ i=1,2,3,4$ be measurable spaces. Let $X_i:(\Omega,\mathscr{F})\rightarrow(\Omega_i,\mathscr{F}_i),\ i=1,2,3,4$ be random objects, and $P[X_i]$ the probability on $(\Omega_i,\mathscr{F}_i)$ induced by $X_i$, $i=1,2,3,4$. Then $(\Omega_i,\mathscr{F}_i,P[X_i]),\ i=1,2,3,4$ are probability spaces. Let $Y_1(\omega_2)$ and $Y_2(\omega_2)$ be random variables on $(\Omega_2,\mathscr{F}_2,P[X_2])$ with $E_{[X_2]}(Y_1),E_{[X_2]}(Y_2)\in \mathbb{R}$. Let $A_i\in\mathscr{F}_i$, $i=3,4$, where $P[X_3\in A_3]>0$. Let $\mathds{1}_{A_i}$ be the indicator of $A_i$, where $A_i\in\mathscr{F}_i$, $i=1,2,3,4$.\\

1. For $\alpha\in \mathbb{R}$,
$$E_{[X_2,X_4\in A_4]}(\alpha Y_1|X_3\in A_3)=\alpha E_{[X_2,X_4\in A_4]}(Y_1|X_3\in A_3),$$
$$E_{[X_2,X_4\in A_4]}(\alpha Y_1|X_1,X_3\in A_3)=\alpha E_{[X_2,X_4\in A_4]}(Y_1|X_1,X_3\in A_3),\ \ a.e.\{P[X_1,X_3\in A_3]\}.$$

2. $$\begin{array}{l}E_{[X_2,X_4\in A_4]}(Y_1+Y_2|X_3\in A_3)\\
=E_{[X_2,X_4\in A_4]}(Y_1|X_3\in A_3)+E_{[X_2,X_4\in A_4]}(Y_2|X_3\in A_3),\end{array}$$
$$\begin{array}{l}E_{[X_2,X_4\in A_4]}(Y_1+Y_2|X_1,X_3\in A_3)\\
=E_{[X_2,X_4\in A_4]}(Y_1|X_1,X_3\in A_3)+E_{[X_2,X_4\in A_4]}(Y_2|X_1,X_3\in A_3),\\\end{array}$$
$a.e.\{P[X_1,X_3\in A_3]\}.$\\

3. If  $Y_1\leq Y_2,\ \ a.e.\{P[X_2]\}$, then
$$\begin{array}{l}E_{[X_2,X_4\in A_4]}(Y_1|X_3\in A_3)\leq E_{[X_2,X_4\in A_4]}(Y_2|X_3\in A_3),\end{array}$$
$$\begin{array}{l}E_{[X_2,X_4\in A_4]}(Y_1|X_1,X_3\in A_3)\leq E_{[X_2,X_4\in A_4]}(Y_2|X_1,X_3\in A_3),\end{array}$$
$a.e.\{P[X_1,X_3\in A_3]\}$.  Especially,
$$\begin{array}{l}|E_{[X_2,X_4\in A_4]}(Y_1|X_3\in A_3)|\leq E_{[X_2,X_4\in A_4]}(|Y_1||X_3\in A_3),\end{array}$$
$$\begin{array}{l}|E_{[X_2,X_4\in A_4]}(Y_1|X_1,X_3\in A_3)|\leq E_{[X_2,X_4\in A_4]}(|Y_1||X_1,X_3\in A_3),\end{array}$$
$a.e.\{P[X_1,X_3\in A_3]\}$.
\end{thm}

\begin{col}\label{corollary6.5.1}Let $\Lambda$ be a nonempty index set. Let $(\Omega,\mathscr{F},P)$ be a given probability space, and $(\Omega_{i},\mathscr{F}_{i}),\ i\in \Lambda$ measurable spaces. Let $X_i:(\Omega,\mathscr{F})\rightarrow(\Omega_i,\mathscr{F}_i),\ i\in \Lambda$ be random objects. Given $I_i\subset \Lambda,\ i=1,2,3,4$ that are pairwise disjoint. Let $Y_1(\omega_{I_2})$ and $Y_2(\omega_{I_2})$ be random variables on $(\Omega_{I_2},\mathscr{F}_{I_2},P[X_{I_2}])$ with $E_{[X_{I_2}]}(Y_1),E_{[X_{I_2}]}(Y_2)\in \mathbb{R}$. Let $A_{I_i}\in\mathscr{F}_{I_i}$, $i=3,4$, where $P[X_{I_3}\in A_{I_3}]>0$. Let $\mathds{1}_{A_{I_i}}$ be the indicator of $A_{I_i}$, where $A_{I_i}\in\mathscr{F}_{I_i}$, $i=1,2,3,4$.\\

1. For $\alpha\in \mathbb{R}$,
$$E_{[X_{I_2},X_{I_4}\in A_{I_4}]}(\alpha Y_1|X_{I_3}\in A_{I_3})=\alpha E_{[X_{I_2},X_{I_4}\in A_{I_4}]}(Y_1|X_{I_3}\in A_{I_3}),$$
$$E_{[X_{I_2},X_{I_4}\in A_{I_4}]}(\alpha Y_1|X_{I_1},X_{I_3}\in A_{I_3})=\alpha E_{[X_{I_2},X_{I_4}\in A_{I_4}]}(Y_1|X_{I_1},X_{I_3}\in A_{I_3}),$$
$a.e.\{P[X_{I_1},X_{I_3}\in A_{I_3}]\}$.\\

2. $$\begin{array}{l}E_{[X_{I_2},X_{I_4}\in A_{I_4}]}(Y_1+Y_2|X_{I_3}\in A_{I_3})\\
=E_{[X_{I_2},X_{I_4}\in A_{I_4}]}(Y_1|X_{I_3}\in A_{I_3})+E_{[X_{I_2},X_{I_4}\in A_{I_4}]}(Y_2|X_{I_3}\in A_{I_3}),\end{array}$$
$$\begin{array}{l}E_{[X_{I_2},X_{I_4}\in A_{I_4}]}(Y_1+Y_2|X_{I_1},X_{I_3}\in A_{I_3})\\
=E_{[X_{I_2},X_{I_4}\in A_{I_4}]}(Y_1|X_{I_1},X_{I_3}\in A_{I_3})+E_{[X_{I_2},X_{I_4}\in A_{I_4}]}(Y_2|X_{I_1},X_{I_3}\in A_{I_3}),\end{array}$$
$a.e.\{P[X_{I_1},X_{I_3}\in A_{I_3}]\}.$\\

3. If  $Y_1\leq Y_2,\ \ a.e.\{P[X_{I_2}]\}$, then
$$\begin{array}{l}E_{[X_{I_2},X_{I_4}\in A_{I_4}]}(Y_1|X_{I_3}\in A_{I_3})\leq E_{[X_{I_2},X_{I_4}\in A_{I_4}]}(Y_2|X_{I_3}\in A_{I_3}),\end{array}$$
$$\begin{array}{l}E_{[X_{I_2},X_{I_4}\in A_{I_4}]}(Y_1|X_{I_1},X_{I_3}\in A_{I_3})\leq E_{[X_{I_2},X_{I_4}\in A_{I_4}]}(Y_2|X_{I_1},X_{I_3}\in A_{I_3}),\end{array}$$
$a.e.\{P[X_{I_1},X_{I_3}\in A_{I_3}]\}$.  Especially,
$$\begin{array}{l}|E_{[X_{I_2},X_{I_4}\in A_{I_4}]}(Y_1|X_{I_3}\in A_{I_3})|\leq E_{[X_{I_2},X_{I_4}\in A_{I_4}]}(|Y_1||X_{I_3}\in A_{I_3}),\end{array}$$
$$\begin{array}{l}|E_{[X_{I_2},X_{I_4}\in A_{I_4}]}(Y_1|X_{I_1},X_{I_3}\in A_{I_3})|\leq E_{[X_{I_2},X_{I_4}\in A_{I_4}]}(|Y_1||X_{I_1},X_{I_3}\in A_{I_3}),\end{array}$$
$a.e.\{P[X_{I_1},X_{I_3}\in A_{I_3}]\}$.
\end{col}

\begin{thm}\label{theorem6.6}Let $(\Omega,\mathscr{F},P)$ be a given probability space, and let $(\Omega_2,\mathscr{F}_2)$ be measurable space. Let $\mathscr{G}_1$ and $\mathscr{G}_2$ be sub $\sigma$-fields of $\mathscr{F}_2$ with $\mathscr{G}_1\subset\mathscr{G}_2$. Let $X_2:(\Omega,\mathscr{F})\rightarrow(\Omega_2,\mathscr{F}_2)$ be a random object, then the random objects $X_{2\mathscr{G}_i}:(\Omega,\mathscr{F})\rightarrow(\Omega_2,\mathscr{G}_i),i=1,2$ are denoted by $X_{2i},\ i=1,2$, respectively. Let $Y(\omega_2)$ be a random variable on $(\Omega_2,\mathscr{F}_2,P[X_2])$ with $E_{[X_2]}(Y)\in \mathbb{R}$. Then
$$E_{[X_{2}]}(E_{[X_2]}(Y|X_{22})|X_{21})=E_{[X_{22}]}(E_{[X_2]}(Y|X_{22})|X_{21})=E_{[X_2]}(Y|X_{21}).$$
\end{thm}

\begin{comm}\label{comment6.6.1}In Theorem \ref{theorem6.6}, let $X_2=I_0$, $X_{21}=I_{\mathscr{G}_1}$ and $X_{22}=I_{\mathscr{G}_2}$, where $\mathscr{G}_1$ and $\mathscr{G}_2$ are sub $\sigma$-fields of $\mathscr{F}$ with $\mathscr{G}_1\subset\mathscr{G}_2$. Then
$$E(E(Y|\mathscr{G}_2)|\mathscr{G}_1)=E(Y|\mathscr{G}_1),$$
which is Theorem 7.2.1 (10) in \cite{yan:2004} (or Theorem 5.5.10 (b) in \cite{Robert:2000}).
\end{comm}

\begin{thm}\label{theorem6.7}Let $(\Omega,\mathscr{F},P)$ be a given probability space, and let $(\Omega_i,\mathscr{F}_i),\ i=1,2,3$ be measurable spaces. Let $X_i:(\Omega,\mathscr{F})\rightarrow(\Omega_i,\mathscr{F}_i),\ i=1,2,3$ be random objects, and $P[X_i]$ the probability on $(\Omega_i,\mathscr{F}_i)$ induced by $X_i$, $i=1,2,3$. Then $(\Omega_i,\mathscr{F}_i,P[X_i]),\ i=1,2,3$ are probability spaces. Let $Y_n(\omega_2),n=1,2,3,\cdots,$ be random variables on $(\Omega_2,\mathscr{F}_2,P[X_2])$ with $E_{[X_2]}(Y_n)\in \mathbb{R},\ n=1,2,3,\cdots$.\\

1. If $Y_n\leq Y_{n+1},\ a.e.\{P[X_2]\},\ n=1,2,3,\cdots$,
$$\lim\limits_{n\to \infty}Y_n=Y,\ a.e.\{P[X_2]\},\ \  and
\ \ \sup\limits_n\{E_{[X_2]}(Y_n)\}\in \mathbb{R},$$
then for each fixed $A_i\in\mathscr{F}_i$, $i=1,3$, $E_{[X_2,X_3\in A_3]}(Y|X_1\in A_1)\in \mathbb{R}$, and
\begin{equation}\label{eqn6.7.1}E_{[X_2,X_3\in A_3]}(Y_n|X_1\in A_1)\uparrow E_{[X_2,X_3\in A_3]}(Y|X_1\in A_1).\end{equation}
Especially,
$$E_{[X_2,X_3\in A_3]}(Y_n)\uparrow E_{[X_2,X_3\in A_3]}(Y),\ \ and \ \ E_{[X_2]}(Y_n)\uparrow E_{[X_2]}(Y).$$

2. If $Y_n\geq Y_{n+1},\ a.e.\{P[X_2]\},\ n=1,2,3,\cdots$, $$\lim\limits_{n\to \infty}Y_n=Y,\ a.e.\{P[X_2]\}, \ \ and
\ \ \inf\limits_n\{E_{[X_2]}(Y_n)\}\in \mathbb{R},$$
then for each fixed $A_i\in\mathscr{F}_i$, $i=1,3$, $E_{[X_2,X_3\in A_3]}(Y|X_1\in A_1)\in \mathbb{R}$, and
$$E_{[X_2,X_3\in A_3]}(Y_n|X_1\in A_1)\downarrow E_{[X_2,X_3\in A_3]}(Y|X_1\in A_1).$$
Especially,
$$E_{[X_2,X_3\in A_3]}(Y_n)\downarrow E_{[X_2,X_3\in A_3]}(Y),\ \ and \ \ E_{[X_2]}(Y_n)\downarrow E_{[X_2]}(Y).$$
\end{thm}

\begin{col}\label{corollary6.7.1}Let $(\Omega,\mathscr{F},P)$ be a given probability space, $(\Omega_{i},\mathscr{F}_{i}),\ i\in \Lambda$ measurable spaces, where $\Lambda$ is a nonempty index set. Let $X_i(\omega):(\Omega,\mathscr{F})\rightarrow(\Omega_i,\mathscr{F}_i),\ i\in \Lambda$ be random objects. Let $I_i\subset\Lambda,\ i=1,2,3$, which are pairwise disjoint. Let $Y_n(\omega_{I_2}),n=1,2,3,\cdots$ be random variables on $(\Omega_{I_2},\mathscr{F}_{I_2},P[X_{I_2}])$ with $E_{[X_{I_2}]}(Y_n)\in \mathbb{R},\ n=1,2,3,\cdots$.\\

1. If $Y_n\leq Y_{n+1},\ a.e.\{P[X_{I_2}]\},\ n=1,2,3,\cdots$,
$$\lim\limits_{n\to \infty}Y_n=Y,\ a.e.\{P[X_{I_2}]\},\ \  and
\ \ \sup\limits_n\{E_{[X_{I_2}]}(Y_n)\}\in \mathbb{R},$$
then for each fixed $A_{I_i}\in\mathscr{F}_{I_i}$, $i=1,3$, $E_{[X_{I_2},X_{I_3}\in A_{I_3}]}(Y|X_{I_1}\in A_{I_1})\in \mathbb{R}$, and
$$E_{[X_{I_2},X_{I_3}\in A_{I_3}]}(Y_n|X_{I_1}\in A_{I_1})\uparrow E_{[X_{I_2},X_{I_3}\in A_{I_3}]}(Y|X_{I_1}\in A_{I_1}).$$
Especially,
$$E_{[X_{I_2},X_{I_3}\in A_{I_3}]}(Y_n)\uparrow E_{[X_{I_2},X_{I_3}\in A_{I_3}]}(Y),\ \ and \ \ E_{[X_{I_2}]}(Y_n)\uparrow E_{[X_{I_2}]}(Y).$$

2. If $Y_n\geq Y_{n+1},\ a.e.\{P[X_{I_2}]\},\ n=1,2,3,\cdots$,
$$\lim\limits_{n\to \infty}Y_n=Y,\ a.e.\{P[X_{I_2}]\},\ \  and
\ \ \inf\limits_n\{E_{[X_{I_2}]}(Y_n)\}\in \mathbb{R},$$
then for each fixed $A_{I_i}\in\mathscr{F}_{I_i}$, $i=1,3$, $E_{[X_{I_2},X_{I_3}\in A_{I_3}]}(Y|X_{I_1}\in A_{I_1})\in \mathbb{R}$, and
$$E_{[X_{I_2},X_{I_3}\in A_{I_3}]}(Y_n|X_{I_1}\in A_{I_1})\downarrow E_{[X_{I_2},X_{I_3}\in A_{I_3}]}(Y|X_{I_1}\in A_{I_1}).$$
Especially,
$$E_{[X_{I_2},X_{I_3}\in A_{I_3}]}(Y_n)\downarrow E_{[X_{I_2},X_{I_3}\in A_{I_3}]}(Y),\ \ and \ \ E_{[X_{I_2}]}(Y_n)\downarrow E_{[X_{I_2}]}(Y).$$
\end{col}

\begin{thm}\label{theorem6.8}Let $(\Omega,\mathscr{F},P)$ be a given probability space, and let $(\Omega_i,\mathscr{F}_i),\ i=1,2,3,4$ be measurable spaces. Let $X_i:(\Omega,\mathscr{F})\rightarrow(\Omega_i,\mathscr{F}_i),\ i=1,2,3,4$ be random objects, and $P[X_i]$ the probability on $(\Omega_i,\mathscr{F}_i)$ induced by $X_i$, $i=1,2,3,4$. Then $(\Omega_i,\mathscr{F}_i,P[X_i]),\ i=1,2,3,4$ are probability spaces. Let $Y_n(\omega_2),n=1,2,3,\cdots,$ be random variables on $(\Omega_2,\mathscr{F}_2,P[X_2])$ with $E_{[X_2]}(Y_n)\in \mathbb{R},\ n=1,2,3,\cdots$.\\

1. If $Y_n\leq Y_{n+1},\ a.e.\{P[X_2]\},\ n=1,2,\cdots$,
$$\lim\limits_{n\to \infty}Y_n=Y,\ a.e.\{P[X_2]\},\ \  and\ \ \sup\limits_n\{E_{[X_2]}(Y_n)\}\in \mathbb{R},$$
then $E_{[X_2]}(Y)\in \mathbb{R}$, and for $A_i\in\mathscr{F}_i$, $i=3,4$,
$$E_{[X_2,X_4\in A_4]}(Y_n|X_1,X_3\in A_3)\uparrow E_{[X_2,X_4\in A_4]}(Y|X_1,X_3\in A_3),\ a.e.\{P[X_1,X_3\in A_3]\}$$
provided $P[X_3\in A_3]>0$ while it is evident provided $P[X_3\in A_3]=0$.\\

2. If $Y_n\geq Y_{n+1},\ a.e.\{P[X_2]\},\ n=1,2,\cdots$,
$$\lim\limits_{n\to \infty}Y_n=Y,\ a.e.\{P[X_2]\},\ \  and\ \ \inf\limits_n\{E_{[X_2]}(Y_n)\}\in \mathbb{R},$$
then $E_{[X_2]}(Y)\in \mathbb{R}$, and for $A_i\in\mathscr{F}_i$, $i=3,4$,
$$E_{[X_2,X_4\in A_4]}(Y_n|X_1,X_3\in A_3)\downarrow E_{[X_2,X_4\in A_4]}(Y|X_1,X_3\in A_3),\ a.e.\{P[X_1,X_3\in A_3]\}$$
provided $P[X_3\in A_3]>0$ while it is evident provided $P[X_3\in A_3]=0$.
\end{thm}

\begin{col}\label{corollary6.8.1}Let $(\Omega,\mathscr{F},P)$ be a given probability space, $(\Omega_{i},\mathscr{F}_{i}),\ i\in \Lambda$ measurable spaces, where $\Lambda$ is a nonempty index set. Let $X_i:(\Omega,\mathscr{F})\rightarrow(\Omega_i,\mathscr{F}_i),\ i\in \Lambda$ be random objects. Let $I_i\subset\Lambda,\ i=1,2,3,4$, which are pairwise disjoint. Let $Y_n(\omega_{I_2}),n=1,2,3,\cdots$ be random variables on $(\Omega_{I_2},\mathscr{F}_{I_2},P[X_{I_2}])$ with $E_{[X_{I_2}]}(Y_n)\in \mathbb{R},\ n=1,2,3,\cdots$.\\

1. If $Y_n\leq Y_{n+1},\ a.e.\{P[X_{I_2}]\},\ n=1,2,\cdots$,
$$\lim\limits_{n\to \infty}Y_n=Y,\ a.e.\{P[X_{I_2}]\},\ \  and\ \ \sup\limits_n\{E_{[X_{I_2}]}(Y_n)\}\in \mathbb{R},$$
then $E_{[X_{I_2}]}(Y)\in \mathbb{R}$, and for $A_{I_i}\in\mathscr{F}_{I_i}$, $i=3,4$,
$$E_{[X_{I_2},X_{I_4}\in A_{I_4}]}(Y_n|X_{I_1},X_{I_3}\in A_{I_3})\uparrow E_{[X_{I_2},X_{I_4}\in A_{I_4}]}(Y|X_{I_1},X_{I_3}\in A_{I_3}),$$
$a.e.\{P[X_{I_1},X_{I_3}\in A_{I_3}]\}$ provided $P[X_{I_3}\in A_{I_3}]>0$ while it is evident provided $P[X_{I_3}\in A_{I_3}]=0$.\\

2. If $Y_n\geq Y_{n+1},\ a.e.\{P[X_{I_2}]\},\ n=1,2,\cdots$,
$$\lim\limits_{n\to \infty}Y_n=Y,\ a.e.\{P[X_{I_2}]\},\ \  and\ \ \inf\limits_n\{E_{[X_{I_2}]}(Y_n)\}\in \mathbb{R},$$
then $E_{[X_{I_2}]}(Y)\in \mathbb{R}$, and for $A_{I_i}\in\mathscr{F}_{I_i}$, $i=3,4$,
$$E_{[X_{I_2},X_{I_4}\in A_{I_4}]}(Y_n|X_{I_1},X_{I_3}\in A_{I_3})\downarrow E_{[X_{I_2},X_{I_4}\in A_{I_4}]}(Y|X_{I_1},X_{I_3}\in A_{I_3}),$$
$a.e.\{P[X_{I_1},X_{I_3}\in A_{I_3}]\}$ provided $P[X_{I_3}\in A_{I_3}]>0$ while it is evident provided $P[X_{I_3}\in A_{I_3}]=0$.
\end{col}

\begin{comm}\label{comment6.8.1}Theorem \ref{theorem6.7}, Theorem \ref{theorem6.8} and their corollarys are called Monotone Convergence Theorem of $E$-integral.
\end{comm}

\begin{thm}\label{theorem6.9}Let $(\Omega,\mathscr{F},P)$ be a given probability space, and let $(\Omega_i,\mathscr{F}_i),\ i=1,2,3,4$ be measurable spaces. Let $X_i:(\Omega,\mathscr{F})\rightarrow(\Omega_i,\mathscr{F}_i),\ i=1,2,3,4$ be random objects, and $P[X_i]$ the probability on $(\Omega_i,\mathscr{F}_i)$ induced by $X_i$, $i=1,2,3,4$. Then $(\Omega_i,\mathscr{F}_i,P[X_i]),\ i=1,2,3,4$ are probability spaces. Let $Y_n(\omega_2),n=1,2,3,\cdots,$ be random variables on $(\Omega_2,\mathscr{F}_2,P[X_2])$ with $E_{[X_2]}(Y_n)\in \mathbb{R},\ n=1,2,3,\cdots$.\\

1. If there exists a random variable $g(\omega_2)$ on $(\Omega_2,\mathscr{F}_2,P[X_2])$ with $E_{[X_2]}(g)\in \mathbb{R}$ such that
$$Y_n\geq g,\ a.e.\{P[X_2]\},\ n=1,2,3,\cdots,\ \ and \ \ \sup\limits_n\{E_{[X_2]}(Y_n)\}\in \mathbb{R},$$
then for each fixed $A_i\in\mathscr{F}_i,\ i=3,4$, $E_{[X_2,X_4\in A_4]}(\liminf_{n\to \infty} Y_n|X_3\in A_3)\in \mathbb{R}$,
$$E_{[X_2,X_4\in A_4]}(\liminf\limits_{n\to \infty} Y_n|X_3\in A_3)\leq \liminf\limits_{n\to \infty} E_{[X_2,X_4\in A_4]}(Y_n|X_3\in A_3),$$
and
$$E_{[X_2,X_4\in A_4]}(\liminf\limits_{n\to \infty} Y_n|X_1,X_3\in A_3)\leq \liminf\limits_{n\to \infty} E_{[X_2,X_4\in A_4]}(Y_n|X_1,X_3\in A_3),$$
$a.e.\{P[X_1,X_3\in A_3]\}$ provided $P[X_1,X_3\in A_3]>0$.\\

2. If there exists a random variable $g(\omega_2)$ on $(\Omega_2,\mathscr{F}_2,P[X_2])$ with $E_{[X_2]}(g)\in \mathbb{R}$ such that
$$Y_n\leq g,\ a.e.\{P[X_2]\},\ n=1,2,3,\cdots,\ \ and \ \ \inf\limits_n\{E_{[X_2]}(Y_n)\}\in \mathbb{R},$$
then for each fixed $A_i\in\mathscr{F}_i,\ i=3,4$, $E_{[X_2,X_4\in A_4]}(\limsup_{n\to \infty} Y_n|X_3\in A_3)\in \mathbb{R}$,
$$E_{[X_2,X_4\in A_4]}(\limsup\limits_{n\to \infty} Y_n|X_3\in A_3)\geq \limsup\limits_{n\to \infty} E_{[X_2,X_4\in A_4]}(Y_n|X_3\in A_3),$$
and
$$E_{[X_2,X_4\in A_4]}(\limsup\limits_{n\to \infty} Y_n|X_1,X_3\in A_3)\geq \limsup\limits_{n\to \infty} E_{[X_2,X_4\in A_4]}(Y_n|X_1,X_3\in A_3),$$
$a.e.\{P[X_1,X_3\in A_3]\}$ provided $P[X_1,X_3\in A_3]>0$.
\end{thm}

\begin{col}\label{corollary6.9.1}Let $(\Omega,\mathscr{F},P)$ be a given probability space, $(\Omega_{i},\mathscr{F}_{i}),\ i\in \Lambda$ measurable spaces, where $\Lambda$ is a nonempty index set. Let $X_i:(\Omega,\mathscr{F})\rightarrow(\Omega_i,\mathscr{F}_i),\ i\in \Lambda$ be random objects. Let $I_i\subset\Lambda,\ i=1,2,3,4$, which are pairwise disjoint. Let $Y_n(\omega_{I_2}),n=1,2,3,\cdots$ be random variables on $(\Omega_{I_2},\mathscr{F}_{I_2},P[X_{I_2}])$ with $E_{[X_{I_2}]}(Y_n)\in \mathbb{R},\ n=1,2,3,\cdots$.\\

1. If there exists a random variable $g(\omega_{I_2})$ on $(\Omega_{I_2},\mathscr{F}_{I_2},P[X_{I_2}])$ with $E_{[X_{I_2}]}(g)\in \mathbb{R}$ such that
$$Y_n\geq g,\ a.e.\{P[X_{I_2}]\},\ n=1,2,3,\cdots,\ \ and \ \ \sup\limits_n\{E_{[X_{I_2}]}(Y_n)\}\in \mathbb{R},$$
then for each fixed $A_{I_i}\in\mathscr{F}_{I_i},\ i=3,4$, $E_{[X_{I_2},X_{I_4}\in A_{I_4}]}(\liminf_{n\to \infty} Y_n|X_{I_3}\in A_{I_3})\in \mathbb{R}$,
$$E_{[X_{I_2},X_{I_4}\in A_{I_4}]}(\liminf\limits_{n\to \infty} Y_n|X_{I_3}\in A_{I_3})\leq \liminf\limits_{n\to \infty} E_{[X_{I_2},X_{I_4}\in A_{I_4}]}(Y_n|X_{I_3}\in A_{I_3}),$$
and
$$E_{[X_{I_2},X_{I_4}\in A_{I_4}]}(\liminf\limits_{n\to \infty} Y_n|X_{I_1},X_{I_3}\in A_{I_3})\leq \liminf\limits_{n\to \infty} E_{[X_{I_2},X_{I_4}\in A_{I_4}]}(Y_n|X_{I_1},X_{I_3}\in A_{I_3}),$$
$a.e.\{P[X_{I_1},X_{I_3}\in A_{I_3}]\}$ provided $P[X_{I_1},X_{I_3}\in A_{I_3}]>0$.\\

2. If there exists a random variable $g(\omega_{I_2})$ on $(\Omega_{I_2},\mathscr{F}_{I_2},P[X_{I_2}])$ with $E_{[X_{I_2}]}(g)\in \mathbb{R}$ such that
$$Y_n\leq g,\ a.e.\{P[X_{I_2}]\},\ n=1,2,3,\cdots,\ \ and \ \ \inf\limits_n\{E_{[X_{I_2}]}(Y_n)\}\in \mathbb{R},$$
then for each fixed $A_{I_i}\in\mathscr{F}_{I_i},\ i=3,4$, $E_{[X_{I_2},X_{I_4}\in A_{I_4}]}(\limsup_{n\to \infty} Y_n|X_{I_3}\in A_{I_3})\in \mathbb{R}$,
$$E_{[X_{I_2},X_{I_4}\in A_{I_4}]}(\limsup\limits_{n\to \infty} Y_n|X_{I_3}\in A_{I_3})\geq \limsup\limits_{n\to \infty} E_{[X_{I_2},X_{I_4}\in A_{I_4}]}(Y_n|X_{I_3}\in A_{I_3}),$$
and
$$E_{[X_{I_2},X_{I_4}\in A_{I_4}]}(\limsup\limits_{n\to \infty} Y_n|X_{I_1},X_{I_3}\in A_{I_3})\geq \limsup\limits_{n\to \infty} E_{[X_{I_2},X_{I_4}\in A_{I_4}]}(Y_n|X_{I_1},X_{I_3}\in A_{I_3}),$$
$a.e.\{P[X_{I_1},X_{I_3}\in A_{I_3}]\}$ provided $P[X_{I_1},X_{I_3}\in A_{I_3}]>0$.
\end{col}

\begin{comm}\label{comment6.9.1}Theorem \ref{theorem6.9} and its corollary are called Fatou's Theorem of $E$-integral or Fatou's Lemma of $E$-integral.
\end{comm}

\begin{thm}\label{theorem6.10}Let $(\Omega,\mathscr{F},P)$ be a given probability space, and let $(\Omega_i,\mathscr{F}_i),\ i=1,2,3,4$ be measurable spaces. Let $X_i:(\Omega,\mathscr{F})\rightarrow(\Omega_i,\mathscr{F}_i),\ i=1,2,3,4$ be random objects, and $P[X_i]$ the probability on $(\Omega_i,\mathscr{F}_i)$ induced by $X_i$, $i=1,2,3,4$. Then $(\Omega_i,\mathscr{F}_i,P[X_i]),\ i=1,2,3,4$ are probability spaces. Let $Y$, $Y_n(\omega_2),n=1,2,3,\cdots,$ be random variables on $(\Omega_2,\mathscr{F}_2,P[X_2])$.\\

1. If
$$\lim_{n\to \infty}Y_n=Y,\ \ a.e.\{P[X_2]\},$$
and there exists a nonnegative random variable $g(\omega_2)$ on $(\Omega_2,\mathscr{F}_2,P[X_2])$ with $E_{[X_2]}(g)\in \mathbb{R}$ such that
$$|Y_n|\leq g,\ \ a.e.\{P[X_2]\},\ \ n=1,2,3,\cdots,$$
then for each fixed $A_i\in\mathscr{F}_i,\ i=3,4$, $E_{[X_2,X_4\in A_4]}(Y|X_3\in A_3)\in \mathbb{R}$,
$$\lim_{n\to \infty}E_{[X_2,X_4\in A_4]}(Y_n|X_3\in A_3)=E_{[X_2,X_4\in A_4]}(Y|X_3\in A_3),$$
and
$$\lim_{n\to \infty}E_{[X_2,X_4\in A_4]}(Y_n|X_1,X_3\in A_3)=E_{[X_2,X_4\in A_4]}(Y|X_1,X_3\in A_3),\ \ a.e.\{P[X_1,X_3\in A_3]\}$$
provided $P[X_1,X_3\in A_3]>0$.\\

2. If
$$\lim_{n\to \infty}Y_n=Y,\ \ in\ \  P[X_2],$$
and there exists a nonnegative random variable $g(\omega_2)$ on $(\Omega_2,\mathscr{F}_2,P[X_2])$ with $E_{[X_2]}(g)\in \mathbb{R}$ such that
$$|Y_n|\leq g,\ \ a.e.\{P[X_2]\},\ \ n=1,2,3,\cdots,$$
then for each fixed $A_i\in\mathscr{F}_i,\ i=3,4$, $E_{[X_2,X_4\in A_4]}(Y|X_3\in A_3)\in \mathbb{R}$,
$$\lim_{n\to \infty}E_{[X_2,X_4\in A_4]}(Y_n|X_3\in A_3)=E_{[X_2,X_4\in A_4]}(Y|X_3\in A_3),$$
and
$$\lim_{n\to \infty}E_{[X_2,X_4\in A_4]}(Y_n|X_1,X_3\in A_3)=E_{[X_2,X_4\in A_4]}(Y|X_1,X_3\in A_3),\ \ in\ P[X_1,X_3\in A_3]$$
provided $P[X_1,X_3\in A_3]>0$.
\end{thm}

\begin{col}\label{corollary6.10.1}Let $(\Omega,\mathscr{F},P)$ be a given probability space, $(\Omega_{i},\mathscr{F}_{i}),\ i\in \Lambda$ measurable spaces, where $\Lambda$ is a nonempty index set. Let $X_i:(\Omega,\mathscr{F})\rightarrow(\Omega_i,\mathscr{F}_i),\ i\in \Lambda$ be random objects. Let $I_i\subset\Lambda,\ i=1,2,3,4$, which are pairwise disjoint. Let $Y(\omega_{I_2})$, $Y_n(\omega_{I_2}),n=1,2,3,\cdots$ be random variables on $(\Omega_{I_2},\mathscr{F}_{I_2},P[X_{I_2}])$.\\

1. If
$$\lim_{n\to \infty}Y_n=Y,\ \ a.e.\ P[X_{I_2}],$$
and there exists a nonnegative random variable $g(\omega_{I_2})$ on $(\Omega_{I_2},\mathscr{F}_{I_2},P[X_{I_2}])$ with $E_{[X_{I_2}]}(g)\in \mathbb{R}$ such that
$$|Y_n|\leq g,\ \ a.e.\{P[X_{I_2}]\},\ \ n=1,2,3,\cdots,$$
then for each fixed $A_{I_i}\in\mathscr{F}_{I_i},\ i=3,4$, $E_{[X_{I_2},X_{I_4}\in A_{I_4}]}(Y|X_{I_3}\in A_{I_3})\in \mathbb{R}$,
$$\lim_{n\to \infty}E_{[X_{I_2},X_{I_4}\in A_{I_4}]}(Y_n|X_{I_3}\in A_{I_3})=E_{[X_{I_2},X_{I_4}\in A_{I_4}]}(Y|X_{I_3}\in A_{I_3}),$$
and
$$\lim_{n\to \infty}E_{[X_{I_2},X_{I_4}\in A_{I_4}]}(Y_n|X_{I_1},X_{I_3}\in A_{I_3})=E_{[X_{I_2},X_{I_4}\in A_{I_4}]}(Y|X_{I_1},X_{I_3}\in A_{I_3}),$$
$a.e.\ P[X_{I_1},X_{I_3}\in A_{I_3}]$ provided $P[X_{I_1},X_{I_3}\in A_{I_3}]>0$.\\

2. If
$$\lim_{n\to \infty}Y_n=Y,\ \ in\ \  P[X_{I_2}],$$
and there exists a nonnegative random variable $g(\omega_{I_2})$ on $(\Omega_{I_2},\mathscr{F}_{I_2},P[X_{I_2}])$ with $E_{[X_{I_2}]}(g)\in \mathbb{R}$ such that
$$|Y_n|\leq g,\ \ a.e.\{P[X_{I_2}]\},\ \ n=1,2,3,\cdots,$$
then for each fixed $A_{I_i}\in\mathscr{F}_{I_i},\ i=3,4$, $E_{[X_{I_2},X_{I_4}\in A_{I_4}]}(Y|X_{I_3}\in A_{I_3})\in \mathbb{R}$,
$$\lim_{n\to \infty}E_{[X_{I_2},X_{I_4}\in A_{I_4}]}(Y_n|X_{I_3}\in A_{I_3})=E_{[X_{I_2},X_{I_4}\in A_{I_4}]}(Y|X_{I_3}\in A_{I_3}),$$
and
$$\lim_{n\to \infty}E_{[X_{I_2},X_{I_4}\in A_{I_4}]}(Y_n|X_{I_1},X_{I_3}\in A_{I_3})=E_{[X_{I_2},X_{I_4}\in A_{I_4}]}(Y|X_{I_1},X_{I_3}\in A_{I_3}),$$
$in\ P[X_{I_1},X_{I_3}\in A_{I_3}]$ provided $P[X_{I_1},X_{I_3}\in A_{I_3}]>0$.
\end{col}

\begin{comm}\label{comment6.10.1}Theorem \ref{theorem6.10} and its corollary are called Dominated Convergence Theorem of $E$-integral.
\end{comm}

\begin{thm}\label{theorem6.11}Let $(\Omega,\mathscr{F},P)$ be a given probability space, and let $(\Omega_i,\mathscr{F}_i)$, $i=1,2,3$ be measurable spaces. Let $\mathscr{G}$ be a sub $\sigma$-field of $\mathscr{F}_2$. Let $X_i:(\Omega,\mathscr{F})\rightarrow(\Omega_i,\mathscr{F}_i)$, $i=1,2,3$ be random objects. The random object $X_{2\mathscr{G}}(\omega):(\Omega,\mathscr{F})\rightarrow(\Omega_2,\mathscr{G})$ is denoted by $X_{21}$. Let $Y(\omega_2)$ be a random variable on $(\Omega_2,\mathscr{F}_2,P[X_2])$ with $E_{[X_2]}(Y)\in \mathbb{R}$, and $Z(\omega_2)$ a random variable on $(\Omega_2,\mathscr{G},P[X_{21}])$ with $E_{[X_2]}(ZY)\in \mathbb{R}$. Then for fixed $A_i\in\mathscr{F}_i,\ i=1,3$ with $P[X_1\in A_1,X_3\in A_3]>0$,
$$E_{[X_2,X_3\in A_3]}(ZY|X_{21},X_1\in A_1)=ZE_{[X_2,X_3\in A_3]}(Y|X_{21},X_1\in A_1),\ \ a.e.\{P[X_{21},X_1\in A_1]\}.$$
\end{thm}

\begin{comm}\label{comment6.11.1}(i) In Theorem \ref{theorem6.11}, let $A_3=\Omega_3$, then
$$E_{[X_2]}(ZY|X_{21},X_1\in A_1)=ZE_{[X_2]}(Y|X_{21},X_1\in A_1),\ \ a.e.\{P[X_{21},X_1\in A_1]\}.$$

(ii) In Theorem \ref{theorem6.11}, let $A_1=\Omega_1$, then
$$E_{[X_2,X_3\in A_3]}(ZY|X_{21})=ZE_{[X_2,X_3\in A_3]}(Y|X_{21}),\ \ a.e.\{P[X_{21}]\}.$$

(iii) In Theorem \ref{theorem6.11}, let $A_1=\Omega_1$, $A_3=\Omega_3$, then
$$E_{[X_2]}(ZY|X_{21})=ZE_{[X_2]}(Y|X_{21}),\ \ a.e.\{P[X_{21}]\}.$$

(iv) In Theorem \ref{theorem6.11}, let $(\Omega_2,\mathscr{F}_2)=(\Omega,\mathscr{F})$, $X_2=I_{\mathscr{F}}=I_0$, $X_{21}=I_{\mathscr{G}}$, $A_1=\Omega_1$, $A_3=\Omega_3$, then
$$E(ZY|\mathscr{G})=E_{[I_0]}(ZY|I_{\mathscr{G}})=ZE_{[I_0]}(Y|I_{\mathscr{G}})=ZE(Y|\mathscr{G}),\ \ a.e.\{P\},$$
which is Theorem 7.2.1 (9) in \cite{yan:2004} or Theorem 5.5.11 (a) in \cite{Robert:2000}.
\end{comm}

Let $(\Omega,\mathscr{F},P)$ be a probability space, $\mathscr{L}^p(\Omega,\mathscr{F},P)$, $(1\leq p<+\infty)$, the set of all random variables $f:\Omega\rightarrow\mathbb{R}$ on $(\Omega,\mathscr{F})$ such that $E(|f|^p)\in\mathbb{R}$, and $\mathscr{L}^\infty(\Omega,\mathscr{F},P)$, $( p=+\infty)$, the set of all random variables $f:\Omega\rightarrow\mathbb{R}$ on $(\Omega,\mathscr{F})$ that are essentially bounded. Then we have H$\ddot{\rm{o}}$lder's inequality of $E$-integral below.

\begin{thm}\label{theorem6.12}Let $(\Omega,\mathscr{F},P)$ be a given probability space, and let $(\Omega_i,\mathscr{F}_i),\ i=1,2,3,4$ be measurable spaces. Let $X_i:(\Omega,\mathscr{F})\rightarrow(\Omega_i,\mathscr{F}_i)$, $i=1,2,3,4$ be random objects, and $P[X_i]$ the probability on $(\Omega_i,\mathscr{F}_i)$ induced by $X_i$, $i=1,2,3,4$. Then $(\Omega_i,\mathscr{F}_i,P[X_i]),\ i=1,2,3,4$ are probability spaces. Let $p$ and $q$ satisfy $1\leq p\leq+\infty$, $1\leq q\leq+\infty$, and $\displaystyle\frac{1}{p}+\frac{1}{q}=1$. Let $Y(\omega_2)\in\mathscr{L}^p(\Omega_2,\mathscr{F}_2,P[X_2])$ and $Z(\omega_2)\in\mathscr{L}^q(\Omega_2,\mathscr{F}_2,P[X_2])$.\\

1. For each fixed $A_i\in\mathscr{F}_i,\ i=3,4$ with $P[X_3\in A_3,X_4\in A_4]>0$,
$$\begin{array}{l}E_{[X_2,X_4\in A_4]}(YZ|X_3\in A_3)\in \mathbb{R};\\\\
E_{[X_2,X_4\in A_4]}(|YZ||X_3\in A_3)\\
\leq (E_{[X_2,X_4\in A_4]}(|Y|^p|X_3\in A_3))^{\frac{1}{p}}\cdot(E_{[X_2,X_4\in A_4]}(|Z|^q|X_3\in A_3))^{\frac{1}{q}},(1<p,q<+\infty);\\\\
E_{[X_2,X_4\in A_4]}(|YZ||X_3\in A_3)\leq \|Y\|_{\infty}\cdot E_{[X_2,X_4\in A_4]}(|Z||X_3\in A_3),\ \ (p=+\infty,q=1).
\end{array}$$

2. For each fixed $A_i\in\mathscr{F}_i,\ i=3,4$ with $P[X_3\in A_3,X_4\in A_4]>0$, there exists $E_{[X_2,X_4\in A_4]}(YZ|X_1,X_3\in A_3)$, and
$$\begin{array}{l}E_{[X_2,X_4\in A_4]}(|YZ||X_1,X_3\in A_3)\leq \|Y\|_\infty\cdot E_{[X_2,X_4\in A_4]}(|Z||X_1,X_3\in A_3),\\
a.e.\{P[X_1,X_3\in A_3]\},\ \ (p=+\infty,q=1).
\end{array}$$
Furthermore if there exists $P[X_2,X_4\in A_4|X_1,X_3\in A_3]$, then
$$\begin{array}{l}E_{[X_2,X_4\in A_4]}(|YZ||X_1,X_3\in A_3)\\
\leq (E_{[X_2,X_4\in A_4]}(|Y|^p|X_1,X_3\in A_3))^{\frac{1}{p}}\cdot(E_{[X_2,X_4\in A_4]}(|Z|^q|X_1,X_3\in A_3))^{\frac{1}{q}}, \\
a.e.\{P[X_1,X_3\in A_3]\},\ \ (1<p,\ q<+\infty).
\end{array}$$
\end{thm}

\begin{col}\label{corollary6.12.1}Let $(\Omega,\mathscr{F},P)$ be a given probability space, $(\Omega_{i},\mathscr{F}_{i}),\ i\in \Lambda$ measurable spaces, where $\Lambda$ is a nonempty index set. Let $X_i:(\Omega,\mathscr{F})\rightarrow(\Omega_i,\mathscr{F}_i),\ i\in \Lambda$ be random objects. Let $I_i\subset\Lambda,\ i=1,2,3,4$, which are pairwise disjoint. Let $p$ and $q$ satisfy $1\leq p\leq+\infty$, $1\leq q\leq+\infty$, and $\displaystyle\frac{1}{p}+\frac{1}{q}=1$. Let $Y(\omega_{I_2})\in\mathscr{L}^p(\Omega_{I_2},\mathscr{F}_{I_2},P[X_{I_2}])$ and $Z(\omega_{I_2})\in\mathscr{L}^q(\Omega_{I_2},\mathscr{F}_{I_2},P[X_{I_2}])$.\\

1. For each fixed $A_{I_i}\in\mathscr{F}_{I_i},\ i=3,4$ with $P[X_{I_3}\in A_{I_3},X_{I_4}\in A_{I_4}]>0$,
$$\begin{array}{l}E_{[X_{I_2},X_{I_4}\in A_{I_4}]}(YZ|X_{I_3}\in A_{I_3})\in \mathbb{R};\\\\
E_{[X_{I_2},X_{I_4}\in A_{I_4}]}(|YZ||X_{I_3}\in A_{I_3})\\
\leq (E_{[X_{I_2},X_{I_4}\in A_{I_4}]}(|Y|^p|X_{I_3}\in A_{I_3}))^{\frac{1}{p}}\cdot(E_{[X_{I_2},X_{I_4}\in A_{I_4}]}(|Z|^q|X_{I_3}\in A_{I_3}))^{\frac{1}{q}},(1<p,q<+\infty);\\\\
E_{[X_{I_2},X_{I_4}\in A_{I_4}]}(|YZ||X_{I_3}\in A_{I_3})\leq \|Y\|_{\infty}\cdot E_{[X_{I_2},X_{I_4}\in A_{I_4}]}(|Z||X_{I_3}\in A_{I_3}),\ \ (p=+\infty,q=1).
\end{array}$$

2. For each fixed $A_{I_i}\in\mathscr{F}_{I_i},\ i=3,4$ with $P[X_{I_3}\in A_{I_3},X_{I_4}\in A_{I_4}]>0$, there exists $E_{[X_{I_2},X_{I_4}\in A_{I_4}]}(YZ|X_{I_1},X_{I_3}\in A_{I_3})$, and
$$\begin{array}{l}E_{[X_{I_2},X_{I_4}\in A_{I_4}]}(|YZ||X_{I_1},X_{I_3}\in A_{I_3})\leq \|Y\|_\infty\cdot E_{[X_{I_2},X_{I_4}\in A_{I_4}]}(|Z||X_{I_1},X_{I_3}\in A_{I_3}),\\
a.e.\{P[X_{I_1},X_{I_3}\in A_{I_3}]\},\ \ (p=+\infty,\ q=1).\end{array}$$
Furthermore if there exists $P[X_{I_2},X_{I_4}\in A_{I_4}|X_{I_1},X_{I_3}\in A_{I_3}]$, then
$$\begin{array}{l}E_{[X_{I_2},X_{I_4}\in A_{I_4}]}(|YZ||X_{I_1},X_{I_3}\in A_{I_3})\\
\leq (E_{[X_{I_2},X_{I_4}\in A_{I_4}]}(|Y|^p|X_{I_1},X_{I_3}\in A_{I_3}))^{\frac{1}{p}}\cdot(E_{[X_{I_2},X_{I_4}\in A_{I_4}]}(|Z|^q|X_{I_1},X_{I_3}\in A_{I_3}))^{\frac{1}{q}},\\
a.e.\{P[X_{I_1},X_{I_3}\in A_{I_3}]\},\ \ (1<p,\ q<+\infty).
\end{array}$$
\end{col}

By Corollary \ref{corollary6.12.1} and Theorem \ref{theorem2.11.3} we have the following corollary.

\begin{col}\label{corollary6.12.2}Let $(\Omega,\mathscr{F},P)$ be a given probability space, and $X_i(\omega),\ i\in \mathbb{N}$ random variables on $(\Omega,\mathscr{F},P)$. Let $I_i\subset\mathbb{N},\ i=1,2,3,4$, which are pairwise disjoint. Let $p$ and $q$ satisfy $1\leq p\leq+\infty$, $1\leq q\leq+\infty$, and $\displaystyle\frac{1}{p}+\frac{1}{q}=1$. Let $Y(x_{I_2})\in\mathscr{L}^p(\mathbb{R}^{I_2},\mathscr{B}(\mathbb{R}^{I_2}),P[X_{I_2}])$ and $Z(x_{I_2})\in\mathscr{L}^q(\mathbb{R}^{I_2},\mathscr{B}(\mathbb{R}^{I_2}),P[X_{I_2}])$.\\

1. For each fixed $A_i\in\mathscr{B}(\mathbb{R}),\ i\in I_3+I_4$ with $P[X_i\in A_i,\ i\in I_3+I_4]>0$,
$$\begin{array}{l}E_{[X_i,i\in I_2;X_i\in A_i,i\in I_4]}(YZ|X_i\in A_i,i\in I_3)\in \mathbb{R};\\\\
E_{[X_i,i\in I_2;X_i\in A_i,i\in I_4}(|YZ||X_i\in A_i,i\in I_3)\\
\leq (E_{[X_i,i\in I_2;X_i\in A_i,i\in I_4]}(|Y|^p|X_i\in A_i,i\in I_3))^{\frac{1}{p}}\cdot(E_{[X_i,i\in I_2;X_i\in A_i,i\in I_4]}(|Z|^q|X_i\in A_i,i\in I_3))^{\frac{1}{q}},\\
\ \ \ (1<p,q<+\infty);\\\\
E_{[X_i,i\in I_2;X_i\in A_i,i\in I_4]}(|YZ||X_i\in A_i,i\in I_3)\leq \|Y\|_{\infty}\cdot E_{[X_i,i\in I_2;X_i\in A_i,i\in I_4]}(|Z||X_i\in A_i,i\in I_3),\\
\ \ \ (p=+\infty,q=1).
\end{array}$$

2. For each fixed $A_i\in\mathscr{B}(\mathbb{R}),\ i\in I_3+I_4$ with $P[X_i\in A_i,\ i\in I_3+I_4]>0$, there exists $E_{[X_i,i\in I_2;X_i\in A_i,i\in I_4]}(YZ|X_i,i\in I_1;X_i\in A_i,i\in I_3)$, and
$$\begin{array}{l}E_{[X_i,i\in I_2;X_i\in A_i,i\in I_4]}(|YZ||X_i,i\in I_1;X_i\in A_i,i\in I_3)\\
\leq (E_{[X_i,i\in I_2;X_i\in A_i,i\in I_4]}(|Y|^p|X_i,i\in I_1;X_i\in A_i,i\in I_3))^{\frac{1}{p}}\\
(E_{[X_i,i\in I_2;X_i\in A_i,i\in I_4]}(|Z|^q|X_i,i\in I_1;X_i\in A_i,i\in I_3))^{\frac{1}{q}},\ \  a.e.\{P[X_i,i\in I_1;X_i\in A_i,i\in I_3]\},\\
(1<p,\ q<+\infty);\\\\
E_{[X_i,i\in I_2;X_i\in A_i,i\in I_4]}(|YZ||X_i,i\in I_1;X_i\in A_i,i\in I_3)\\
\leq \|Y\|_\infty E_{[X_i,i\in I_2;X_i\in A_i,i\in I_4]}(|Z||X_i,i\in I_1;X_i\in A_i,i\in I_3),\ \  a.e.\{P[X_i,i\in I_1;X_i\in A_i,i\in I_3]\},\\
(p=+\infty,q=1).
\end{array}$$
\end{col}

The following Theorem is Minkowski's inequality of $E$-integral.

\begin{thm}\label{theorem6.13}Let $(\Omega,\mathscr{F},P)$ be a given probability space, and let $(\Omega_i,\mathscr{F}_i),\ i=1,2,3,4$ be measurable spaces. Let $X_i:(\Omega,\mathscr{F})\rightarrow(\Omega_i,\mathscr{F}_i)$, $i=1,2,3,4$ be random objects, and $P[X_i]$ the probability on $(\Omega_i,\mathscr{F}_i)$ induced by $X_i$, $i=1,2,3,4$. Then $(\Omega_i,\mathscr{F}_i,P[X_i]),\ i=1,2,3,4$ are probability spaces. Let $p$ satisfy $1\leq p<+\infty$, and  $Y(\omega_2),Z(\omega_2)\in\mathscr{L}^p(\Omega_2,\mathscr{F}_2,P[X_2])$.\\

1. For each fixed $A_i\in\mathscr{F}_i,\ i=3,4$ with $P[X_3\in A_3,X_4\in A_4]>0$,
$$\begin{array}{l}\{E_{[X_2,X_4\in A_4]}(|Y+Z|^p|X_3\in A_3)\}^\frac{1}{p}\\
\leq\{E_{[X_2,X_4\in A_4]}(|Y|^p|X_3\in A_3)\}^\frac{1}{p}+ \{E_{[X_2,X_4\in A_4]}(|Z|^p|X_3\in A_3)\}^\frac{1}{p}.
\end{array}$$

2. For each fixed $A_i\in\mathscr{F}_i,\ i=3,4$ with $P[X_3\in A_3,X_4\in A_4]>0$, if there exists $P[X_2,X_4\in A_4|X_1,X_3\in A_3]$, then
$$\begin{array}{l}\{E_{[X_2,X_4\in A_4]}(|Y+Z|^p|X_1,X_3\in A_3)\}^\frac{1}{p}\\
\leq\{E_{[X_2,X_4\in A_4]}(|Y|^p|X_1,X_3\in A_3)\}^\frac{1}{p}+ \{E_{[X_2,X_4\in A_4]}(|Z|^p|X_1,X_3\in A_3)\}^\frac{1}{p},\\
a.e.\{P[X_1,X_3\in A_3]\}.
\end{array}$$
\end{thm}

\begin{col}\label{corollary6.13.1}Let $(\Omega,\mathscr{F},P)$ be a given probability space, $(\Omega_{i},\mathscr{F}_{i}),\ i\in \Lambda$ measurable spaces, where $\Lambda$ is a nonempty index set. Let $X_i:(\Omega,\mathscr{F})\rightarrow(\Omega_i,\mathscr{F}_i),\ i\in \Lambda$ be random objects. Let $I_i\subset\Lambda,\ i=1,2,3,4$, which are pairwise disjoint. Let $p$ satisfy $1\leq p<+\infty$, and  $Y(\omega_{I_2}),Z(\omega_{I_2})\in\mathscr{L}^p(\Omega_{I_2},\mathscr{F}_{I_2},P[X_{I_2}])$.\\

1. For each fixed $A_{I_i}\in\mathscr{F}_{I_i},\ i=3,4$ with $P[X_{I_3}\in A_{I_3},X_{I_4}\in A_{I_4}]>0$,
$$\begin{array}{l}\{E_{[X_{I_2},X_{I_4}\in A_{I_4}]}(|Y+Z|^p|X_{I_3}\in A_{I_3})\}^\frac{1}{p}\\
\leq\{E_{[X_{I_2},X_{I_4}\in A_{I_4}]}(|Y|^p|X_{I_3}\in A_{I_3})\}^\frac{1}{p}+ \{E_{[X_{I_2},X_{I_4}\in A_{I_4}]}(|Z|^p|X_{I_3}\in A_{I_3})\}^\frac{1}{p}.
\end{array}$$

2. For each fixed $A_{I_i}\in\mathscr{F}_{I_i},\ i=3,4$ with $P[X_{I_3}\in A_{I_3},X_{I_4}\in A_{I_4}]>0$, if there exists $P[X_{I_2},X_{I_4}\in A_{I_4}|X_{I_1},X_{I_3}\in A_{I_3}]$, then
$$\begin{array}{l}\{E_{[X_{I_2},X_{I_4}\in A_{I_4}]}(|Y+Z|^p|X_{I_1},X_{I_3}\in A_{I_3})\}^\frac{1}{p}\\
\leq\{E_{[X_{I_2},X_{I_4}\in A_{I_4}]}(|Y|^p|X_{I_1},X_{I_3}\in A_{I_3})\}^\frac{1}{p}+ \{E_{[X_{I_2},X_{I_4}\in A_{I_4}]}(|Z|^p|X_{I_1},X_{I_3}\in A_{I_3})\}^\frac{1}{p},\\
a.e.\{P[X_{I_1},X_{I_3}\in A_{I_3}]\}.
\end{array}$$
\end{col}

By Corollary \ref{corollary6.13.1} and Theorem \ref{theorem2.11.3} we have the following corollary.

\begin{col}\label{corollary6.13.2} Let $(\Omega,\mathscr{F},P)$ be a given probability space, and $X_i(\omega),\ i\in \mathbb{N}$ random variables on $(\Omega,\mathscr{F},P)$. Let $I_i\subset\mathbb{N},\ i=1,2,3,4$, which are pairwise disjoint. Let $p$ satisfy $1\leq p<+\infty$, and  $Y(x_{I_2}),Z(x_{I_2})\in\mathscr{L}^p(\mathbb{R}^{I_2},\mathscr{B}(\mathbb{R}^{I_2}),P[X_{I_2}])$. Then for each fixed $A_i\in\mathscr{B}(\mathbb{R}),\ i\in I_3+I_4$ with $P[X_i\in A_i,\ i\in I_3+I_4]>0$,
$$\begin{array}{l}\{E_{[X_i,\ i\in I_2;X_i\in A_i,\ i\in I_4]}(|Y+Z|^p|X_i\in A_i,\ i\in I_3)\}^\frac{1}{p}\\
\leq\{E_{[X_i,\ i\in I_2;X_i\in A_i,\ i\in I_4]}(|Y|^p|X_i\in A_i,\ i\in I_3)\}^\frac{1}{p}\\
+ \{E_{[X_i,\ i\in I_2;X_i\in A_i,\ i\in I_4]}(|Z|^p|X_i\in A_i,\ i\in I_3)\}^\frac{1}{p},
\end{array}$$
and
$$\begin{array}{l}\{E_{[X_i,\ i\in I_2;X_i\in A_i,\ i\in I_4]}(|Y+Z|^p|X_i,\ i\in I_1;X_i\in A_i,\ i\in I_3)\}^\frac{1}{p}\\
\leq\{E_{[X_i,\ i\in I_2;X_i\in A_i,\ i\in I_4]}(|Y|^p|X_i,\ i\in I_1;X_i\in A_i,\ i\in I_3)\}^\frac{1}{p}\\
+ \{E_{[X_i,\ i\in I_2;X_i\in A_i,\ i\in I_4]}(|Z|^p|X_i,\ i\in I_1;X_i\in A_i,\ i\in I_3)\}^\frac{1}{p},\\
a.e.\{P[X_i,\ i\in I_1;X_i\in A_i,\ i\in I_3]\}.
\end{array}$$
\end{col}

The following Theorem is Jensen's inequality of $E$-integral.

\begin{thm}\label{theorem6.14}Let $(\Omega,\mathscr{F},P)$ be a given probability space, and let $(\Omega_i,\mathscr{F}_i),\ i=1,2,3$ be measurable spaces. Let $X_i:(\Omega,\mathscr{F})\rightarrow(\Omega_i,\mathscr{F}_i)$, $i=1,2,3$ be random objects, and $P[X_i]$ the probability on $(\Omega_i,\mathscr{F}_i)$ induced by $X_i$, $i=1,2,3$. Then $(\Omega_i,\mathscr{F}_i,P[X_i]),\ i=1,2,3$ are probability spaces. Suppose that $\varphi:\mathbb{R}\rightarrow\mathbb{R}$ is convex, in the sense that $\varphi(tx+(1-t)y)\leq t\varphi(x)+(1-t)\varphi(y)$ holds for $\forall x,y\in \mathbb{R}$ and $\forall t\in [0,1]$. Let $Y$ be a random variable on $(\Omega_2,\mathscr{F}_2,P[X_2])$ with $E_{[X_2]}(Y)\in\mathbb{R}$, and $A_3\in\mathscr{F}_3$ with $P[X_3\in A_3]>0$. If there exists $P[X_2|X_1,X_3\in A_3]$, then
$$\varphi(E_{[X_2]}(Y|X_1,X_3\in A_3))\leq E_{[X_2]}(\varphi(Y)|X_1,X_3\in A_3), \ \ a.e.\{P[X_1,X_3\in A_3]\}.$$
\end{thm}

\begin{col}\label{corollary6.14.1}Let $(\Omega,\mathscr{F},P)$ be a given probability space, $(\Omega_{i},\mathscr{F}_{i}),\ i\in \Lambda$ measurable spaces, where $\Lambda$ is a nonempty index set. Let $X_i:(\Omega,\mathscr{F})\rightarrow(\Omega_i,\mathscr{F}_i),\ i\in \Lambda$ be random objects. Let $I_i\subset\Lambda,\ i=1,2,3$, which are pairwise disjoint. Suppose that $\varphi:\mathbb{R}\rightarrow\mathbb{R}$ is convex, in the sense that $\varphi(tx+(1-t)y)\leq t\varphi(x)+(1-t)\varphi(y)$ holds for $\forall x,y\in \mathbb{R}$ and $\forall t\in [0,1]$. Let $Y$ be a random variable on $(\Omega_{I_2},\mathscr{F}_{I_2},P[X_{I_2}])$ with $E_{[X_{I_2}]}(Y)\in\mathbb{R}$, and $A_{I_3}\in\mathscr{F}_{I_3}$ with $P[X_{I_3}\in A_{I_3}]>0$. If there exists $P[X_{I_2}|X_{I_1},X_{I_3}\in A_{I_3}]$, then
$$\varphi(E_{[X_{I_2}]}(Y|X_{I_1},X_{I_3}\in A_{I_3}))\leq E_{[X_{I_2}]}(\varphi(Y)|X_{I_1},X_{I_3}\in A_{I_3}), \ \ a.e.\{P[X_{I_1},X_{I_3}\in A_{I_3}]\}.$$
\end{col}

By Corollary \ref{corollary6.14.1} and Theorem \ref{theorem2.11.3} we have the following corollary.

\begin{col}\label{corollary6.14.2}Let $(\Omega,\mathscr{F},P)$ be a given probability space, and $X_i(\omega),\ i\in \mathbb{N}$ random variables on $(\Omega,\mathscr{F},P)$. Let $I_i\subset\mathbb{N},\ i=1,2,3$, which are pairwise disjoint. Suppose that $\varphi:\mathbb{R}\rightarrow\mathbb{R}$ is convex, in the sense that $\varphi(tx+(1-t)y)\leq t\varphi(x)+(1-t)\varphi(y)$ holds for $\forall x,y\in \mathbb{R}$ and $\forall t\in [0,1]$. Let $Y$ be a random variable on $(\mathbb{R}^{I_2},\mathscr{B}(\mathbb{R}^{I_2}),P[X_{I_2}])$ with $E_{[X_{I_2}]}(Y)\in\mathbb{R}$, and $A_i\in\mathscr{B}(\mathbb{R}),i\in I_3$ with $P[X_i\in A_i,i\in I_3]>0$. Then
$$\begin{array}{l}\varphi(E_{[X_i,i\in I_2]}(Y|X_i,i\in I_1;X_i\in A_i,i\in I_3))\leq E_{[X_i,i\in I_2]}(\varphi(Y)|X_i,i\in I_1;X_i\in A_i,i\in I_3),\\\\
a.e.\{P[X_i,i\in I_1;X_i\in A_i,i\in I_3]\}.\end{array}$$
\end{col}

%\textbf{Declarations of interest: none}

\textbf{Funding: This research did not receive any specific grant from funding agencies in the public, commercial, or not-for-profit sectors.}

\textbf{Acknowledgements:}

\textbf{Conflicts of Interest: The authors declare no conflict of interest.}


\begin{thebibliography}{4}

\bibitem{Albert:2000}  \textsc{Albert, R.} and \textsc{Barab\'{a}si, A. L.} (2000). Topology of evolving networks: Local events and universality. \textit{Physical Review Letters}, \textbf{85}, 5234-5237.

\bibitem{Agborsangaya:2012} \textsc{Agborsangaya, C. B., Lau, D., Lahtinen, M., Cooke, T.} and \textsc{Johnson, J. A.}  (2012). Multimorbidity prevalence and patterns across socioeconomic determinants: a cross-sectional survey.  \textit{BMC Public Health}, 12:201.

\bibitem{Anselin:2000}  \textsc{Anselin, L.} (2000). The Alchemy of Statistics, or Creating Data Where No Data Exist. \textit{Annals of The Association of American Geographers}, \textbf{90}(3), 586-592.

\bibitem{Robert:2000}  \textsc{Ash, Robert B.}  (2000). \textit{Probability and Measure Theory}, 2nd ed. Academic Press.

\bibitem{Baldi:2001} \textsc{Baldi, P.} and \textsc{Brunak, S.} (2001). \textit{Bioinformatics: The Machine Learning Approach}, 2nd ed. MIT Press, Cambridge.

\bibitem{Barnett:2012} \textsc{Barnett, K., Mercer, S. W., Norbury, M., Watt, G., Wyke, S.} and \textsc{Guthrie, B.}  (2012). Epidemiology of multimorbidity and implications for health care, research, and medical education: a cross-sectional study. \textit{The Lancet}, \textbf{380}:37-43.

\bibitem{Begun:1983} \textsc{Begun, Janet M., Hall, W.J., Huang, Wei-Min} and \textsc{Wellner, Jon A.} (1983). Information and Asymptotic Efficiency In Parametric-Nonparametric Models, \textit{The Annals of Statistics}, \textbf{11}(2), 432-452.

\bibitem{Billingsley:1995}  \textsc{Billingsley, Patrick}  (1995). \textit{Probability and Measure}, 3rd ed. Wiley-Interscience.

\bibitem{Barnard:1995}  \textsc{Barnard, K., Duygulu, P., de Freitas, N., Forsyth, D., Blei, D., and Jordan, M. I.}  (1995). Matching Words and Pictures, \textit{Journal of Machine Learning Research}, \textbf{3}, 1107-1135.

\bibitem{Bogachev:2007} \textsc{Bogachev, V.I.}  (2007). \textit{Measure Theory}, Vol. \textbf{I} and \textbf{II}. Springer-Verlag Berlin Heidelberg.

\bibitem{Bollen:1989} \textsc{Bollen, K. A.} (1989).  \textit{Structural Equations with Latent Variables}.  Wiley Series in Probability and Mathematical Statistics: Applied Probability and Statistics.  Wiley, New York. MR0996025. https://doi.org/10.1002/9781118619179.

\bibitem{Bongers:2021} \textsc{Bongers, S., Forr\'{e}, P., Peters, J.} and \textsc{Mooij, J. M.} (2021) Foundations of structural causal models with cycles and latent variables. \textit{Ann. Statist.} \textbf{49}(5): 2885-2915.

\bibitem{Buhlmann:2014} \textsc{B\"{u}hlmann, P., Peters, J.} and \textsc{Ernest, J.} (2014).  CAM: Causal additive models, high-dimensional order search and penalized regression. \textit{Ann. Statist.} \textbf{42} 2526-2556. MR3277670. https://doi.org/10.1214/14-AOS1260.

\bibitem{Brezger:2007} \textsc{Brezger, A., Fahrmeir, L.} and \textsc{Hennerfeind, A.} (2007). Adaptive Gaussian Markov random fields with applications in human brain mapping. \textit{Journal of the Royal Statistical Society}, Series C (Applied Statistics), \textbf{56}, 327-345.

\bibitem{Kailai:2000} \textsc{Chung, K. L.} (2001). \textit{A Course in Probability Theory}, 3rd ed. Academic Press.


\bibitem{Donald:1980} \textsc{Cohn, D. L.} (1980). \textit{Measure Theory}. Birkh$\ddot{\rm a}$user Boston, USA.

\bibitem{Derin:1984} \textsc{Derin, H., Elliott, H., Cristi, R.} and \textsc{Geman, D.} (1984). Bayes Smoothing Algorithms for Segmentation of Binary Images Modeled by Markov Random Fields. \textit{IEEE Transactions on Pattern Analysis and Machine Intelligence}, \textbf{6}, 707-720.

\bibitem{Duncan:1975} \textsc{Duncan, O. D.} (1975).  \textit{Introduction to Structural Equation Models: Studies in Population}. Academic Press, New York. MR0398558.


\bibitem{Durrett:2019}  \textsc{Durrett, R.} (1995) \textit{Probability (Theory and Examples)}, 5th ed. Cambridge University Press.


\bibitem{Feng:2010} \textsc{Feng, W., Jia, J.} and \textsc{Liu, Z. Q.} (2010).  Self-Validated Labeling of Markov Random Fields for Image Segmentation. \textit{IEEE Transactions on Pattern Analysis and Machine Intelligence}, \textbf{32}(10), 1871-1887.

\bibitem{Fortin:2005} \textsc{Fortin, M., Bravo, G., Hudon, C., Vanasse, A.} and \textsc{Lapointe, L.} (2005). Prevalence of multimorbidity among adults seen in family practice. \textit{The Annals of Family Medicine}, May 2005, \textbf{3}(3), 223-228.

\bibitem{Gao:2000} \textsc{Gao, G. S.} (2000). \textit{Topological Space Theory}. Science Press, China.

\bibitem{Geman:1984} \textsc{Geman, S.} and \textsc{Geman, D.} (1984). Stochastic Relaxation, Gibbs Distributions, and the Bayesian Restoration of Images. \textit{IEEE Transactions on Pattern Analysis and Machine Intelligence}, \textbf{6}, 721-741.

\bibitem{Gine:2016} \textsc{Gin\'{e}, E.} and \textsc{Nickl, R.}  (2016). \textit{Mathematical Foundations of Infinite-Dimensional Statistical Models}. Cambridge University Press.

\bibitem{Glynn:2011} \textsc{Glynn, L. G., Valderas, J. M., Healy, P., Burke, E., Newell, J., Gillespie, P.}, et al.  (2011). The prevalence of multimorbidity in primary care and its effect on health care utilization and cost. \textit{Family Practice},  \textbf{28}(5),  516-523.


\bibitem{Duncan:1973} \textsc{Goldberger, A. S.} and  \textsc{Duncan, O. D.} (1973).  \textit{Structural Equation Models in the Social Sciences}. Seminar Press, New York.


\bibitem{Gravier:2000} \textsc{Gravier, G., Sigelle, M.} and \textsc{Chollet, G.} (2000). A markov random field model for automatic speech recognition. \textit{In: Proceedings of the International Conference on Pattern Recognition}, pp. 3258-3261.

\bibitem{Haavelmo:1943} \textsc{Haavelmo, T.} (1943). The statistical implications of a system of simultaneous equations. \textit{Econometrica}, \textbf{11}, 1-12. MR0007954 https://doi.org/10.2307/1905714.

\bibitem{Hedstrom:2010} \textsc{Hedstr\"{o}m, P. and Ylikoski} (2010). Causal Mechanisms in the Social Sciences. \textit{Annual Review of Sociology}, \textbf{36}, 49-67.

\bibitem{Held:1997} \textsc{Held, K., Kops, E. R., Krause, B. J., Wells, W. M. I., Kikinis, R.} and \textsc{M\"{u}ller-G\"{a}rtner, H.W.} (1997).  Markov random field segmentation of brain MR images. \textit{IEEE Transactions on Medical Imaging}, \textbf{16}(6), 878-886.

\bibitem{Hofmann:1998} \textsc{Hofmann, T.} and \textsc{Puzicha, J.} (1998).  Statistical models for co-occurrence data. \textit{A.I. Memo 1625}. Massachusetts Institute of Technology.

\bibitem{King:1997} \textsc{King, G.} (1997).   \textit{A Solution to The Ecological Inference Problem: Reconstructing Individual Behavior From Aggregate Data}. Princeton University Press.

\bibitem{Llorente:2010} \textsc{Llorente, A., Manmatha, R.} and \textsc{R\"{u}ger, S. M.}(2010). Image retrieval using Markov Random Fields and global image features. \textit{In: Proceedings of the ACM International Conference on Image and Video Retrieval}, pp. 243-250.

\bibitem{Maathuis:2009} \textsc{Maathuis, M. H., Kalisch, M.} and \textsc{B\"{u}hlmann, P.} (2009).  Estimating high-dimensional intervention effects from observational data. \textit{Ann. Statist.} \textbf{37}, 3133-3164. MR2549555, https://doi.org/10.1214/09-AOS685.

\bibitem{Maglogiannis:2009} \textsc{Maglogiannis, I., Vouyioukas, D.} and \textsc{Aggelopoulos, C.} (2009). Face Detection and Recognition of Natural Human Emotion Using Markov Random Fields, \textit{Personal and Ubiquitous Computing}, \textbf{13}(1), 95-101.


\bibitem{Mardia:1988} \textsc{Mardia, K. V.} (1988). Multi-dimensional multivariate Gaussian Markov random fields with application to image processing. \textit{Journal of Multivariate Analysis}, \textbf{24}, 265-284.

\bibitem{Marengoni:2008} \textsc{Marengoni, A., Winblad, B., Karp, A.} and \textsc{Fratiglioni, L.}  (2008). Prevalence of chronic diseases and multimorbidity among the elderly population in Sweden. \textit{ American Journal of Public Health}, \textbf{98}, no. 7 (July 1, 2008): 1198-1200.

\bibitem{Marengoni:2011} \textsc{Marengoni, A., Angleman, S., Melis, R., Mangialasche, F., Karp, A., Garmen, A.}, et al.  (2011). Aging with multimorbidity: a systematic review of the literature. \textit{Ageing Research Reviews}, \textbf{10}(4), September 2011, 430-439.

\bibitem{Mason:1953} \textsc{Mason, S.J.} (1953). Feedback theory-Some properties of signal flow graphs.  \textit{In Proceedings of the IRE} \textbf{41}, 1144-1156. IEEE.

\bibitem{Mason:1956} \textsc{Mason, S.J.} (1956). Feedback theory-Further properties of signal flow graphs.  \textit{In Proceedings of the IRE} \textbf{44}, 920-926. IEEE.

\bibitem{McPhail:2016} \textsc{McPhail, S. M.}  (2016). Multimorbidity in chronic disease: impact on health care resources and costs. \textit{Risk Management and Healthcare Policy}, Volume 2016:9, 143-156.

\bibitem{Mignotte:2000} \textsc{Mignotte, M., Collet, C., P\'{e}rez, P.} and \textsc{Bouthemy, P.} (2000). Markov Random Field and Fuzzy Logic Modeling in Sonar Imagery: Application to the Classification of Underwater Floor. \textit{Computer Vision and Image Understanding}, \textbf{79}(1), 4-24.

\bibitem{Mooij:2013}   \textsc{Mooij, J. M.} and \textsc{Heskes, T.}(2013). Cyclic causal discovery from continuous equilibrium data. \textit{In Proceedings of the 29th Conference on Uncertainty in Artificial Intelligence (UAI-13)}. (A. Nicholson and P. Smyth, eds.) 431-439. AUAI Press, Corvallis, OR, USA.

\bibitem{Mooij:2016}  \textsc{Mooij, J. M., Peters, J., Janzing, D., Zscheischler, J.} and \textsc{Sch\"{o}lkope, B.} (2016). Distinguishing cause from effect using observational data: Methods and benchmarks.  \textit{J. Mach. Learn. Res.} \textbf{17} Paper No. 32, 102. MR3491126.

\bibitem{Mori:1999} \textsc{Mori, Y., Takahashi, H., and Oka, R.} (1999). Image-to-word transformation based on dividing and vector
quantizing images with words. \textit{In First International Workshop on Multimedia Intelligent Storage
and Retrieval Management} (in conjunction with ACM Multimedia Conference 1999), Orlando,
Florida.


\bibitem{Oakes:1981} \textsc{Oakes, D.} (1981). Survival times: aspects of partial likelihood. \textit{Internal.Statist. Rev.} \textbf{49}, 235-264.

\bibitem{Oliver:2012} \textsc{Oliver, D.}  (2012). 21st century health services for an ageing population: 10 challenges for general practice. \textit{British Journal of General Practice}, \textbf{62}(601), 396-397.

\bibitem{Openshaw:1983} \textsc{Openshaw, S.}  (1983). The modifiable areal unit problem. \textit{CATMOG}. \textbf{38} (Norwich, UK: Geo Books).

\bibitem{Pearl:2009}  \textsc{Pearl, J.} (2009).  \textit{Causality: Models, Reasoning, and Inference}, 2nd ed. Cambridge Univ. Press, Cambridge. MR2548166. https://doi.org/10.1017/CBO9780511803161.

\bibitem{Peters:2016}  \textsc{Peters, J., B\"{u}hlmann, P.} and \textsc{Meinshausen, N.} (2016). Causal inference by using invariant prediction: identification and confidence intervals.  \textit{J. R. Statist. Soc. B} \textbf{78}, Part 5, 947-1012.

\bibitem{Peters:2017} \textsc{Peters, J., Janzing, D.} and \textsc{Sch\"{o}lkopf, B.} (2017).  \textit{Elements of Causal Inference: Foundations and Learning Algorithms}. Adaptive Computation and Machine Learning. MIT Press, Cambridge, MA. https://doi.org/10.1017/CBO9780511803161. MR3822088.


\bibitem{Peters:2014}  \textsc{Peters, J., Mooij, J. M., Janzing, D.} and \textsc{Sch\"{o}lkope, B.} (2014). Causal discovery with continuous additive noise models.  \textit{J. Mach. Learn. Res.} \textbf{15}, 2009-2053. MR3231600.

\bibitem{Perez:1998} \textsc{P\'{e}rez, P.} (1998). Markov Random Fields and Images. \textit{CWI Quarterly}, \textbf{11}, 413-437.

\bibitem{Pollard:2001} \textsc{Pollard, D.} (2001). \textit{A User's Guide to Mesure Theoretic Probability}. Cambridge University Press.

\bibitem{Robinson:1950} \textsc{Robinson, W. S.}  (1950). Ecological correlations and the behavior of individuals. \textit{American Sociological Review}, \textbf{15}(601), 351-357.

\bibitem{Segev:2021} \textsc{Segev, E.} (2021).  \textit{Semantic Network Analysis in Social Sciences}. London: Routledge.


\bibitem{Spirtes:2000} \textsc{Spirtes, P., Glymour, C.} and \textsc{Scheines, R.} (2000).  \textit{Causation, Prediction, and Search}, 2nd ed. Adaptive Computation and Machine Learning. MIT Press, Cambridge, MA. MR1815675.

\bibitem{Stainvas:2003} \textsc{Stainvas, I.} and \textsc{Lowe, D.} (2003). A Generative Model for Separating Illumination and Reflectance from Images. \textit{Journal of Machine Learning Research}, \textbf{4}, 1499-1519.

\bibitem{Tate:2001}  \textsc{Tate, N.} and \textsc{Atkinson, P. M.} (2001). \textit{Modelling Scale in Geographical Information Science}, 1st ed. Wiley.

\bibitem{Torrione:2008} \textsc{Torrione, P. A.} and \textsc{Collins, L.} (2008). Application of Markov random fields to landmine detection in ground penetrating radar data. \textit{In: Harmon, R.S., Holloway, J.H.J., Broach, J.T. (eds.) Detection and Sensing of Mines, Explosive Objects, and Obscured Targets XIII}, vol. \textbf{6953}.

\bibitem{Trianni:2009}  \textsc{Trianni, G.} and \textsc{Gamba, P.} (2009). Fast damage mapping in case of earthquakes using multitemporal SAR data. \textit{Journal of Real-Time Image Processing}, \textbf{4}, 195-203.

\bibitem{van der vaart:2000}  \textsc{van der vaart, A. W.} and \textsc{Wellner, J. A.} (2000). \textit{Weak Convergence and Empirical Processes: With Applications to Statistics}. Springer.

\bibitem{Wang:2021}  \textsc{Wang, J. Y. and Yu, H.}  (2022). \textit{Supplement to ``Probability of Co-occurrence and E-Integral"}.

\bibitem{Whittle:1992} \textsc{Whittle, P.} (1992). \textit{Probability via Expectation}, 3th edition. Springer-Verlag, New York. First edition 1970, under the tilte ``Probability".

\bibitem{Whittle:2000} \textsc{Whittle, P.} (2000). \textit{Probability via Expectation}, 4th edition. Springer.

\bibitem{Wright:1921}  \textsc{Wright, S.} (1921). Correlation and causation. \textit{J. Agric. Res.} \textbf{20}, 557-585.

\bibitem{yan:2004} \textsc{Yan, J. A.} (2004). \textit{Measure Theory Handout}. Science Press, China.

\bibitem{Zheng:2004}  \textsc{Zheng, H., Daoudi, M.} and \textsc{Jedynak, B.} (2004). Blocking adult images based on statistical skin detection. \textit{Electronic Letters on Computer Vision and Image Analysis}, \textbf{4}(2), 1-14.

\bibitem{Zheng:2004}  \textsc{Zheng, Y., Li, H.} and \textsc{Doermann, D.}(2004). Machine Printed Text and Handwriting Identification in Noisy Document Images. \textit{IEEE Transactions on Pattern Analysis and Machine Intelligence}, \textbf{26}, 337-353.

\bibitem{Zhu:2008} \textsc{Zhu, J., Nie, Z., Zhang, B.} and \textsc{Wen, J. R.} (2008). Dynamic Hierarchical Markov Random Fields for Integrated Web Data Extraction, \textit{Journal of Machine Learning Research}, \textbf{9}, 1583-1614.

\end{thebibliography}
\end{document}